%% file: main.tex
\ifdefined\surveyarxiv
\else
\documentclass[Afour,sageh,times]{sagej}
\fi
\usepackage[T1]{fontenc}
\usepackage{amsmath,amssymb}
\usepackage{textcomp}
\usepackage{needspace}
\usepackage{xurl}
\usepackage{graphicx}
\usepackage{booktabs}
\usepackage{tabularx}
\usepackage{array}
\usepackage[table]{xcolor}
\usepackage{pifont}
\usepackage{tikz}
\usetikzlibrary{arrows.meta,calc}

\usepackage[hidelinks]{hyperref}
\usepackage{survey-contents}

\definecolor{surveygreen}{HTML}{168344}
\definecolor{surveyred}{HTML}{CB3434}
\definecolor{surveyhighlight}{HTML}{EEF6F2}
\newcommand{\surveyyes}{\textcolor{surveygreen}{\ding{51}}}
\newcommand{\surveyno}{\textcolor{surveyred}{\ding{55}}}
\newcommand{\roboagentlab}{RoboAgentLab}
\hypersetup{
  pdftitle={RoboAgentLab: A Taxonomy and Open Research Platform for Agentic Robotics Across Acting, Building, and Evolving},
  pdfauthor={RoboAgentLab Team}
}

\title{RoboAgentLab: A Taxonomy and Open Research Platform\\for Agentic Robotics Across Acting, Building, and Evolving}
\ifdefined\surveyteamsignature
\author{RoboAgentLab Team}
\newcommand{\surveyauthornotes}{}
\runninghead{RoboAgentLab Team}
\hypersetup{pdfauthor={RoboAgentLab Team}}
\else
\input{author-metadata}
\runninghead{RoboAgentLab Team}
\fi
\ifdefined\surveyarxiv
\else
\usepackage{survey-title}
\fi

\begin{document}


\begin{abstract}
\begingroup Foundation models are extending the scope of robotic intelligence beyond task execution to the development and improvement of robot code and capabilities. This emerging paradigm, termed \textbf{agentic robotics}, where coding agents write and refine code for robots, calls for a unified account of what agents decide and how their decisions affect robot performance. We introduce a taxonomy organized around three complementary roles: \textbf{Agent as Actor}, which directs robot behavior; \textbf{Agent as Infrastructure}, which constructs resources and directs development workflows; and \textbf{Agent as Evolver}, which revises and retains system changes through feedback. We synthesize the design trade-offs across these roles by examining observation access, action authority, and what execution can reuse. Moving model computation into development changes the balance between online adaptation and preparation cost, while the value of retained programs, memories, and components depends on how they affect later runs. We introduce RoboAgentLab, an open research platform that records robot execution, memory reuse, and feedback-informed software revision. We identify future directions in efficient agentic decision-making, reliable evaluation of retained improvements, and training agents to continue robotics research.\parfillskip=0pt plus 0.5\linewidth\par\endgroup
\end{abstract}

\keywords{Agentic robotics, robot learning, embodied agents, taxonomy, research platform.}

\newcommand{\surveyteaser}{%
\begin{minipage}{\textwidth}
\centering
\captionsetup{skip=4pt,hypcap=false}
\includegraphics[width=\textwidth]{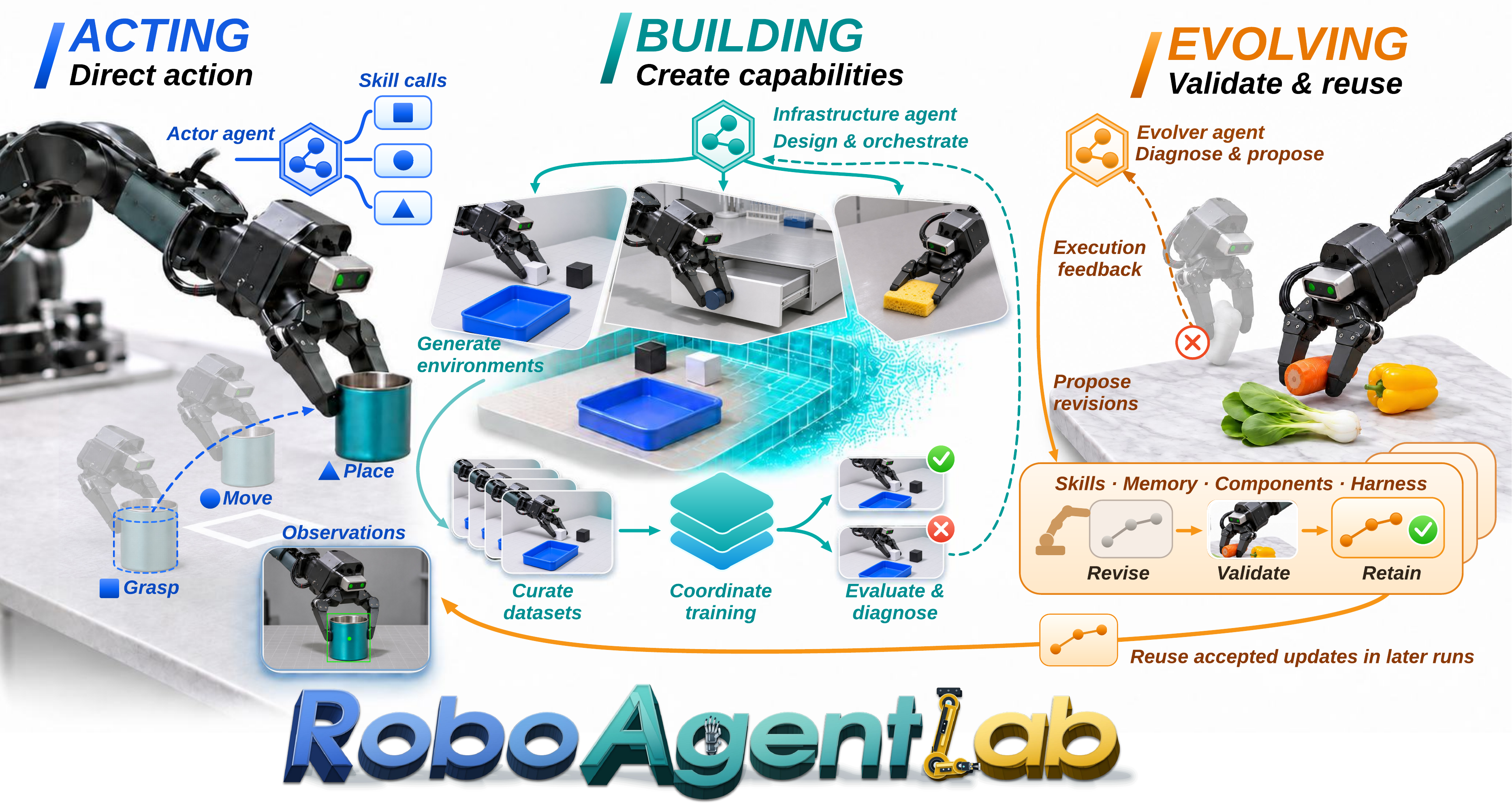}
\captionof{figure}{\textbf{Agentic robotics across acting, building, and evolving.}
\textbf{Acting:} the Agent as Actor directs robot behavior through observation and action interfaces.
\textbf{Building:} the Agent as Infrastructure constructs environments and datasets and orchestrates training, evaluation, and diagnosis.
\textbf{Evolving:} the Agent as Evolver uses experimental feedback to select update targets, revise and validate system changes, and retain them for later use.}
\label{fig:teaser}
\end{minipage}%
}

\providecommand{\surveyteaser}{}
\maketitle
\surveycontents

  \twocolumn[{%
    \begin{minipage}{\textwidth}
      \renewenvironment{figure*}[1][]{\par\centering\captionsetup{type=figure,hypcap=true,font={sf,small},labelfont=bf,labelsep=period}}{\par}%

\input{figures/timeline-linked}%
    \end{minipage}\vspace{\dbltextfloatsep}%
  }]%

\input{sections/01_introduction}

\input{sections/02_foundations_and_taxonomy}

\input{sections/03_agent_as_policy}

\input{sections/04_agent_as_infrastructure}

\input{sections/05_agent_as_self_improver}

\input{sections/06_platform}
\input{sections/07_challenges_and_future_directions}
\input{sections/08_conclusion}

\section*{Acknowledgements}
\addcontentsline{toc}{section}{Acknowledgements}
\begingroup OpenAI ChatGPT/Codex and Anthropic Claude assisted literature retrieval and the preparation of schematic illustrations. We searched across publication venues and followed references in related reviews to reduce the risk of bias and omissions in model-assisted searches. To guard against errors in model output, we checked original method descriptions and available implementation evidence before deciding which studies to include and how to classify them. We take responsibility for the accuracy, originality, source attribution, and final content of this article.\parfillskip=0pt plus 0.1\linewidth\par\endgroup

\section*{Author Contributions}
\addcontentsline{toc}{section}{Author Contributions}
Jingxiang Guo, Jiaqi Bai, Weize Li, and Yujie Zang contributed equally to the investigation and writing of this review and share first authorship.

Haoyuan Deng, Haichao Liu, Puming Jiang, Huang Huang, Peiqi Yu, Zihan Ding, Quanyi Li, Zeyi Li, Jiayi Li, Yanzhe Chen, Yiyuan Pan, Shangtao Li, Siyuan Cen, Haoyi Duan, Chen Gao, Jindou Jia, Jiachen Li (University of Texas at Austin), Zhigen Zhao, Kaixin Li, Kevin Qinghong Lin, Kevin Yuchen Ma, Anxing Xiao, Cunjun Yu, Kaiyuan Chen, Yongxi Lai, Shuangyu Xie, Xiaopan Zhang, Haoyu Zhao, Uksang Yoo, Zehao Wang, Yuhang Zheng, Xinyue Zhu, Kaichen Zhou, and Zhenyu Wei investigated the literature and edited the manuscript.

Mike Zheng Shou, Ziwei Wang, Yan Wu, Mingyu Ding, Zhuo Xu, Jiachen Li (Georgia Institute of Technology), Yilun Du, Yuke Zhu, Jiajun Wu, David Hsu, and Ken Goldberg contributed to the investigation, reviewed the literature in their areas of expertise, and provided technical edits.

Jianfei Yang and Shuicheng Yan, the co-corresponding authors, contributed to the conception and supervision of the review, project administration, and manuscript editing.

\section*{Statements and Declarations}
\addcontentsline{toc}{section}{Statements and Declarations}
\subsection*{Ethical considerations}
Ethical approval was not required. This review synthesizes published work, and the accompanying robot simulation and software experiments involved no human participants or animals.

\subsection*{Consent to participate}
Not applicable. No human participants were recruited for this study.

\subsection*{Consent for publication}
Not applicable. No identifiable individual-level data are reported in this article or its accompanying materials.

\subsection*{Declaration of conflicting interest}
David Hsu and Ken Goldberg serve on the Advisory Board of \emph{The International Journal of Robotics Research}. The authors declare no other potential conflicts of interest with respect to the research, authorship, or publication of this article.

\subsection*{Funding}
\begingroup This work was supported by the NUS Start-up Grant (A-0010106-00-00) and NAII Seed Funding (A-8004365-00-00).\parfillskip=0pt plus 0.1\linewidth\par\endgroup

\subsection*{Data availability}
Appendix~\ref{app:taxonomy} links the classified studies to their references. Figure~\ref{fig:platform-paired} reports measurements for all 18 ManiSkill runs in the paired comparison. Appendix~\ref{app:platform-data} reports the results of memory reuse, replay, and prompt revision. The \href{https://github.com/RoboAgentLab/RoboAgentLab/tree/908d969e8368dd909f24bf1373d45d0edd74e033}{RoboAgentLab source code at revision \texttt{908d969e8}} is publicly available. The project also releases selected \href{https://huggingface.co/datasets/RoboAgentLab/pi-embodied-policy-interactions}{policy interaction records} and \href{https://huggingface.co/datasets/RoboAgentLab/pi-embodied-rd-interactions}{research and development interaction records}.

\setlength{\bibsep}{0.12pt}
\renewcommand{\bibfont}{\refsize\advance\baselineskip by -0.16pt\relax}
\bibliographystyle{SageH-survey}
\bibliography{references}

\clearpage
\appendix
\input{sections/09_appendix}

\end{document}

%% file: author-metadata.tex
\author{%
  \mbox{Jingxiang Guo\textsuperscript{1*}},
  \mbox{Jiaqi Bai\textsuperscript{2*}},
  \mbox{Weize Li\textsuperscript{1*}},
  \mbox{Yujie Zang\textsuperscript{1*}},
  \mbox{Haoyuan Deng\textsuperscript{2}},
  \mbox{Haichao Liu\textsuperscript{2}},
  \mbox{Puming Jiang\textsuperscript{1}},
  \mbox{Huang Huang\textsuperscript{3}},
  \mbox{Peiqi Yu\textsuperscript{5}},
  \mbox{Zihan Ding\textsuperscript{2}},
  \mbox{Quanyi Li\textsuperscript{1}},
  \mbox{Zeyi Li\textsuperscript{1}},
  \mbox{Jiayi Li\textsuperscript{2}},
  \mbox{Yanzhe Chen\textsuperscript{1}},
  \mbox{Yiyuan Pan\textsuperscript{5}},
  \mbox{Shangtao Li\textsuperscript{5}},
  \mbox{Siyuan Cen\textsuperscript{5}},
  \mbox{Haoyi Duan\textsuperscript{9}},
  \mbox{Chen Gao\textsuperscript{1}},
  \mbox{Jindou Jia\textsuperscript{2}},
  \mbox{Jiachen Li\textsuperscript{7}},
  \mbox{Zhigen Zhao\textsuperscript{11}},
  \mbox{Kaixin Li\textsuperscript{1}},
  \mbox{Kevin Qinghong Lin\textsuperscript{10}},
  \mbox{Kevin Yuchen Ma\textsuperscript{1,12}},
  \mbox{Anxing Xiao\textsuperscript{1}},
  \mbox{Cunjun Yu\textsuperscript{6}},
  \mbox{Kaiyuan Chen\textsuperscript{4}},
  \mbox{Yongxi Lai\textsuperscript{1}},
  \mbox{Shuangyu Xie\textsuperscript{4}},
  \mbox{Xiaopan Zhang\textsuperscript{11}},
  \mbox{Haoyu Zhao\textsuperscript{1}},
  \mbox{Uksang Yoo\textsuperscript{4}},
  \mbox{Zehao Wang\textsuperscript{11}},
  \mbox{Yuhang Zheng\textsuperscript{1}},
  \mbox{Xinyue Zhu\textsuperscript{4}},
  \mbox{Kaichen Zhou\textsuperscript{3,8,13}},
  \mbox{Zhenyu Wei\textsuperscript{14}},
  \mbox{Mike Zheng Shou\textsuperscript{1}},
  \mbox{Ziwei Wang\textsuperscript{2,15}},
  \mbox{Yan Wu\textsuperscript{12,15}},
  \mbox{Mingyu Ding\textsuperscript{14}},
  \mbox{Zhuo Xu\textsuperscript{7}},
  \mbox{Jiachen Li\textsuperscript{11}},
  \mbox{Yilun Du\textsuperscript{8}},
  \mbox{Yuke Zhu\textsuperscript{6,7}},
  \mbox{Jiajun Wu\textsuperscript{3}},
  \mbox{David Hsu\textsuperscript{1}},
  \mbox{Ken Goldberg\textsuperscript{4}},
  \mbox{Jianfei Yang\textsuperscript{2,15\textdagger}},
  \mbox{Shuicheng Yan\textsuperscript{1\textdagger}}%
}

\newcommand{\surveyauthornotes}{%
  \begingroup\small\normalfont\sffamily\raggedright
  \textsuperscript{*}Equal contribution.\quad
  \textsuperscript{\textdagger}Co-corresponding authors:
  Jianfei Yang (\href{mailto:jianfei.yang@ntu.edu.sg}{jianfei.yang@ntu.edu.sg});
  Shuicheng Yan (\href{mailto:yansc@nus.edu.sg}{yansc@nus.edu.sg}).\par
  \endgroup
}

%% file: figures/timeline-linked.tex
\begin{figure*}[t]
  \centering
  \includegraphics[width=\textwidth]{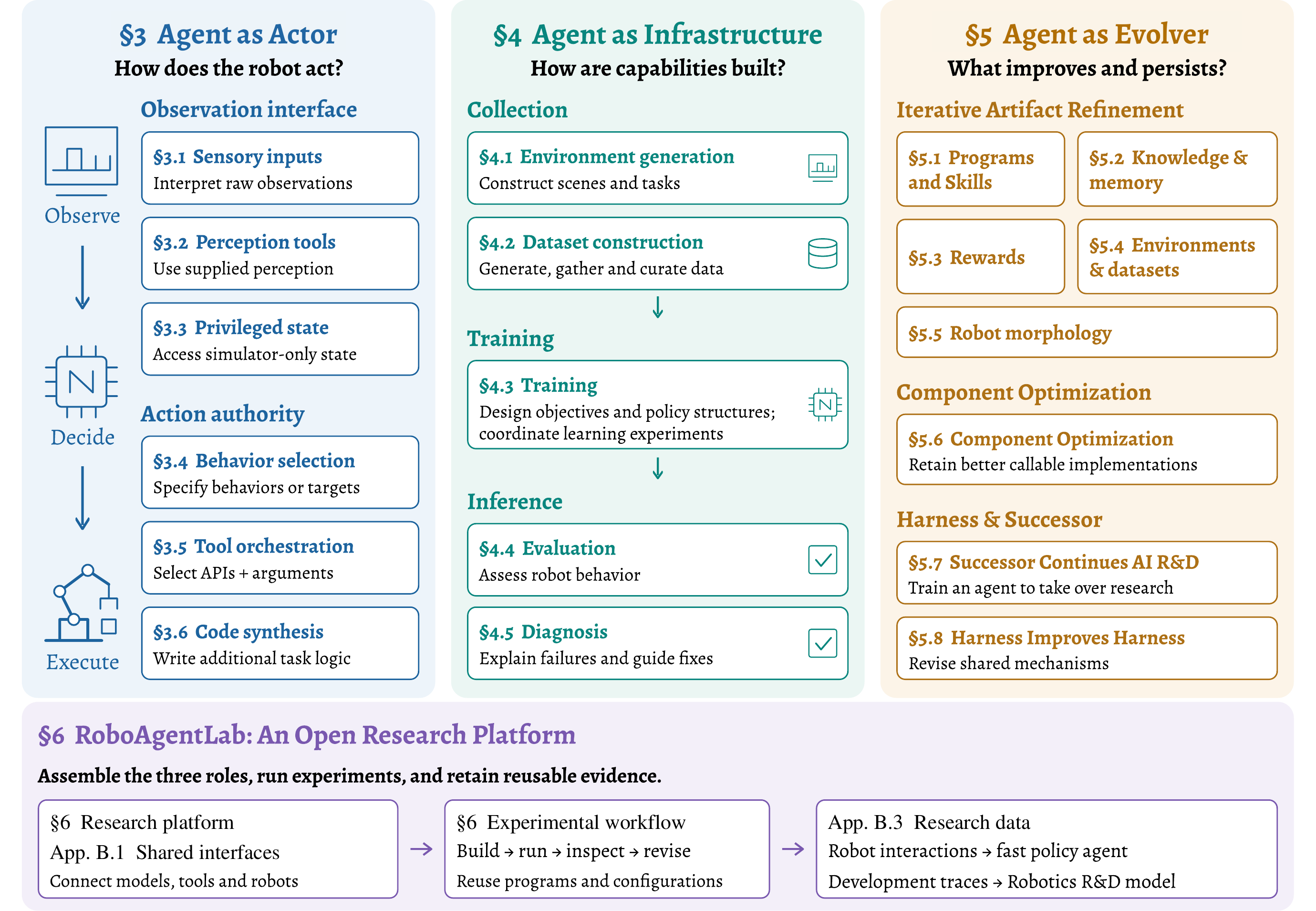}
  \caption{Reading map of the survey. Actor is organized by
  observation interfaces and action authority; Infrastructure
  by collection, training, and inference; and Evolver by
  retained update targets. RoboAgentLab connects these roles
  through a shared experimental platform. References point to the corresponding sections and appendices.}
  \label{fig:reading-map}
\end{figure*}

%% file: sections/01_introduction.tex
\section{Introduction}
\label{sec:introduction}

\begingroup Foundation models are increasingly enabling intelligent agents to direct robot behavior and participate in the development of robotic capabilities~\citep{firoozi2025foundation}. These agents can select skills, generate executable programs, coordinate learning workflows, and revise system components through feedback~\citep{saycan2022,grid66,jin2026nautilus,liang2026robofoundry}. Together, these advances suggest an emerging paradigm of \emph{agentic robotics}, in which agents assume responsibilities across the robotics lifecycle. This paradigm raises a fundamental question: how do intelligent agents contribute to robotic systems, and how can their roles be systematically characterized?\parfillskip=0pt plus 0.5\linewidth\par\endgroup

\begingroup Research in agentic robotics spans robot control, automated capability development, and feedback-driven system improvement. These directions are closely related but address different aspects of robotic intelligence: agents may direct existing capabilities, support the creation of new ones, or modify system components based on experience. Existing surveys examine foundation-model integration, executable skills, self-improvement mechanisms, and supporting robotic infrastructure~\citep{embodiedsurvey2025,weightsskills2026,wang2026agenticroboticssurvey,qin2026towardsagenticrobots,huang2026embodiedrsisurvey}. Li et al.\ classify executable programs for physical tasks by their role and by the evidence used to judge their outcomes~\citep{li2026worldtestsuite}. Building on these perspectives, we focus on the responsibilities delegated to agents and the relationships between their decisions, the resources they develop, and the changes retained for subsequent use (Table~\ref{tab:related-surveys}).\parfillskip=0pt plus 0.5\linewidth\par\endgroup

\begingroup We organize agentic robotics around three complementary roles (Figure~\ref{fig:reading-map}). \emph{Agent as Actor} directs robot behavior through skill selection, tool orchestration, and code synthesis, with observation access and action authority defining distinct interfaces. \emph{Agent as Infrastructure} supports capability development through environment and data generation, training, evaluation, and diagnosis. \emph{Agent as Evolver} uses feedback to revise and retain executable artifacts, callable components, and the agent harness. These responsibilities can coexist within a system, linking execution with resource development and validated revisions.\parfillskip=0pt plus 0.5\linewidth\par\endgroup

\begin{table*}[t]
\centering
\caption{Comparison with surveys of agentic robotics and robot self-improvement.}
\label{tab:related-surveys}
\small
\setlength{\tabcolsep}{1pt}
\renewcommand{\arraystretch}{1.16}
\renewcommand{\tabularxcolumn}[1]{m{#1}}
\begin{tabularx}{\textwidth}{@{}>{\raggedright\arraybackslash}X
>{\centering\arraybackslash}m{0.073\textwidth}
>{\centering\arraybackslash}m{0.112\textwidth}
>{\centering\arraybackslash}m{0.100\textwidth}
>{\centering\arraybackslash}m{0.120\textwidth}
>{\centering\arraybackslash}m{0.103\textwidth}
>{\centering\arraybackslash}m{0.062\textwidth}@{}}
\toprule
\textbf{Survey} & \multicolumn{3}{c}{\textbf{Roles covered}} & \multicolumn{3}{c}{\textbf{Analytical structure}} \\
\cmidrule(lr){2-4}\cmidrule(l){5-7}
& \textbf{Agent as Actor} & \textbf{Agent as Infrastructure} & \textbf{Agent as Evolver} & \textbf{\shortstack{Observation \&\\action axes}} & \textbf{Development lifecycle} & \textbf{Update targets} \\
\midrule
Towards Embodied Agentic AI~\citep{embodiedsurvey2025} & \surveyyes & \surveyyes & \surveyno & \surveyno & \surveyno & \surveyno \\
Multimodal Embodied Agents~\citep{chen2026papav} & \surveyyes & \surveyno & \surveyyes & \surveyno & \surveyno & \surveyno \\
Toward Recursive Self-Improvement in Physical AI~\citep{luo2026physicalrsi} & \surveyno & \surveyyes & \surveyyes & \surveyno & \surveyno & \surveyyes \\
Weights or Skills?~\citep{weightsskills2026} & \surveyyes & \surveyyes & \surveyyes & \surveyno & \surveyno & \surveyyes \\
Agentic Robotics~\citep{wang2026agenticroboticssurvey} & \surveyyes & \surveyyes & \surveyyes & \surveyno & \surveyno & \surveyyes \\
Self-Evolving AI for Humanoids~\citep{nguyen2026selfevolvinghumanoids} & \surveyyes & \surveyyes & \surveyyes & \surveyno & \surveyno & \surveyyes \\
Toward Self-Improving Robotic Agents~\citep{zi2026selfimprovingrobotic} & \surveyyes & \surveyyes & \surveyyes & \surveyno & \surveyno & \surveyyes \\
Towards Agentic Robots~\citep{qin2026towardsagenticrobots} & \surveyyes & \surveyyes & \surveyyes & \surveyno & \surveyyes & \surveyyes \\
Recursive Self Improvement for Embodied Agents~\citep{huang2026embodiedrsisurvey} & \surveyyes & \surveyyes & \surveyyes & \surveyno & \surveyyes & \surveyyes \\
When the World Is the Test Suite~\citep{li2026worldtestsuite} & \surveyyes & \surveyyes & \surveyyes & \surveyno & \surveyyes & \surveyyes \\
\midrule
\rowcolor{surveyhighlight}
\textbf{\roboagentlab\ (ours)} & \surveyyes & \surveyyes & \surveyyes & \surveyyes & \surveyyes & \surveyyes \\
\bottomrule
\end{tabularx}
\par\smallskip
{\footnotesize\raggedright
\surveyyes: dedicated treatment or explicit classification; \surveyno: absent or incidental coverage.\par}
\end{table*}

\begingroup Online coordination keeps model reasoning available during execution, whereas reusable programs can move that computation into development~\citep{hu2026coap,wang2026sear}.
Comparing these approaches requires accounting for observation access, action authority, preparation effort, and response latency.
The value of generated environments, datasets, and retained revisions likewise depends on their effects on subsequent robot behavior~\citep{tang2026video2world,meng2025lyra}.
Evaluation must therefore link development decisions to later outcomes, distinguishing artifact validity from task performance and testing retained changes under new physical conditions.\parfillskip=0pt plus 0.5\linewidth\par\endgroup

\begingroup We introduce \roboagentlab{}, an open research platform for inspecting agent decisions through execution and development records. ManiSkill runs document tool orchestration, code synthesis, and memory reuse. A separate development session links execution failures to a reviewed prompt revision and follow-up runs (Section~\ref{sec:platform}).\parfillskip=0pt plus 0.5\linewidth\par\endgroup

Our contributions are:
\begin{itemize}
\setlength{\parfillskip}{0pt plus 0.5\linewidth}
    \item \textbf{A role-based taxonomy} of agent responsibilities across robot execution, capability development, and iterative improvement, characterized through decision interfaces, development workflows, and retained system changes (Sections~\ref{sec:taxonomy} to~\ref{sec:self-improvement}).

    \item \textbf{A cross-role synthesis} of design trade-offs and evaluation requirements, covering decision efficiency, validation of retained improvements, and continued robotics experimentation (Section~\ref{sec:challenges}).

    \item \textbf{\roboagentlab{}}, an open research platform that links robot actions to the observations, tool calls, and code that produced them (Section~\ref{sec:platform}).
\end{itemize}

%% file: sections/02_foundations_and_taxonomy.tex
\section{Foundations and Taxonomy}
\label{sec:taxonomy}

\subsection{Definition and Scope}
\label{sec:scope}
\begingroup Building on the working definitions in Section~\ref{sec:introduction}, we classify systems by the responsibilities assigned to their agents. These responsibilities may concern robot behavior, data collection, environment construction, or learning experiments, and are organized into three roles: \emph{Actor}, \emph{Infrastructure}, and \emph{Evolver}. An Evolver uses feedback to revise skills, components, or workflows and retains validated changes for later runs.\parfillskip=0pt plus 0.1\linewidth\par\endgroup

We cover physical and simulated robots, together with tools and workflows built to develop their capabilities. Using a foundation model alone is insufficient: ordinary VLA inference, implicit reasoning, fixed scoring, and fixed training or post-training do not establish an agent role. The agent must decide robot behavior or a development intervention. When a system combines agent-directed and fixed processes, we classify only the responsibilities demonstrated by its agent. Human guidance and approval may remain part of the workflow. In collaborative manipulation, CollaBot uses a vision-language model to select grasp points for a robot team within a fixed planning procedure~\citep{song2025collabot}. Related evaluation resources illustrate this boundary: SimuGuard audits simulated contact failures through physics checks and replay~\citep{zhou2026simuguard}, and ContinualDrive compares updated planners with their deployed checkpoints~\citep{li2026continualdrive}. In the text-based ALFWorld configuration of System1-Agents, Jev selects among admissible household commands from a textual observation; AI2-THOR renders the resulting trajectory for replay~\citep{system1agents2026}. For video repair, including manipulation scenes~\citep{chen2026physloop}, a robot-domain role requires evidence that the repaired output informs a robotics decision. In robot video simulation, Agentic LongLive selects reference images to preserve object appearance and location as scripted actions bring objects back into view~\citep{liu2026agenticlonglive}.

\begingroup \emph{Literature selection.} We identified candidate works through targeted web searches of arXiv, publication venues, and official project pages, and by following references in related reviews. We last updated the main literature search on 26 October 2026 and subsequently added selected works. We considered peer-reviewed papers, preprints, and documented research prototypes. We checked original method descriptions and available implementation evidence against the scope above, consolidated versions and aliases, and assigned taxonomy roles to documented configurations. Inclusion and placement required evidence of agent decisions within the scope above. The resulting synthesis connects these decisions to the three-role taxonomy.\parfillskip=0pt plus 0.1\linewidth\par\endgroup

\subsection{Role-Based Taxonomy}
\label{sec:roles}
The full taxonomy in Appendix~\ref{app:taxonomy} compares Actor observation and action interfaces (Figure~\ref{fig:role-taxonomy}a), Infrastructure development workflows (Figure~\ref{fig:role-taxonomy-infrastructure}b), and Evolver update targets (Figure~\ref{fig:role-taxonomy-evolver}c).

\subsubsection{Agent as Actor}
\begingroup The agent directs robot behavior through observation and action interfaces. Here, \emph{observation} denotes information available to the agent; \emph{perception} interprets sensor measurements, while \emph{state estimation} infers robot or environment state. Observation interfaces ($O$) are classified as initially supplied: sensory inputs are $O1$ and agent-built perception remains $O1$; supplied perception tools or outputs, including estimates of object poses or scene state, are $O2$; simulator privilege takes precedence as $O3$.
Action authority ($A$) describes which action decisions the agent makes: $A1$ specifies behaviors, continuous targets, or trajectories; Code/Model denotes its fixed programmed or learned executor, not the deciding agent. $A2$ selects existing execution APIs, supplies arguments, and coordinates their use; $A3$ writes code with additional logic, including conditions, loops, or intermediate computations. $A3$ may reuse APIs and construct programs before deployment (a Python API call alone still constitutes tool invocation).
\parfillskip=0pt plus 0.1\linewidth\par\endgroup

\subsubsection{Agent as Infrastructure}
\begingroup The agent automates robot development by constructing or organizing resources and workflows across collection, training, and inference. In \emph{Collection}, it builds scenes and tasks or generates, gathers, and curates robot data. In \emph{Training}, it designs learning objectives or policy structures and coordinates experiments. In \emph{Inference}, it constructs or selects evaluation procedures, assesses robot execution, and investigates failures to guide further development. The agent can construct a workflow in one pass or refine it with human guidance, making decisions about its design, configuration, or operation.\parfillskip=0pt plus 0.1\linewidth\par\endgroup

\subsubsection{Agent as Evolver}
The agent uses feedback to revise the system and retains changes for subsequent use (Figure~\ref{fig:self-improver-tree}). \emph{Iterative Artifact Refinement} revises programs and skills, stored knowledge, rewards, environments and datasets, or robot-body designs. The account follows the feedback, the resulting revision, its validation, and its effect when reused. In streaming video question answering, MORSE revises the code that stores and retrieves observations and tests each revision on held-out streams~\citep{zhang2026morse}. Environment and data construction supplies initial resources; refinement changes these artifacts in response to execution or development outcomes.

\begingroup \emph{Component Optimization} changes a callable component's implementation. \emph{Model Trains Model} directs learning or adaptation of a functional model, including capability distillation into a policy or perception model. Weight-space programming can repair a PPO policy from a failed simulated driving episode~\citep{takehi2026weightspace} or construct transformer weights from scratch for maze solving~\citep{elsayed2026written}. In robotics, the agent revises selected parameters of an existing policy using task feedback~\citep{gong2026roboticsweightediting,zhang2026debugweights,wang2026vlaweightediting}. \emph{Model Revises Tool Code} uses feedback to revise a tool's internal algorithm. Infrastructure Training organizes learning workflows; component optimization concerns the learned capability retained for later use. Reward revision changes the learning objective, while model optimization changes the learned component.\parfillskip=0pt plus 0.1\linewidth\par\endgroup

\begingroup \emph{Harness Improves Harness} revises the agent's supporting software: input representation, task and interaction state management, tool interfaces, or execution workflows. Updating a stored lesson changes an artifact; changing its storage or retrieval mechanism changes the harness. Revising a tool's algorithm changes a component; revising its calling interface changes the harness. \emph{Successor AI R\&D} trains an agent to take over the harness and continue research decisions; the cited examples concern AI R\&D beyond robotics.\parfillskip=0pt plus 0.1\linewidth\par\endgroup

\subsubsection{Connections Between Roles}
\begingroup An Infrastructure agent can call an Actor agent to collect demonstrations or test behavior. A Evolver can use Infrastructure to construct practice environments and evaluate revisions, with the Actor providing execution feedback. Accepted updates enter the affected modules in the next cycle. $A3$ constructs executable task logic; Programs and Skills follows its feedback-guided revision and reuse; tool-code optimization changes the implementation of a called component. Construction and improvement can occur within the same system, with each role supported by its corresponding decisions and retained outputs.\parfillskip=0pt plus 0.1\linewidth\par\endgroup

%% file: sections/03_agent_as_policy.tex



\section{Agent as Actor}\label{sec:policy}

\begin{figure}[!t]
\centering
\setlength{\abovecaptionskip}{4pt}
\includegraphics[width=\linewidth,trim=4bp 2bp 4bp 2bp,clip]{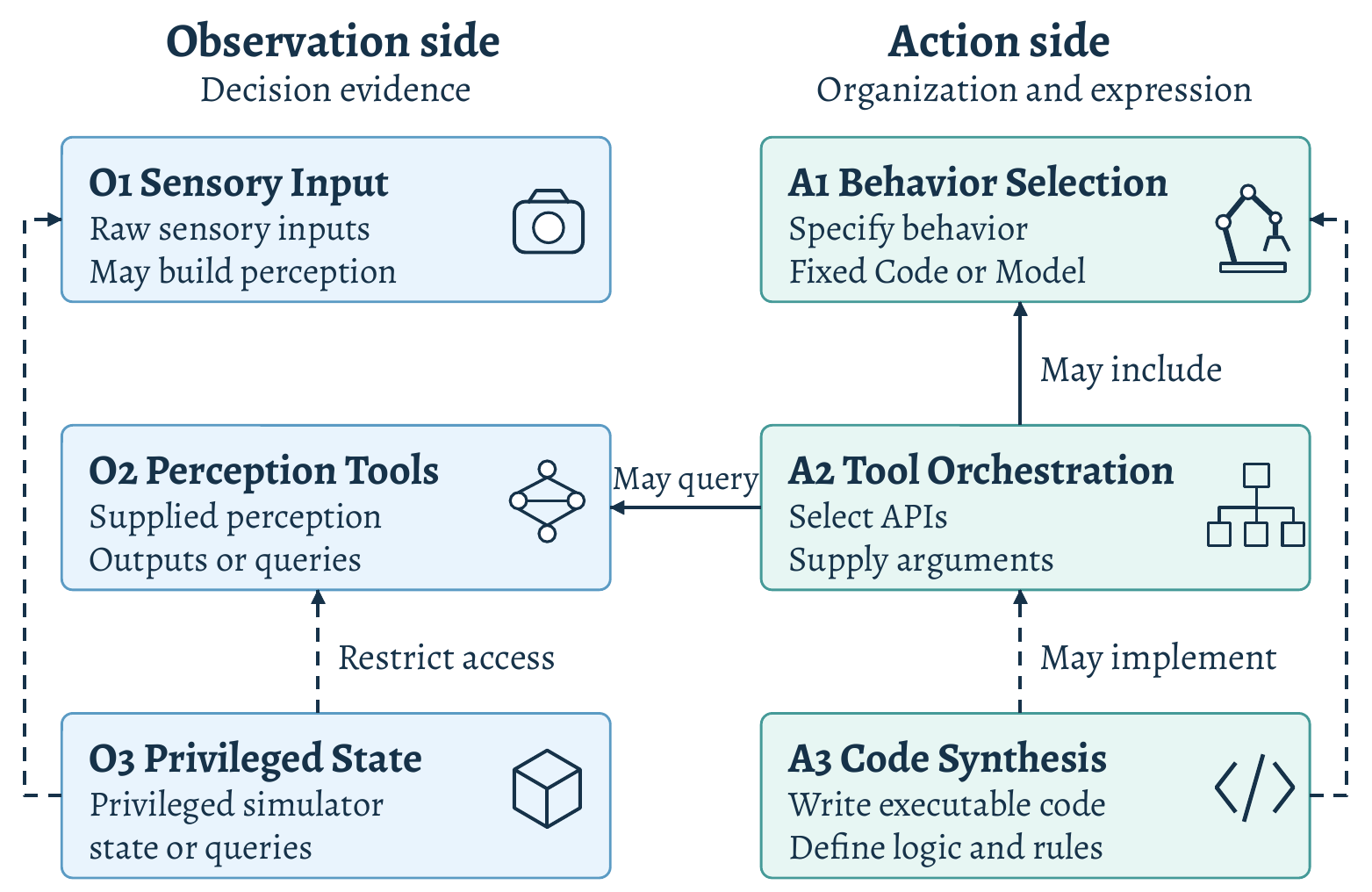}
\caption{Observation and action interfaces. Restricting privileged access yields O1 or O2; arrows on the action side indicate possible inclusion, implementation, or querying.}
\label{fig:policy-axes}
\end{figure}

An agent serves as a policy when it directly shapes robot behavior by selecting actions, coordinating planning and control components, or synthesizing executable control logic. Its decisions may concern task planning, motion planning, or low-level control. An action may be a motion command or a task-level behavior; execution is the process of carrying it out.
We organize these systems along the two complementary axes in Figure~\ref{fig:policy-axes}. The observation axis distinguishes sensory inputs (O1), supplied perception tools or outputs (O2), and privileged state (O3). The action axis distinguishes behavior selection through a fixed executor (A1), tool orchestration (A2), and code synthesis (A3). We classify each evaluated configuration by its observation interface and action authority; a system can occupy multiple cells across experiments. For observation, classification follows the initially supplied interface; for action, it follows the authority exercised by the agent. Training the underlying skills and retaining validated revisions for later runs are addressed in Agent as Infrastructure (Section~\ref{sec:infra-training}) and Agent as Evolver (Section~\ref{sec:self-improvement}), respectively.

\begin{figure*}[!t]
\centering
\includegraphics[width=\textwidth]{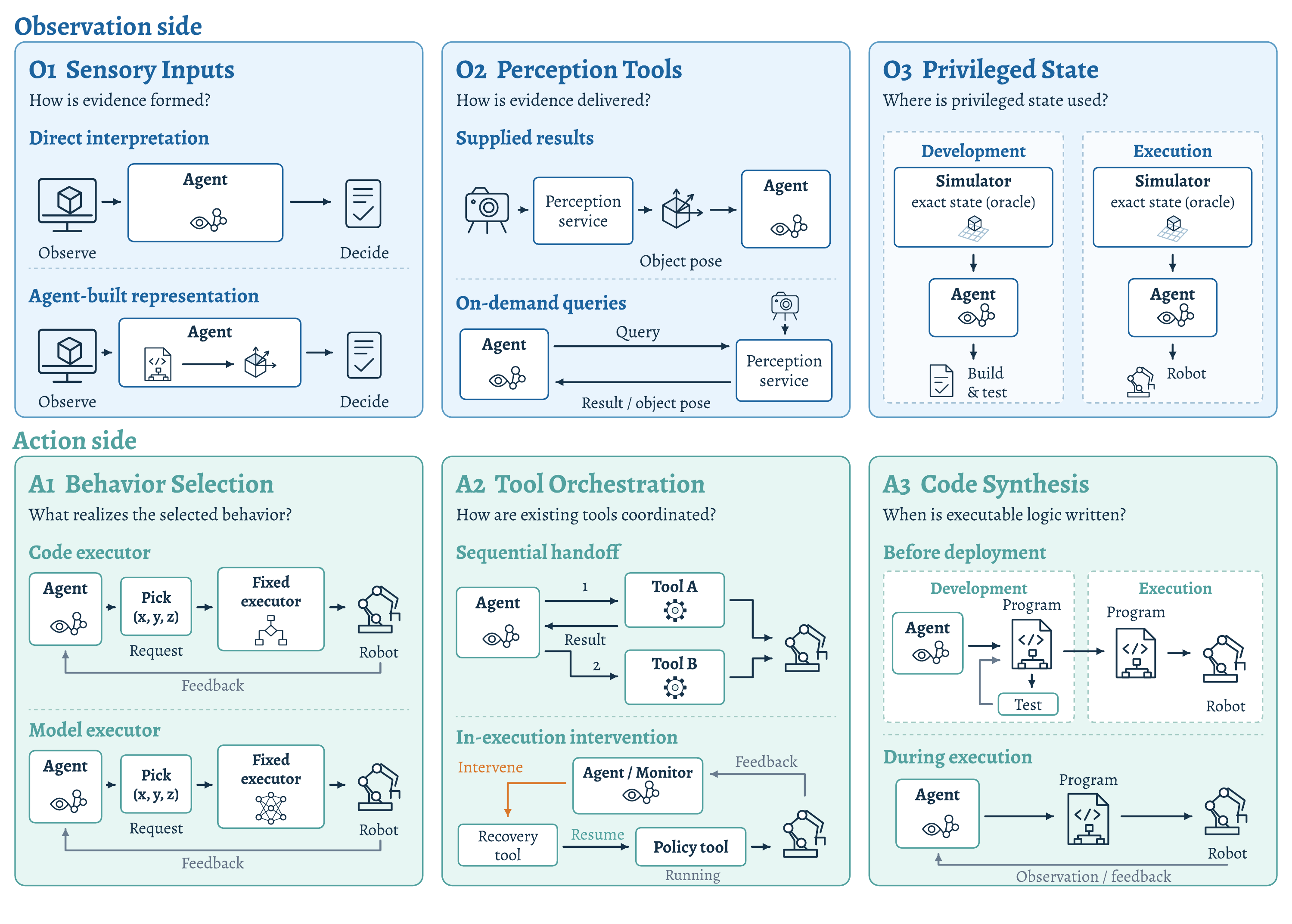}
\caption{Mechanisms within Agent as Actor. Observation-side comparisons concern evidence formation (O1), delivery of supplied perception (O2), and the stage of privileged-state use (O3). Action-side comparisons concern fixed executor type (A1), tool handoff and intervention (A2), and code-generation timing (A3). A system may combine mechanisms within a panel.}
\label{fig:policy-mechanisms}
\end{figure*}

Figure~\ref{fig:policy-mechanisms} illustrates these policy mechanisms.


\subsection{Policies with Sensory Inputs}
\label{sec:policy-o1}
Sensory-input policies ground decisions in observations such as camera images, depth, proprioception, and tactile measurements. The agent must extract information relevant to the requested action without an initially supplied interface that already identifies objects, estimates their poses, or answers semantic scene queries. 
The central challenge is bridging perceptually uncertain observations and the spatial precision required for physical interaction.
PIVOT addresses part of this gap by drawing candidate actions on the current image and iteratively refining the candidates selected by a vision-language model~\citep{nasiriany2024pivot}. The visual interface connects a discrete model response to continuous action targets. 
This shifts the bottleneck from unconstrained action prediction to the coverage and discriminability of visually represented action candidates.
Additional sensing can reveal failures that are difficult to diagnose visually. TacZero supplies camera images, robot state, and three-axis tactile readings to a general-purpose VLM that requests end-effector targets and gripper commands~\citep{tanaka2026taczero}. In its real-world peg-insertion study, numerical tactile input increased success from 10/20 trials without touch to 15/20. This peg-insertion study illustrates how contact evidence can guide decisions to continue, retract, or adjust a motion. Contact Harness uses textual skills to guide the VLM in combining images, wrist wrench, and tactile readings where available to judge whether contact is intended or the object is jammed~\citep{shan2026contactharness}. Sensor representation, interpretation, and feedback timing all affect whether the extra evidence changes a useful decision.

Temporal context supplies another form of grounding. RoboICL structures interaction around observations, actions, execution receipts, and subsequent observations, combining demonstration context with recent execution history~\citep{roboicl2026}. Such context helps distinguish the state the agent intended to reach from the outcome it actually observed. The direct action-selection configuration in SenseNova's LIBERO evaluation similarly uses frozen, task-matched preparation notes during evaluation~\citep{sensenova2026embodied}. These examples separate adaptation through context and feedback from changes to model parameters or persistent revisions of the agent.

\begingroup An agent may also construct its own perceptual representation from sensory data (Table~\ref{tab:policy-o1}). In RAX's sensory configurations, generated programs process images and depth to recover geometry needed for manipulation~\citep{huang2026rax}. These configurations remain O1 because the agent constructs the perception routines from the supplied sensory inputs. The resulting behavior may nevertheless fall under A3, as discussed in Section~\ref{sec:policy-a3}. Thus, O1 identifies the sensory starting point from which the agent builds representations for subsequent decisions.\parfillskip=0pt plus 0.1\linewidth\par\endgroup

\input{tables/policy-o1-comparison}

\subsection{Policies with Perception Tools}
\label{sec:policy-o2}
Perception-tool policies receive semantic or geometric evidence through an interface supplied to the agent. This interface may expose a previously constructed scene representation, return estimates of object poses or scene state alongside sensor measurements, or provide services that the agent queries when additional information is needed (Table~\ref{tab:policy-o2}). Classification follows the evidence supplied to the agent, whether text, annotated images, or numerical records. Geometry used internally by a fixed motion executor remains outside the agent's observation interface.

Queryable representations make scene information accessible at the level of task reasoning. NLMap connects open-vocabulary object queries to a scene representation, allowing a planner to check the availability and location of task-relevant objects~\citep{chen2022nlmap}. ConceptGraphs integrates observations into a 3D representation that supports semantic and spatial queries~\citep{gu2023conceptgraphs}. SayPlan uses a collapsed 3D scene graph to retrieve the relevant subgraph before planning~\citep{sayplan2023}. LLM-State maintains object attributes and action summaries over supplied detections for replanning~\citep{grid54}, while HUME tests hypotheses about missing world knowledge through scene graphs and visual queries~\citep{audit260706501}. These interfaces reduce the amount of visual interpretation that the agent must repeat, while making decisions dependent on the coverage, granularity, and freshness of the supplied representation. A missing object or outdated relation can invalidate a plan even when the subsequent reasoning is internally consistent.

For manipulation, perceptual outputs must also connect to the coordinates and tolerances used by control. MotorMind's Executor receives localization offsets, uncertainty, and disagreement between views, derived from supplied grounding procedures, together with robot and execution state~\citep{motormind2026}. This configuration belongs to O2 even though it also exposes camera images. The uncertainty and disagreement are operational evidence: they can motivate another observation or a revised target before committing to contact. MIGU combines semantic and pointing evidence in a grounding belief used by a fixed execute-or-clarify rule~\citep{lu2026migu}. ApBot grounds manual-derived actions in detected appliance controls and checks outcomes through visual feedback~\citep{zhang2025robot}. In RAX's supplied-Gemini configuration, semantic localization is likewise provided as a service, while generated programs use depth and geometric checks to refine the resulting estimates~\citep{huang2026rax}. This differs from RAX configurations in which the agent constructs perception from the initial sensory interface.

The key trade-off is between reusing available perceptual evidence and actively acquiring new information. Cached representations reduce repeated computation but risk stale state, whereas on-demand queries improve observability at additional latency and failure risk.
Robust execution requires the agent to relate returned estimates to subsequent outcomes, especially when occlusion, contact, or object motion changes the scene~\citep{liu2026unimanip}. Whether the agent then selects a fixed behavior, orchestrates additional services, or writes a control program is a separate distinction developed in Sections~\ref{sec:policy-a1}--\ref{sec:policy-a3}.




\input{tables/policy-o2-comparison}

\subsection{Policies with Privileged State}
\label{sec:policy-o3}
\begingroup Privileged-state policies receive information that would not be directly available through the intended deployment interface, such as exact simulator object poses, hidden articulation variables, or oracle task state. Privilege is relative to the deployment setting and the phase being evaluated. Simulation alone does not imply O3: a simulated robot may expose only rendered images and ordinary proprioception. Conversely, a structured state vector can contain privileged quantities even when it resembles an estimated state used on a real robot. State hidden inside the executor or used only to score an episode must also be distinguished from information available to the agent.\parfillskip=0pt plus 0.1\linewidth\par\endgroup

Privileged interfaces isolate reasoning and composition from some perception errors. APPL's evaluated runtime agent and skill policies receive structured simulator state, including object poses, articulation state, and goal information~\citep{jiang2026appl}. This supports explicit arguments and numerical stopping conditions, but its composition results must be interpreted under that observation assumption. CaP-X makes the perception distinction explicit through interface tiers: its S1 setting supplies ground-truth masks and poses, whereas S2 uses perceptual outputs estimated from RGB-D observations~\citep{capx}. Such comparisons help identify how much a reported policy depends on reliable state access rather than on its planning or programming mechanism alone.

\begingroup Privilege can also be confined to development (Table~\ref{tab:policy-o3}). Embodied Turing Machines (ETM) uses oracle information while constructing and diagnosing programs, but its deployed library estimates state from camera images and proprioception without oracle access~\citep{hu2026coap}. In a separate controlled analysis, the same program achieved 63\% success with measured state and 83\% with oracle state, revealing a substantial perception bottleneck. The development and deployment configurations therefore receive separate observation classifications.\parfillskip=0pt plus 0.1\linewidth\par\endgroup

Reporting the phase and recipient of each state signal is essential for interpreting generalization. Transferring a privileged-state result to physical manipulation requires replacing oracle quantities with estimates and assessing how errors propagate through selection, stopping, and recovery. Transfer evaluation should test the resulting behavior under uncertain and delayed observations.


\input{tables/policy-o3-comparison}

\subsection{Behavior Selection}
\label{sec:policy-a1}
In behavior selection, the agent issues requests to a fixed executor. It may choose the behavior, arguments, duration, or stopping conditions, but the executor retains responsibility for realizing each request. The request can cover an entire task or a single motion, and the executor may be programmed or learned (Table~\ref{tab:policy-a1}). 
The distinction from A2 is not the number of available skills, but whether the agent selects behaviors within a predefined execution protocol or dynamically determines how execution components are composed and coordinated.

Feedback determines when the agent should revise its next request. Inner Monologue returns success assessments, scene descriptions, and human answers after skill execution, enabling the language model to replan~\citep{innermonologue}. Hi Robot incorporates situated human corrections into commands issued to a low-level policy~\citep{grid22}. KnowNo uses calibrated uncertainty over candidate next steps to request assistance when the choice remains ambiguous~\citep{grid49}. These mechanisms address different gaps: discrepancies between expected and observed outcomes, changes in the user's intent, and uncertainty about the intended action. Their immediate effect is to alter task execution; retaining a revised policy or decision procedure across tasks is a separate form of improvement.

\subsubsection{Code-based Execution}
\label{sec:policy-a1-code}
Code-based executors admit the widest range of request sizes, down to individual motions. At the coarse end, SayPlan requests a whole-task plan over a 3D scene graph and corrects infeasible steps with feedback from a scene-graph simulator before execution~\citep{sayplan2023}. HUME supplies symbolic hypotheses to a fixed planner~\citep{audit260706501}; ApBot selects parameterized appliance macros expanded by fixed transition rules, using visual feedback on hardware and simulator state in simulation~\citep{zhang2025robot}. PIVOT draws candidate actions on the image, asks a VLM to pick the best, and refines the candidates over several rounds, so a model that only outputs text ends up selecting continuous targets~\citep{nasiriany2024pivot}. TacZero gives a general-purpose VLM camera images, robot state, and three-axis tactile readings, and has it output end-effector targets and gripper commands for a low-level controller; with numerical tactile input it inserted a peg in 15 of 20 real trials, against 10 of 20 without~\citep{tanaka2026taczero}. At the finest grain, GPT-6-Astra keeps an entire RoboMME episode in context and outputs every end-effector waypoint and gripper command, succeeding in 92.1\% of 800 test episodes, against 79.1\% for a hierarchical system that pairs the same model with a benchmark-trained policy~\citep{xref261017350}. Its 63 failures were mostly premature completions and mistimed or unintended button presses. As requests shrink, the model absorbs work once left to planners and controllers, and the reported failures shift from infeasible plan steps to misjudged timing and completion.

Navigent illustrates how a fixed executor can interrupt a navigation plan~\citep{zhou2026navigent}. The agent proposes short sequences of forward and turning commands. A geometric critic checks measured motion against the commands and tracks agent-selected landmarks, halting the sequence when it detects a discrepancy so the agent can reobserve or backtrack. The hardware evaluation uses RGB-D and wheel odometry (O1), while the Habitat benchmarks supply the agent with ground-truth self-pose (O3). On the physical robot, Navigent succeeded in 27 of 30 trials across six instructions.





\input{tables/policy-a1-comparison}

\subsubsection{Model-based Execution}
\label{sec:policy-a1-model}
With a learned executor, the agent's own choice among behaviors is what makes the system A1; ordinary VLA inference or a fixed scoring rule makes no such choice. Because the agent cannot read how a learned behavior works, it must rely on what the interface reports about it. SayCan obtains this report at decision time, as a number: value functions associated with each skill estimate whether it can succeed from the current state, and the estimate is combined with the language model's judgment of the skill's usefulness~\citep{saycan2022}. VLAs-as-Tools obtains it during execution, through explicit tool selection and in-execution progress feedback that let the agent replan when an event occurs instead of polling~\citep{vlaastools2026}. APPL writes it at construction time, as a description~\citep{jiang2026appl}. Each frozen diffusion policy in its library carries a description of the structural prior used to train it, its demonstrated training support, its handoff conditions, and verification evidence. For every call, the runtime agent chooses a policy, typed arguments, a duration of at most 300 steps, and numeric stop conditions that the executor checks after each step. In five long-horizon ManiSkill tasks, where the agent and the policies read privileged structured state from the simulator, removing these descriptions from an otherwise unchanged library reduced success on unseen skill compositions from 8 to 2 of 16. When the library was replaced by a single fine-tuned VLA that the same agent invoked by skill name, 6 of the 16 compositions succeeded, and success on task-level shifts, such as resuming a partly completed task, fell from 92.5\% to 70.0\%. These systems differ in what the selector knows, not in the authority it holds. APPL's runtime agent, for example, cannot write control code or modify the library, and its memory is reset between episodes.

A fundamental challenge is the mismatch between the states a skill is designed to handle and those produced by preceding skills.
\citet{rui2026semantichandoff} report selected navigation, grasping, placement, and door-opening skills that succeed in 77--100\% of trials from clean skill-boundary snapshots, yet their composed rollouts frequently stall from the states earlier skills leave behind. Text2Motion checks handoffs during planning: it treats the learned Q-functions of its skills as feasibility estimates and resolves geometric dependencies that span a skill sequence, solving 82\% of its long-horizon problems where earlier language-based planners solved 13\%~\citep{grid67}. APPL addresses them during training (Section~\ref{sec:infra-training}), overlapping each skill's training segments with its neighbors' so that a successor can take over from states its predecessor actually reaches. Both prepare for handoffs before execution; the orchestrators in Section~\ref{sec:policy-a2} can also respond to mismatched handoffs while the task runs.






\subsection{Tool Orchestration}
\label{sec:policy-a2}

In tool orchestration, the agent decides which existing components to call, with what arguments, and when control passes between them; it does not write the logic inside them. Whether the agent selects or orchestrates, it must know where each callable behavior works and whether that behavior can start from the state its predecessor leaves. Orchestration makes both questions harder. When one process trains every policy for a single executor, it can record this information during training, as APPL does with the interfaces discussed in Section~\ref{sec:policy-a1}~\citep{jiang2026appl}. RoboHarness instead wraps independently developed VLAs, world-action models, RL policies, and task-and-motion planners as callable skills~\citep{grid27}. Because these policies were developed separately, it treats their capability boundaries as uncertain and their handoffs as a source of distribution mismatch, reasons over memory to route subtasks to policies, and chains policies through a plug-in Memory Bridge rather than joint retraining. Orchestration can also operate over elementary tools. ZAP's harness offers no grasp macros: the model works through five tools, steering a background segmenter with text queries and building every approach, grasp, lift, and release from end-effector pose and gripper calls~\citep{pikus2026pretrainingknowswhatsimulation}. Much of the performance depends on the harness: in the same study, GPT-6-Astra succeeded in 96.7\% of real-robot trials in this harness and in 36.7\% in the Robocurve harness.

An orchestrator also decides which component holds the critical path (Table~\ref{tab:policy-a2}). VoLo's VLM steers a VLA or world-action model mid-rollout as an interruptible tool, alongside vision models and action primitives, because the physical world does not pause for reasoning~\citep{grid23}. Spotter reverses the roles: the embodied model executes continuously, a lightweight local screener watches for errors, and the VLM intervenes only when one is detected, then returns control~\citep{li2026spotter}. With GPT as the VLM, this raised $\pi_{0.5}$ from 53\% to 83\% on a real robot; with Qwen, a successful episode took about 70\% less time than under a VLM-led baseline using the same model.

The orchestrator must also decide what evidence to gather and when to trust it. RegenHarness distinguishes a model's proposal, a controller's termination, and verified task completion, commits progress to trusted task state only through a version-bound gate, and checks the original user goal before reporting completion~\citep{si260927612}. ARNA lets the agent define task-specific workflows that query perception modules as needed before it selects navigation actions~\citep{lange2025arna}, so evidence is requested rather than streamed. Recova monitors progress, invokes a learned or programmatic recovery, verifies that the scene was restored, and resumes; when no recovery fits, a human demonstration resolves the failure~\citep{recova2026}.

Contact Harness separates the timing of tool calls from the timing of contact regulation~\citep{shan2026contactharness}. The VLM requests selected sensor readings, sets force limits according to the task phase, and issues motion commands. During each command, the impedance controller tracks a reference pose adjusted using real-time force feedback. On a Rokae arm, the system achieved full insertion in 136 of 150 trials across 15 connector types with one shared insertion prompt, averaging 378 seconds per trial.

Coordination at run time should be separated from retained changes to the harness. With the same capability library, DynaHarness reaches 74.0\% under full dynamic execution against 63.9\% under one-step replanning~\citep{dynaharness2026}. That gain comes from coordination: a slow model proposes capabilities and arguments, while a fast layer grounds and monitors commands, refuses unresolved actions, substitutes capabilities, and requests replans. Its failure attribution and regression-checked admission of revised capabilities are claims about retained improvement, assessed by the criteria of Section~\ref{sec:self-improvement}. Classification follows the evaluated configuration in the same way: ALRM runs one ReAct loop either by selecting tools and supplying their arguments or by generating control code directly~\citep{grid43}. SEAR runs the same frontier models live in the control loop and as code writers, and finds that some models evolve best by acting live and others by writing code first~\citep{wang2026sear}.






\input{tables/policy-a2-comparison}

\subsection{Code Synthesis}
\label{sec:policy-a3}

Code synthesis grants the agent authority over executable control logic rather than merely tool arguments, enabling it to construct conditional behavior, intermediate computation, and recovery procedures. Such logic may be synthesized online or before deployment.
What the agent writes, and against which interface, determines what testing can establish. Code as Policies generates Python that processes perception outputs, parameterizes supplied control primitives, performs geometric computation with standard numerical libraries, and recursively defines functions it has not been given~\citep{grid66}; the program is the behavior. ProgPrompt writes plans as programs over imported actions and objects whose assertions check preconditions, such as being close to the fridge before opening it, and trigger recovery actions when they fail~\citep{grid65}. The supplied interface sets much of the difficulty: across 12 models, CaP-X finds that success rises with human-crafted abstractions and falls as they are removed, and that multi-turn interaction, structured execution feedback, and visual differencing narrow the gap over low-level primitives~\citep{capx}.

%

AutoTAMP lies on the other side of the boundary. Its language model writes a signal-temporal-logic specification and a fixed planner computes the motion~\citep{chen2023autotamp}; the output specifies behavior for a fixed executor instead of implementing logic that runs, which keeps AutoTAMP in A1 Code. VoxPoser also leaves motion to a planner, but its generated code queries a vision-language model and composes 3D value maps from which the planner synthesizes trajectories~\citep{voxposer2023}; that computation places VoxPoser in A3. SetItUp induces Python functions that generate spatial-relation graphs for fixed diffusion models~\citep{xu2024set}, while RAPID constructs geometry resolvers and optimization costs for demonstration-derived manipulation programs~\citep{rapid2026}. Generated code can also implement part of a learned skill: APPL's construction agent writes the observation transforms and action coordinates of each skill policy~\citep{jiang2026appl}, so testing one candidate means training a policy (Section~\ref{sec:infra-training}).

When the code is written matters as much as what it contains (Table~\ref{tab:policy-a3}). AGP keeps the agent in the loop throughout execution: it interprets visual evidence, writes programs, issues motion commands, and revises its actions after physical outcomes, succeeding in at least 80\% of trials in seven of eight real-world task configurations~\citep{agp}. ETM moves every model call into development~\citep{hu2026coap}. A coding agent builds a shared library under rules that forbid episode-specific hacks, and the deployed code reaches 70.24\% on 42 bimanual RoboDojo tasks with no model in the loop. As Section~\ref{sec:policy-o3} discusses, oracle state shapes its development but not its deployment. Development also has costs that a final success rate hides. Coding agents in EmbodiedSWE's simulation benchmark solve long-horizon dexterous tasks, but their solutions need substantial iterative interaction and are usually specialized to individual task instances~\citep{embodiedswe2026}. People can be part of the loop as well: in ChatGPT for Robotics, users check generated code by inspection or in simulation and iterate before deploying it~\citep{vemprala2024chatgpt}. Section~\ref{sec:self-improvement} examines how generated programs are revised, validated, and retained for subsequent use.

The observation and action axes interact. Code authority can supply perception that the interface lacks: RAX's generated programs recover geometry from images and depth~\citep{huang2026rax}. It does not remove the dependence on observation quality, since ETM's deployed program succeeds 20 points less often on its own state estimates than on oracle state (Section~\ref{sec:policy-o3}). Action authority also determines when model latency is paid: Spotter keeps its VLM off the critical path during execution (Section~\ref{sec:policy-a2}), whereas ETM makes every model call during development. A result obtained under one observation interface and action authority is therefore weak evidence about another.








\input{tables/policy-a3-comparison}

%% file: tables/policy-o1-comparison.tex
\begin{table}[!htbp]
\centering
\caption{Sensory-input policies: direct sensory interpretation and agent-built perceptual representations.}
\label{tab:policy-o1}
\begingroup
\small
\makeatletter\def\NAT@spacechar{\nobreakspace}\makeatother
\setlength{\tabcolsep}{3pt}
\renewcommand{\arraystretch}{1.10}
\begin{tabularx}{\linewidth}{@{}>{\raggedright\arraybackslash}p{0.42\linewidth}>{\raggedright\arraybackslash}p{20pt}>{\raggedright\arraybackslash}p{0.20\linewidth}>{\raggedright\arraybackslash}X@{}}
\toprule
\rowcolor[HTML]{E4F4FA}
\textbf{Work} & \textbf{Link} & \textbf{Action interface} & \textbf{Evidence formation} \\
\midrule
PIVOT \citep{nasiriany2024pivot} & \href{https://arxiv.org/abs/2402.07872}{\includegraphics[height=8pt]{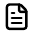}}\hspace{3pt}\makebox[8pt]{--} & A1 Code & Direct \\
TacZero \citep{tanaka2026taczero} & \href{https://arxiv.org/abs/2610.07621}{\includegraphics[height=8pt]{figures/icons/paper.pdf}}\hspace{3pt}\makebox[8pt]{--} & A1 Code & Direct \\
RoboICL \citep{roboicl2026} & \href{https://arxiv.org/abs/2609.34261}{\includegraphics[height=8pt]{figures/icons/paper.pdf}}\hspace{3pt}\href{https://github.com/Mosi-AI/RoboICL}{\includegraphics[height=8pt]{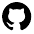}} & A1 Code & Direct \\
SenseNova-RoboRSI \citep{sensenova2026embodied} & \href{https://arxiv.org/abs/2610.22471}{\includegraphics[height=8pt]{figures/icons/paper.pdf}}\hspace{3pt}\href{https://github.com/OpenSenseNova/SenseNova-RoboRSI}{\includegraphics[height=8pt]{figures/icons/github.pdf}} & A1 Code & Direct \\
RAX \citep{huang2026rax} & \href{https://arxiv.org/abs/2610.22496}{\includegraphics[height=8pt]{figures/icons/paper.pdf}}\hspace{3pt}\href{https://github.com/robot-agent-exploration/rax/tree/real/franka}{\includegraphics[height=8pt]{figures/icons/github.pdf}} & A3 & Agent-built \\
\bottomrule
\end{tabularx}
\endgroup
\end{table}

%% file: tables/policy-o2-comparison.tex
\begin{table}[!htbp]
\caption{Policies using supplied perception and queries.}
\label{tab:policy-o2}
\begingroup
\small
\setlength{\tabcolsep}{3pt}
\renewcommand{\arraystretch}{1.10}
\begin{tabularx}{\linewidth}{@{}>{\raggedright\arraybackslash}p{0.30\linewidth}>{\raggedright\arraybackslash}p{0.095\linewidth}>{\raggedright\arraybackslash}p{0.19\linewidth}>{\raggedright\arraybackslash}X@{}}
\toprule
\rowcolor[HTML]{E4F4FA}
\textbf{Work} & \textbf{Link} & \textbf{Action interface} & \textbf{Perception access} \\
\midrule
NLMap–SayCan \citep{chen2022nlmap} & \href{https://arxiv.org/abs/2209.09874}{\includegraphics[height=8pt]{figures/icons/paper.pdf}}\hspace{3pt}\makebox[8pt]{--} & A1 Model & Queries \\
SayPlan \citep{sayplan2023} & \href{https://arxiv.org/abs/2307.06135}{\includegraphics[height=8pt]{figures/icons/paper.pdf}}\hspace{3pt}\makebox[8pt]{--} & A1 Code & Queries \\
ConceptGraphs \citep{gu2023conceptgraphs} & \href{https://arxiv.org/abs/2309.16650}{\includegraphics[height=8pt]{figures/icons/paper.pdf}}\hspace{3pt}\href{https://github.com/concept-graphs/concept-graphs}{\includegraphics[height=8pt]{figures/icons/github.pdf}} & A1 Code & Queries \\
HUME \citep{audit260706501} & \href{https://arxiv.org/abs/2607.06501}{\includegraphics[height=8pt]{figures/icons/paper.pdf}}\hspace{3pt}\href{https://github.com/threefruits/HUME.git}{\includegraphics[height=8pt]{figures/icons/github.pdf}} & A1 Code & Queries \\
MotorMind \citep{motormind2026} & \href{https://arxiv.org/abs/2609.38078}{\includegraphics[height=8pt]{figures/icons/paper.pdf}}\hspace{3pt}\href{https://github.com/Motor-Mind/MotorMind-Code}{\includegraphics[height=8pt]{figures/icons/github.pdf}} & A1 Code & Results \\
RAX \citep{huang2026rax} & \href{https://arxiv.org/abs/2610.22496}{\includegraphics[height=8pt]{figures/icons/paper.pdf}}\hspace{3pt}\href{https://github.com/robot-agent-exploration/rax/tree/real/franka}{\includegraphics[height=8pt]{figures/icons/github.pdf}} & A3 & Queries \\
ZAP\newline\citep{pikus2026pretrainingknowswhatsimulation} & \href{https://arxiv.org/abs/2610.35817}{\includegraphics[height=8pt]{figures/icons/paper.pdf}}\hspace{3pt}\makebox[8pt]{--} & A2 & Queries and geometry \\
CaPtain \citep{fan2026captain} & \href{https://arxiv.org/abs/2610.40799}{\includegraphics[height=8pt]{figures/icons/paper.pdf}}\hspace{3pt}\href{https://github.com/waltstephen/CaPtain}{\includegraphics[height=8pt]{figures/icons/github.pdf}} & A3 & Queries \\
\bottomrule
\end{tabularx}
\endgroup
\end{table}

%% file: tables/policy-o3-comparison.tex
\begin{table}[!htbp]
\caption{Privileged state in development and execution.}
\label{tab:policy-o3}
\begingroup
\small
\setlength{\tabcolsep}{3pt}
\renewcommand{\arraystretch}{1.10}
\begin{tabularx}{\linewidth}{@{}>{\raggedright\arraybackslash}p{0.30\linewidth}>{\raggedright\arraybackslash}p{0.095\linewidth}>{\raggedright\arraybackslash}p{0.19\linewidth}>{\raggedright\arraybackslash}X@{}}
\toprule
\rowcolor[HTML]{E4F4FA}
\textbf{Work} & \textbf{Link} & \textbf{Action interface} & \textbf{Access phase} \\
\midrule
APPL \citep{jiang2026appl} & \href{https://arxiv.org/abs/2609.35690}{\includegraphics[height=8pt]{figures/icons/paper.pdf}}\hspace{3pt}\makebox[8pt]{--} & A1 Model & Execution \\
Navigent \citep{zhou2026navigent} & \href{https://arxiv.org/abs/2611.34249}{\includegraphics[height=8pt]{figures/icons/paper.pdf}}\hspace{3pt}\href{https://github.com/MarsLab-Navigent/Navigent}{\includegraphics[height=8pt]{figures/icons/github.pdf}} & A1 Code & Simulation execution \\
CaP-X \citep{capx} & \href{https://arxiv.org/abs/2603.22435}{\includegraphics[height=8pt]{figures/icons/paper.pdf}}\hspace{3pt}\href{https://github.com/capgym/cap-x}{\includegraphics[height=8pt]{figures/icons/github.pdf}} & A3 & Development \\
ETM \citep{hu2026coap} & \href{https://arxiv.org/abs/2610.12369}{\includegraphics[height=8pt]{figures/icons/paper.pdf}}\hspace{3pt}\makebox[8pt]{--} & A3 & Development \\
\bottomrule
\end{tabularx}
\endgroup
\end{table}

%% file: tables/policy-a1-comparison.tex
\begin{table}[!t]
\centering
\captionsetup{font={rm,normalsize},textfont=rm,labelfont=normalfont,labelsep=colon,justification=centering}
\caption{Behavior selection by fixed executor type.}
\label{tab:policy-a1}
\begingroup
\small
\setlength{\tabcolsep}{3pt}
\renewcommand{\arraystretch}{1.10}
\begin{tabularx}{\linewidth}{@{}>{\raggedright\arraybackslash}p{0.35\linewidth}>{\raggedright\arraybackslash}p{0.095\linewidth}>{\raggedright\arraybackslash}p{0.20\linewidth}>{\raggedright\arraybackslash}X@{}}
\toprule
\rowcolor[HTML]{E2F7F1}
\textbf{Work} & \textbf{Link} & \textbf{Observation interface} & \textbf{Executor type} \\
\midrule
SayCan \citep{saycan2022} & \href{https://arxiv.org/abs/2204.01691}{\includegraphics[height=8pt]{figures/icons/paper.pdf}}\hspace{3pt}\href{https://github.com/google-research/google-research/tree/master/saycan}{\includegraphics[height=8pt]{figures/icons/github.pdf}} & O2 & Model \\
AutoTAMP \citep{chen2023autotamp} & \href{https://arxiv.org/abs/2306.06531}{\includegraphics[height=8pt]{figures/icons/paper.pdf}}\hspace{3pt}\href{https://github.com/yongchao98/AutoTAMP}{\includegraphics[height=8pt]{figures/icons/github.pdf}} & O3 & Code \\
SayPlan \citep{sayplan2023} & \href{https://arxiv.org/abs/2307.06135}{\includegraphics[height=8pt]{figures/icons/paper.pdf}}\hspace{3pt}\makebox[8pt]{--} & O2 & Code \\
PIVOT \citep{nasiriany2024pivot} & \href{https://arxiv.org/abs/2402.07872}{\includegraphics[height=8pt]{figures/icons/paper.pdf}}\hspace{3pt}\makebox[8pt]{--} & O1 & Code \\
ApBot \citep{zhang2025robot} & \href{https://arxiv.org/abs/2505.20424}{\includegraphics[height=8pt]{figures/icons/paper.pdf}}\hspace{3pt}\href{https://github.com/zhangj1an/ApBot}{\includegraphics[height=8pt]{figures/icons/github.pdf}} & O2 & Code \\
TacZero \citep{tanaka2026taczero} & \href{https://arxiv.org/abs/2610.07621}{\includegraphics[height=8pt]{figures/icons/paper.pdf}}\hspace{3pt}\makebox[8pt]{--} & O1 & Code \\
Astra-RoboMME \citep{xref261017350} & \href{https://arxiv.org/abs/2610.17350}{\includegraphics[height=8pt]{figures/icons/paper.pdf}}\hspace{3pt}\href{https://github.com/jackbjed/embodied-astra-robomme}{\includegraphics[height=8pt]{figures/icons/github.pdf}} & O1 & Code \\
Inner Monologue \citep{innermonologue} & \href{https://arxiv.org/abs/2207.05608}{\includegraphics[height=8pt]{figures/icons/paper.pdf}}\hspace{3pt}\makebox[8pt]{--} & O2 & Model \\
Text2Motion \citep{grid67} & \href{https://arxiv.org/abs/2303.12153}{\includegraphics[height=8pt]{figures/icons/paper.pdf}}\hspace{3pt}\makebox[8pt]{--} & O2 & Model \\
Hi Robot \citep{grid22} & \href{https://arxiv.org/abs/2502.19417}{\includegraphics[height=8pt]{figures/icons/paper.pdf}}\hspace{3pt}\makebox[8pt]{--} & O1 & Model \\
VLAs-as-Tools \citep{vlaastools2026} & \href{https://arxiv.org/abs/2605.13119}{\includegraphics[height=8pt]{figures/icons/paper.pdf}}\hspace{3pt}\makebox[8pt]{--} & O1 & Model \\
APPL \citep{jiang2026appl} & \href{https://arxiv.org/abs/2609.35690}{\includegraphics[height=8pt]{figures/icons/paper.pdf}}\hspace{3pt}\makebox[8pt]{--} & O3 & Model \\
\bottomrule
\end{tabularx}
\endgroup
\end{table}

%% file: tables/policy-a2-comparison.tex
\begin{table}[!htbp]
\caption{Tool orchestration through motion commands, skill handoff, intervention, evidence requests, and verification during task execution.}
\label{tab:policy-a2}
\begingroup
\small
\setlength{\tabcolsep}{3pt}
\renewcommand{\arraystretch}{1.10}
\begin{tabularx}{\linewidth}{@{}>{\raggedright\arraybackslash}p{0.32\linewidth}>{\raggedright\arraybackslash}p{0.095\linewidth}>{\raggedright\arraybackslash}p{0.20\linewidth}>{\raggedright\arraybackslash}X@{}}
\toprule
\rowcolor[HTML]{E2F7F1}
\textbf{Work} & \textbf{Link} & \textbf{Observation interface} & \textbf{Coordination} \\
\midrule
RoboHarness (Huang) \citep{grid27} & \href{https://arxiv.org/abs/2607.18060}{\includegraphics[height=8pt]{figures/icons/paper.pdf}}\hspace{3pt}\makebox[8pt]{--} & O2 / O3 & Handoff \\
VoLo \citep{grid23} & \href{https://arxiv.org/abs/2606.07723}{\includegraphics[height=8pt]{figures/icons/paper.pdf}}\hspace{3pt}\href{https://github.com/NVlabs/VoLoAgent}{\includegraphics[height=8pt]{figures/icons/github.pdf}} & O1 & Intervention \\
Spotter \citep{li2026spotter} & \href{https://arxiv.org/abs/2609.36808}{\includegraphics[height=8pt]{figures/icons/paper.pdf}}\hspace{3pt}\href{https://github.com/zqc3117/Spotter}{\includegraphics[height=8pt]{figures/icons/github.pdf}} & O1 & Intervention \\
RegenHarness \citep{si260927612} & \href{https://arxiv.org/abs/2609.27612}{\includegraphics[height=8pt]{figures/icons/paper.pdf}}\hspace{3pt}\makebox[8pt]{--} & O2 & Verified completion \\
ARNA \citep{lange2025arna} & \href{https://arxiv.org/abs/2506.17462}{\includegraphics[height=8pt]{figures/icons/paper.pdf}}\hspace{3pt}\makebox[8pt]{--} & O2 & Perception queries \\
Contact Harness \citep{shan2026contactharness} & \href{https://arxiv.org/abs/2611.29377}{\includegraphics[height=8pt]{figures/icons/paper.pdf}}\hspace{3pt}\makebox[8pt]{--} & O1 & Sensor queries and force limits \\
Recova \citep{recova2026} & \href{https://arxiv.org/abs/2610.01178}{\includegraphics[height=8pt]{figures/icons/paper.pdf}}\hspace{3pt}\makebox[8pt]{--} & O1 & Intervention \\
DynaHarness \citep{dynaharness2026} & \href{https://arxiv.org/abs/2609.40306}{\includegraphics[height=8pt]{figures/icons/paper.pdf}}\hspace{3pt}\makebox[8pt]{--} & O2 / O3 & Command monitoring \\
SEAR \citep{wang2026sear} & \href{https://arxiv.org/abs/2610.18792}{\includegraphics[height=8pt]{figures/icons/paper.pdf}}\hspace{3pt}\href{https://github.com/Self-Evolving-Agentic-Robotics/SEAR}{\includegraphics[height=8pt]{figures/icons/github.pdf}} & O2 & Live tool selection \\
ZAP\newline\citep{pikus2026pretrainingknowswhatsimulation} & \href{https://arxiv.org/abs/2610.35817}{\includegraphics[height=8pt]{figures/icons/paper.pdf}}\hspace{3pt}\makebox[8pt]{--} & O2 & Motion-tool sequencing \\
\bottomrule
\end{tabularx}
\endgroup
\end{table}

%% file: tables/policy-a3-comparison.tex
\begin{table}[!htbp]
\caption{Code synthesis during development and execution.}
\label{tab:policy-a3}
\begingroup
\small
\setlength{\tabcolsep}{3pt}
\renewcommand{\arraystretch}{1.10}
\begin{tabularx}{\linewidth}{@{}>{\raggedright\arraybackslash}p{0.32\linewidth}>{\raggedright\arraybackslash}p{0.095\linewidth}>{\raggedright\arraybackslash}p{0.20\linewidth}>{\raggedright\arraybackslash}X@{}}
\toprule
\rowcolor[HTML]{E2F7F1}
\textbf{Work} & \textbf{Link} & \textbf{Observation interface} & \textbf{Generation phase} \\
\midrule
Code as Policies \citep{grid66} & \href{https://arxiv.org/abs/2209.07753}{\includegraphics[height=8pt]{figures/icons/paper.pdf}}\hspace{3pt}\href{https://github.com/google-research/google-research/tree/master/code_as_policies}{\includegraphics[height=8pt]{figures/icons/github.pdf}} & O2 / O3 & Execution \\
ProgPrompt \citep{grid65} & \href{https://arxiv.org/abs/2209.11302}{\includegraphics[height=8pt]{figures/icons/paper.pdf}}\hspace{3pt}\href{https://github.com/NVlabs/progprompt-vh}{\includegraphics[height=8pt]{figures/icons/github.pdf}} & O2 / O3 & Execution \\
VoxPoser \citep{voxposer2023} & \href{https://arxiv.org/abs/2307.05973}{\includegraphics[height=8pt]{figures/icons/paper.pdf}}\hspace{3pt}\href{https://github.com/huangwl18/VoxPoser}{\includegraphics[height=8pt]{figures/icons/github.pdf}} & O2 / O3 & Execution \\
ChatGPT for Robotics \citep{vemprala2024chatgpt} & \href{https://doi.org/10.1109/ACCESS.2024.3387941}{\includegraphics[height=8pt]{figures/icons/paper.pdf}}\hspace{3pt}\makebox[8pt]{--} & O2 / O3 & Development \\
SetItUp \citep{xu2024set} & \href{https://doi.org/10.1177/02783649251378198}{\includegraphics[height=8pt]{figures/icons/paper.pdf}} & O2 & Development \\
AGP \citep{agp} & \href{https://arxiv.org/abs/2609.12541}{\includegraphics[height=8pt]{figures/icons/paper.pdf}}\hspace{3pt}\href{https://github.com/agent-as-policy-2026/agent-as-policy}{\includegraphics[height=8pt]{figures/icons/github.pdf}} & O1 / O2 & Execution \\
EmbodiedSWE \citep{embodiedswe2026} & \href{https://arxiv.org/abs/2609.27308}{\includegraphics[height=8pt]{figures/icons/paper.pdf}}\hspace{3pt}\href{https://github.com/EmbodiedSWE/EmbodiedSWE}{\includegraphics[height=8pt]{figures/icons/github.pdf}} & O3 & Development \\
RAPID \citep{rapid2026} & \href{https://arxiv.org/abs/2609.30249}{\includegraphics[height=8pt]{figures/icons/paper.pdf}} & O2 / O3 & Development \\
ETM \citep{hu2026coap} & \href{https://arxiv.org/abs/2610.12369}{\includegraphics[height=8pt]{figures/icons/paper.pdf}}\hspace{3pt}\makebox[8pt]{--} & O3 & Development \\
CaPtain \citep{fan2026captain} & \href{https://arxiv.org/abs/2610.40799}{\includegraphics[height=8pt]{figures/icons/paper.pdf}}\hspace{3pt}\href{https://github.com/waltstephen/CaPtain}{\includegraphics[height=8pt]{figures/icons/github.pdf}} & O2 & Execution \\
\bottomrule
\end{tabularx}
\endgroup
\end{table}

%% file: sections/04_agent_as_infrastructure.tex
\section{Agent as Infrastructure}
\label{sec:infrastructure}

\begin{figure}[!t]
\centering
\includegraphics[width=\columnwidth]{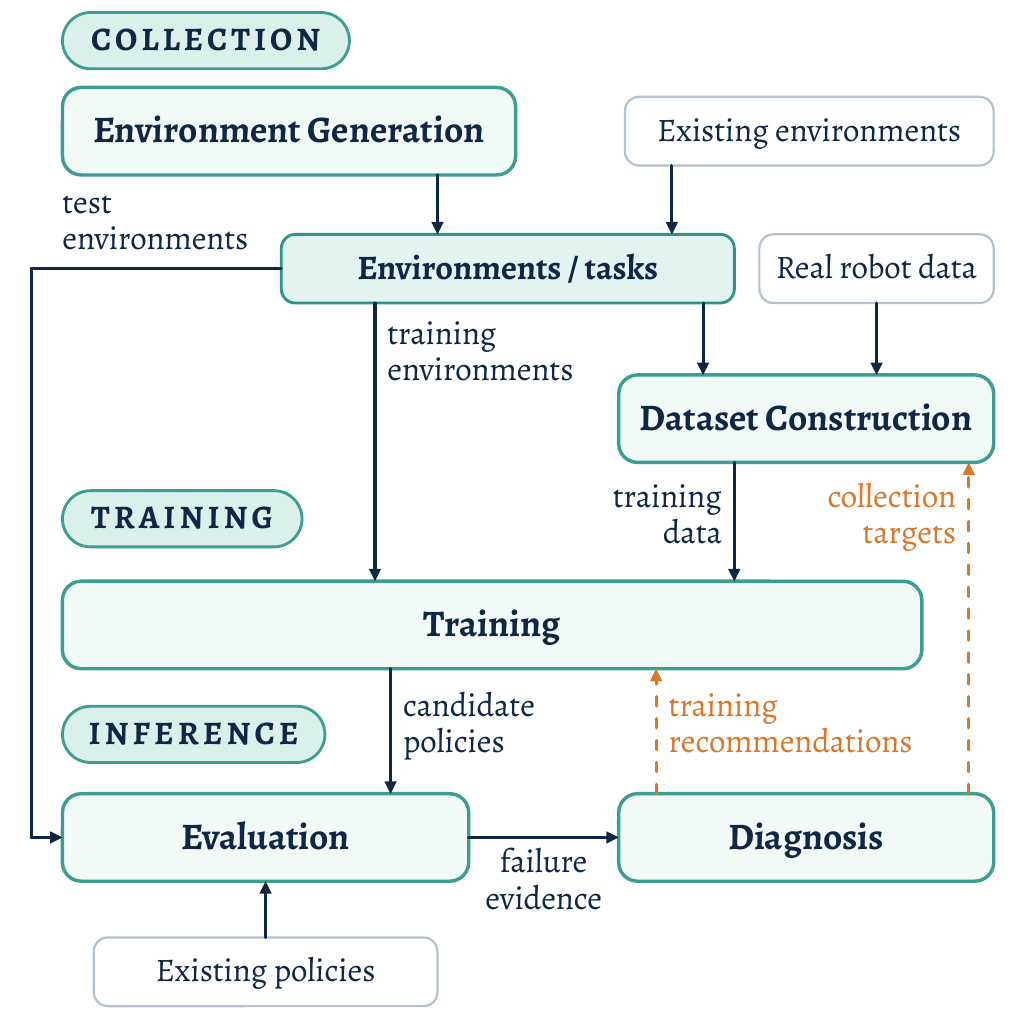}
\caption{Agent as Infrastructure across Collection, Training and Inference. Teal boxes are the five subsections and the shared environment resource; white boxes are external inputs. Solid arrows pass environments, data, policies or evidence; dashed arrows carry diagnostic suggestions. Systems can combine different branches of this workflow.}
\label{fig:infrastructure-overview}
\end{figure}

Figure~\ref{fig:infrastructure-overview} relates environment generation, dataset construction, training, evaluation and diagnosis across the collection--training--inference workflow. The inference stage covers robot operation and assessment in simulation or on physical robots. Figures~\ref{fig:infra-environment}--\ref{fig:infra-diagnosis} show the agent decisions within each subsection.

\subsection{Environment Generation}\label{sec:infra-environment}

\begin{figure}[!t]
\centering
\includegraphics[width=\columnwidth]{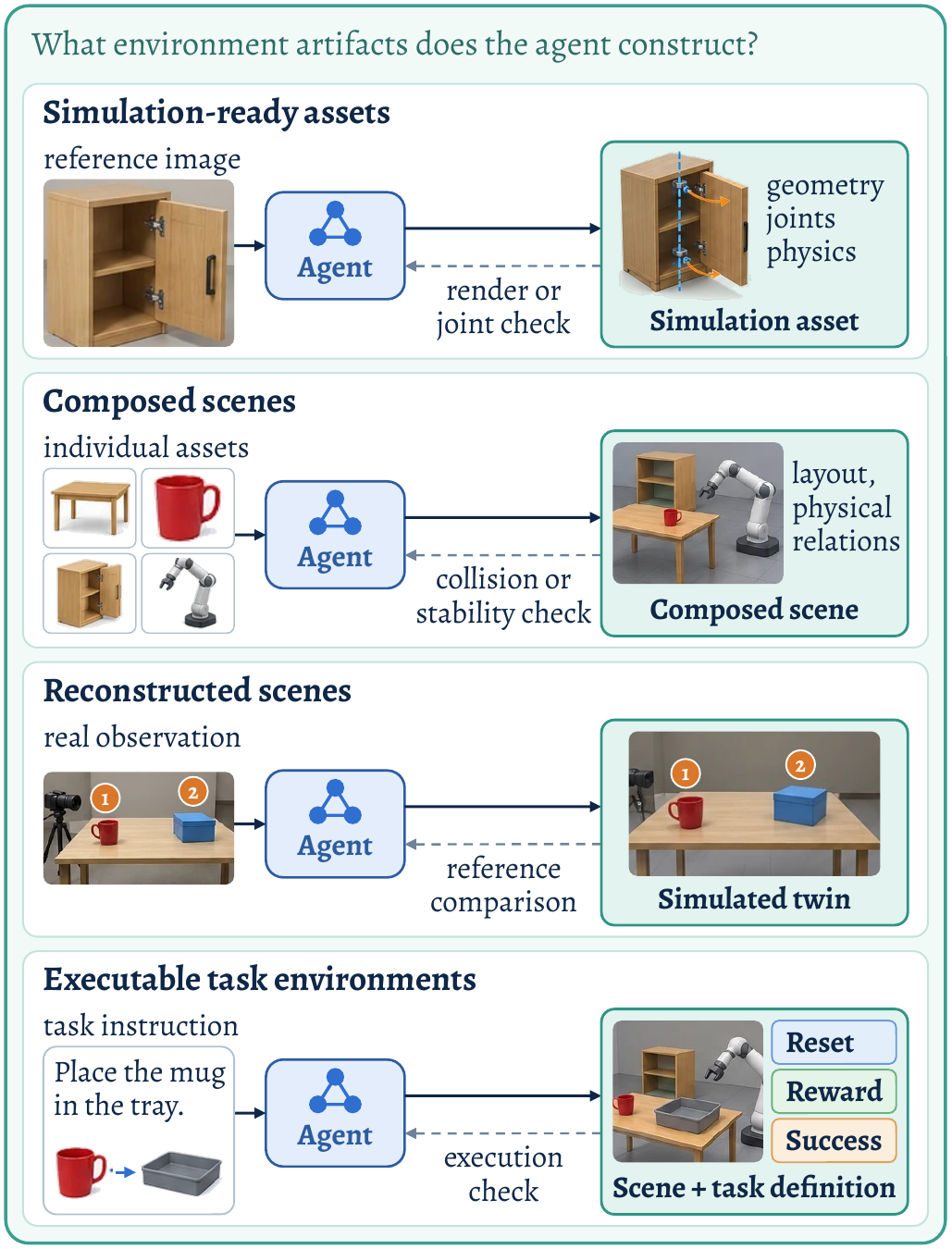}
\caption{Environment Generation. Each row pairs an input with the artifact the agent constructs; dashed arrows indicate optional feedback to the agent. Blue boxes denote agents and their decisions, and teal boxes outputs.}
\label{fig:infra-environment}
\end{figure}

Environment Generation supplies the Collection stage with simulator-ready artifacts for robot training, evaluation, and data production. We organize works by the artifacts they construct: simulation-ready assets, composed scenes, reconstructions of observed scenes, and executable task environments (Figure~\ref{fig:infra-environment} and Table~\ref{tab:infra-environment}). Construction is single-pass when at most fixed filters follow one generation, human-assisted when people supply content or approval, and feedback-driven when simulator, critic, or execution evidence returns to the agent. Within each group, works run broadly from no check or visual comparison toward acceptance through robot execution, followed by boundary cases. Revisions retained across runs are discussed under Environments and Datasets in Section~\ref{sec:self-improvement}.

\subsubsection{Simulation-Ready Asset Construction}


Asset builders differ in which evidence returns to the agent. In Articulate-Anything~\citep{le2024articulateanything}, vision-language actors rewrite URDF-compiling code until a critic, comparing simulator renders with the input image or video, rates the result above a threshold, although the critic tends to overestimate success. SimFoundry~\citep{simfoundry2026} adapts this actor-critic loop for joint parameters. Articraft~\citep{articraft2026} instead revises its asset program from execution errors, harness checks, its own tests, and measurement probes, with only qualitative robot interaction in Isaac Sim~\citep{isaacsim}. 3DCodeVerse~\citep{gao2026_3dcodeverse} revises link meshes and a URDF from measured joint-pose collisions and a VLM judge's render critique, but its only robot test is a preliminary contour check that a static last frame also passes. DiagGen~\citep{diaggen2026} uses simulated interactions for diagnosis: its diagnostic agent designs grasp probes for a deformable Genesis asset~\citep{genesis2024} and traces failures to a pipeline stage from which construction may rerun. None of these checks tests a downstream robot task.

\subsubsection{Scene Composition and Physical Validation}

\begingroup Scene composers arrange assets from language, task, or image specifications, with physical validity serving as a key acceptance criterion. Single-pass composers let fixed stages repair or filter a foundation model's plan. EmbodiedGen V2~\citep{wang2026embodiedgenv2} realizes an instruction-conditioned scene graph through quality gates and reachability-constrained placement, V-CAGE~\citep{vcage2026} optimizes collisions out of an inpainted layout, and V-Dreamer~\citep{he2026vdreamer} settles its layout in Genesis. Validation of human-assisted composition is not clearly documented: Genie Sim~\citep{geniesim2026} describes but does not measure conversational refinement or report a scene check, and EmbodiedClaw~\citep{embodiedclaw2026} builds environments from user reference images under a step verifier with unstated criteria.\parfillskip=0pt plus 0.1\linewidth\par\endgroup

Feedback-driven composers differ in where physics enters the loop. Solver or settling failures return to the agent in PhyScensis~\citep{physcensis2026} and RoboLab~\citep{yang2026robolabhighfidelitysimulationbenchmark}, although RoboLab's policy benchmark is manually authored. In SAGE~\citep{sage2026}, an Isaac Sim critic checks each edit and reports failures to the agent. The agents of SceneSmith~\citep{scenesmith2026} may query Drake collision checks~\citep{drake2019} but leave residual penetrations and fallen objects to fixed post-processing. For composers intended to support learning or robot interaction, the reported checks vary in how directly they assess the intended use. Adaptive Environment Generation~\citep{si260206366} edits layouts from a navigation policy's trajectory but leaves training on them to future work. GIF~\citep{xu2026gif} judges passive settling rather than robot execution, GenPHRI~\citep{wang2026genphri} gates only its robot stages in simulation, and Awomo-SimDataEngine~\citep{physicalrsiteam2026awomo} keeps its world gate separate from the episode-success predicate.

\subsubsection{Real-to-Simulation Reconstruction and Refinement}

Reconstruction aims to recover a particular observed scene, but works compare against that reference in different ways. RAPID combines VLM-estimated object properties with supplied reconstruction tools to build demonstration-derived verification scenes and writes filters for feasible variants~\citep{rapid2026}. Agentic RSR~\citep{li2026agenticrsr} compares its renders with frozen real views and receives scalar depth and color errors as its only numeric feedback. Recova~\citep{recova2026} fits its twin so that a replayed trajectory aligns with real camera views, without a reported fidelity metric. HomeBody's project page~\citep{huh2026homebody} documents no agent decision beyond room dimensions and no check beyond visual comparison. DexAgent~\citep{wang2026dexagent} accepts reconstructions on property-specific verifiers that it selects or writes. Other twins are accepted or revised according to whether a task or recorded behavior succeeds in simulation. As its prompts prescribe, the reconstruction agent of ARSTAG~\citep{arstag2026} rescaled and reoriented an object after collection failures in one reported case. Agentic Real2Sim~\citep{agenticreal2sim2026} uses a parameter sweep to select an object-position offset that enables the fixed recorded trajectory to grasp the object, and Real2Gym~\citep{real2gym2026} accepts a seed environment once an \mbox{adjusted replay completes the task.}

In Video2World~\citep{tang2026video2world} and CoDimRecon~\citep{xie2026codimrecon}, the agent also specifies the robot motion that exercises its reconstruction, though CoDimRecon exercises only deformables, in reduced trial scenes, and reports no downstream robot use. Real-world outcomes drive construction in RoboSimStudio~\citep{chen2026robosimstudio}, SimEX~\citep{simex2026}, and F4R~\citep{f4r2026}, whose VLM chooses among configured labels, but acceptance rests on qualitative judgment or failure reproduction rather than measured fidelity. SimFoundry~\citep{simfoundry2026} tests its VLM-supplied physical parameters only through real-world evaluation, where operator-tuned simulations preserved policy rankings. Elsewhere, rollout failures prompt revisions to programs or collected experience, but not to the scene, as in RAPID~\citep{rapid2026} (whose coding agent only filters sampled scene variants), SUN~\citep{wang2026sunagenticrobotpolicy}, and EmbodiRSI~\citep{embodirsi2026}.

\subsubsection{Executable Task Environments}

Task generators write the code or configuration that defines a task, typically its success conditions and often its initial state, rewards, and scene. Single-pass generators provide limited validation of the generated task definition. Gen2Sim~\citep{gen2sim2023} applies no explicit check, and RoboGen~\citep{wang2023robogen} applies only a VLM asset filter and collision push-apart, yet later inspection found incorrect scenes in 13 of its 155 tasks. SEES~\citep{si260932698} has an LLM write only success predicates, which its authors validate offline on Atomic-Seen rollouts rather than on the subtasks it trains. When the same model writes a task and its check, a pass alone does not establish that the task matches the instruction, as in GenSim's execution check~\citep{gensim2023} and in AnyTask~\citep{anytask2025}. Feedback may then revise the artifact toward the agent's own check, as the glm-skills-embodied repository~\citep{hxwxss2026glmskills} reports, or revise the check itself. GRS~\citep{grs2024} can repair its generated unit tests, and RoboPlayground~\citep{roboplayground2026} can \mbox{rewrite parts of its success predicate.}

People enter at specification or acceptance in HARBOR~\citep{li2026harbor}, GenSim2~\citep{hua2024gensim2}, RPG~\citep{rpg2026}, and RoboWits~\citep{robowits2026}, although RoboWits admits mutated tasks without executing a solution. Execution evidence drives revision in SUN~\citep{wang2026sunagenticrobotpolicy}, EmbodiedSmith~\citep{qin2026embodiedsmithscalingembodieddata}, and FATE~\citep{wei2026fate}, which repair the task program, the skill plan and robot placement, or scene geometry and solver settings, respectively. Their acceptance stays conditional on the executor and its success test. In Scene2Demo~\citep{liu2026scene2demo}, HumanoidGen~\citep{humanoidgen2025}, Video2Policy~\citep{ye2025video2policy}, Eurekaverse~\citep{liang2024eurekaverse}, and OMNI-EPIC~\citep{si240515568}, whose learner is a six-action R2D2 robot and whose construction repairs only code errors, later loops serve demonstration generation or policy training rather than construction repair. The synthetic-task generator of EmbodiedSWE~\citep{embodiedswe2026} admits environments on checks around a teleport solution rather than a robot-executed one. SimWorld Studio~\citep{self260509423} and Embodied-BenchForge~\citep{si260913082} build tasks for agents that act through discrete \mbox{actions rather than robot controllers.}

\begingroup\emergencystretch=1em Across the four groups, passing a construction check seldom shows that a robot can complete the intended task. The coding-agent systems benchmarked in Video2World~\citep{tang2026video2world} reach family-averaged build rates of 87.6--93.1\% but task-success rates of only 1.5--25.5\%. In FATE~\citep{wei2026fate}, 93.8\% of statically generated tasks pass scene-validity checks, yet only 65.4\% prove feasible, against 92.1\% with execution-based repair. Dataset Construction then decides \mbox{which of these executions become data.}\par\endgroup

\input{tables/infra-environment-comparison}

\subsection{Dataset Construction}
\label{sec:infra-dataset}

\begin{figure}[!t]
\centering
\includegraphics[width=\columnwidth]{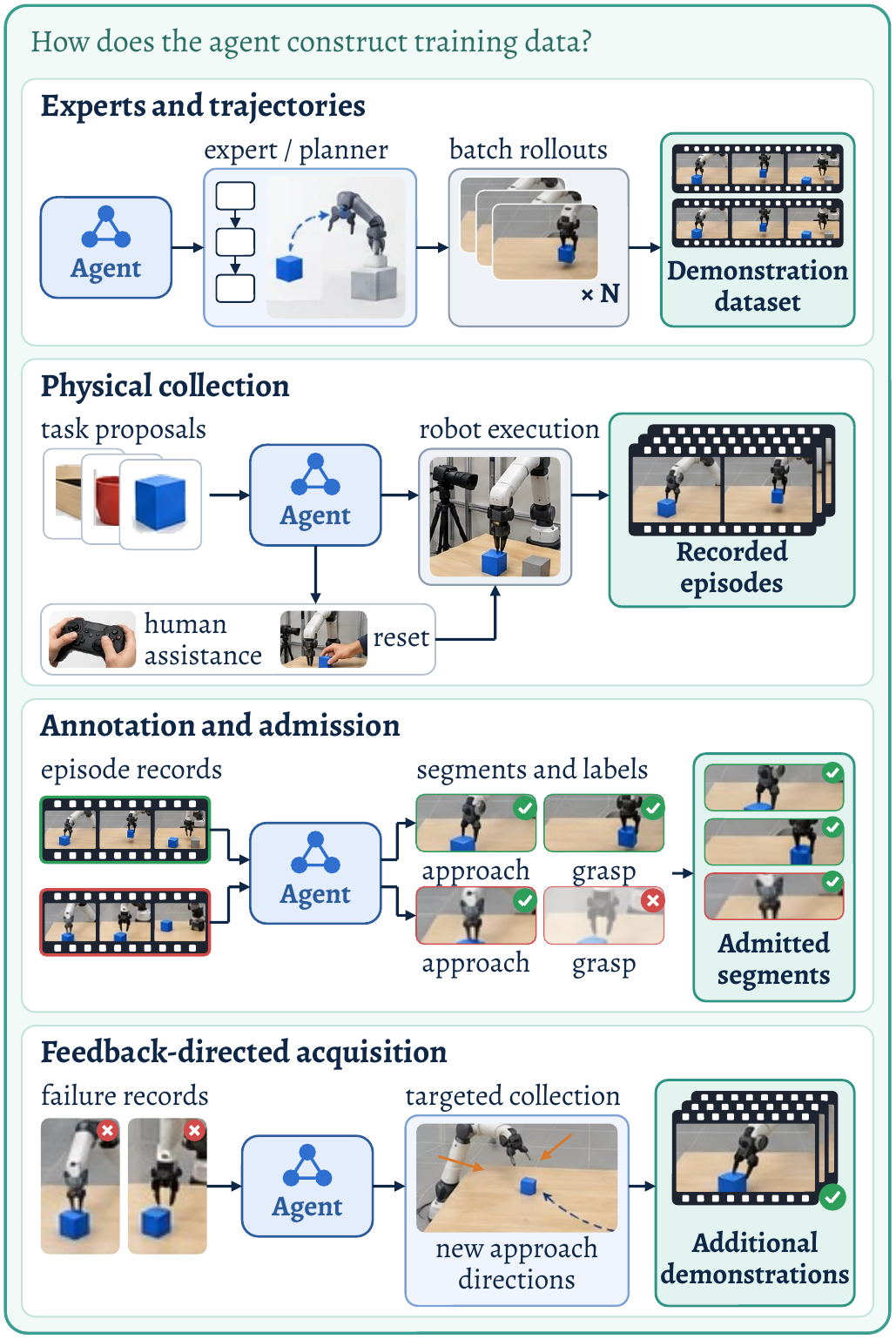}
\caption{Dataset Construction. Agents specify experts, organize physical collection, annotate and admit segments, and direct targeted training data acquisition.}
\label{fig:infra-dataset}
\end{figure}

Dataset Construction turns executable environments and robot interactions into supervision for learning. We group works by the decisions that determine the resulting data: constructing experts and trajectories, organizing physical collection, annotating and admitting examples, and directing further acquisition from feedback (Figure~\ref{fig:infra-dataset} and Table~\ref{tab:infra-dataset}). Within each group, we compare what the agent specifies, what executes or checks that specification, and which outputs become training data. A valid environment is only a starting point: demonstrations must also achieve the intended behavior, carry usable supervision, and cover states relevant to the learner. Retained feedback-driven dataset revisions are discussed under \mbox{Environments and Datasets in Section~\ref{sec:self-improvement}.}

\subsubsection{Expert Programs and Trajectory Synthesis}

\begingroup Synthetic collection begins by specifying an expert whose executions supply action labels. GenSim~\citep{gensim2023} generates task code from which demonstrations can be produced offline, but its execution checks do not capture every mismatch between the instruction and the behavior. RoboGen~\citep{wang2023robogen} instead selects motion planning, trajectory optimization, or reinforcement learning for each subtask and generates the required supervision. GenSim2~\citep{hua2024gensim2} grounds solver constraints in rendered scenes and annotated keypoints, with optional rejection sampling and human review. Video2Policy~\citep{ye2025video2policy} uses video-derived task specifications and feedback-refined rewards to train experts, then retains successful rollouts for behavior cloning. Scaling Up and Distilling Down~\citep{grid64} combines language-guided task decomposition, motion planning, and generated success checks in simulation, retaining verified demonstrations for policy distillation. In these workflows, the model specifies the problem or solver; the numerical planner or learner produces the trajectory.\parfillskip=0pt plus 0.1\linewidth\par\endgroup

\begingroup Execution feedback can also change the expert itself. RoboTwin 2.0~\citep{robotwin2agent2025} combines execution logs with visual failure localization to repair task programs across repeated trials. HumanoidGen~\citep{humanoidgen2025} searches constraint-chain programs through simulation, preserving executable prefixes and exploring alternatives after failure. AnyTask's ViPR~\citep{anytask2025} revises plans from parallel rollout videos and dense logs, whereas Scene2Demo~\citep{liu2026scene2demo} uses visual critiques to revise action flows and inter-subtask transitions. These loops differ from acceptance-only feedback: V-CAGE~\citep{vcage2026} instantiates manipulation templates and retains a complete trajectory only when a VLM accepts every subtask. Visual acceptance filters semantic failures according to the judging model; physical feasibility is assessed through execution checks.\parfillskip=0pt plus 0.1\linewidth\par\endgroup

\begingroup Video-conditioned generators differ in what they transfer from the visual reference. V-Dreamer~\citep{he2026vdreamer} generates manipulation videos from a stabilized scene and lifts tracked motion into three-dimensional trajectories. DexAgent~\citep{wang2026dexagent} uses a human video to guide grasp and motion synthesis, allowing the agent to select or revise tools when pose-specific interaction tests fail. It then adapts the accepted trajectory to randomized initial states and rechecks each episode. Video plausibility supplies a motion prior; embodiment-specific feasibility and task completion require additional checks.\parfillskip=0pt plus 0.1\linewidth\par\endgroup

Scaling an accepted expert requires preserving the task while varying its instances. ARSTAG~\citep{arstag2026} lets its collection agent configure task-consistent randomization and perturbation recovery around state-based experts. GenPHRI~\citep{wang2026genphri} collects demonstrations across synthesized human bodies and applies fixed trajectory perturbations for policy learning. GIF~\citep{xu2026gif} turns an agent-generated object relation into randomized pick-and-place demonstrations, rejecting failed plans and unstable final relations. Awomo-SimDataEngine~\citep{physicalrsiteam2026awomo} separates world validity from episode success and exports only successful episodes with aligned observations and actions. Thus, agent-generated targets can support data production even when augmentation and rejection are fixed procedures.

The surrounding collection workflow is another agent responsibility. EmbodiedClaw~\citep{embodiedclaw2026} plans trajectory-collection and data-conversion operations through reusable skills. Genie Sim~\citep{geniesim2026} combines LLM-assisted task assembly with motion-planned collection, while retaining teleoperation for difficult tasks. The glm-skills-embodied \mbox{repository~\citep{hxwxss2026glmskills}} documents agent-written experts and replay-compatible HDF5 exports. Workflow completion and format validity establish usable artifacts, but do not alone show \mbox{that training on them improves a policy.}

\subsubsection{Physical Collection and Assisted Data Workflows}

\begingroup Agentic systems extend data construction to physical robot interaction, where task selection, trajectory acquisition, outcome verification, and environment recovery introduce additional decisions and human costs. SOAR~\citep{soar2024} uses VLMs to propose manipulation goals and evaluate autonomous real-world rollouts, while hindsight goal relabeling enables learning from trajectories that do not achieve their original goals. AutoRT~\citep{autort2024} combines VLM-based scene understanding with LLM task generation and safety filtering to organize large-scale physical collection through autonomous execution and human teleoperation.\parfillskip=0pt plus 0.1\linewidth\par\endgroup

Beyond selecting tasks and verifying outcomes, physical collection must restore usable starting states and coordinate human assistance. RoboClaw~\citep{roboclaw} and HERO~\citep{hero2026} support autonomous execution and recovery, with HERO combining hierarchical capability selection, success verification, and inverse-task resets. RoboGene~\citep{robogene2026} generates diverse and physically feasible tasks for human demonstration collection, whereas REVOLVE~\citep{revolvecollection2026} assists task-specific collection-code generation and integrates human validation, automated resets, and corrective execution. These methods distribute collection responsibilities between agents and people. Human interventions, outcome reliability, and downstream data utility provide \mbox{common dimensions for comparison.}

\begingroup Direct agent execution provides another source of physical demonstrations. MOKA~\citep{liu2024moka} records successful visually grounded executions for policy distillation, with people checking predictions and final success. Manipulate-Anything~\citep{xref240618915} verifies subtasks and replans from failed states, so retained demonstrations can include recovery, although its physical collection still uses manual resets after failure. Tether~\citep{liang2026tether} instead warps a small set of human demonstrations and uses a VLM to choose feasible next tasks and judge outcomes within a task set designed for natural resets. Waddle~\citep{waddle2026introducing} demonstrates requested repeated pick-and-place collection, but its announcement leaves dataset filtering and downstream learning unquantified. These sources differ in both who supplies the initial behavior and who certifies the recorded result.\parfillskip=0pt plus 0.1\linewidth\par\endgroup

Failure changes collection from replaying nominal behavior to obtaining corrective supervision. Zero2Skill~\citep{wang2026zero2skill} requests human assistance when reliability is insufficient and uses language corrections to refine collection and verification criteria; its evaluated collect--train--deploy cycle does not establish repeated autonomous iteration. Recova~\citep{recova2026} generates task and recovery data in a digital twin, then adds verified \mbox{physical recovery} and human demonstrations during deployment. Phoenix~\citep{xia2025phoenix} reuses motion-corrected interaction trajectories to improve motion prediction, but its correction model is initially trained with human intervention and annotation. Reduced intervention during collection therefore needs to be assessed alongside \mbox{the supervision that made collection possible.}

\begingroup Intervention can supply training experience while adapting collection to the learner. UniIntervene~\citep{deng2026uniintervene} uses a vision-language backbone to predict action consequences and value risk, triggering memory-guided recovery when progress deteriorates. Recovery transitions enter replay for online reinforcement learning, although the intervention modules are learned offline and remain fixed during target-task training. UniIntervene++~\citep{lin2026uniintervenepp} provides a learned-intervention comparison: a Double-DQN scheduler learns online to allocate control among the RL policy, trajectory correction, and a task-structured CodePolicy. Unassisted probes expose the policy's current competence, while assisted transitions enter the demonstration buffer used in mixed off-policy updates. Across five physical tasks, it reports 89.67\% success with 0.77\% human-controlled steps. This links data acquisition to changing policy competence; reduced human control still includes automated assistance, whose share must be distinguished from autonomous execution.\parfillskip=0pt plus 0.1\linewidth\par\endgroup

\subsubsection{Annotation and Dataset Admission}

\begingroup Annotations can specify how to interact with an asset or describe what happened in a trajectory. AnnotateAnything~\citep{lu2026annotateanything} grounds semantic descriptions, parts, and affordances in three-dimensional interaction anchors, then uses physics-based generation and validation to obtain executable action labels. AnyTask~\citep{anytask2025} lets generated programs choose when and what simulator information to log as dense language annotations. EmbodiedSWE-Gen~\citep{embodiedswe2026} similarly diversifies verified solutions through agent-authored variants, while much of its progress and situation supervision is derived from programs and simulator states by fixed processing. Such labels inherit the specification's semantics; simulator access does not independently validate its interpretation of the instruction.\parfillskip=0pt plus 0.1\linewidth\par\endgroup

Admission need not be decided at whole-episode granularity. ARS~\citep{hu2026ars} inspects selected video windows, verifies proposed task events in a separate reasoning context, and estimates progress curves that select action chunks for behavior cloning. Robo-COP~\citep{zhang2026robocop} aligns scripted and VLA actions in a common stream, then uses text and video judges to segment, label, and admit skill demonstrations. A successful segment can survive a failed episode, while successful episodes can contain rejected segments. The resulting supervision preserves useful partial behavior, but its reliability depends on whether the judges inspect the evidence needed \mbox{to distinguish progress from apparent success.}

\subsubsection{Feedback-Directed Acquisition and Data Utility}

Beyond correcting individual examples, feedback can change which examples are sought next. EmbodiRSI~\citep{embodirsi2026} combines recovery from learner-reached states with targeted recollection in configurations implicated by rollout failures. SEAR's auto-research configuration~\citep{wang2026sear} lets the agent curate code-policy demonstrations, choose tasks needing more coverage, and add recovery or perturbed-start rollouts before VLA fine-tuning. RoboLineage~\citep{luo2026robolineage} separates raw capture, post-rollout interpretation, training admission, and failure-pool routing; dataset-health analysis then identifies coverage gaps and subsequent collection targets. These mechanisms make the retained distribution responsive to execution evidence. AgentChord's recovery-augmented execution graph~\citep{xu2026agentchord} illustrates the boundary: generating corrective behavior alone does not establish how that behavior \mbox{becomes an admitted training example.}

Across the four groups, collection yield, label reliability, and downstream learning answer different questions. Tether~\citep{liang2026tether} checks its success judgments against manual annotations, reporting 98.4\% precision and 89.6\% recall over 1,946 play attempts. EmbodiRSI~\citep{embodirsi2026} compares adaptive and static acquisition at 400 retained simulated trajectories per subtask in two scenes, obtaining 79.9\% versus 59.3\% scene-balanced simulation success; this tests the combined acquisition strategy rather than isolating either feedback path. Zero2Skill~\citep{wang2026zero2skill} compares policies trained on equal numbers of valid demonstrations and removes collection-time assistance at evaluation. Such controls connect the generated dataset to autonomous policy performance. Training then determines how that supervision is converted into a learned policy.

\input{tables/infra-dataset-comparison}

\subsection{Training}\label{sec:infra-training}

\begin{figure}[!t]
\centering
\includegraphics[width=\columnwidth]{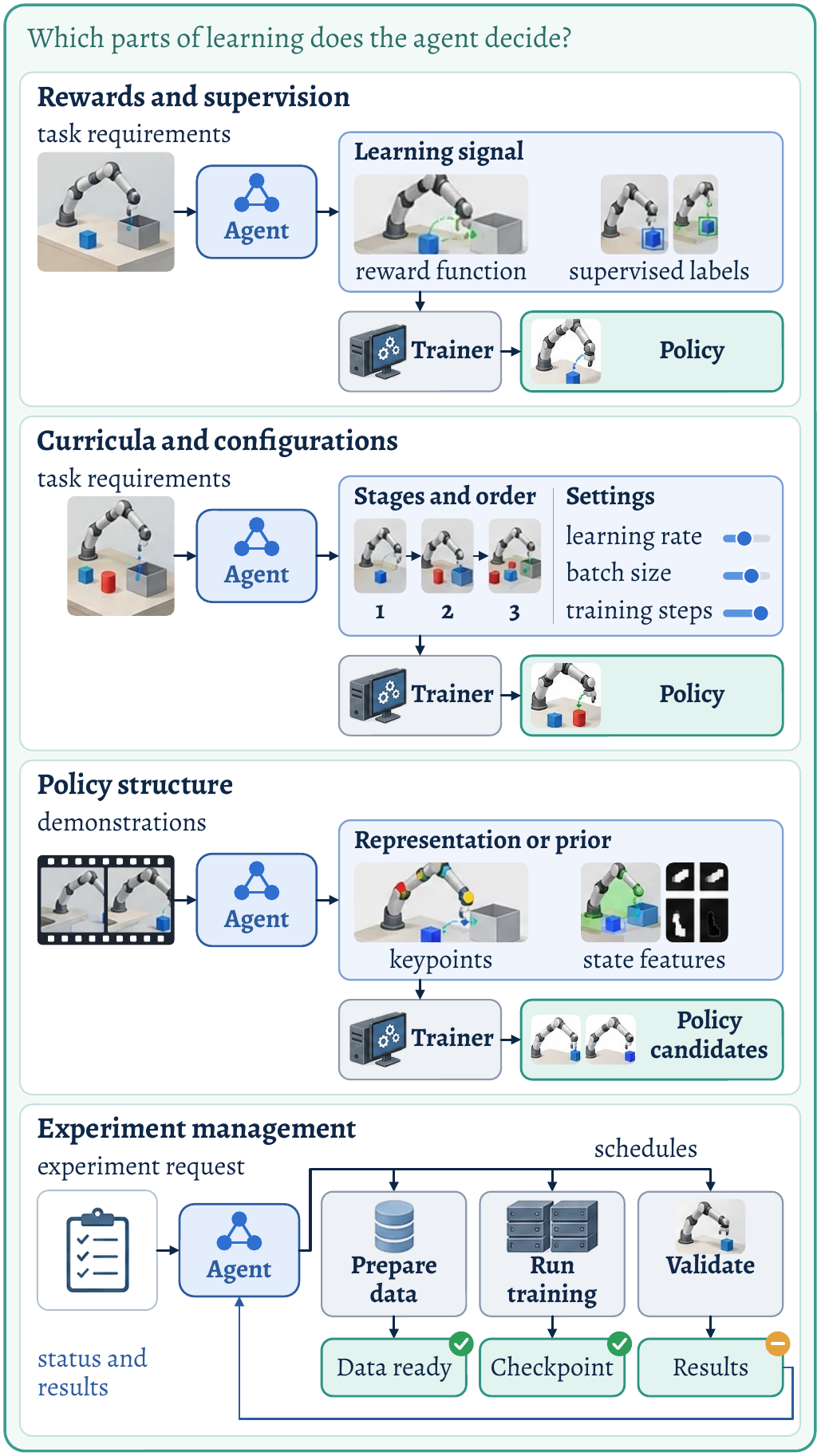}
\caption{Training. Agents decide learning signals, curricula, settings, policy structures, and experiment schedules; trainers and jobs execute these decisions.}
\label{fig:infra-training}
\end{figure}

\begingroup Training infrastructure turns task requirements into learning procedures. Agents specify supervision, construct curricula, select training configurations, design policy structure, and coordinate experiments (Figure~\ref{fig:infra-training} and Table~\ref{tab:infra-training}). We organize these responsibilities by the decisions delegated to the agent and the feedback used to make them. Training tools and optimizers execute the resulting specifications. This subsection follows the learning workflow; Section~\ref{sec:self-improvement} examines how its outputs become retained improvements to rewards, environments, and learned components.\parfillskip=0pt plus 0.5\linewidth\par\endgroup

\subsubsection{Reward Specification and Training Supervision}

\begingroup Agents connect task descriptions to learning signals through several forms of supervision. In its multi-task learning configuration, LARG$^2$ generates task-dependent reward code and repairs execution errors before training~\citep{perez2023larg2}. Text2Reward incorporates human corrections after inspection of learned behavior~\citep{text2reward2023}, while Eureka uses training statistics to revise reward programs and a supplied task-fitness function to select candidates~\citep{ma2023eureka}. The feedback thus progresses from program validity to observed behavior and measured task performance. Code as Reward instead constructs visual subtask checks, tests them on expert and random trajectories, and leaves their assembly into a reward class to the user~\citep{venuto2024codeasreward}. In CaPtain, coding agents provide demonstrations by choosing what to inspect and which robot commands to execute next, producing trajectories for supervised fine-tuning~\citep{fan2026captain}.\parfillskip=0pt plus 0.5\linewidth\par\endgroup

\begingroup Structured outputs separate semantic choices from numerical fitting. ELEMENTAL prompts VLMs to design reward features from demonstrations and environment specifications, then fits feature weights with inverse reinforcement learning; its manipulation setup uses expert-selected keyframes~\citep{elemental2025}. Self-Alignment similarly delegates feature and parameterization proposals to an LLM before ranking-based numerical adjustment~\citep{zeng2024selfalignment}. ReCoVLA uses VLM failure descriptors to select semantic reward components, compiles them into predefined terms, and trains recovery residuals around a frozen VLA~\citep{hu2026recovla}. These interfaces let agents express task-relevant objectives while established learning procedures determine their numerical realization.\parfillskip=0pt plus 0.5\linewidth\par\endgroup

\begingroup Behavioral diagnosis can guide the next training intervention. AHA supplies visual failure explanations for reward refinement~\citep{aha2024}; RDA combines subtask decomposition, visual rollout assessment, and reward revision~\citep{lee2026rda}. Agent-guided Policy Search (AGPS) intervenes during exploration: a detector triggers an agent to supply corrective waypoints and spatial bounds, and RL updates the policy using the resulting experience alongside human demonstrations and manual resets~\citep{chen2026agps}. The shared mechanism is a path from observed failure to a changed learning signal or exploration target. Retained reward revisions are discussed in Section~\ref{sec:si-rewards}.\parfillskip=0pt plus 0.5\linewidth\par\endgroup

\subsubsection{Curricula and Training Configuration}

\begingroup Curricula determine which problems a policy learns and how experience carries between stages. CurricuLLM proposes an ordered subtask sequence, generates stage-specific reward and goal code, and selects policies using rollout statistics; each stage initializes from the preceding selected policy~\citep{curricullm2024}. CRAFT uses visual judgments to advance or refine a stage within a prescribed refinement budget~\citep{choi2025craft}. Eurekaverse revises terrain-generating programs from learning feedback while retaining within-terrain difficulty rules~\citep{liang2024eurekaverse}. These approaches adapt different parts of a curriculum: the task sequence, the current stage, or the distribution of environments encountered by the learner.\parfillskip=0pt plus 0.5\linewidth\par\endgroup

\begingroup Task discovery also lets agents choose what to learn beyond a predetermined sequence. Agentic Skill Discovery proposes tasks from the environment and skill history, evolves reward code using success-rate feedback, and checks candidate successes with a VLM before admitting skills to a library~\citep{zhao2024agenticskilldiscovery}. Each skill is learned against a fixed sampled success function. The propose--learn--validate cycle expands the available training problems, while accepted skills become resources for subsequent tasks.\parfillskip=0pt plus 0.5\linewidth\par\endgroup

\begingroup Training configurations specify the conditions under which optimization proceeds. DrEureka proposes domain-randomization parameters and ranges from language-model reasoning informed by perturbation tests, then trains policies under the proposed distributions~\citep{ma2024dreureka}. AURA exposes reward, randomization, and training specifications through structured configurations and validation tools, combining automated feedback within iterations with user feedback between iterations~\citep{auracurriculum2025}. Both make experimental conditions accessible to revision without requiring the agent to implement the underlying optimizer.\parfillskip=0pt plus 0.5\linewidth\par\endgroup

\begingroup Configuration search adds a decision about where to spend further training effort. GigaWorld-Policy-0.5 reports an AutoResearch workflow for a pick-the-fruit study: propose hyperparameters, conduct short pilot runs, select candidates by validation metrics, and extend promising runs~\citep{gigaworldpolicy052026}. Early learning outcomes thus guide the allocation of further training effort to promising configurations.\parfillskip=0pt plus 0.5\linewidth\par\endgroup

\subsubsection{Policy Structure and Inductive Bias}

\begingroup Agents can also choose the representations and priors through which a policy learns. LGA converts a task description into a state abstraction that masks irrelevant features before imitation learning~\citep{peng2024lga}. KALM proposes visual keypoints, retains those consistent across demonstrations, and trains policies in keypoint-centric action frames~\citep{fang2025kalm}. These choices encode assumptions about which scene properties matter and how actions should generalize from limited demonstrations.\parfillskip=0pt plus 0.5\linewidth\par\endgroup

\begingroup APPL makes such assumptions candidates for experimental comparison~\citep{jiang2026appl}. It segments demonstrations into skills, proposes priors for each skill, and implements them in input representations, action coordinates, or auxiliary losses. Separate diffusion policies are trained with a shared recipe and budget; development results inform an additional proposal and policy selection. On six MetaWorld tasks with two demonstrations each, the first proposal alone exceeds a predefined relational prior by 18.1 points in out-of-distribution success, and searching over priors with selection on development states raises success to 89.6\%, against 37.9\% for that prior; test states were not available during design. A covered-assembly failure also shows why verification states must reflect runtime skill handoffs. The chosen priors enter the policy's runtime interface, linking training-time design to behavior selection (Section~\ref{sec:policy-a1}).\parfillskip=0pt plus 0.5\linewidth\par\endgroup

\begin{figure}[!t]
\centering
\includegraphics[width=\columnwidth]{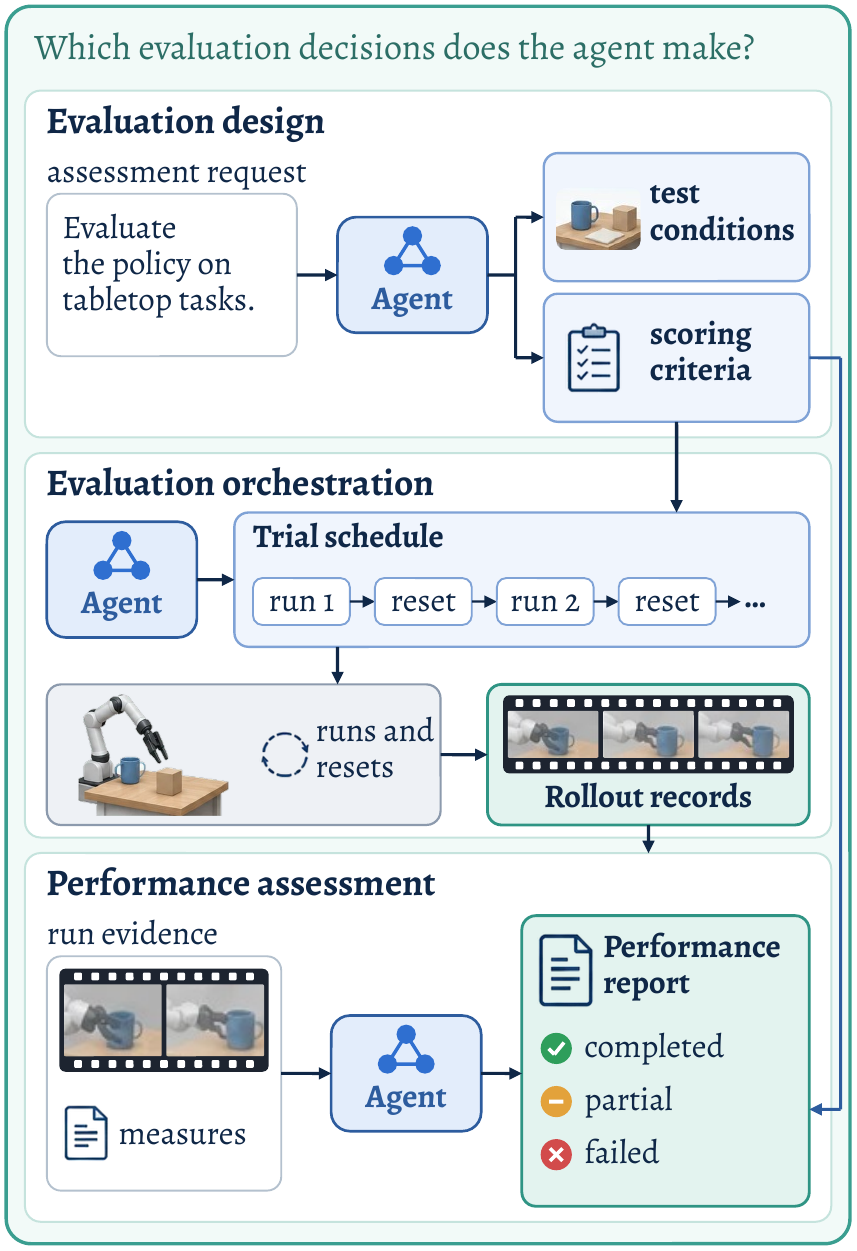}
\caption{Evaluation. Agents design tests and scoring, schedule trials and resets, and assess rollout evidence.}
\label{fig:infra-evaluation}
\end{figure}

\subsubsection{Experiment Execution and Management}

\begingroup Experiment management connects data, learners, compute, and evaluation. Nautilus uses coding agents to construct and validate adapters and execution environments around typed policy, benchmark, and robot contracts, supporting reproduction, evaluation, and initial fine-tuning~\citep{jin2026nautilus}. ARSTAG coordinates demonstration conversion, training launch, and evaluation through specialized agents and asynchronous execution~\citep{arstag2026}. These systems make learning workflows executable across heterogeneous components, with training schedules and selection procedures supplied by the workflow.\parfillskip=0pt plus 0.5\linewidth\par\endgroup

\begingroup Other systems expose the experiment itself to revision. HARBOR coordinates task construction, reward design, randomization, and RL integration through stage-specific agents, executable gates, parallel trials, and human interventions~\citep{li2026harbor}. ENPIRE first establishes reset, verification, and safety interfaces with human guidance; agents then modify training code and test policy-improvement hypotheses through those interfaces~\citep{self260619980}. Stable interfaces allow successive experiments to change a learning procedure while preserving a common execution and assessment framework.\parfillskip=0pt plus 0.5\linewidth\par\endgroup

\begingroup Agents may also decide when learning is needed and whether to adopt its result. Robo-COP identifies target skills and decides whether to train, then checks target performance and regressions before adopting a candidate, with a fallback available~\citep{zhang2026robocop}. SEAR's auto-research branch connects demonstration collection and curation, VLA fine-tuning, evaluation, and subsequent experimental choices~\citep{wang2026sear}. Such workflows use evidence from current performance to initiate and assess another learning intervention.\parfillskip=0pt plus 0.5\linewidth\par\endgroup

\begingroup Efficient experimentation depends on obtaining useful feedback before committing to full training. CARD screens reward candidates by their ranking of stored successful and unsuccessful trajectories~\citep{card2024}; EvoNav progresses from analytical proxies through lightweight training to full policy training~\citep{evonav2026}. Final assessment can then use held-out tasks, as in SEAR, or verifier-controlled hidden tests under resource limits, as in RLE-Bench~\citep{wang2026sear,ma2026rlebench}. Accounting for the complete development cycle also changes the cost picture: Eurekaverse reports about 24 hours on eight A6000 GPUs alongside roughly US\$15 in model API charges~\citep{liang2024eurekaverse}, while DrEureka evaluates policies from 16 randomization configurations on hardware~\citep{ma2024dreureka}. Training comparisons should therefore report candidate-search and selection costs alongside final policy performance.\parfillskip=0pt plus 0.5\linewidth\par\endgroup

\input{tables/infra-training-comparison}

\subsection{Evaluation}\label{sec:infra-evaluation}

Evaluation measures robot policy performance across tasks and conditions, helping identify strengths and limitations. Fixed task suites offer limited flexibility for new evaluation goals, aggregate scores provide limited insight into specific behaviors, and testing new policy--benchmark combinations often requires repeated manual integration~\citep{chen2026manipevalagent,jin2026nautilus}. We organize agent contributions to evaluation into three functions: evaluation design adapts tests and scoring procedures; evaluation orchestration coordinates trials; and performance assessment interprets policy behavior and outcomes. The functions in Figure~\ref{fig:infra-evaluation} and Table~\ref{tab:infra-evaluation} can support policy development or standalone testing, with results interpreted according to the test conditions, scoring criteria, and assistance provided during execution.

\subsubsection{Evaluation Design}
\begingroup Evaluation design specifies which policy capabilities to test, under what conditions, and how performance will be measured. Agents translate assessment goals into executable task programs and evaluation tools, and develop evaluators that check task success or provide scores and feedback.\parfillskip=0pt plus 0.15\linewidth\par\endgroup

ManipEvalAgent~\citep{chen2026manipevalagent} selects evaluation aspects from user questions and simulator capabilities, then retrieves or generates task programs and evaluation tools. Its Plan Agent refines subsequent tests using trial results.

To score policy trials, ENPIRE~\citep{self260619980} uses coding agents to synthesize binary success checks from successful and failed demonstrations, with human feedback guiding refinement. MEMENTO~\citep{sygkounas2026memento} evolves evaluator programs on fixed initial-policy rollouts to guide policy search with fitness scores and feedback; task success is measured separately. Both approaches hold their evaluation functions fixed during \mbox{subsequent policy optimization.}

\subsubsection{Evaluation Orchestration}
Agents orchestrate evaluation by managing the operations needed to run specified tests. They connect policies to test environments through compatible interfaces, coordinate evaluation jobs and collect their results, and arrange scene resets between physical trials.

XPolicyLab~\citep{xpolicylab2026} equips coding agents to construct and audit policy adapters under human configuration and review. Nautilus~\citep{jin2026nautilus} assembles policy and benchmark modules from user requests, using shared interface contracts and fixed checks to guide agent corrections. Comparisons with published benchmark results assess protocol fidelity. As supporting infrastructure, the unified evaluation harness vla-eval~\citep{choi2026vlaeval} connects policy servers to benchmarks and records evaluation configurations and per-episode metrics.

EmbodiedClaw~\citep{embodiedclaw2026} coordinates trials by composing evaluation workflows and revising them using execution feedback. ARSTAG~\citep{arstag2026} assigns simulation evaluation to a learning agent that returns evaluation summaries and failure diagnostics to its coordinator. Hermes Embodied~\citep{hermesembodied2026} provides a repository prototype with scripts for simulation \mbox{evaluation and checkpoint comparison.}

\begingroup HALTER~\citep{jiang2026halter} uses an LLM to select and sequence learned reset skills that restore the required scene relations between physical trials. It checks skill effects and the final scene, requesting human intervention when verification fails.\parfillskip=0pt plus 0.1\linewidth\par\endgroup

\begin{figure}[!t]
\centering
\includegraphics[width=\columnwidth]{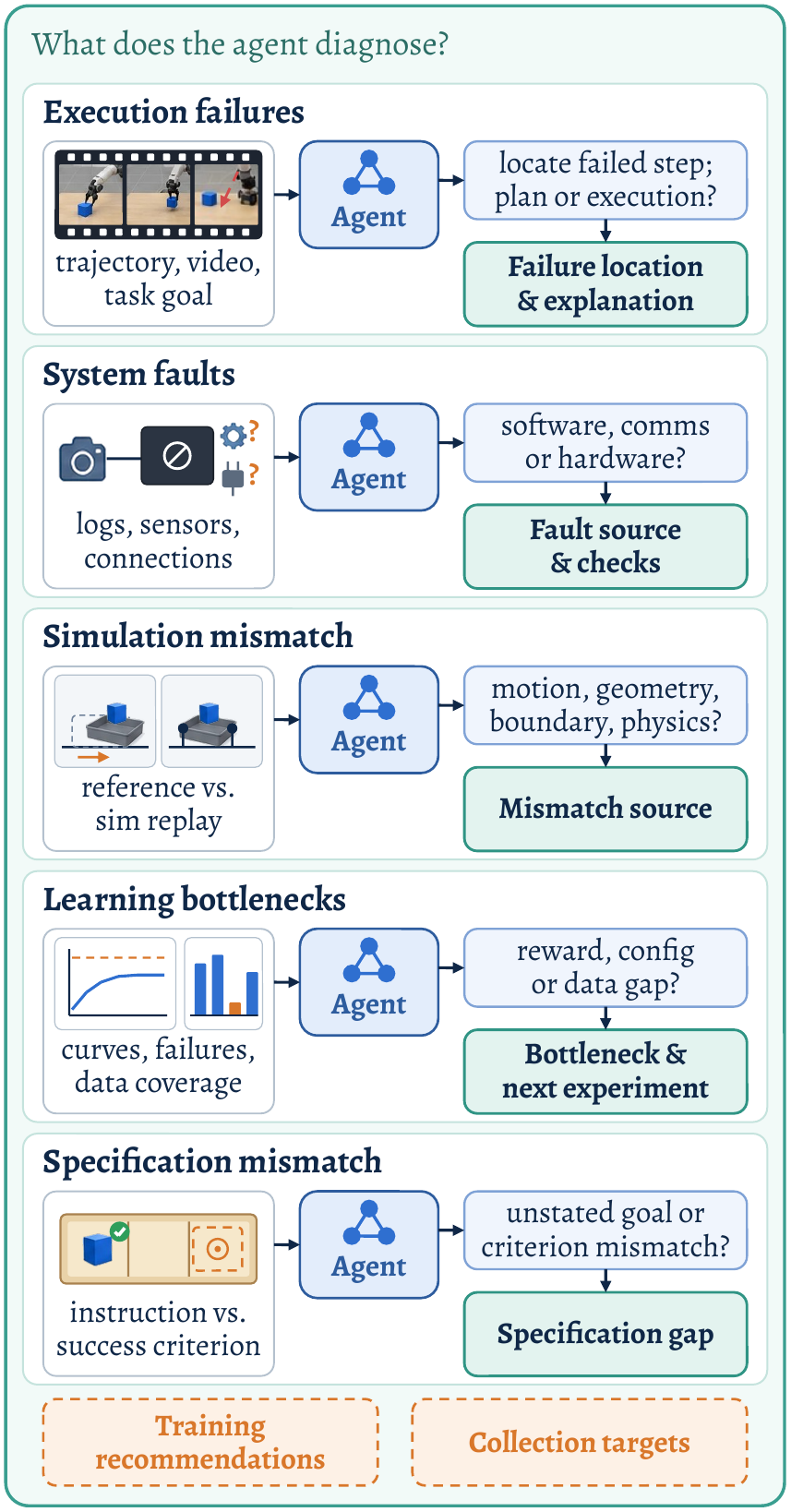}
\caption{Diagnosis. Agents diagnose execution failures, system faults, simulation mismatch, learning bottlenecks, and specification mismatch. Findings identify causes and guide further training or data collection.}
\label{fig:infra-diagnosis}
\end{figure}

\subsubsection{Performance Assessment}
\begingroup Performance assessment interprets policy behavior and task outcomes according to the evaluation criteria. Agents analyze observations and execution records during or after trials to judge task completion and progress, identify behavioral weaknesses, and check whether policy updates improve or worsen performance.\parfillskip=0pt plus 0.1\linewidth\par\endgroup

\begingroup SceneSmith~\citep{scenesmith2026} combines geometric queries with visual inspection to judge task completion. Agreement with human annotations assesses the reliability of these success judgments, while policy success rates describe performance on the generated tasks. ManipEvalAgent's Plan Agent~\citep{chen2026manipevalagent} interprets simulator measurements and visual judgments to answer the user's questions about policy capabilities.\parfillskip=0pt plus 0.1\linewidth\par\endgroup

Beyond final outcomes, HARBOR~\citep{li2026harbor} analyzes metrics and rendered rollouts to identify behavioral problems such as collisions during policy tuning. EmbodiRSI's monitor agent~\citep{embodirsi2026} uses images and simulator states during training data collection to assess task progress, request recovery, or declare task completion. The trained policy is tested without agent takeover, planner recovery, or human correction.

\begingroup HALTER~\citep{jiang2026halter} assesses task progress by estimating the fraction of skills whose intended effects were achieved at least once during a rollout, including effects later undone or hidden from view. ROVER~\citep{schroeder2025rover} recursively segments videos into subtasks to estimate frame-level task progress. Evaluation uses simulator-derived progress references for expert and perturbed non-expert RoboCasa trajectories, and frame number as a progress proxy for real-world OXE videos.\parfillskip=0pt plus 0.1\linewidth\par\endgroup

RoboLineage's Policy Evaluation Agent~\citep{luo2026robolineage} uses task criteria and rollout records to identify failure phases and behavioral weaknesses, and detect performance regressions~\mbox{relative to the parent policy.}

Agents can adapt evaluation to new tasks and assessment goals while reducing the manual work needed to set up and repeat trials. By interpreting task outcomes alongside execution records, they provide a more detailed account of what a policy can accomplish, where execution fails, and whether policy updates improve performance.










\input{tables/infra-evaluation-comparison}

\subsection{Diagnosis}\label{sec:infra-diagnosis}




\input{tables/infra-diagnosis-comparison}

\begingroup Diagnosis asks why an outcome departed from intended behaviour and what should be checked or changed next.
We distinguish it from detection, which establishes that something went wrong, and from corrective control, which changes execution. Diagnosis develops an explanation supported by evidence.
The agentic setting differs in two respects: the evidence is heterogeneous, ranging from logs and videos to code and training curves, and the agent can act on its findings by revising software, simulation models, or learning pipelines.\parfillskip=0pt plus 0.1\linewidth\par\endgroup

Figure~\ref{fig:infra-diagnosis} poses one diagnostic question for each of five targets, moving from a single execution through the execution stack, the simulator, and the learning pipeline to the task specification, and Table~\ref{tab:infra-diagnosis} lists a representative work for each.
We discuss the targets in this order.
Each paragraph starts from the target's diagnostic question, describes how the representative work answers it, and then compares related works along one dimension.

\subsubsection{Execution failures}
\label{sec:diag-execution}

The question here is which step failed and whether the plan or its execution was at fault.
REFLECT answers this question in a fixed order~\citep{grid47}.
It summarizes multisensory experience hierarchically, verifies subgoals one by one, and retrieves the event-based summary when a subgoal fails.
If all subgoals succeed but the task still fails, it examines the plan against the final state, and the resulting explanation guides a planner that generates a correction plan.

Related works differ mainly in how the explanation is used.
AHA predicts success or failure and adds a free-form explanation; it is trained on data generated from seven failure modes, and its feedback supports replanning in PRoC3S and reward revision in Eureka~\citep{aha2024}.
RoboFAC uses failure localization to produce corrections at two levels: high-level corrections specify or revise the recovery subtask sequence, and low-level corrections guide end-effector motion~\citep{ye2025robofac}.
Across four real-world tasks, success within five attempts averages 61.25\% with low-level corrections and 50.00\% with high-level corrections.
All three works explain from evidence recorded during the failed execution.
By contrast, Argon's developer demonstration adds a check before acting: its prompt requires the agent to review logs and footage from a failed SO-101 attempt, diagnose the failure, and dry-run the proposed fix before a hardware retry~\citep{rao2026argon}.
The demonstration shows faster and more accurate placement over successive iterations, although it is not a controlled evaluation of diagnosis.

\begingroup Across these works, the main difference lies in the next step that an explanation enables: a correction plan in REFLECT, replanning or reward revision in AHA, subtask- or motion-level corrections in RoboFAC, and a checked fix in Argon.\parfillskip=0pt plus 0.1\linewidth\par\endgroup

\subsubsection{System faults}
\label{sec:diag-system}
\begingroup The question here is whether a failure originates in the execution stack rather than in the policy.
SPINE addresses this question with a debugger that separates software, hardware, and hybrid faults and iterates through diagnosis, repair, and validation until the teleoperation pipeline and profile-defined readiness checks pass~\citep{ham2026spinediagnosis}.
For software faults, the agent edits code, configuration, or environment.
For physical faults, it presents one action at a time from a hardware playbook, and a novice operator supplies the requested observations and carries out the action.\parfillskip=0pt plus 0.1\linewidth\par\endgroup

\begingroup Related systems differ in how much authority the agent holds.
ROS Help Desk adapts its explanations to the user's expertise and executes a fix only after the user requests it; it is evaluated on lidar and image drops, delays, corruption, and node crashes injected into a simulated TurtleBot3~\citep{roshelpdesk2025}.
By contrast, AURA is advisory: one agent turns telemetry anomalies into a structured description, a second reasons with the operator over curated external sources, and the operator makes the final diagnosis and decides on any action on the vehicle~\citep{auradiagnostics2025}.
Operator-validated cases are then stored in a vector database as context for later diagnoses.\parfillskip=0pt plus 0.1\linewidth\par\endgroup

\begingroup The same kind of fault also arises in evaluation stacks, where it can be mistaken for a policy failure.
When reproducing published scores across six VLA codebases, vla-eval found that using the wrong proprioceptive state source lowered X-VLA's LIBERO success from 97.8\% to 42\%, that confusing absolute and delta actions produced 0\% success, and that OpenVLA applies an undocumented centre crop at evaluation~\citep{choi2026vlaeval}.
Shared interfaces are one response: XPolicyLab makes adapter conformance checkable through staged gates and supports agent-assisted integration under human supervision~\citep{xpolicylab2026}.\parfillskip=0pt plus 0.1\linewidth\par\endgroup

\begingroup Across these works, the agent's role ranges from executing repairs, as in SPINE and ROS Help Desk, to advising an operator who keeps control, as in AURA; the same fault types should also be ruled out before a policy is judged to have failed.\parfillskip=0pt plus 0.1\linewidth\par\endgroup

\subsubsection{Simulation mismatch}
\label{sec:diag-simulation}
\begingroup The question here is which modelling choice explains a gap between real and simulated behaviour: motion, geometry, boundary conditions, or physical parameters.
RoboSimStudio traces such a gap back from a real failure~\citep{chen2026robosimstudio}.
A policy optimized in simulation raised the wiped area from 0\% to 100\% over 14 iterations, but one real rollout displaced a bar support that the simulation held fixed.
Given the failure video and a description, the agent replaced the fixed constraint with a free rigid body, evaluated the revised simulation under different wiping trajectories and contact conditions, and reproduced the bar's motion and tipping.\parfillskip=0pt plus 0.1\linewidth\par\endgroup

Related works differ in what a passed check establishes.
Agentic RSR treats an agent's assignment of mesh roles as a reviewable physical hypothesis~\citep{li2026agenticrsr}.
A physics-checking session then runs settling, simplified-grasp, or articulated-motion probes and tests candidates with changed collision geometry, pose, or physical parameters, and a runner repeats the relevant probes after each revision.
CoDimRecon checks whether each reconstructed asset supports its intended interaction, revising motion first and then geometry and boundary conditions, numerics, and material~\citep{xie2026codimrecon}.
For example, paper springback after release led to a plastic hinge-bending model: one second after the gripper opened, the paper retained 92.5\% of its peak fold angle, compared with 3.7\% under elastic bending.

Across these works, a passed check establishes different things: that an observed failure can be reproduced (RoboSimStudio), that a scene is plausible for its task (Agentic RSR), or that a material model yields the intended behaviour (CoDimRecon).
In addition, these workflows revise modelling assumptions as well as numerical parameters: RoboSimStudio changes a support constraint, and CoDimRecon changes the bending model.

\subsubsection{Learning bottlenecks}
\label{sec:diag-learning}

The question here is why training fails to improve the policy.
The candidate explanations are implementation faults, reward or configuration problems, and data gaps, and the evidence includes training curves, rendered behaviour, failures, and data coverage.
HARBOR addresses this question by running training as gated stages~\citep{li2026harbor}.
When a gate fails, the stage agent receives the failed check, the error message, and the observed values for repair.
After parallel trials finish, the main agent analyses their outcomes, proposes the next configuration, and distils recurring patterns into an experience memory for later rounds.
Users can audit outputs, correct a stage, and resume from the last passing gate, and a stage that remains unresolved after its retry budget is escalated to them.

\begingroup Related works differ in who chooses the next experiment and in what form.
In a quadruped RL study of more than 70 experiments in 14 waves, the human usually chose the broad topic of each phase and the agent chose the implementation; in the process, the agent traced PhysX deadlocks to terrain sets containing boxes and stair-like primitives~\citep{khandelwal2026agentrl}.
In autoresearch-robotics, the agent works within a fixed training-time budget and keeps a change only if evaluation success improves, using rendered start, middle, and end frames as additional evidence~\citep{autoresearchrobotics2026}.
ENPIRE extends this loop to physical robots~\citep{self260619980}.
Its coding agents first construct environment APIs for reset and verification through procedural tools and human feedback, using a few minutes of success and failure demonstrations to synthesize an outcome checker.
They then analyse logged trajectories, video, and rewards to modify training code.\parfillskip=0pt plus 0.1\linewidth\par\endgroup

By contrast, two methods turn an explanation into a reward through a fixed interface.
Reflection-Based Task Adaptation has a VLM analyse failure videos and trajectory logs and select from a library of parameterized reward functions~\citep{li2025reflectiveadaptation}.
In ReCoVLA, a deterministic compiler maps the VLM's recovery descriptor to stage-gated rewards for a residual recovery policy~\citep{hu2026recovla}.
On three simulated tabletop tasks, this raises average success from 36.7\% to 66.7\% over a fine-tuned $\pi_{0.5}$ policy without recovery.
On the data side, RoboLineage examines coverage and imbalance, and F4R uses failure-associated conditions to target simulation data~\citep{luo2026robolineage,f4r2026}.

Across these works, the next experiment ranges from a configuration chosen by the agent to a reward produced through a fixed interface, and people set the direction or audit the stages where the workflow requires it.

\subsubsection{Specification mismatch}
\label{sec:diag-specification}

\begingroup The question here is whether a failure belongs to the task rather than the policy because the instruction omits a condition that the success checker enforces.
BenchMend addresses this question by separating the information available for acting from the information available for auditing~\citep{zhang2026benchmend}.
The acting agent sees the instruction, camera views, and proprioception, but not the checker code or demonstrations, and its completion judgements are compared with the benchmark's verdicts.
The authors then review every disagreement manually, and a repair step, which may consult task definitions, edits the instruction minimally while keeping the checker and recorded demonstrations unchanged.
On LIBERO and RoboTwin, the edits add missing goal information such as a target location or orientation.\parfillskip=0pt plus 0.2\linewidth\par\endgroup

In this target, human reviewers play a central role: they distinguish instruction--criterion mismatches from agent mistakes and judge what goal information is missing.

\subsubsection{Evidence and human participation}
\label{sec:diag-evidence}

\begingroup The five targets differ in what is diagnosed, but they can be compared by the evidence that supports a diagnosis.
We distinguish three forms of support: an explanation consistent with recorded evidence, a targeted diagnostic test, and an improved outcome after an intervention.
REFLECT, AHA, and RoboFAC ground their explanations in recorded executions.
Agentic RSR's probes, CoDimRecon's behavioural tests, and RoboSimStudio's reproduction of an observed failure are targeted tests.
RoboFAC's corrections, ReCoVLA's rewards, and BenchMend's instruction edits are evaluated through downstream outcomes.
Such outcomes support the chosen intervention, but they do not by themselves establish that the proposed explanation identifies the real-world cause.
Stronger causal support would come from testing diagnostic hypotheses against new evidence, using pre-specified predictions that distinguish alternative explanations; developing such prospective tests is a direction for diagnostic infrastructure.\parfillskip=0pt plus 0.1\linewidth\par\endgroup

The quality of a diagnosis also depends on the agent's own habits.
EmbodiedSWE observes that agents sometimes rely on printed simulator state instead of rendered observations and may lock into an incorrect task decomposition, treating later failures as tuning problems~\citep{embodiedswe2026}.
In RHO, a single mutation step with unlimited internal turns reaches 13--30\% across models on the seven Robosuite tasks of CaP-Bench, compared with 70\% for the repository selected after evolutionary search.
The authors conclude that repeated reflective search, rather than one extended session, is needed to close this gap~\citep{rho2026}.

People take part in two ways across the targets.
They supply observations or carry out physical interventions, as in RoboSimStudio and SPINE, or they retain authority over the diagnosis, the research direction, or the interpretation of the task, as in AURA, the quadruped study, and BenchMend.

%% file: tables/infra-environment-comparison.tex
\begin{table}[!tb]
\centering
\caption{Environment generation by constructed artifact.}
\label{tab:infra-environment}
\begingroup
\small
\makeatletter
\def\NAT@spacechar{\nobreakspace}
\makeatother
\setlength{\tabcolsep}{2pt}
\renewcommand{\arraystretch}{1.10}
\begin{tabularx}{\linewidth}{@{}>{\raggedright\arraybackslash}p{0.40\linewidth}>{\raggedright\arraybackslash}p{20pt}>{\raggedright\arraybackslash}p{0.24\linewidth}>{\raggedright\arraybackslash}X@{}}
\toprule
\rowcolor[HTML]{DEF4F0}
\textbf{Work} & \textbf{Link} & \textbf{Subproblem} & \textbf{Output} \\
\midrule
Articulate-Anything \citep{le2024articulateanything} & \href{https://proceedings.iclr.cc/paper_files/paper/2025/hash/2c8047bf3ed8ef6905351608d641f02f-Abstract-Conference.html}{\includegraphics[height=8pt]{figures/icons/paper.pdf}}\hspace{3pt}\href{https://github.com/vlongle/articulate-anything}{\includegraphics[height=8pt]{figures/icons/github.pdf}} & Assets & Articulated asset \\
Articraft \citep{articraft2026} & \href{https://arxiv.org/abs/2605.15187}{\includegraphics[height=8pt]{figures/icons/paper.pdf}}\hspace{3pt}\href{https://github.com/articraftresearch/Articraft}{\includegraphics[height=8pt]{figures/icons/github.pdf}} & Assets & Asset program \\
SceneSmith \citep{scenesmith2026} & \href{https://arxiv.org/abs/2602.09153}{\includegraphics[height=8pt]{figures/icons/paper.pdf}}\hspace{3pt}\href{https://github.com/nepfaff/scenesmith}{\includegraphics[height=8pt]{figures/icons/github.pdf}} & Scenes & Scene layout \\
EmbodiedGen V2 \citep{wang2026embodiedgenv2} & \href{https://arxiv.org/abs/2607.07459}{\includegraphics[height=8pt]{figures/icons/paper.pdf}}\hspace{3pt}\href{https://github.com/HorizonRobotics/EmbodiedGen}{\includegraphics[height=8pt]{figures/icons/github.pdf}} & Scenes & Scene layout \\
Agentic Real2Sim \citep{agenticreal2sim2026} & \href{https://arxiv.org/abs/2607.19190}{\includegraphics[height=8pt]{figures/icons/paper.pdf}}\hspace{3pt}\href{https://github.com/agentic-real2sim/agentic_real2sim}{\includegraphics[height=8pt]{figures/icons/github.pdf}} & Reconstruction & Simulated twin \\
RAPID \citep{rapid2026} & \href{https://arxiv.org/abs/2609.30249}{\includegraphics[height=8pt]{figures/icons/paper.pdf}} & Reconstruction & Verification scenes \\
RoboSimStudio \citep{chen2026robosimstudio} & \href{https://arxiv.org/abs/2610.18835}{\includegraphics[height=8pt]{figures/icons/paper.pdf}}\hspace{3pt}\href{https://github.com/robosimstudio/robosimstudio}{\includegraphics[height=8pt]{figures/icons/github.pdf}} & Reconstruction & Simulated twin \\
GenSim \citep{gensim2023} & \href{https://arxiv.org/abs/2310.01361}{\includegraphics[height=8pt]{figures/icons/paper.pdf}}\hspace{3pt}\href{https://github.com/liruiw/GenSim}{\includegraphics[height=8pt]{figures/icons/github.pdf}} & Tasks & Task program \\
Gen2Sim \citep{gen2sim2023} & \href{https://arxiv.org/abs/2310.18308}{\includegraphics[height=8pt]{figures/icons/paper.pdf}}\hspace{3pt}\href{https://github.com/pushkalkatara/Gen2Sim}{\includegraphics[height=8pt]{figures/icons/github.pdf}} & Tasks & Task definition \\
\bottomrule
\end{tabularx}
\endgroup
\end{table}

%% file: tables/infra-dataset-comparison.tex
\begin{table}[!tb]
\centering
\caption{Dataset construction: collection and admission.}
\label{tab:infra-dataset}
\begingroup
\small
\makeatletter
\def\NAT@spacechar{\nobreakspace}
\makeatother
\setlength{\tabcolsep}{2pt}
\renewcommand{\arraystretch}{1.10}
\begin{tabularx}{\linewidth}{@{}>{\raggedright\arraybackslash}p{0.40\linewidth}>{\raggedright\arraybackslash}p{20pt}>{\raggedright\arraybackslash}p{0.21\linewidth}>{\raggedright\arraybackslash}X@{}}
\toprule
\rowcolor[HTML]{DEF4F0}
\textbf{Work} & \textbf{Link} & \textbf{Subproblem} & \textbf{Output} \\
\midrule
GenSim2 \citep{hua2024gensim2} & \href{https://arxiv.org/abs/2410.03645}{\includegraphics[height=8pt]{figures/icons/paper.pdf}}\hspace{3pt}\href{https://github.com/GenSim2/GenSim2}{\includegraphics[height=8pt]{figures/icons/github.pdf}} & Experts & Demonstrations \\
RoboTwin 2.0 \citep{robotwin2agent2025} & \href{https://arxiv.org/abs/2506.18088}{\includegraphics[height=8pt]{figures/icons/paper.pdf}}\hspace{3pt}\href{https://github.com/RoboTwin-Platform/RoboTwin}{\includegraphics[height=8pt]{figures/icons/github.pdf}} & Experts & Demonstrations \\
AutoRT \citep{autort2024} & \href{https://arxiv.org/abs/2401.12963}{\includegraphics[height=8pt]{figures/icons/paper.pdf}}\hspace{3pt}\makebox[8pt]{--} & Physical collection & Robot episodes \\
SOAR \citep{soar2024} & \href{https://arxiv.org/abs/2407.20635}{\includegraphics[height=8pt]{figures/icons/paper.pdf}}\hspace{3pt}\href{https://github.com/rail-berkeley/soar}{\includegraphics[height=8pt]{figures/icons/github.pdf}} & Physical collection & Robot episodes \\
UniIntervene \citep{deng2026uniintervene} & \href{https://arxiv.org/abs/2606.12372}{\includegraphics[height=8pt]{figures/icons/paper.pdf}}\hspace{3pt}\makebox[8pt]{--} & Assisted collection & Recovery transitions \\
UniIntervene++ \citep{lin2026uniintervenepp} & \href{https://arxiv.org/abs/2610.03620}{\includegraphics[height=8pt]{figures/icons/paper.pdf}}\hspace{3pt}\makebox[8pt]{--} & Double-DQN intervention & Online / assisted replay \\
ARS \citep{hu2026ars} & \href{https://arxiv.org/abs/2609.34484}{\includegraphics[height=8pt]{figures/icons/paper.pdf}}\hspace{3pt}\href{https://github.com/midea-ai/ars}{\includegraphics[height=8pt]{figures/icons/github.pdf}} & Annotation / admission & Action chunks \\
Robo-COP \citep{zhang2026robocop} & \href{https://arxiv.org/abs/2610.09228}{\includegraphics[height=8pt]{figures/icons/paper.pdf}}\hspace{3pt}\makebox[8pt]{--} & Annotation / admission & Skill segments \\
EmbodiRSI \citep{embodirsi2026} & \href{https://arxiv.org/abs/2609.38905}{\includegraphics[height=8pt]{figures/icons/paper.pdf}}\hspace{3pt}\makebox[8pt]{--} & Targeted acquisition & Recovery data \\
SEAR \citep{wang2026sear} & \href{https://arxiv.org/abs/2610.18792}{\includegraphics[height=8pt]{figures/icons/paper.pdf}}\hspace{3pt}\href{https://github.com/Self-Evolving-Agentic-Robotics/SEAR}{\includegraphics[height=8pt]{figures/icons/github.pdf}} & Targeted acquisition & Additional data \\
\bottomrule
\end{tabularx}
\endgroup
\end{table}

%% file: tables/infra-training-comparison.tex
\begin{table}[!tb]
\centering
\caption{Agent decisions in training.}
\label{tab:infra-training}
\begingroup
\small
\makeatletter
\def\NAT@spacechar{\nobreakspace}
\makeatother
\setlength{\tabcolsep}{2pt}
\renewcommand{\arraystretch}{1.10}
\begin{tabularx}{\linewidth}{@{}>{\raggedright\arraybackslash}p{0.40\linewidth}>{\raggedright\arraybackslash}p{20pt}>{\raggedright\arraybackslash}p{0.21\linewidth}>{\raggedright\arraybackslash}X@{}}
\toprule
\rowcolor[HTML]{DEF4F0}
\textbf{Work} & \textbf{Link} & \textbf{Subproblem} & \textbf{Output} \\
\midrule
Eureka \citep{ma2023eureka} & \href{https://arxiv.org/abs/2310.12931}{\includegraphics[height=8pt]{figures/icons/paper.pdf}}\hspace{3pt}\href{https://github.com/eureka-research/Eureka}{\includegraphics[height=8pt]{figures/icons/github.pdf}} & Supervision & Reward code \\
Text2Reward \citep{text2reward2023} & \href{https://arxiv.org/abs/2309.11489}{\includegraphics[height=8pt]{figures/icons/paper.pdf}}\hspace{3pt}\href{https://github.com/xlang-ai/text2reward}{\includegraphics[height=8pt]{figures/icons/github.pdf}} & Supervision & Reward code \\
CurricuLLM \citep{curricullm2024} & \href{https://arxiv.org/abs/2409.18382}{\includegraphics[height=8pt]{figures/icons/paper.pdf}}\hspace{3pt}\href{https://github.com/labicon/CurricuLLM}{\includegraphics[height=8pt]{figures/icons/github.pdf}} & Curriculum / settings & Task sequence \\
DrEureka \citep{ma2024dreureka} & \href{https://arxiv.org/abs/2406.01967}{\includegraphics[height=8pt]{figures/icons/paper.pdf}}\hspace{3pt}\href{https://github.com/eureka-research/DrEureka}{\includegraphics[height=8pt]{figures/icons/github.pdf}} & Curriculum / settings & Randomization \\
LGA \citep{peng2024lga} & \href{https://arxiv.org/abs/2402.18759}{\includegraphics[height=8pt]{figures/icons/paper.pdf}}\hspace{3pt}\makebox[8pt]{--} & Policy structure & State abstraction \\
KALM \citep{fang2025kalm} & \href{https://arxiv.org/abs/2410.23254}{\includegraphics[height=8pt]{figures/icons/paper.pdf}}\hspace{3pt}\makebox[8pt]{--} & Policy structure & Keypoints \\
APPL \citep{jiang2026appl} & \href{https://arxiv.org/abs/2609.35690}{\includegraphics[height=8pt]{figures/icons/paper.pdf}}\hspace{3pt}\makebox[8pt]{--} & Policy structure & Skill priors \\
Nautilus \citep{jin2026nautilus} & \href{https://arxiv.org/abs/2605.11665}{\includegraphics[height=8pt]{figures/icons/paper.pdf}}\hspace{3pt}\makebox[8pt]{--} & Experiments & Run interfaces \\
ARSTAG \citep{arstag2026} & \href{https://arxiv.org/abs/2609.24563}{\includegraphics[height=8pt]{figures/icons/paper.pdf}}\hspace{3pt}\makebox[8pt]{--} & Experiments & Training runs \\
\bottomrule
\end{tabularx}
\endgroup
\end{table}

%% file: tables/infra-evaluation-comparison.tex
\begin{table}[!tb]
\centering
\caption{Evaluation design, orchestration and assessment.}
\label{tab:infra-evaluation}
\begingroup
\small
\makeatletter
\def\NAT@spacechar{\nobreakspace}
\makeatother
\setlength{\tabcolsep}{2pt}
\renewcommand{\arraystretch}{1.10}
\begin{tabularx}{\linewidth}{@{}>{\raggedright\arraybackslash}p{0.42\linewidth}>{\raggedright\arraybackslash}p{20pt}>{\raggedright\arraybackslash}p{0.21\linewidth}>{\raggedright\arraybackslash}X@{}}
\toprule
\rowcolor[HTML]{DEF4F0}
\textbf{Work} & \textbf{Link} & \textbf{Subproblem} & \textbf{Output} \\
\midrule
ManipEvalAgent \citep{chen2026manipevalagent} & \href{https://proceedings.iclr.cc/paper_files/paper/2026/file/8833c8aa10542d24d693bbaf6a4598f5-Paper-Conference.pdf}{\includegraphics[height=8pt]{figures/icons/paper.pdf}}\hspace{3pt}\makebox[8pt]{--} & Design & Test programs \\
ENPIRE \citep{self260619980} & \href{https://arxiv.org/abs/2606.19980}{\includegraphics[height=8pt]{figures/icons/paper.pdf}}\hspace{3pt}\href{https://github.com/NVlabs/ENPIRE}{\includegraphics[height=8pt]{figures/icons/github.pdf}} & Design & Success checks \\
Nautilus \citep{jin2026nautilus} & \href{https://arxiv.org/abs/2605.11665}{\includegraphics[height=8pt]{figures/icons/paper.pdf}}\hspace{3pt}\makebox[8pt]{--} & Orchestration & Test interfaces \\
HALTER \citep{jiang2026halter} & \href{https://arxiv.org/abs/2609.19413}{\includegraphics[height=8pt]{figures/icons/paper.pdf}}\hspace{3pt}\makebox[8pt]{--} & Orchestration & Reset sequences \\
SceneSmith \citep{scenesmith2026} & \href{https://arxiv.org/abs/2602.09153}{\includegraphics[height=8pt]{figures/icons/paper.pdf}}\hspace{3pt}\href{https://github.com/nepfaff/scenesmith}{\includegraphics[height=8pt]{figures/icons/github.pdf}} & Assessment & Success judgments \\
ROVER \citep{schroeder2025rover} & \href{https://arxiv.org/abs/2508.01943}{\includegraphics[height=8pt]{figures/icons/paper.pdf}}\hspace{3pt}\makebox[8pt]{--} & Assessment & Task progress \\
\bottomrule
\end{tabularx}
\endgroup
\end{table}

%% file: tables/infra-diagnosis-comparison.tex
\begin{table}[!b]
\centering
\caption{Diagnosis by target (Figure~\ref{fig:infra-diagnosis}).}
\label{tab:infra-diagnosis}
\begingroup
\small
\makeatletter
\def\NAT@spacechar{\nobreakspace}
\makeatother
\setlength{\tabcolsep}{2pt}
\renewcommand{\arraystretch}{1.10}
\begin{tabularx}{\linewidth}{@{}>{\raggedright\arraybackslash}p{0.42\linewidth}>{\raggedright\arraybackslash}p{20pt}>{\raggedright\arraybackslash}p{0.23\linewidth}>{\raggedright\arraybackslash}X@{}}
\toprule
\rowcolor[HTML]{DEF4F0}
\textbf{Work} & \textbf{Link} & \textbf{Subproblem} & \textbf{Output} \\
\midrule
REFLECT \citep{grid47} & \href{https://arxiv.org/abs/2306.15724}{\includegraphics[height=8pt]{figures/icons/paper.pdf}}\hspace{3pt}\href{https://github.com/real-stanford/reflect}{\includegraphics[height=8pt]{figures/icons/github.pdf}} & Execution failures & Failure explanation \\
RoboFAC \citep{ye2025robofac} & \href{https://arxiv.org/abs/2505.12224}{\includegraphics[height=8pt]{figures/icons/paper.pdf}}\hspace{3pt}\href{https://github.com/MINT-SJTU/RoboFAC}{\includegraphics[height=8pt]{figures/icons/github.pdf}} & Execution failures & Recovery instruction \\
ROS Help Desk \citep{roshelpdesk2025} & \href{https://arxiv.org/abs/2507.07846}{\includegraphics[height=8pt]{figures/icons/paper.pdf}}\hspace{3pt}\makebox[8pt]{--} & System faults & Fault explanation \\
SPINE (2026) \citep{ham2026spinediagnosis} & \href{https://arxiv.org/abs/2607.13049}{\includegraphics[height=8pt]{figures/icons/paper.pdf}}\hspace{3pt}\makebox[8pt]{--} & System faults & Fault source \\
RoboSimStudio \citep{chen2026robosimstudio} & \href{https://arxiv.org/abs/2610.18835}{\includegraphics[height=8pt]{figures/icons/paper.pdf}}\hspace{3pt}\href{https://github.com/robosimstudio/robosimstudio}{\includegraphics[height=8pt]{figures/icons/github.pdf}} & Simulation mismatch & Model mismatch \\
HARBOR \citep{li2026harbor} & \href{https://arxiv.org/abs/2606.08610}{\includegraphics[height=8pt]{figures/icons/paper.pdf}}\hspace{3pt}\makebox[8pt]{--} & Learning bottlenecks & Next experiment \\
BenchMend \citep{zhang2026benchmend} & \href{https://tianqi-zh.github.io/BenchMend/}{\includegraphics[height=8pt]{figures/icons/paper.pdf}}\hspace{3pt}\makebox[8pt]{--} & Specification mismatch & Goal clarification \\
\bottomrule
\end{tabularx}
\endgroup
\end{table}

%% file: sections/05_agent_as_self_improver.tex
\section{Agent as Evolver}
\label{sec:self-improvement}
\begin{figure*}[!t]
\centering
\setlength{\abovecaptionskip}{4pt}
\includegraphics[width=\linewidth,trim=8.5bp 9bp 9bp 7.5bp,clip]{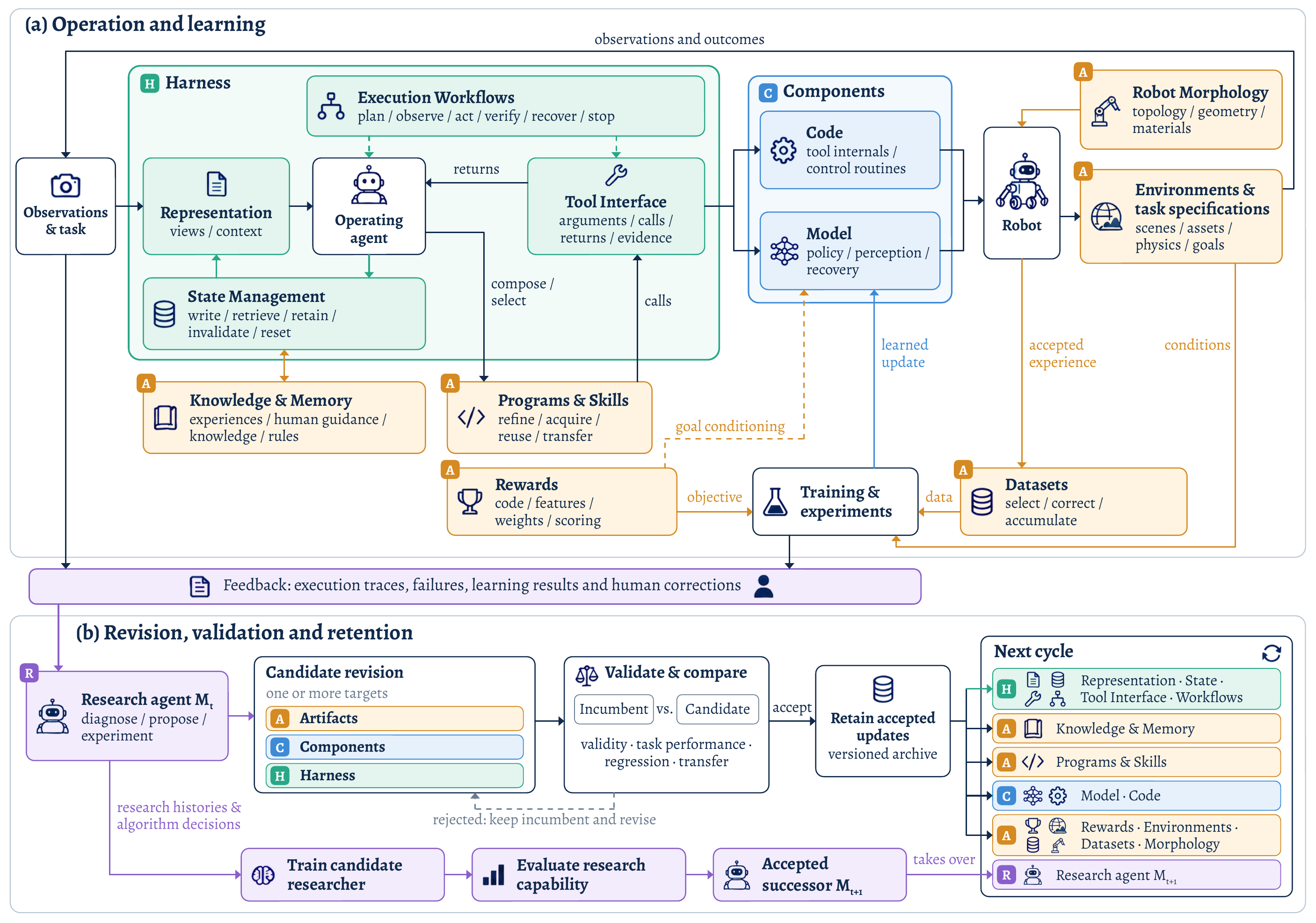}
\caption{Evolver updates across successive operating cycles. (a) The operating agent acts through a harness of representation, state management, tool interface and execution workflows, using artifacts and components during operation and learning. (b) Feedback guides candidate revisions, which are compared with the incumbent; accepted versions are retained and re-enter their modules in the next cycle. A successor evaluated for research capability can take over subsequent research. Badges mark update targets: A, artifacts; C, components; H, harness; R, research agent.}
\label{fig:self-improver-tree}
\end{figure*}

\subsection{Programs and Skills}
\label{sec:si-artifact}
\begingroup In A3 code synthesis (Section~\ref{sec:policy-a3}), an Actor writes control logic and may debug it repeatedly to complete the current task. The agent also acts as an Evolver when it uses execution feedback to select a repair, validates the revised program or skill, and retains it for later runs~\citep{self260700272,huang2026rax}. A retained routine can then be adapted or composed with other skills on new tasks~\citep{capx,rats2026}. We examine how agents refine programs, admit skills to a library, and test whether those skills improve subsequent performance.\parfillskip=0pt plus 0.1\linewidth\par\endgroup



\subsubsection{Program Refinement}

Execution feedback helps an agent locate faults in a task program and revise its control logic, parameters, or action sequence. Constructing this feedback requires both execution evidence and criteria for judging task achievement. ASPIRE exposes primitive-level multimodal traces, including API return statuses and visual observations, for the agent to inspect execution and diagnose failures~\citep{self260700272}. CaP-X supplements execution logs with a visual differencing module that describes scene changes and estimates task completion~\citep{capx}. RAX uses benchmark evaluators to judge task completion during simulation exploration, while its BEHAVIOR-1K setting additionally employs a separate review model to report unmet goals and likely failure causes~\citep{huang2026rax}. RAPID repairs demonstration-derived primitives and their composition through simulation checks~\citep{rapid2026}, whereas SetItUp uses example-consistency checks to guide regeneration of \mbox{reusable arrangement functions~\citep{xu2024set}.}

Such feedback supports targeted program edits and broader searches over alternative implementations.
Parameter-level refinement adjusts exposed controller settings, such as the scaling of an existing controller's action outputs~\citep{benamor2025sasprompt}.
Editing controller code permits changes to motion phases and recovery logic, including adding grasp retries after failed executions~\citep{kumar2026aor}.
Structured representations expose task stages and execution steps for modification: typed task programs support revisions to stage objectives and completion conditions~\citep{wang2026sunagenticrobotpolicy}, while action-flow methods revise primitive sequences, action parameters, and navigation waypoints~\citep{liu2026scene2demo}.
The scope of an edit can also extend beyond a single routine.
RHO uses tool-enabled coding agents to inspect, test, and modify multiple files within a candidate policy repository~\citep{rho2026}.
MEMENTO combines sequential local revisions with larger mutations and crossover, allowing targeted changes and broader program variations to be explored within the same search process~\citep{sygkounas2026memento}.

The refinement loop can also adapt what evidence it collects to evaluate a program and guide the next revision.
When rollout feedback leaves task outcomes or failure causes unresolved, agents can generate diagnostic procedures or select experiments that test alternative explanations.
Fly-by-Code synthesizes attempt-specific diagnostic programs after task execution, acquiring informative views to check unobserved outcomes and conducting targeted physical probes in simulation to investigate failure causes.
The resulting evidence informs completion assessment and subsequent task-program revision~\citep{lee2026flybycodeembodiedcodingagents}.
Experiment selection can likewise be conditioned on the changes under consideration.
EmbodiedRSI maintains hypotheses about code and skill revisions and selects experiments by their expected reduction in hypothesis uncertainty relative to execution cost.
Matched trials compare code-only, skill-only, and joint changes; their outcomes update support for the hypotheses and guide the next experiments and revisions~\citep{song2026embodiedrsiactivecontinualrobot}.

\subsubsection{Skill Acquisition}
\begingroup Skill acquisition expands an agent's callable repertoire with reusable behaviors, which may be abstracted from programs or learned from demonstrations and interaction. It begins with deciding which behaviors to add. HALP lets a language planner request new manipulation skills, which are learned from human demonstrations~\citep{parakh2023halp}. Agentic Skill Discovery proposes tasks from scene and robot descriptions, whereas RATs selects learnable practice goals through self-directed play~\citep{zhao2024agenticskilldiscovery,rats2026}.\parfillskip=0pt plus 0.1\linewidth\par\endgroup

\begingroup The next step converts experience into callable implementations. CaP-X compiles shared helper functions from task programs in a single development-time pass~\citep{capx}. RoboCoder retains corrected action programs, Lifelong Robot Library Learning abstracts recent code into library skills through wake--sleep cycles, and SkillComposer compresses recurring successful call sequences into macros~\citep{li2024robocoder,self240618746,woods2026skillcomposer}. Agentic Skill Discovery instead has the agent specify tasks and rewards, then trains the policies with reinforcement learning~\citep{zhao2024agenticskilldiscovery}. These methods yield different callable artifacts: code routines, macros, and learned policies. Their policy-training and reward-design mechanisms are separate update targets.\parfillskip=0pt plus 0.1\linewidth\par\endgroup

\begingroup Retention governs which candidate behaviors are accepted into a reusable skill library. RATs verifies goal- and step-level progress before saving code skills, and Agentic Skill Discovery screens learned behaviors with an independent vision-language model~\citep{rats2026,zhao2024agenticskilldiscovery}. HALP relies on human task-outcome checks~\citep{parakh2023halp}. RAPID admits primitives after verification across specified effect--environment pairs, allowing later demonstration segments to reuse them~\citep{rapid2026}. Zetta retains validated critic--recovery programs around a frozen policy, while Recova develops recovery programs in digital twins and trains recovery policies for deployment, requesting human demonstrations for unsupported failures~\citep{zetta2026,recova2026}. ASPIRE also stores validated repair knowledge as retrievable guidance for later synthesis~\citep{self260700272,nvlabs2026aspirecode}. Callable skills and synthesis guidance are distinct artifacts. Project reports also describe persistent or shared code skills~\citep{physicalrsi2026,noematrix2026roborsi,skillclaw2026,waddle2026introducing}. Admission establishes acceptance under the tested conditions; cross-task reuse and cumulative improvement require separate evidence, discussed below.\parfillskip=0pt plus 0.1\linewidth\par\endgroup

\subsubsection{Skill Reuse and Transfer}
\begingroup Once a skill library has been constructed, the next problem is ensuring that it is reusable and transferable. ASPIRE retains revised programs for LIBERO/Robosuite, whereas its BEHAVIOR configuration generates policies for individual evaluation seeds from frozen repair guidance~\citep{self260700272,nvlabs2026aspirecode}. HarnessPAI separates maintained code from repair notes, while Maestro uses prior programs and execution analyses as generation context~\citep{si260929166,shi2025maestro}. EmbodiedEvo separates executable Skills from diagnostic Lessons and evaluates their frozen inheritance on unseen layouts of the same task, while local programs may still adapt~\citep{li2026embodiedevo}. ASENA and RoboSkill retain executable routines alongside textual records for later use, while RoboFoundry distinguishes persistent composite skills from trial-local corrections~\citep{cheng2026asena,roboskill2026,liang2026robofoundry}.\parfillskip=0pt plus 0.1\linewidth\par\endgroup

\begingroup Archived source programs can guide development on new tasks~\citep{embodiedswe2026,kuroki2026closedloop}, while Memory Transfer Planning adapts examples from a statically constructed memory~\citep{xref250924160}. WetRobo reports helper reuse across tasks with further laboratory-specific adaptation~\citep{oikawa2026wetrobo}. Complementing these update mechanisms, ManiSkillFormer connects reusable motion templates to scene geometry through task-conditioned geometric contracts. Agents fill human-defined schemas while accounting for subsequent skill requirements; runtime perception grounds the requested geometry~\citep{yu2026maniskillformer}. This supports reuse and composition, but persistent feedback-driven skill revision is not evaluated.\parfillskip=0pt plus 0.1\linewidth\par\endgroup

\subsubsection{Improvement Validation}
\label{sec:si-artifact-validation}
Acceptance must distinguish executable code from improved behavior. SkillComposer checks intent and program structure before execution, without establishing task success~\citep{woods2026skillcomposer}. RPG compares candidates with the incumbent on paired trials across 22 tasks: mean success must improve, with at most one additional failure per five trials on each task. Merged revisions are retested; failed integration retains the incumbent~\citep{rpg2026}. The two checks assess candidate eligibility and integrated-system improvement separately.

Generalization requires trials not used for revision selection. BiGym~2.0 evaluates frozen, independently developed task programs on hidden same-distribution initializations, not cross-task inheritance~\citep{zhang2026bigym2}. SEAR compares inherited code-and-memory libraries with empty-library runs and time-matched restarts~\citep{wang2026sear}. Matching current-task development time does not equalize prior library-building cost or isolate skill-code benefits.

\begingroup Retention includes regression checks on affected capabilities. ETM combines default-off extensions, compatibility checks on test fixtures, and old-task retests~\citep{hu2026coap}; these establish evidence only within the tested conditions. LYRA incorporates human acceptance and re-evaluation on previous tasks~\citep{meng2025lyra}. Tests must also exercise the modified behavior: in the 123-round case study, upstream failures often prevented later skills from being tested, while accepted repairs did not solve the target task~\citep{si260931760}.\parfillskip=0pt plus 0.1\linewidth\par\endgroup

\input{tables/si-programs-comparison}

\subsection{Knowledge and Memory}\label{sec:si-memory}



\subsubsection{Scope and Retained Knowledge}
\label{sec:si-memory-scope}

Knowledge and Memory concerns feedback-informed revisions to reusable lessons, procedural guidance, applicability conditions, and symbolic or interaction-grounded knowledge. Following Section~\ref{sec:scope}, inclusion requires a documented connection between the agent's decisions, the revised contents, and their subsequent use. The comparison therefore follows how experience is interpreted, which knowledge is retained or corrected, and how that knowledge informs a later decision. Robot execution, human corrections, and observations of human activity provide different evidence sources \mbox{and should remain distinguishable.}

The retained object's function determines its relation to neighboring categories. Programs and Skills (Section~\ref{sec:si-artifact}) concerns executable behaviors; knowledge can instead guide the generation or selection of a new behavior. Component Optimization (Section~\ref{sec:si-model}) concerns the implementation of a functional model or code component. Revisions to the mechanisms organizing the agent's knowledge inputs, managing persistence, or retrieving stored contents concern Harness Improves Harness (Section~\ref{sec:si-harness}). Knowledge contents can evolve under fixed retrieval and storage procedures; changes to those procedures belong to harness improvement.

\subsubsection{Experience-Grounded Memory}
\label{sec:si-memory-experience}

Experience-grounded memories differ in what they retain, how the retained evidence changes a later decision, and where it remains reusable. The following examples distinguish scene-specific navigation priors, interaction-derived world properties, policy-competence records, and transferable diagnostic guidance. Updating these contents does not by itself demonstrate a change to the mechanisms governing retention or retrieval (Section~\ref{sec:si-harness-state-management}).

ME-Brain~\citep{si260924271} separates the record of an attempt from the experience drawn from it. The Cognitive Core interprets action outcomes, diagnoses failures, and stores state changes and subtask results alongside their visual and action evidence. Task events are then consolidated into longer-term success conditions and recurring failure patterns. Later planning retrieves related events through task relationships and reusable experience through semantic similarity; action execution can also recover finer-grained historical states. The proposed hierarchy therefore connects individual outcomes to guidance across tasks. Its reported evaluations assess the integrated system, without separately measuring the contribution of long-term experience abstraction.

\begingroup EvoMemNav~\citep{si260603509} retains raw views and goal-conditioned navigation statistics on a shared graph. Visited regions, revisits, dead ends, and STOP/RESELECT judgments update exploration-utility and stopping-reliability priors. Similar-goal retrieval refines candidate ordering and supplies verification hints; decay and conflict suppression limit unreliable history. Episode-local records reset, whereas scene-level priors persist across subtasks and episodes within that environment. On the GOAT-Bench VAL-UNSEEN ablation subset (episode index 1 in each of 36 scenes), adding reflection-driven continual memory adaptation (RDCMA) to the same graph and coarse-to-fine policy raises overall success rate (SR) from 60.8\% to 64.8\% and success weighted by inverse path length (SPL) from 40.8\% to 43.1\%. Perception modules and the VLM backbone are shared across conditions. Success requires stopping within 1\,m of the goal; SPL additionally weights path efficiency. This simulation result tests RDCMA jointly, without separating Episode-STM from Scene-LTM; the subset is smaller than the full benchmark, and scene-specific persistence does not establish cross-scene memory transfer.\parfillskip=0pt plus 0.1\linewidth\par\endgroup


\begingroup MessyMem~\citep{banwasi2026messymem} retains interaction-derived properties and failure explanations in a spatial scene graph linked to keyframes. Action evidence distinguishes a missed grasp from a locked fixture; new observations correct stale properties. Later planning retrieves graph evidence and visually reranked keyframes to select a primitive and target, reusing knowledge across successive tasks within the environment. In the continuous 25-task, over-three-hour simulation without memory reset, mean task progress is 80.0\% (95\% CI: 77.4--82.6), versus 65.2\% without keyframes (SG+IA) and 47.9\% without interaction-analyzer updates (SG+KF), over 50 trials per method. Each trial starts with a fixture-observation trajectory that leaves contents hidden; tasks then follow completion or failure of the preceding task. All methods share the planner, perception stack, and action primitives. Task progress measures fulfillment of task-specific criteria rather than whole-task success. These controls support complementary structured and visual memory in this sequential setting; they do not establish transfer between independent environments or sessions. Physical demonstrations use operator-gated primitives, and detection categories are predefined.\parfillskip=0pt plus 0.1\linewidth\par\endgroup

RoboRouter~\citep{chen2026roborouter} instead distills experience into evidence about policy competence. An evaluator summarizes successful executions and specific failure behaviors from rollout frames and task-specific measurements. A recorder saves these summaries in a multimodally indexed database and revises itemized routing notes, merging duplicate entries. For a later task, similarity search and reranking retrieve comparable executions, helping the router choose the policy most likely to succeed under the current conditions. Online outcomes refresh this evidence as tasks continue. The ablations examine retrieval, evaluation, and updating, but do not remove all memory in a single control; optional researcher corrections are an additional feedback source.

\begingroup Teach-and-Grow~\citep{nie2026teachandgrow} separates reusable Skill Blocks from experience records of task context, outcome, diagnosis, repair, and supporting evidence. Records guide later composition, grasp selection, and outcome checks across tasks; humans can inspect and correct overgeneralized rules or incompatible tool assumptions. This makes diagnostic guidance available beyond the originating attempt, but the controlled save/reload and related-task studies on LIBERO evaluate Skill Block persistence and library expansion. LIBERO-Plus tests broader robustness. Neither establishes an isolated Experience-Memory retrieval effect, and Skill Block validation should not be treated as a separate quantitative admission gate for experience records.\parfillskip=0pt plus 0.1\linewidth\par\endgroup

\begingroup Reuse can extend beyond the original scene, but its scope depends on the protocol. EmbodiedEvo~\citep{li2026embodiedevo} distills operating conditions, failure modes, and recovery advice into Lessons alongside executable Skills, then freezes both for fresh agents in unseen scenes. Matched tests additionally examine cross-model inheritance and transfer from related or mismatched source tasks to twelve target tasks, without target-task exploration. These results concern the combined Skills/Lessons package; even the matched raw-history comparison does not isolate diagnostic Lessons. Agentic RSR~\citep{li2026agenticrsr} instead transfers a task policy and summaries of effective actions, failure causes, and perception difficulties from simulation to a fresh robot-side conversation. Current observations must re-ground object locations, and physical outcomes may update the memory. Its experiments assess the coupled policy-and-memory handoff on the reconstructed tasks, without establishing a separate memory effect or reuse on unrelated tasks.\parfillskip=0pt plus 0.1\linewidth\par\endgroup

Robo-COP~\citep{zhang2026robocop} shows how experience about a policy can outlive the conditions that produced it. Episode reflection turns execution logs and outcome judgments into notes on successful strategies and known failures, which guide later attempts at the task. Advice to use scripted motion after repeated VLA grasp failures may become outdated after fine-tuning. Following candidate adoption or reversion, the system marks such advice as unverified for the current checkpoint, preserving the earlier evidence while allowing the agent to reconsider it. The notes persist across attempts at the same task; held-out evaluation freezes both policy and memory on new initializations. Frozen-policy and fixed-schedule controls also adapt memory, so the comparisons do not isolate checkpoint reconciliation or establish cross-task memory transfer.

DexAgent~\citep{wang2026dexagent} also reuses prior work, but saves parameterized Python skills and verifiers for later human-video conversion. Its persistent artifacts implement reconstruction and trajectory-generation procedures. They complement the experience summaries discussed here and are covered under Programs and Skills and Component Optimization (Sections~\ref{sec:si-artifact} and~\ref{sec:si-model}).

\subsubsection{Human Feedback and Structured Knowledge}
\label{sec:si-memory-human}

MEMORA~\citep{yu2026memora} forms persistent knowledge from egocentric videos of human activity. It revises object records as new observations arrive and consolidates repeated activity into routines and preferences linked to supporting episodes. Later planning retrieves this knowledge together with object states and locations. Its physical demonstration uses a fixed rule-based controller in the same scene, providing a qualitative check of memory-grounded planning rather than a controlled robot benchmark.

Human corrective feedback can be transformed into persistent knowledge that guides robot behavior beyond individual interactions. Language-based approaches, such as DROC~\citep{zha2023droc} and Zero2Skill~\citep{wang2026zero2skill}, retain corrections as reusable guidance but differ in how they organize and apply this knowledge. DROC addresses corrections at both short- and long-term timescales. In the short term, it selects relevant interaction history to interpret human feedback and revise the current plan or skill. For long-term reuse, an LLM-based knowledge extractor distills completed interactions into a task-indexed knowledge base containing task constraints, user preferences, skill parameters, revised plans or code, and updated object states. When encountering new tasks, a knowledge retriever first identifies semantically related past tasks and further uses DINOv2/CLIP-based visual matching to select object-specific knowledge. The retrieved information is then incorporated into the task planner or skill composer to guide subsequent decisions. In contrast, Zero2Skill emphasizes the explicit applicability of retained corrections during autonomous data collection. When repeated collection or reset failures trigger human intervention, an LLM parser distinguishes temporary adjustments from corrections intended for future use. Persistent corrections are stored as structured Markdown rules specifying their triggers, required adjustments, applicability scopes, and original human utterances. These rules are consulted in subsequent collection rounds to modify execution strategies or verification criteria when the relevant conditions are met. Thus, while DROC emphasizes retrieving relevant past knowledge based on task semantics and object similarity, Zero2Skill retains explicit conditions governing when corrective guidance should be reapplied.

Human corrections can also be retained as procedural, executable, or symbolic knowledge, enabling different forms of reuse. ARCHITECT~\citep{chen2026architect} combines human language corrections with robot execution traces to revise generated policy code, while distilling reusable insights into a persistent skill library containing natural-language rules, usage guidelines, and code patterns. The accumulated guidance is incorporated into the coding agent's context during subsequent program synthesis, allowing earlier corrections to inform new robot programs without model fine-tuning. LYRA~\citep{meng2025lyra}, by comparison, incorporates human-guided corrections and refinements directly into parameterized executable skills. It retains both skill implementations and successful task–code examples in external memory, indexed by skill descriptions and task instructions, respectively. For new tasks, retrieval-augmented generation selects relevant skill interfaces and examples from two vector databases, while natural-language hints from users can help identify useful skills overlooked by similarity-based retrieval. Whereas ARCHITECT primarily preserves guidance for generating new programs, LYRA also retains executable skills that can be directly reused and composed. InterPreT~\citep{han2024interpret} adopts a complementary symbolic approach by transforming human feedback about task goals and action feasibility into reusable predicates and action constraints. An LLM-based Reasoner identifies task-relevant predicates and extracts supervision labels from language feedback, while executable predicate functions ground these concepts in robot observations. Combining feedback-derived constraints with robot interaction data, a fixed learning procedure further infers symbolic action preconditions and effects. The resulting predicates and operators form a reusable PDDL domain, allowing future planning to explicitly evaluate world states and action applicability rather than retrieve previous language corrections. Together, these methods illustrate how different representations of human corrective knowledge—from language-based rules and procedural guidance to executable skills and symbolic models—determine how acquired experience is retained and reused in subsequent robot decisions.

\subsubsection{Knowledge Validation and Maintenance}
\label{sec:si-memory-validation}

KnowBody~\citep{lou2026knowbody} connects agent-selected diagnostic actions to validated body knowledge and conditional action--effect rules. When observations leave competing explanations, the VLM selects an observation or diagnostic motion with an expected discriminating effect. Qualified interaction measurements support candidate body estimates or rules; fixed fitting and independent validation govern publication, and subsequent episodes use the accepted versions. The retained body estimates and action--effect rules place this configuration under Knowledge and Memory. When body estimates change, dependent rules are re-evaluated before reuse. The reported reduction in reasoning rounds is conditioned on successful completions and does not by itself measure improvement across all attempts.

EmbodiedRSI~\citep{song2026embodiedrsiactivecontinualrobot} also retains evidence at three levels: execution records, code--skill hypotheses, and reusable knowledge abstracted across experiments. Later experiments retrieve hypotheses and distilled knowledge, establishing reuse beyond the current trajectory. A separate memory manager is trained with PPO to select retention, merging, abstraction, compression, forgetting, or skipping according to subsequent task improvement, uncertainty reduction, and interaction cost. The revised knowledge contents and the learned memory-selection policy are distinct update targets, with the robot foundation model held fixed. The reported harness evolves in simulation, and its learned memory policy is then evaluated through physical transfer.

These cases expose two different maintenance questions: whether a knowledge item is supported under its current conditions, and how the system chooses which evidence to retain or transform. KnowBody revises and validates contents under a supplied procedure; EmbodiedRSI additionally learns a policy over memory actions. Comparing them requires identifying the retained contents and the status of the surrounding mechanisms separately.

\input{tables/si-memory-comparison}

\subsubsection{Evidence of Persistent Improvement}
\label{sec:si-memory-evidence}

Evaluation should first establish what survives a task boundary and where it is used again. Reuse within a scene, transfer to a new task, and transfer across scenes or embodiments provide different evidence and should be reported separately. To attribute a gain to retained knowledge, useful controls include clearing memory while preserving the executor and interaction budget, comparing raw histories with distilled contents, and varying retrieval while holding the knowledge base fixed. Such controls distinguish the effect of useful knowledge from additional attempts, longer context, or changes to the underlying controller.

\begingroup Success rate alone also leaves the cost and reliability of retention unresolved. Later-task evaluation should account for failed acquisition attempts, human corrections, robot interactions, model calls, and retrieval overhead, alongside preservation of earlier capabilities. ME-Brain's integrated physical results and MEMORA's qualitative robot use~\citep{yu2026memora} establish different evidence from a controlled test of cross-task memory benefit. KnowBody's successful-completion reasoning curves likewise answer a narrower question than all-attempt completion rates. These distinctions connect the mechanisms reviewed here to the challenge of retaining knowledge without accumulating errors (Section~\ref{sec:challenges}): persistent revisions require both traceable support and evaluation of their later consequences.\parfillskip=0pt plus 0.1\linewidth\par\endgroup

\subsection{Rewards}
\label{sec:si-rewards}

\begingroup Reward evolution changes the objective used to learn or realize robot behavior: reward code, features, weights, activation conditions, or a reward model's scoring instructions. The change must respond to feedback and persist for subsequent learning or execution. Whereas Section~\ref{sec:infra-training} examines the training workflow, here we compare the feedback guiding reward revisions, the permitted edits, and the criteria for retaining them. The resulting objective remains distinct from the policy trained under it.\parfillskip=0pt plus 0.1\linewidth\par\endgroup

\subsubsection{Performance-Guided Search over Reward Code}

Policy training reveals consequences of a reward that its code alone cannot expose. Self-Refined LLM~\citep{selfrefinedreward2023} returns training statistics and task-performance feedback, including signal-temporal-logic success tests, for reward revision. Eureka~\citep{ma2023eureka} similarly alternates LLM proposals and training, using task fitness for selection and reward-component statistics for reflection. ROSKA~\citep{si241213492} additionally carries forward the learned policy, mixing inherited parameters with random initialization and expanding the reward population when progress stalls. Its changed learner initialization means that gains cannot be attributed to better reward proposals alone.

DrEureka~\citep{ma2024dreureka} conditions reward search on safety instructions; its later domain-randomization design is a separate intervention. STRIDE~\citep{xref250204692} uses training feedback to revise terms and coefficients for simulated humanoid locomotion. Both retain revised objectives for conventional RL backends to optimize.

The provenance of the task need not determine the feedback used for reward revision. Video2Policy~\citep{ye2025video2policy} derives manipulation tasks from internet videos, but its reward-reflection loop uses training and evaluation logs to generate new functions from the best previous candidate. Its final evaluation uses human-written ground-truth success functions withheld from training and manual assessment of the learned behaviors. It thus illustrates video-grounded task construction followed by log-driven reward refinement, rather than direct \mbox{visual diagnosis of failed rollouts.}

\input{tables/si-rewards-comparison}

\subsubsection{Preferences and Behavioral Diagnosis}

\begingroup Human feedback can specify what a performance score misses. In Text2Reward's interactive configurations~\citep{text2reward2023}, users inspect learned behavior and request reward corrections before retraining. In AURA~\citep{auracurriculum2025}, reports of jerky motion lead to penalties on action and velocity changes. AURA separates these between-iteration human corrections from automated checks within an iteration.\parfillskip=0pt plus 0.1\linewidth\par\endgroup

Preferences can also govern candidate selection. In REvolve's human-assisted settings~\citep{revolve2024}, people compare videos and supply feedback for an LLM to mutate or recombine reward components; a population database retains reward--policy pairs and their feedback. Eureka also supports human-guided revisions~\citep{ma2023eureka}. CoRe~\citep{core2026} instead selects LLM-revised reward code by consistency with VLM-derived trajectory preferences. This formal-reward search is distinct from fitting its residual reward model to preference supervision.

Diagnosis identifies a concrete edit target. AgenticRL~\citep{si260603963} combines UAV failure-stage, geometric, and safety measurements with visual context; a stalled racing policy motivates progress and anti-stalling terms. HARBOR~\citep{li2026harbor} uses diagnostics and rollout videos to revise reward code, retaining both code and iteration history. Its stage-based stacking rewards combine diagnostic revisions with supplied human heuristics.

FATE~\citep{wei2026fate} embeds reward repair in task construction: failed execution traces guide revisions to SAC manipulation rewards, retained in further attempts and the accepted task. Its scene-rescaling and planning-constraint edits are separate changes; combined repair results do not isolate the reward contribution.

\subsubsection{Visual Failure Diagnosis and Motion Discrepancies}

\begingroup Visual feedback can reveal behavior obscured by training scores. AHA~\citep{aha2024} supplies failure explanations to Eureka's reward reflection. CRAFT~\citep{choi2025craft} combines visual analysis with component curves to revise terms and weights within a curriculum stage. Its bounded refinement can advance with the best available policy, so stage progression does not certify successful reward repair.\parfillskip=0pt plus 0.1\linewidth\par\endgroup

Visual diagnosis supports different reward representations. Reflective Self-Adaptation~\citep{li2025reflectiveadaptation} analyzes failure videos, selects parameterized reward components, and combines them through AND, IF, and OR relationships. Trajectory-state logs ground weights and thresholds in a structured configuration reused by subsequent PPO updates. Its examples successively add obstacle avoidance and orientation maintenance as new failures emerge. RDA~\citep{lee2026rda} instead produces subtask-level completion scores and explanations from rollout videos, then revises reward code and the subtask descriptions guiding it. The selected reward, policy, and analysis are checkpointed for further search. Its package-pushing example exposes a failure that task completion alone misses: moving the object with the torso instead of the intended bilateral hand contact. These methods therefore differ both in the representation being revised and in how diagnostic evidence \mbox{identifies deficiencies in the objective.}

\begingroup Other visual evidence describes progress or motion discrepancies. RE-GoT~\citep{regot2025} interprets rollouts under a task graph to guide reward edits. SDS~\citep{sds2024} compares quadruped behavior with a demonstration, returning pose grids, foot-contact plots, and training statistics alongside the selected reward for the next proposal.\parfillskip=0pt plus 0.1\linewidth\par\endgroup

Video2Reward~\citep{video2reward2024} provides a complementary, measured feedback channel. It converts target-video keypoint trajectories into textual input for reward generation, then compares projected simulated trajectories with the target using FastDTW. Per-joint discrepancies, training-performance feedback, and the selected reward guide the next round of reward-code generation and policy learning. Here the discrepancy computation is fixed, while the LLM revises the objective in response. This separates motion matching from VLM-authored causal explanations: both can inform retained reward changes, but they measure different aspects of behavior and \mbox{support different improvement claims.}

\subsubsection{Constrained Calibration and Coupled Design}

Reward revision need not involve unrestricted code generation. ELEMENTAL~\citep{elemental2025} uses differences between demonstrated and learned feature counts to prompt VLM feature revisions, then refits weights through inverse reinforcement learning; expert-selected keyframes initialize its manipulation specification. Self-Alignment~\citep{zeng2024selfalignment} instead responds when consistent trajectory rankings accompany low task success: the LLM identifies parameters and directions that constrain numerical search. The agent thus revises either the reward representation or the calibration search space.

SUN~\citep{wang2026sunagenticrobotpolicy} keeps accepted task semantics fixed and adjusts reward weights using a deterministic behavioral-cloning-prior rollout. This feedback pathway differs from initial weight selection and residual-action-authority calibration. RoboMoRe~\citep{si250600276} instead alternates revisions to morphology--reward pairs. Debate2Create~\citep{qiu2026debate2create} adds control-oriented and criterion-specific critiques to co-design, while simulator scores, not judge preferences, rank retained pairs. Their body/reward ablations help distinguish reward contributions from joint-design gains (Section~\ref{sec:si-morphology}).

Under a human-set research agenda, Khandelwal and Gupta's coding agent adapts existing quadruped reward terms and revises weights after excessive penalties collapse learning~\citep{khandelwal2026agentrl}. Larger weighted returns can nevertheless accompany worse velocity tracking. This illustrates why behavioral comparisons must use criteria that remain stable while the reward changes.

\subsubsection{Acceptance, Retention, and Subsequent Use}

Acceptance determines which objectives survive. Agentic Skill Discovery~\citep{zhao2024agenticskilldiscovery} holds a sampled success function fixed while evolving a skill's reward, then uses a separate VLM to inspect executions marked successful. This can reject behavior that satisfies the generated predicate for the wrong reason; accepted policies and rewards remain paired as reusable skill options.

\begingroup Trajectory checks can reject rewards before costly policy training. Code as Reward~\citep{venuto2024codeasreward} corrects visual subtask programs from false positives on expert and random trajectories; humans assemble the programs into the training reward. CARD~\citep{card2024} instead revises code when average per-step return fails to rank successful trajectories above unsuccessful ones. EvoNav~\citep{evonav2026} screens navigation rewards through prescribed analytical rankings, lightweight policy trials, and full training. The shared design reserves costly learning for objectives that pass cheaper checks.\parfillskip=0pt plus 0.1\linewidth\par\endgroup

Demo2Reward~\citep{demo2reward2026} revises scoring instructions rather than reward code. A meta-VLM uses demonstrations and misclassified examples to propose replacements, selected against human success labels with an emphasis on rejecting false positives. The chosen instruction remains fixed during subsequent policy learning. Unlike CoRe's formal-reward search, the retained change concerns how a visual model interprets success.

Retained rewards need not require policy retraining. InterEvolve~\citep{audit261002196} searches in simulation with a frozen forward--backward controller: an LLM revises reward terms and stage structure, while CMA-ES tunes constants. Repeated fixed-verifier tests govern replacement, and accepted programs enter a reusable skill library. Physical deployment replays the resulting latent prompts rather than searching online. The retained program combines objectives with stage ordering, extending reward evolution to what an existing controller is asked to accomplish.

Across these mechanisms, evaluation on new behavior tests the reliability of a retained objective. Comparisons should preserve reward versions and use independent task criteria, matching starting policies and development budgets when isolating reward changes. That budget includes rejected candidates, rollouts, and human feedback. Section~\ref{sec:challenges} develops evaluation criteria for improvements to behavior and its development process.

\subsection{Environments and Datasets}
\label{sec:si-environments-datasets}

Environments and datasets determine which situations a robot encounters and which experience becomes supervision. Their evolution builds on construction (Sections~\ref{sec:infra-environment} and~\ref{sec:infra-dataset}) through feedback-informed revisions retained for later use. We compare changes to simulation environments, task specifications, and training examples, distinguishing these artifacts from the behavior programs and learned components that use them.

\subsubsection{Feedback-Informed Refinement of Environment Artifacts}
\begingroup Feedback-informed refinement uses execution, collection, or training feedback to revise scenes, assets, or simulator configurations for more reliable simulation, data collection, and policy training. Each purpose requires different evidence: simulations should reproduce task-relevant behavior, collection environments should support feasible demonstrations, and training environments should improve learning. Revised artifacts must also be retained for reuse beyond the local repair.\parfillskip=0pt plus 0.1\linewidth\par\endgroup

\begingroup For simulation, refinement adds credibility when physical behavior is tested through interaction, beyond appearance or code validity. DiagGen revises geometry and materials using agent-selected grasp probes, then exports accepted assets for later simulation~\citep{diaggen2026}. CoDimRecon checks motion, geometry, and numerical settings before revising a deformable asset's material model; in a paper-folding example, it adds plastic bending after observing springback, without claiming to recover measured properties~\citep{xie2026codimrecon}. SimEX compares real and simulated traces under fixed-policy replay, revises the simulator with a coding agent, and retains corrections for screening toolbox repairs~\citep{simex2026}. These tests support specific physical behaviors or diagnostic reuse, not general fidelity or training-data benefits.\parfillskip=0pt plus 0.1\linewidth\par\endgroup

\begingroup For data collection, refinement improves usability when it resolves barriers exposed by execution feedback. ARSTAG routes grasp-generation failures to a reconstruction agent, which, in a nut-placement example, follows prescribed recovery guidance to adjust scale, orientation, mass, and support contacts before collection resumes~\citep{arstag2026}. FATE resizes assets after execution failures alongside changing planning constraints or rewards; only the resizing revises the environment~\citep{wei2026fate}. EmbodiedSmith also replaces and repositions objects before revalidation and replanning, although its fixed-scene benchmark does not isolate their benefit~\citep{qin2026embodiedsmithscalingembodieddata}. Such repairs support collection feasibility, not necessarily physical fidelity: an enlarged object may become graspable without matching its real dimensions. Nor does completed collection alone show policy-learning gains.\parfillskip=0pt plus 0.1\linewidth\par\endgroup

For training, the test is whether feedback-revised environments improve policies under evaluation separate from revision. Eurekaverse revises terrain-generating programs using terrain statistics and policy-training results, retaining environment libraries across iterations~\citep{liang2024eurekaverse}. The LLM edits only environment specifications; policy updates, selection, and within-terrain difficulty progression follow predefined procedures. On held-out parkour tasks, variants lacking policy feedback or using only initial or final environments underperform the full curriculum~\citep{liang2024eurekaverse}. This supports learning utility under the tested protocol, not reconstruction accuracy. Across these purposes, a retained feedback-driven edit is an improvement only when purpose-specific evidence supports its later use.

\subsubsection{Task-Specification Refinement}
Feedback can expose defects in a task specification without changing scene geometry. Self-CriTeach revises generated Planning Domain Definition Language (PDDL) domains from planner errors, correcting predicates and action schemas before using the validated domain to generate problem--plan training pairs~\citep{si250921543}. The retained domain defines state transitions for producing further supervision, rather than prescribing a robot's control sequence. Planner validation checks consistency on sampled symbolic problems but does not establish physical executability. This complements Eurekaverse's changes to training environments by revising how learning problems are specified.

BenchMend instead repairs benchmark instructions after human review of disagreements between an agent's assessment and the native success checker~\citep{zhang2026benchmend}. An instruction-refining agent adds missing goal information; revised instructions are retained and affected tasks or episodes are rerun without changing demonstrations or the checker. Whereas Self-CriTeach's domain supports training-data generation, BenchMend changes the specification presented at evaluation. Higher scores after clarification therefore need not indicate improved policy capability.

\input{tables/si-environments-comparison}

\subsubsection{Failure-Driven Dataset Revision}
Failures can also identify missing supervision while task definitions remain fixed. Recova accumulates separate task and recovery datasets across rounds~\citep{recova2026}. Successful task executions and human task demonstrations enter the task dataset; human recovery demonstrations and autonomous recoveries passing a scene-restoration check enter the recovery dataset. Failed policy segments remain available for analysis but are excluded from training. Subsequent fine-tuning consumes these accumulated datasets. The agent's recovery choices and outcome checks thus determine which experience becomes supervision, while people still supply demonstrations for unsupported failures.

\begingroup Skill-Space Shooting makes segment-level admission explicit~\citep{si260938178}. A fixed detector triggers intervention, and a VLM selects and verifies a corrective skill. Only accepted skill-executed repair segments from rollouts judged complete enter the accumulated training set; learner actions and rejected trials are excluded. The next behavioral-cloning round combines these repairs with the initial demonstrations. Where Recova routes accepted experience to distinct policy datasets, this method concentrates the task-policy update on skill-executed corrections. Initial demonstrations and scene resets still require people.\parfillskip=0pt plus 0.1\linewidth\par\endgroup

Across these cases, the retained revision changes either the problem presented to the robot or the experience used to improve it. Comparisons should track what survives, its acceptance conditions, and its subsequent use. Independent task criteria and development-cost accounting are needed to distinguish useful revisions from more data, easier specifications, or additional human effort.

\subsection{Robot Morphology}
\label{sec:si-morphology}

\begingroup Morphology refinement changes the robot-body description that determines which behaviors a controller can realize. The retained artifact may specify a modular topology, geometric dimensions, or the placement of passive and actuated material. Its quality is assessed through the behavior of a controller trained or optimized for that body. Consequently, the observed fitness depends jointly on the proposed morphology, the controller-training procedure, and the evaluation environment. The following methods differ in what the agent can change and what information survives each design evaluation.\parfillskip=0pt plus 0.1\linewidth\par\endgroup

RoboMorph~\citep{si240708626} searches modular robot structures expressed in a grammar. The LLM proposes complete designs and rationales; SAC-trained controllers supply locomotion scores for elite selection. Selected designs and rationales become examples for later proposals, so feedback reaches the generator through these exemplars, with numerical scores omitted from its context. Search uses a shared reward and controller-training recipe across bodies on four terrains, with learning curves reported over eight search seeds. For the final comparison with RoboGrammar on three shared terrains, the authors reconstruct selected bodies, restore self-collisions, and train each body's controller from scratch eight times. Additional MPC tests assess how these bodies perform under another control procedure. This comparison supports body quality across controllers; the contribution of LLM proposals within the search would be clearer with a matched non-LLM evolutionary baseline.

RoboMoRe~\citep{si250600276} co-designs morphology parameters and reward programs. The agent receives a partially masked robot description and environment code without the original reward. A coarse stage evaluates body--reward combinations through SAC; a fine stage alternates their revision using prior evaluations and accepts higher-scoring pairs. The selected body and reward are retained. Across eight MuJoCo tasks, selection uses task fitness divided by body volume, incorporating material use into the design objective. Four ablation conditions---neither intervention, morphology only, reward only, and both---compare their contributions under a shared training protocol, with a controller trained for each pair. Joint optimization achieves the highest efficiency on seven tasks; morphology-only is slightly higher on Swimmer. The comparison shows how reward design and body geometry jointly shape the behaviors learned by the controller.

\begingroup Auto-Robotist~\citep{si260525832} retains voxel bodies and reusable design knowledge. An LLM edits elite bodies using retrieved structural rules alongside genetic mutations; each candidate receives a fixed PPO training budget. Fitness and parent-relative gains update a library of structural archetypes, positive and negative rules, and evaluated examples through Add, Diagnose, and Merge operations. Seven EvoGym tasks test discovery and transfer from $5\times5$ to $10\times10$ bodies. Cross-scale transfer preserves the archetypes and rules, discards raw source-scale observations, and accumulates new evidence at the target scale. A skill-only condition removes the source-body exemplar to examine transfer through rules. Separate Walker and Pusher ablations remove diagnosis, merging, or genetic mutation, or omit detailed rules and examples from proposal prompts. The larger losses on Pusher and smaller effects on near-ceiling Walker show that the library's benefits depend on the task. Convergence is measured in candidate evaluations, with transferred knowledge built through an earlier source-scale search.\parfillskip=0pt plus 0.1\linewidth\par\endgroup

Three further studies broaden this comparison. Debate2Create~\citep{qiu2026debate2create} alternates design proposals, control-oriented critiques, and revised morphology--reward pairs. Simulator scores rank its retained archive; criterion-specific judges supply textual feedback for subsequent proposals. Its body/reward crossover and budget-matched zero-shot comparisons test co-design and iterative feedback while still retraining a controller for each candidate. Kawaharazuka et al.~\citep{kawaharazuka2026efficient} instead interleave LLM proposals with black-box optimization of base positions, joint types, and link lengths. Evaluated samples and Pareto solutions provide feedback under fixed inverse-kinematics position-error and torque objectives, enabling design-search comparisons without reward or policy training. AID-SR~\citep{chen2026aidsr} revises continuum-robot descriptions using simulator diagnostics and semantic critiques, retaining qualified designs for subsequent PPO evaluation and demonstrating three fabricated robots. Its feedback primarily establishes physical validity; downstream task performance is evaluated separately rather than returned through a demonstrated reward-driven design loop.

\begingroup These approaches trade the scope of design feedback against the cost of obtaining it. Controller-based search tests task behavior but trains a controller for each candidate; kinematic search evaluates reachability and torque without policy training. Reusable design rules can carry experience across searches, while fresh controller training tests whether selected bodies depend on the controller used during selection. Comparisons should count body evaluations, training effort, and prior search experience together. Budget-matched tests across controllers, tasks, and fabricated bodies would establish when feedback-guided design produces reusable physical advantages.\parfillskip=0pt plus 0.5\linewidth\par\endgroup

\input{tables/si-morphology-comparison}

\subsection{Component Optimization}
\label{sec:si-model}

\subsubsection{Model}
\label{sec:si-model-training}

\begingroup Model-level component optimization concerns agent-directed improvements to learned robotic components through data selection, supervision design, architecture search, or parameter adaptation. EmbodiRSI~\citep{embodirsi2026} uses agents to diagnose policy failures, generate corrective demonstrations, and prioritize underperforming task configurations, progressively updating a pretrained VLA through advantage-conditioned behavior cloning. Reflective Self-Adaptation~\citep{li2025reflectiveadaptation} instead employs VLM-based failure analysis to construct structured rewards for PPO, complemented by quality-guided SFT on successful trajectories. Both retain updated policy checkpoints for subsequent execution and refinement. TacEvo~\citep{si260630109} extends optimization to model architecture: an LLM proposes tactile-network mutations and crossovers, while supervised training and validation guide MAP-Elites selection. Promising architectures are archived and reused as candidates for further search.\parfillskip=0pt plus 0.1\linewidth\par\endgroup

\begingroup Other approaches improve modular components or govern model updates. ReCoVLA~\citep{hu2026recovla} uses VLM-diagnosed failures to select recovery rewards, training failure-specific residual policies through simulated PPO while keeping the base VLA frozen. Trained residuals are retained in a library and dispatched during deployment. Robo-COP~\citep{zhang2026robocop} lets an agent identify useful execution demonstrations, decide when to fine-tune a VLA, and specify target skills; candidate checkpoints are adopted only after verification against existing capabilities, with the incumbent retained when a candidate is rejected. RACaP~\citep{racap2026} additionally distills validated frontier-agent ReAct decisions into role-specific LoRA adapters of a compact VLM, retaining the resulting model for subsequent tool-based execution, although this distillation does not constitute a recurring autonomous model-improvement loop. These methods distinguish agent-guided learning interventions from the fixed optimization procedures executing them, while highlighting different retained improvements: updated policy weights, selected architectures, modular recovery policies, and distilled decision models.\parfillskip=0pt plus 0.1\linewidth\par\endgroup



\subsubsection{Code}
\label{sec:si-code}

Code-level component optimization retains feedback-informed revisions inside callable tools, such as state estimators, geometric routines, and controllers. The updated artifact is the implementation rather than learned parameters, task-level behavior (Section~\ref{sec:si-artifact}), or exposed calling contracts (Section~\ref{sec:si-harness-tool-interface}). Task-program synthesis (Section~\ref{sec:policy-a3}) contributes when feedback-guided edits are retained inside a reusable component.

\begingroup InterPreT revises Python predicate functions using interpreter errors and predicate labels derived from human language feedback; the retained functions support subsequent symbolic planning~\citep{han2024interpret}. ETM instead improves shared perception routines offline: discrepancies between estimated and privileged simulator states guide geometric checks and branches, with candidates selected through validation~\citep{hu2026coap}. Accepted changes remain in the library used by subsequent frozen executions. HexaShell illustrates a related repair at the pixel-to-geometry boundary: an area check accounts for pixel footprints on the object plane rather than counting occupied grid cells, followed by fresh robot execution~\citep{gao2026hexashell}. Human-language labels, privileged state discrepancies, and execution records thus provide different grounds for revising these components.\parfillskip=0pt plus 0.1\linewidth\par\endgroup

\begingroup Control-side revisions similarly alter what a tool does behind its call interface. In the shared cloth-folding case reported by Physical Coding and URAI, human-steered development replaces a pick-and-place implementation with cloth pinching, coordinated transport, and release while retaining its two-point interface~\citep{physicalcoding2026,ge2026urai}. LITHE moves the revision into the running controller: a supervisory agent probes a one-degree-of-freedom arm, estimates gravity effects from telemetry, and generates C++ compensation code for hot replacement~\citep{lim2026lithe}. The hot-swap mechanism itself remains supporting infrastructure. Whereas the folding tool is revised between trials, LITHE loads new control code while the real-time loop continues.\parfillskip=0pt plus 0.1\linewidth\par\endgroup

Systems differ in which tool internals can change and how the resulting implementations are tested. DexAgent retains parameterized reconstruction, grasp-generation, and verification tools across demonstrations, with property-based simulation checks constraining their outputs~\citep{wang2026dexagent}. RHO's RAI configuration constrains tool names and signatures while permitting body-level fixes such as detection filtering and fallbacks~\citep{rho2026}. RACaP evaluates revisions to contact-mechanics routines under paired parent/candidate tasks, seeds, and budgets; positive net success admits a revision before deployment is frozen~\citep{racap2026}. SimEX takes a complementary route after physical failures, using a corrected simulator to screen toolbox repairs and capability replay to check retained behavior; reusable toolbox changes survive while task scripts are discarded~\citep{simex2026}.

Navigent uses an offline agent to inspect successful and failed navigation trajectories and propose patches to its geometric critic or navigation prompts~\citep{zhou2026navigent}. Candidates are replayed on the same training episodes as the incumbent, and the best patch is retained only if it fixes more failures than it introduces. Accepted revisions are carried into later episodes with the model weights fixed.

\begingroup Tool-internal revisions should be distinguished from simultaneous prompt, memory, interface, or simulator changes; whole-system gains cannot automatically be attributed to code optimization. Conversely, the 123-round study reports accepted local repairs without solving its target task, showing that component-level progress need not compose into end-to-end gains~\citep{si260931760}. Broader acceptance, generalization, and retention protocols are discussed in Section~\ref{sec:si-artifact}.\parfillskip=0pt plus 0.1\linewidth\par\endgroup

\input{tables/si-components-comparison}

\subsection{Successor Continues AI R\&D}
\label{sec:si-successor}

\begingroup This category concerns a change in who performs the next round of model-development research. The current model-based agent contributes to developing a trained successor, and the resulting model assumes responsibility for subsequent research decisions, such as proposing learning algorithms, designing training interventions, or evaluating candidate models. The required evidence connects the development process, the retained successor checkpoint, and its later research activity. Improving a robot policy, distilling a teacher into a task executor, or training a supporting evaluator can improve the system without transferring this responsibility. Similarly, editing an agent's tools or memory mechanisms changes its harness rather than demonstrating a trained model successor.\parfillskip=0pt plus 0.1\linewidth\par\endgroup

EmbodiedSWE~\citep{embodiedswe2026} illustrates the distinction within robotics. Its generated simulation tasks support turn-level PPO training of Seed-2.1-Lite, but the updated agent is studied as a robot-task solver. The reported training improvements and before/after trajectories do not show that this checkpoint takes over task-generation or model-training research. Moreover, held-out evaluation of the post-trained agent was unfinished in the inspected manuscript. EmbodiedRSI~\citep{song2026embodiedrsiactivecontinualrobot} provides a complementary boundary: it selects informative experiments, revises code and skills, and learns memory-management decisions around a frozen robot foundation model. Improving how this system investigates execution failures does not itself establish a trained successor assuming model-training R\&D.

\begingroup Self-Developing~\citep{ishibashi2025selfdeveloping} provides a non-robotics example in which an updated model continues algorithm discovery. Its algorithm factory generates Python model-merging procedures, and development scores of the resulting task models define preferences for DPO updates to the factory. The updated factory then proposes the next round of algorithms, with three iterations evaluated. Throughout this process, the generated procedures operate on the original seed model. The factory therefore carries the continuing research role, while the merged models provide evidence about its proposals. This demonstrates succession within model-merging research for mathematical tasks under a prescribed prompt, preference construction, and factory-training procedure.\parfillskip=0pt plus 0.1\linewidth\par\endgroup

\begingroup Two other non-robotics studies clarify the relationship between continued research and model development. DiscoveryRSI~\citep{zhong2026discoveryrsi} trains a research model from credited investigation histories and applies accepted model and harness updates to fresh scientific questions. Its updated model continues hypothesis generation and refinement in life sciences and batteries, extending the succession mechanism to scientific inquiry. Self-Developing's continuing research instead concerns model-improvement algorithms. PostTrainBench~\citep{rank2026posttrainbench} gives a frontier-model agent control over data, training methods, and experiments for a separate base model. In the reported workflow, the frontier agent retains the research role and the resulting checkpoint is evaluated on held-out tasks. Together, these examples distinguish the object of subsequent research and the model responsible for carrying it forward.\parfillskip=0pt plus 0.1\linewidth\par\endgroup

\begingroup The distinctive opportunity is to retain the ability to choose better experiments, alongside the artifacts those experiments produce. Training a research successor adds a model-update cost, so its value must be measured through the quality and cost of improvements it produces on new development problems. The examples above establish this mechanism outside robotics; extending it to robot development requires following the trained successor through later experiments and their physical outcomes (Section~\ref{sec:future-successor}).\parfillskip=0pt plus 0.5\linewidth\par\endgroup

\input{tables/si-successor-comparison}

\subsection{Harness Improves Harness}
\label{sec:si-harness}

\begin{figure}[!t]
\centering
\includegraphics[width=\columnwidth]{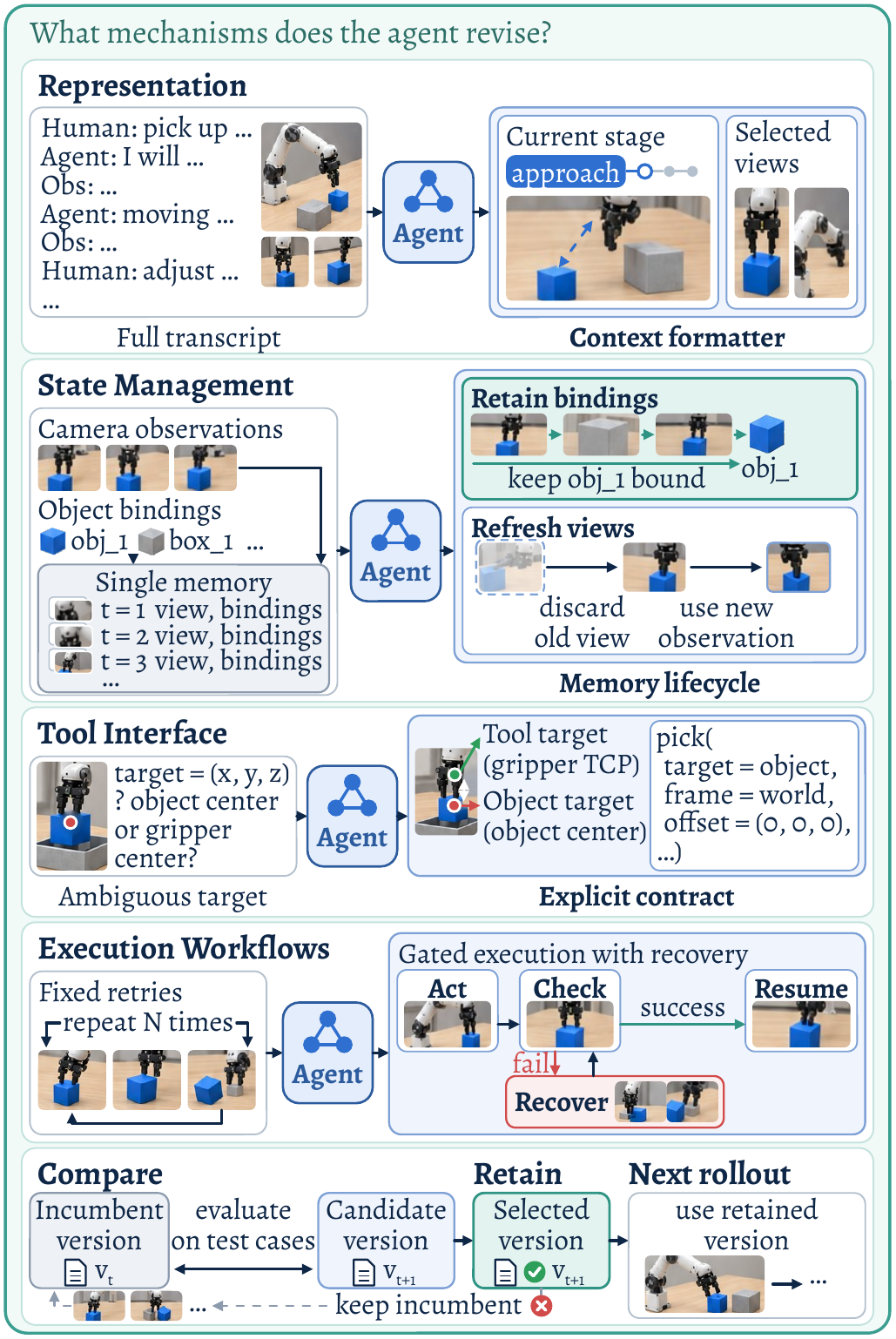}
\caption{Harness Improves Harness. Agents revise input construction, information lifetimes, tool contracts, and execution coordination. Candidate revisions are compared with the incumbent, and accepted versions are retained for later rollouts.}
\label{fig:si-harness}
\end{figure}

Harness improvement changes the mechanisms through which an agent receives information and coordinates robot execution. We organize retained revisions along that interaction chain (Figure~\ref{fig:si-harness}): Representation structures decision inputs, State Management governs information across steps, Tool Interface defines callable operations, and Execution Workflows governs their coordination. These targets complement the preceding artifact and component categories: updating a stored lesson changes knowledge contents, while changing retrieval rules changes the harness; revising a tool algorithm changes its implementation, while revising its arguments or result semantics changes its interface. We compare the concrete mechanism changed, the execution evidence motivating revision, and its subsequent use. Development-time search can produce such retained improvements even \mbox{when the deployed harness is frozen.}

\subsubsection{Representation}
\label{sec:si-harness-representation}

\begingroup Representation revisions change which observations, spatial relations, and execution records reach the agent, and how they are expressed. The persistent change is a rule for constructing decision inputs; individual images, object positions, and progress values are the data processed by that rule.\parfillskip=0pt plus 0.1\linewidth\par\endgroup

Long execution histories can obscure what remains to be done. SHAPER~\citep{grid62} replaces an accumulating transcript with context code that emphasizes the latest subtask and execution count, compresses older rounds, and flags repeated attempts. Rollout critiques of repetition and missing progress cues guide these revisions. For spatial reasoning, the revised context code assembles selected views and RGB-derived crops into the current decision input. RoboFoundry~\citep{liang2026robofoundry} similarly replaces a constant task prompt with the next unfinished stage. Its counting and transfer examples show the revised formatter being reused after partial completion. In both cases, the retained formatter makes task-relevant evidence explicit at the next decision point.

Spatial representations must also connect perception to an actionable target. PhysicalRSI's digit-arrangement example~\citep{physicalrsi2026} identifies digit values, orders them, and assigns placement slots. This makes the symbolic-to-spatial mapping explicit, although the demonstration also changes the task program and does not isolate the representation's contribution. AgentCanvas~\citep{zhou2026agentcanvas} offers a controlled comparison: architecture search revises heading descriptions and introduces tolerance to elevation differences in MapGPT's candidate observations. Adding these edits to the frozen baseline improves held-out navigation; other surviving edits contribute little or reduce performance. An accepted revision can therefore contain both useful and ineffective changes.

\begingroup Visual formatting can affect reasoning cost even when the underlying evidence stays the same. NavGPT-3~\citep{zhou2026navgpt3} replaces a stitched panorama with separate camera views during development. A subsequent study holds tools and scene evidence fixed while varying labels, map orientation, and directional cues. Success rates remain relatively stable, while token use changes; human readability does not predict the cheapest format for the model. AdaHVLA~\citep{si260929204} selects observations, progress records, and corrective context for VLA calls. Its assessments check whether a revised mechanism was activated, but the reported coordination results do not isolate a formatting-only effect.\parfillskip=0pt plus 0.1\linewidth\par\endgroup

\begingroup The comparison rests on a retained change to the input-construction rule and evidence of its later use. Stored knowledge and persistence policies are discussed separately in Sections~\ref{sec:si-memory} and~\ref{sec:si-harness-state-management}.\parfillskip=0pt plus 0.1\linewidth\par\endgroup

\subsubsection{State Management}
\label{sec:si-harness-state-management}

\begingroup State Management governs when task and interaction information is written, retained, retrieved, invalidated, or reset. The evolving object is the lifecycle rule, distinct from episode state values and reusable knowledge contents (Section~\ref{sec:si-memory}). Retention must preserve evidence needed after occlusion without allowing stale observations to misdirect control.\parfillskip=0pt plus 0.1\linewidth\par\endgroup

Different information can require different lifetimes within one task. AdaHVLA~\citep{si260929204} preserves history useful for route reasoning while refreshing visual context that has become stale for local execution. Between rollouts, agents diagnose failures from aligned observations and logs, patch the harness, and separately assess parent and revised executions. Accepted code persists across trials, whereas runtime state is initialized for each rollout. PhysicalRSI~\citep{physicalrsi2026} illustrates the complementary need for retention: its revised harness preserves object--cup correspondences across occlusion. This project demonstration identifies a management rule and its use, without isolating its contribution to performance.

\begingroup Retained information also needs conditions for use and update. In RoboFoundry~\citep{liang2026robofoundry}, occlusion-related failures motivate revised historical-frame retrieval and fallback rules. Old views serve as reminders rather than authority over current observations; fresh rollouts test repairs before accepted rules enter persistent execution code. HyMeS~\citep{zhao2026hymes} instead makes event commitment explicit: rollout traces and stage verdicts guide a coding agent's revision of memory and verification rules. Its selected program updates bindings, repetition counts, and interaction flags, while evaluation freezes the program and VLA and initializes state for each task. RoboFoundry thus emphasizes when historical evidence remains usable, while HyMeS examines what evidence permits a state update. Subsequent action scheduling belongs to Execution Workflows (Section~\ref{sec:si-harness-execution-workflows}).\parfillskip=0pt plus 0.1\linewidth\par\endgroup

An accepted state revision need not improve performance on its own. AgentCanvas~\citep{zhou2026agentcanvas} retains revisions to history-processing and answer-state mechanisms. Yet adding AFlow's retained history tags alone to the frozen MapGPT baseline reduces mean held-out success from 39.4\% to 35.8\% across three evaluation passes, even though the complete selected architecture improves. This comparison tests the edit in isolation, not its interactions with other changes. Retaining a revised rule and demonstrating its independent utility therefore require different evidence.

\subsubsection{Tool Interface}
\label{sec:si-harness-tool-interface}
Tool Interface revisions change what the agent can request and how a request is interpreted. RoboFoundry~\citep{liang2026robofoundry} separates an object-center placement target from the controller's tool-center-point target, keeping the hand--object offset inside the execution interface. This changes the calling convention shared by later programs. NavGPT-3~\citep{zhou2026navgpt3} identifies a mismatch between a prompt describing local movement and the absence of a corresponding tool, then adds that operation. Interface repair therefore concerns agreement between the advertised action and its executable contract, including coordinate conventions and returned evidence.

RACaP~\citep{racap2026} exposes grasp, clearance, target-interpretation, and corrective parameters through typed calls, with readable reports and fresh images informing subsequent choices. Its evolution process can revise argument exposure and reports alongside routing; changes inside grasp or transport algorithms belong separately to Component Optimization. Scale and Selection~\citep{wei2026scaleselection} searches a repository whose interface surface includes tool semantics and per-call motion limits. This establishes the reported optimization scope, but does not identify an accepted edit to every exposed field. A general permission to edit tools should consequently be distinguished from evidence that a particular \mbox{calling contract changed and was retained.}

\subsubsection{Execution Workflows}
\label{sec:si-harness-execution-workflows}

Execution Workflows determine when to observe, act, verify, recover, or stop. RoboFoundry~\citep{liang2026robofoundry} retains changes to replanning cadence, while AdaHVLA~\citep{si260929204} revises goal handoffs and the duration of corrective guidance across VLA calls. Premature goal advancement motivates continuing guidance through a corrective turn and the subsequent forward motion before returning to nominal control. The rules that commit progress evidence are covered in State Management (Section~\ref{sec:si-harness-state-management}); here the revision governs the resulting action sequence. These examples distinguish revising coordination rules between rollouts from ordinary \mbox{state-dependent execution within a rollout.}

Recovery workflows must also specify how control returns to nominal execution. Zetta~\citep{zetta2026} retains critics, recovery logic, and re-entry predicates that require the diagnosed failure to clear and contact to stabilize before VLA resumption. Its transfer studies reuse grasp-retention monitoring and failure-gated retries across tasks. RACaP~\citep{racap2026} separates task-level routing from local transport decisions, returning control and visual reports after physical operations. AgentCanvas~\citep{zhou2026agentcanvas} retains structural edits such as a stop-adviser connection and revised stopping gates. These mechanisms organize interventions and handoffs; the recovery motion itself can \mbox{remain a skill or a tool implementation.}

\begingroup DynaHarness~\citep{dynaharness2026} connects execution governance to revision admission. Its command records distinguish grounding refusals, local capability outcomes, and task completion, allowing failure attribution to target a capability or execution mechanism. Reported changes include a latched completion verdict and revised physical execution conditions; paired checks and broader regression tests determine whether candidates enter the next round. This provides a concrete comparison for validating retained harness changes, while its model-directed authorship of individual revisions remains unspecified. The runtime coordination comparison is discussed separately in Section~\ref{sec:policy-a2}.\parfillskip=0pt plus 0.1\linewidth\par\endgroup

\begingroup Resource use can be revised through coordination guidance. RPG~\citep{rpg2026} changes its system prompt to introduce a bounded perception--action loop and track remaining interaction budget. Between rounds 3 and 4, held-out simulation success rises from 43.2\% to 73.6\% without a recorded skill edit. The revision also changes observation-budget use, so the gain concerns the combined protocol rather than an isolated instruction. Candidate and merged revisions undergo cross-task checks before retention, while model weights and task evaluators remain fixed. This connects harness improvement to a persistent change in how existing capabilities are used.\parfillskip=0pt plus 0.1\linewidth\par\endgroup

Across the four targets, adoption and attribution require separate evidence. Scale and Selection~\citep{wei2026scaleselection} compares challengers with the incumbent on the same cases and rejects ties or regressions; its 100-case campaign reaches 67\% held-out success against a 51\% baseline, while smaller-batch campaigns yield mixed results. AgentCanvas~\citep{zhou2026agentcanvas} uses reruns to assess selection noise and excludes evaluator-information leakage from deployable gains. AdaHVLA~\citep{si260929204} additionally checks whether the revised mechanism was triggered before attributing an observed effect. These controls assess retained execution improvements. Evaluating changes to the improvement process additionally requires tracking revisions to proposal procedures, selection gates, or research infrastructure.
\input{tables/si-harness-comparison}

%% file: tables/si-programs-comparison.tex
\begin{table}[!tb]
\centering
\caption{Representative program and skill revisions.}
\label{tab:si-programs}
\begingroup
\small
\makeatletter
\def\NAT@spacechar{\nobreakspace}
\makeatother
\setlength{\tabcolsep}{2pt}
\renewcommand{\arraystretch}{1.10}
\begin{tabularx}{\linewidth}{@{}>{\raggedright\arraybackslash}p{0.42\linewidth}>{\raggedright\arraybackslash}p{20pt}>{\raggedright\arraybackslash}p{0.22\linewidth}>{\raggedright\arraybackslash}X@{}}
\toprule
\rowcolor[HTML]{FFF1DA}
\textbf{Work} & \textbf{Link} & \textbf{Target} & \textbf{Revision} \\
\midrule
SetItUp \citep{xu2024set} & \href{https://doi.org/10.1177/02783649251378198}{\includegraphics[height=8pt]{figures/icons/paper.pdf}} & Arrangement programs & Example-guided regeneration \\
CaP-X \citep{capx} & \href{https://arxiv.org/abs/2603.22435}{\includegraphics[height=8pt]{figures/icons/paper.pdf}} & Helpers & Extract shared routines \\
RATs \citep{rats2026} & \href{https://arxiv.org/abs/2606.19419}{\includegraphics[height=8pt]{figures/icons/paper.pdf}} & Skills & Acquire through play \\
ASPIRE \citep{self260700272} & \href{https://arxiv.org/abs/2607.00272}{\includegraphics[height=8pt]{figures/icons/paper.pdf}} & Programs & Trace-guided repair \\
SkillComposer \citep{woods2026skillcomposer} & \href{https://arxiv.org/abs/2608.14944}{\includegraphics[height=8pt]{figures/icons/paper.pdf}} & Macros & Compress accepted calls \\
RAPID \citep{rapid2026} & \href{https://arxiv.org/abs/2609.30249}{\includegraphics[height=8pt]{figures/icons/paper.pdf}} & Primitives / strategies & Simulation-guided repair \\
SEAR \citep{wang2026sear} & \href{https://arxiv.org/abs/2610.18792}{\includegraphics[height=8pt]{figures/icons/paper.pdf}}\hspace{3pt}\href{https://github.com/Self-Evolving-Agentic-Robotics/SEAR}{\includegraphics[height=8pt]{figures/icons/github.pdf}} & Programs / skills & Revise and retain libraries \\
EmbodiedEvo \citep{li2026embodiedevo} & \href{https://arxiv.org/abs/2610.18867}{\includegraphics[height=8pt]{figures/icons/paper.pdf}} & Skills & Reuse in new scenes of the same task \\
RAX \citep{huang2026rax} & \href{https://arxiv.org/abs/2610.22496}{\includegraphics[height=8pt]{figures/icons/paper.pdf}} & Programs & Exploration-time repair \\
\bottomrule
\end{tabularx}
\endgroup
\end{table}

%% file: tables/si-memory-comparison.tex
\begin{table}[!b]
\centering
\caption{Representative revisions to retained knowledge and memory.}
\label{tab:si-memory}
\vspace{-1pt}
\begingroup
\small
\makeatletter
\def\NAT@spacechar{\nobreakspace}
\makeatother
\setlength{\tabcolsep}{2pt}
\renewcommand{\arraystretch}{1.10}
\begin{tabularx}{\linewidth}{@{}>{\raggedright\arraybackslash}p{0.42\linewidth}>{\raggedright\arraybackslash}p{20pt}>{\raggedright\arraybackslash}p{0.22\linewidth}>{\raggedright\arraybackslash}X@{}}
\toprule
\rowcolor[HTML]{FFF1DA}
\textbf{Work} & \textbf{Link} & \textbf{Target} & \textbf{Revision} \\
\midrule
ME-Brain \citep{si260924271} & \href{https://arxiv.org/abs/2609.24271}{\includegraphics[height=8pt]{figures/icons/paper.pdf}} & Experience & Abstract event records \\
EvoMemNav \citep{si260603509} & \href{https://arxiv.org/abs/2606.03509}{\includegraphics[height=8pt]{figures/icons/paper.pdf}} & Navigation knowledge & Revise scene priors \\
MessyMem \citep{banwasi2026messymem} & \href{https://arxiv.org/abs/2609.15976}{\includegraphics[height=8pt]{figures/icons/paper.pdf}} & World knowledge & Revise graph properties \\
RoboRouter \citep{chen2026roborouter} & \href{https://arxiv.org/abs/2603.07892}{\includegraphics[height=8pt]{figures/icons/paper.pdf}} & Policy knowledge & Refresh routing notes \\
DROC \citep{zha2023droc} & \href{https://arxiv.org/abs/2311.10678}{\includegraphics[height=8pt]{figures/icons/paper.pdf}} & Guidance & Distill language corrections \\
Zero2Skill \citep{wang2026zero2skill} & \href{https://arxiv.org/abs/2607.14047}{\includegraphics[height=8pt]{figures/icons/paper.pdf}} & Guidance & Retain human corrections \\
InterPreT \citep{han2024interpret} & \href{https://arxiv.org/abs/2405.19758}{\includegraphics[height=8pt]{figures/icons/paper.pdf}} & Symbolic knowledge & Learn predicates and operators \\
KnowBody \citep{lou2026knowbody} & \href{https://arxiv.org/abs/2609.28530}{\includegraphics[height=8pt]{figures/icons/paper.pdf}} & Body knowledge & Validate conditional rules \\
Robo-COP \citep{zhang2026robocop} & \href{https://arxiv.org/abs/2610.09228}{\includegraphics[height=8pt]{figures/icons/paper.pdf}} & Policy advice & Flag stale advice \\
EmbodiedRSI \citep{song2026embodiedrsiactivecontinualrobot} & \href{https://arxiv.org/abs/2610.10498}{\includegraphics[height=8pt]{figures/icons/paper.pdf}} & Experience & Abstract experimental knowledge \\
\bottomrule
\end{tabularx}
\endgroup
\end{table}

%% file: tables/si-rewards-comparison.tex
\begin{table}[!b]
\centering
\caption{Reward revisions by objective and mechanism.}
\label{tab:si-rewards}
\begingroup
\small
\makeatletter
\def\NAT@spacechar{\nobreakspace}
\makeatother
\setlength{\tabcolsep}{2pt}
\renewcommand{\arraystretch}{1.10}
\begin{tabularx}{\linewidth}{@{}>{\raggedright\arraybackslash}p{0.42\linewidth}>{\raggedright\arraybackslash}p{20pt}>{\raggedright\arraybackslash}p{0.22\linewidth}>{\raggedright\arraybackslash}X@{}}
\toprule
\rowcolor[HTML]{FFF1DA}
\textbf{Work} & \textbf{Link} & \textbf{Target} & \textbf{Revision} \\
\midrule
Eureka \citep{ma2023eureka} & \href{https://arxiv.org/abs/2310.12931}{\includegraphics[height=8pt]{figures/icons/paper.pdf}} & Reward code & Training-guided search \\
Text2Reward \citep{text2reward2023} & \href{https://arxiv.org/abs/2309.11489}{\includegraphics[height=8pt]{figures/icons/paper.pdf}} & Reward code & Human corrections \\
REvolve (2024) \citep{revolve2024} & \href{https://arxiv.org/abs/2406.01309}{\includegraphics[height=8pt]{figures/icons/paper.pdf}} & Reward code & Preference-guided search \\
CRAFT \citep{choi2025craft} & \href{https://arxiv.org/abs/2509.14380}{\includegraphics[height=8pt]{figures/icons/paper.pdf}} & Terms / weights & Visual failure analysis \\
Video2Reward \citep{video2reward2024} & \href{https://arxiv.org/abs/2412.05515}{\includegraphics[height=8pt]{figures/icons/paper.pdf}} & Reward code & Motion discrepancies \\
ELEMENTAL \citep{elemental2025} & \href{https://proceedings.mlr.press/v267/chen25at.html}{\includegraphics[height=8pt]{figures/icons/paper.pdf}} & Features & Feature-count feedback \\
Self-Alignment \citep{zeng2024selfalignment} & \href{https://proceedings.mlr.press/v235/zeng24d.html}{\includegraphics[height=8pt]{figures/icons/paper.pdf}} & Weight search & Constrain calibration \\
Demo2Reward \citep{demo2reward2026} & \href{https://arxiv.org/abs/2606.00083}{\includegraphics[height=8pt]{figures/icons/paper.pdf}} & Scoring prompt & Correct label errors \\
\bottomrule
\end{tabularx}
\endgroup
\end{table}

%% file: tables/si-environments-comparison.tex
\begin{table}[!b]
\centering
\caption{Revisions to environments, tasks and datasets.}
\label{tab:si-environments}
\begingroup
\small
\makeatletter
\def\NAT@spacechar{\nobreakspace}
\makeatother
\setlength{\tabcolsep}{2pt}
\renewcommand{\arraystretch}{1.10}
\begin{tabularx}{\linewidth}{@{}>{\raggedright\arraybackslash}p{0.42\linewidth}>{\raggedright\arraybackslash}p{20pt}>{\raggedright\arraybackslash}p{0.22\linewidth}>{\raggedright\arraybackslash}X@{}}
\toprule
\rowcolor[HTML]{FFF1DA}
\textbf{Work} & \textbf{Link} & \textbf{Target} & \textbf{Revision} \\
\midrule
DiagGen \citep{diaggen2026} & \href{https://arxiv.org/abs/2609.23103}{\includegraphics[height=8pt]{figures/icons/paper.pdf}} & Assets & Grasp-guided repair \\
CoDimRecon \citep{xie2026codimrecon} & \href{https://arxiv.org/abs/2609.36024}{\includegraphics[height=8pt]{figures/icons/paper.pdf}} & Physics & Revise material model \\
Eurekaverse \citep{liang2024eurekaverse} & \href{https://arxiv.org/abs/2411.01775}{\includegraphics[height=8pt]{figures/icons/paper.pdf}} & Terrains & Training-guided revision \\
Self-CriTeach \citep{si250921543} & \href{https://arxiv.org/abs/2509.21543}{\includegraphics[height=8pt]{figures/icons/paper.pdf}} & Task domains & Repair planner errors \\
BenchMend \citep{zhang2026benchmend} & \href{https://tianqi-zh.github.io/BenchMend/}{\includegraphics[height=8pt]{figures/icons/paper.pdf}} & Instructions & Clarify missing goals \\
Recova \citep{recova2026} & \href{https://arxiv.org/abs/2610.01178}{\includegraphics[height=8pt]{figures/icons/paper.pdf}} & Datasets & Accumulate accepted trials \\
Skill-Space Shooting \citep{si260938178} & \href{https://arxiv.org/abs/2609.38178}{\includegraphics[height=8pt]{figures/icons/paper.pdf}} & Datasets & Admit repair segments \\
\bottomrule
\end{tabularx}
\endgroup
\end{table}

%% file: tables/si-morphology-comparison.tex
\begin{table}[!tb]
\centering
\caption{Robot-body revisions by target and mechanism.}
\label{tab:si-morphology}
\begingroup
\small
\makeatletter
\def\NAT@spacechar{\nobreakspace}
\makeatother
\setlength{\tabcolsep}{2pt}
\renewcommand{\arraystretch}{1.10}
\begin{tabularx}{\linewidth}{@{}>{\raggedright\arraybackslash}p{0.42\linewidth}>{\raggedright\arraybackslash}p{20pt}>{\raggedright\arraybackslash}p{0.22\linewidth}>{\raggedright\arraybackslash}X@{}}
\toprule
\rowcolor[HTML]{FFF1DA}
\textbf{Work} & \textbf{Link} & \textbf{Target} & \textbf{Revision} \\
\midrule
RoboMorph \citep{si240708626} & \href{https://arxiv.org/abs/2407.08626}{\includegraphics[height=8pt]{figures/icons/paper.pdf}} & Topology & Reuse elite designs \\
RoboMoRe \citep{si250600276} & \href{https://arxiv.org/abs/2506.00276}{\includegraphics[height=8pt]{figures/icons/paper.pdf}} & Body / reward & Alternate joint revisions \\
Auto-Robotist \citep{si260525832} & \href{https://arxiv.org/abs/2605.25832}{\includegraphics[height=8pt]{figures/icons/paper.pdf}} & Voxel bodies & Reuse design rules \\
Debate2Create \citep{qiu2026debate2create} & \href{https://proceedings.mlr.press/v306/qiu26c.html}{\includegraphics[height=8pt]{figures/icons/paper.pdf}} & Body / reward & Critique design pairs \\
AID-SR \citep{chen2026aidsr} & \href{https://www.roboticsproceedings.org/rss22/p194.html}{\includegraphics[height=8pt]{figures/icons/paper.pdf}} & Body descriptions & Repair validity defects \\
\bottomrule
\end{tabularx}
\endgroup
\end{table}

%% file: tables/si-components-comparison.tex
\begin{table}[!tb]
\centering
\caption{Revisions to models and tool code.}
\label{tab:si-components}
\begingroup
\small
\makeatletter
\def\NAT@spacechar{\nobreakspace}
\makeatother
\setlength{\tabcolsep}{2pt}
\renewcommand{\arraystretch}{1.10}
\begin{tabularx}{\linewidth}{@{}>{\raggedright\arraybackslash}p{0.42\linewidth}>{\raggedright\arraybackslash}p{20pt}>{\raggedright\arraybackslash}p{0.22\linewidth}>{\raggedright\arraybackslash}X@{}}
\toprule
\rowcolor[HTML]{E8F2FF}
\textbf{Work} & \textbf{Link} & \textbf{Target} & \textbf{Revision} \\
\midrule
EmbodiRSI \citep{embodirsi2026} & \href{https://arxiv.org/abs/2609.38905}{\includegraphics[height=8pt]{figures/icons/paper.pdf}} & Model: policy & Corrective fine-tuning \\
TacEvo \citep{si260630109} & \href{https://arxiv.org/abs/2606.30109}{\includegraphics[height=8pt]{figures/icons/paper.pdf}} & Model: structure & Architecture search \\
ReCoVLA \citep{hu2026recovla} & \href{https://arxiv.org/abs/2606.09630}{\includegraphics[height=8pt]{figures/icons/paper.pdf}} & Model: recovery & Train residual policies \\
Robo-COP \citep{zhang2026robocop} & \href{https://arxiv.org/abs/2610.09228}{\includegraphics[height=8pt]{figures/icons/paper.pdf}} & Model: policy & Verify before adoption \\
InterPreT \citep{han2024interpret} & \href{https://arxiv.org/abs/2405.19758}{\includegraphics[height=8pt]{figures/icons/paper.pdf}} & Code: predicates & Repair from labels \\
ETM \citep{hu2026coap} & \href{https://arxiv.org/abs/2610.12369}{\includegraphics[height=8pt]{figures/icons/paper.pdf}} & Code: perception & Revise geometric checks \\
LITHE \citep{lim2026lithe} & \href{https://arxiv.org/abs/2603.07442}{\includegraphics[height=8pt]{figures/icons/paper.pdf}} & Code: control & Revise compensation \\
SimEX \citep{simex2026} & \href{https://arxiv.org/abs/2609.38982}{\includegraphics[height=8pt]{figures/icons/paper.pdf}} & Code: tools & Replay-screened repair \\
\bottomrule
\end{tabularx}
\endgroup
\end{table}

%% file: tables/si-successor-comparison.tex
\suppressfloats[t]
\begin{table}[t]
\centering
\caption{Research succession and its boundary cases.}
\label{tab:si-successor}
\begingroup
\small
\makeatletter
\def\NAT@spacechar{\nobreakspace}
\makeatother
\setlength{\tabcolsep}{2pt}
\renewcommand{\arraystretch}{1.10}
\begin{tabularx}{\linewidth}{@{}>{\raggedright\arraybackslash}p{0.42\linewidth}>{\raggedright\arraybackslash}p{20pt}>{\raggedright\arraybackslash}p{0.22\linewidth}>{\raggedright\arraybackslash}X@{}}
\toprule
\rowcolor[HTML]{F1EBFF}
\textbf{Work} & \textbf{Link} & \textbf{Domain} & \textbf{Later role} \\
\midrule
Self-Developing \citep{ishibashi2025selfdeveloping} & \href{https://aclanthology.org/2025.naacl-long.519/}{\includegraphics[height=8pt]{figures/icons/paper.pdf}} & Model merging & Updated factory continues \\
DiscoveryRSI \citep{zhong2026discoveryrsi} & \href{https://arxiv.org/abs/2610.23839}{\includegraphics[height=8pt]{figures/icons/paper.pdf}} & Scientific inquiry & Updated researcher continues \\
PostTrainBench \citep{rank2026posttrainbench} & \href{https://arxiv.org/abs/2603.08640}{\includegraphics[height=8pt]{figures/icons/paper.pdf}} & Post-training & Original researcher remains \\
EmbodiedSWE \citep{embodiedswe2026} & \href{https://arxiv.org/abs/2609.27308}{\includegraphics[height=8pt]{figures/icons/paper.pdf}} & Robot tasks & Updated task solver \\
\bottomrule
\end{tabularx}
\endgroup
\end{table}

%% file: tables/si-harness-comparison.tex
\begin{table}[!tb]
\centering
\caption{Four targets of harness revision (Figure~\ref{fig:self-improver-tree}).}
\label{tab:si-harness}
\begingroup
\small
\makeatletter
\def\NAT@spacechar{\nobreakspace}
\makeatother
\setlength{\tabcolsep}{2pt}
\renewcommand{\arraystretch}{1.10}
\begin{tabularx}{\linewidth}{@{}>{\raggedright\arraybackslash}p{0.42\linewidth}>{\raggedright\arraybackslash}p{20pt}>{\raggedright\arraybackslash}p{0.225\linewidth}>{\raggedright\arraybackslash}X@{}}
\toprule
\rowcolor[HTML]{DEF4F0}
\textbf{Work} & \textbf{Link} & \textbf{Target} & \textbf{Revision} \\
\midrule
SHAPER \citep{grid62} & \href{https://arxiv.org/abs/2608.11350}{\includegraphics[height=8pt]{figures/icons/paper.pdf}} & Representation & Restructure context \\
NavGPT-3 \citep{zhou2026navgpt3} & \href{https://arxiv.org/abs/2610.10787}{\includegraphics[height=8pt]{figures/icons/paper.pdf}} & Representation & Reformat camera views \\
AdaHVLA \citep{si260929204} & \href{https://arxiv.org/abs/2609.29204}{\includegraphics[height=8pt]{figures/icons/paper.pdf}} & State & Revise context lifetimes \\
HyMeS \citep{zhao2026hymes} & \href{https://arxiv.org/abs/2608.09410}{\includegraphics[height=8pt]{figures/icons/paper.pdf}} & State & Revise commit rules \\
RoboFoundry \citep{liang2026robofoundry} & \href{https://arxiv.org/abs/2609.32862}{\includegraphics[height=8pt]{figures/icons/paper.pdf}} & Tool interface & Clarify target coordinates \\
RACaP \citep{racap2026} & \href{https://arxiv.org/abs/2609.29394}{\includegraphics[height=8pt]{figures/icons/paper.pdf}} & Tool interface & Revise arguments / reports \\
Zetta \citep{zetta2026} & \href{https://arxiv.org/abs/2608.16590}{\includegraphics[height=8pt]{figures/icons/paper.pdf}} & Workflows & Revise recovery re-entry \\
RPG \citep{rpg2026} & \href{https://arxiv.org/abs/2610.02204}{\includegraphics[height=8pt]{figures/icons/paper.pdf}} & Workflows & Bound execution loops \\
\bottomrule
\end{tabularx}
\endgroup
\end{table}

%% file: sections/06_platform.tex
\Needspace{8\baselineskip}
\section{\texorpdfstring{\roboagentlab}{RoboAgentLab}: An Open Research Platform}
\label{sec:platform}

\begin{figure*}[!t]
\centering
\includegraphics[width=\textwidth]{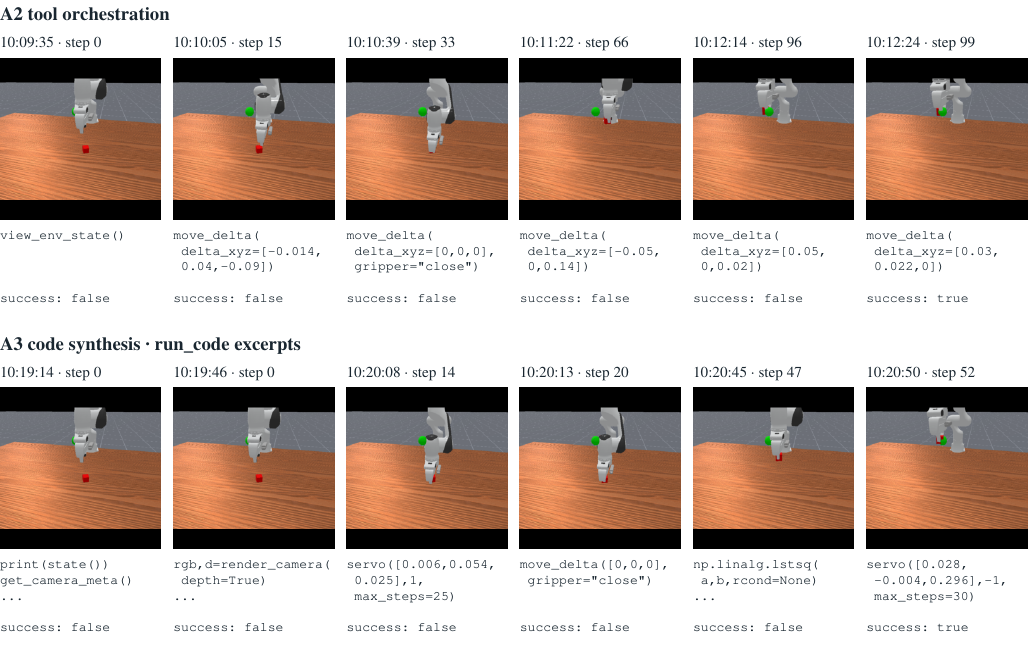}
\caption{Recorded \roboagentlab{} runs on PickCube-v1, seed 0. Each row pairs camera observations with tool calls or program excerpts and the environment's success flag. Timestamps are UTC; step labels count simulator steps. The two runs start from the same state and differ in their access to depth, calibration, and absolute-pose control.}
\label{fig:platform-case}
\end{figure*}

\begingroup \href{https://github.com/RoboAgentLab/RoboAgentLab}{\roboagentlab} connects a coding-agent harness to robot services and persistent experiment records. We ran GPT-6 Astra with a Panda arm in ManiSkill through tool orchestration (A2) and code synthesis (A3). Figure~\ref{fig:platform-case} pairs camera frames with calls and feedback from matched initial states. The agent calls motion tools in A2; in A3, it also writes Python to reconstruct geometry from camera data before issuing motion commands.\parfillskip=0pt plus 0.5\linewidth\par\endgroup

\begingroup Both configurations completed all nine task--seed combinations across picking, pushing, and stacking. They share a primitive manifest and the \texttt{move\_delta} service, but code mode additionally exposes depth, camera calibration, and absolute-pose control. Figure~\ref{fig:platform-paired} in Appendix~\ref{app:platform-integrations} reports each run's motion calls, simulator steps, model calls, tokens, and elapsed time. The resource measurements reflect each configuration's sensing, motion, and decision interfaces. Appendix~\ref{app:platform-interfaces} details the architecture and implemented integrations (Figures~\ref{fig:platform-architecture} and~\ref{fig:platform-integrations}).\parfillskip=0pt plus 0.5\linewidth\par\endgroup

\begingroup An exploration run wrote a memory note that later runs read; a separate coding-agent session used failed memory reads to diagnose and patch the prompt, with external review and follow-up runs. Appendix~\ref{app:platform-data} distinguishes their collection and validation; Section~\ref{sec:future-successor} discusses how such evidence could support training robotics research models.\parfillskip=0pt plus 0.5\linewidth\par\endgroup

%% file: sections/07_challenges_and_future_directions.tex
\Needspace{5\baselineskip}
\section{Challenges and Future Directions}
\label{sec:challenges}
\begingroup The preceding comparisons expose three connected bottlenecks. Actors must respond before their observations become stale; Infrastructure must connect generated resources and experimental choices to robot performance; Evolvers must preserve useful changes without accumulating errors. These bottlenecks motivate efficient decisions, evaluation across successive revisions, and research agents that learn to conduct better experiments.\parfillskip=0pt plus 0.5\linewidth\par\endgroup

\subsection{Distilling Fast and Accurate Robot Decision Models}
\label{sec:future-efficiency}
\begingroup Reducing online deliberation requires deciding which reasoning to learn in advance and which decisions to leave to runtime feedback (Section~\ref{sec:policy}). TinyAstral distills reasoning, tool calls, and recovery trajectories from an evolved teacher harness~\citep{gong2026tinyastral}; ZAP combines supervised learning on self-generated successful rollouts with reinforcement learning~\citep{pikus2026pretrainingknowswhatsimulation}. Comparing these routes and their combination under matched interfaces and training budgets could reveal which supervision transfers to unfamiliar states. Training records should connect observations and tools to decisions, physical outcomes, and recovery attempts. Student rollouts can expose gaps in demonstration coverage, while context and memory selection can reduce the information processed per decision.\parfillskip=0pt plus 0.5\linewidth\par\endgroup

\begingroup Efficiency also depends on the division of decisions between agent and executor. Language-specified subtasks and direct motion targets demand different levels of spatial reasoning and intervention~\citep{grid22,luo2026handson}. Retained skills and selective feedback reduce repeated model calls~\citep{huang2026rax,si2026fewer}, but longer execution intervals risk acting on stale observations. Future work should vary goal representations and execution horizons while controlling the executor and observation access. Handoffs should verify the executor's physical state before retrying or replanning: neither a timeout nor a cancellation request confirms that execution has stopped~\citep{grid2026agentic}.\parfillskip=0pt plus 0.5\linewidth\par\endgroup

\begingroup Evaluate these choices during live execution. Report task success, response latency, and completion time, including communication, retries, and recovery~\citep{qin2026towardsagenticrobots}. SenseNova's proposed protocol keeps the world clock running during deliberation~\citep{sensenova2026embodied}; observation age and missed deadlines can then expose costs hidden by paused simulation. Comparing event-triggered intervention with periodic polling under the same physical-time budget would test whether faster inference or fewer calls improve recovery from unexpected motion and contact.\parfillskip=0pt plus 0.5\linewidth\par\endgroup

\subsection{Benchmarking the Self-Evolving Process}
\label{sec:future-evaluation}
\begingroup As Infrastructure supplies tests and an Evolver selects revisions, evaluation must distinguish better robot capabilities from easier tests or larger budgets (Sections~\ref{sec:infra-evaluation} and~\ref{sec:self-improvement}). Anthropic's internal RoboDojo evaluation compares models under a fixed harness, using each model's highest available reasoning effort~\citep{anthropic2026fable52}; SEAR's independent-restart and no-retained-library controls help isolate accumulated experience~\citep{wang2026sear}. With matched starting conditions and budgets, benchmarks can test whether each revision's gains persist across repeated trials.\parfillskip=0pt plus 0.5\linewidth\par\endgroup

\begingroup Success must be checked independently of the agent's completion claim. Switching the stopping decision from simulator to agent lowered RoboICL's mean score from 83.2 to 63.5 over 50 RoboDojo episodes~\citep{openrua2026revisiting}, while RoboARCade agents prematurely marked \mbox{24--39\%} of attempted subtasks complete~\citep{sharma2026roboarcade}. Time-stamped observations, actions, and physical-state checks can reveal whether a revision changes these errors~\citep{wan2026nofreechecker,xie2026vabar}. Record constraint violations, interventions, and irreversible outcomes alongside visual safety checks before execution and commitment~\citep{zhu2026leash}, charging resets and repairs to the same budget.\parfillskip=0pt plus 0.5\linewidth\par\endgroup

\begingroup Test for regressions by rerunning earlier tasks and for transfer by evaluating the revised system on held-out tasks. The Imitator Game's object-substitution test asks whether the same demonstrated intent can be achieved when a replacement object requires a different manipulation strategy~\citep{zhou2026imitatorgame}. Cross-session tests should also vary user preferences and task priorities, checking whether retained improvements remain useful and obsolete guidance is revised~\citep{grid2026agentic}. Appendix~\ref{app:process-evaluation} provides a protocol for these comparisons across successive versions.\parfillskip=0pt plus 0.5\linewidth\par\endgroup

\subsection{Successor Models that Continue Robotics AI R\&D}
\label{sec:future-successor}
\begingroup Infrastructure agents already organize learning experiments, while Evolvers retain engineering knowledge and revised tools. A further direction is to train the agent that chooses the next experiment (Section~\ref{sec:si-successor}). Development records can connect proposed changes to tests, training runs, deployment outcomes, and subsequent corrections. Training on these records could specialize an open-weight model for robotics R\&D, building on automated model development~\citep{gao2026rsijev,chan2026intelligenceexplosion}. A tractable starting point is a bounded task, such as controller improvement, where each proposed intervention can be tested against its physical consequences.\parfillskip=0pt plus 0.5\linewidth\par\endgroup

\begingroup Once a successor takes over research decisions, the protocol in Section~\ref{sec:future-evaluation} should assess the quality and cost of its improvements on new robotics development problems. Tests should track retained engineering skills, adaptation to unfamiliar robots, and human corrections. Across generations, the central question is whether each successor produces evidence that helps the next model choose more informative experiments.\parfillskip=0pt plus 0.5\linewidth\par\endgroup

%% file: sections/08_conclusion.tex
\Needspace{8\baselineskip}
\section{Conclusion}
\label{sec:conclusion}
\begingroup Agentic robotics connects decisions about current robot behavior with decisions about the resources and revisions used in later runs. Our synthesis identifies three design choices: what observations and action authority to give an Actor, which development experiments Infrastructure should automate, and what an Evolver should retain after testing. Greater authority changes the agent's responsibilities; it does not remove the need for accurate observations and capable execution. Reusable programs and skills can move computation into development, so their value depends on both preparation cost and behavior under new conditions.\parfillskip=0pt plus 0.5\linewidth\par\endgroup

\begingroup The \roboagentlab{} platform makes execution and software revision inspectable through linked records. These records provide a basis for testing how development decisions affect subsequent robot behavior. Timely intervention, independent outcome checks, and transfer tests can then guide which capabilities to reuse and which to revise, and measure progress across tasks, development, and execution.\parfillskip=0pt plus 0.5\linewidth\par\endgroup

%% file: sections/09_appendix.tex
\onecolumn
\section{Full Taxonomy of Agentic Robotics}
\label{app:taxonomy}
\setcounter{figure}{0}
\renewcommand{\thefigure}{\thesection\arabic{figure}}
\renewcommand{\theHfigure}{appendix.\thesection.\arabic{figure}}

\begingroup
\renewenvironment{figure*}[1][]{%
  \par\noindent\begin{minipage}{\textwidth}%
  \captionsetup{type=figure,hypcap=false}%
}{\end{minipage}\par}
\input{figures/role-taxonomy-appendix}
\endgroup

\input{sections/10_platform_appendix}
\input{sections/11_evaluation_appendix}

%% file: sections/10_platform_appendix.tex
\begingroup
\setlength{\intextsep}{6pt}
\setlength{\textfloatsep}{10pt}
\setlength{\floatsep}{8pt}
\section{Platform Implementation and Research Data}
\label{app:platform}
\setcounter{figure}{0}

\subsection{Harness, Services, and Shared Interfaces}
\label{app:platform-interfaces}
\begingroup \roboagentlab{} uses pi for model interaction, tool dispatch, and persistent sessions. Robot extensions translate requests into service calls; a per-robot manifest declares the operations, parameters, and availability of each tool or Python API. Figure~\ref{fig:platform-architecture} shows the implemented architecture, separating planner backends from action mechanisms. Its broader simulator, robot, perception, and policy interfaces are summarized in Figure~\ref{fig:platform-integrations}; the runs below exercise the ManiSkill path.\parfillskip=0pt plus 0.5\linewidth\par\endgroup

\begin{figure}[!htbp]
\centering
\includegraphics[width=.92\textwidth,trim=8bp 8bp 8bp 8bp,clip]{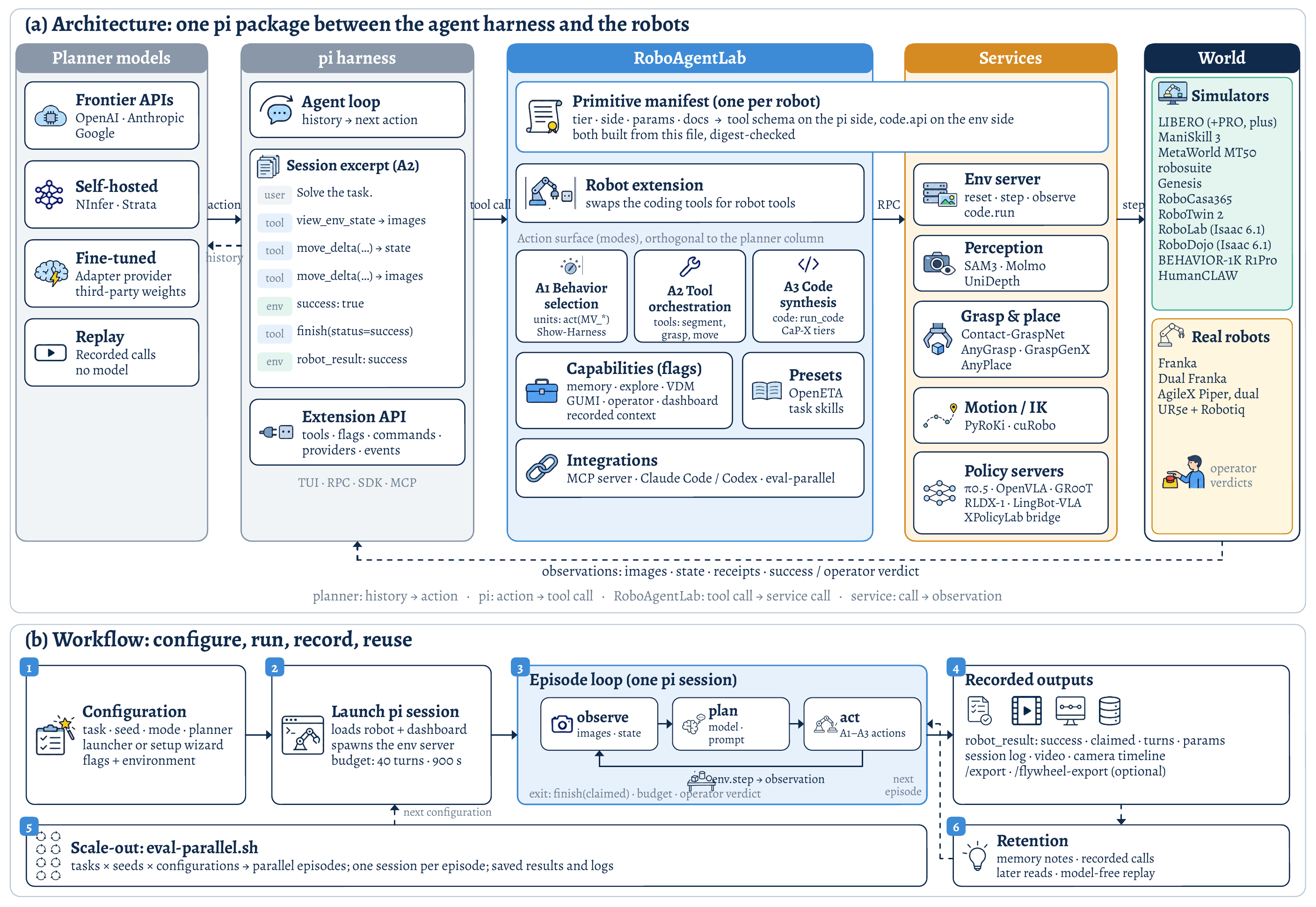}
\caption{\roboagentlab{} implementation and workflow. (a) Planner backends connect through pi and robot extensions to services. The abbreviated session illustrates tool orchestration. (b) Configuration, execution, logging, and reuse. The platform also supports fine-tuned model adapters and optional exports.}
\label{fig:platform-architecture}
\end{figure}

\subsection{Run Configuration and Measurements}
\label{app:platform-integrations}
\begingroup The paired runs used platform revision \texttt{908d969e8}, ManiSkill 3.0.1, SAPIEN 3.0.3, and a Panda arm on an NVIDIA L40S. GPT-6 Astra used high reasoning with a 40-turn, 900-second budget. We ran PickCube-v1, PushCube-v1, and StackCube-v1 at seeds 0--2 once per configuration, with three concurrent workers per batch and unisolated code execution without root.\parfillskip=0pt plus 0.5\linewidth\par\endgroup

\begingroup Both modes initially receive two RGB views and robot state, and observe the environment's success flag during execution. A2 uses \texttt{view\_env\_state} and \texttt{move\_delta}. A3 additionally accesses depth arrays, camera calibration, and absolute-pose \texttt{servo} commands; simulator object poses are disabled. In the run shown in Figure~\ref{fig:platform-case}, generated Python back-projects depth pixels and fits the goal sphere's center before control.\parfillskip=0pt plus 0.5\linewidth\par\endgroup

\begin{figure}[!htbp]
\centering
\includegraphics[width=.92\textwidth]{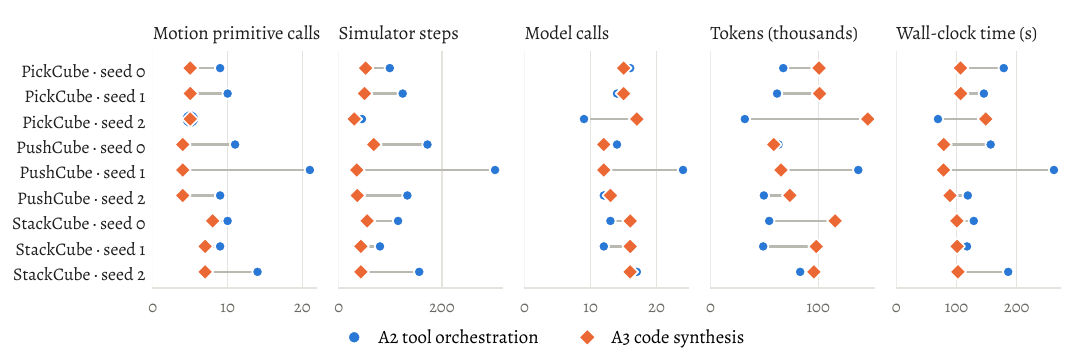}
\caption{Paired ManiSkill runs, all successful by the environment verdict. Lines connect matching task--seed pairs; axes show separate resources. Time includes model waiting. A2 and A3 differ in sensing and motion access.}
\label{fig:platform-paired}
\end{figure}

\subsection{Interaction and Development Records}
\label{app:platform-data}
\begingroup \emph{Interaction records.} Sessions retain observations, model responses, tool arguments, generated programs, and execution feedback. Structured results separately store environment success and the agent's completion claim; all paired runs agree on success (Figure~\ref{fig:platform-paired}). The dashboard displays messages, tool results, and camera frames. Model calls count assistant responses, and elapsed times are derived from session timestamps; episode videos omit planner waiting.\parfillskip=0pt plus 0.5\linewidth\par\endgroup

\begingroup \emph{Retained memory and replay.} One PushCube exploration run wrote a task-family note, which was automatically merged and read by five later runs across seeds and tasks. This records reuse, without a performance-based acceptance gate. Separately, a recorded PickCube tool sequence completed its source seed and one of two additional seeds without model calls; the other replay failed. Retention therefore requires evaluation beyond the episode that produced the asset.\parfillskip=0pt plus 0.5\linewidth\par\endgroup

\begingroup \emph{Development records.} A separate coding-agent session diagnosed five failed memory-file probes in each baseline episode and changed the prompt for an empty memory library. Five of eleven targeted tests failed before the patch; all eleven passed afterward. Four follow-up episodes completed with no failed memory probes. An external coding-agent reviewer accepted the patch, and a collection wrapper linked the request, diff, tests, and runs into one development record.\parfillskip=0pt plus 0.5\linewidth\par\endgroup

\begin{figure}[!htbp]
\centering
\includegraphics[trim=0 141bp 0 0,clip,width=\textwidth,height=.48\textheight,keepaspectratio]{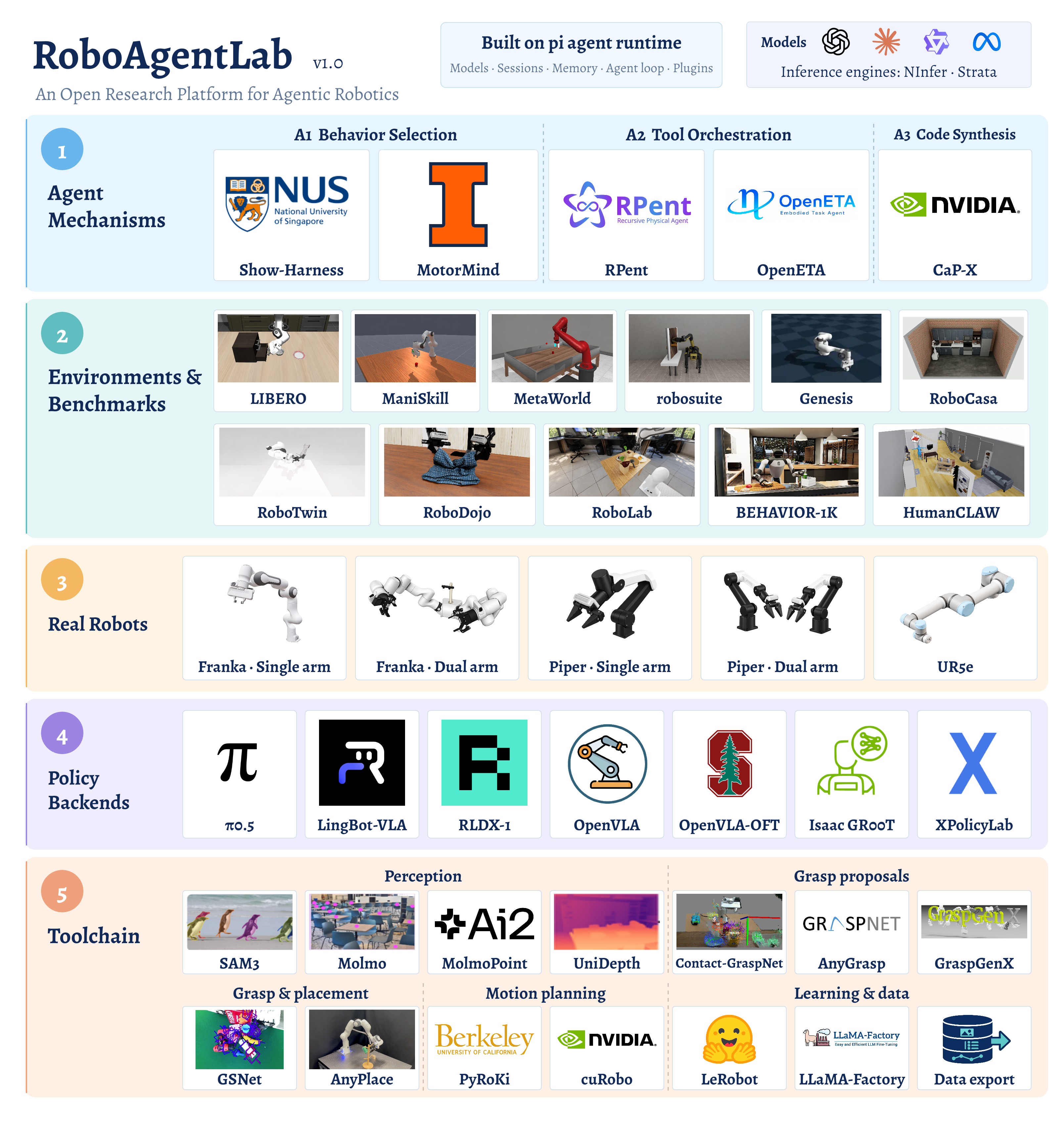}
\caption{Implemented platform interfaces: agent mechanisms, environments and benchmarks, physical robot adapters, policy backends, and tools. The reported runs use the ManiSkill adapter.}
\label{fig:platform-integrations}
\end{figure}
\endgroup

%% file: sections/11_evaluation_appendix.tex
\section{A Protocol for Evaluating Retained Improvements}
\label{app:process-evaluation}

\begingroup The following protocol operationalizes the process comparisons in Section~\ref{sec:future-evaluation}. It combines matched starting conditions with versioned revisions and complete-cycle cost accounting, extending fair-comparison principles~\citep{qin2026towardsagenticrobots} to systems that change between trials.\parfillskip=0pt plus 0.5\linewidth\par\endgroup

\begin{enumerate}
\item \textbf{Fix the starting conditions.} Record the initial model and component versions, task distribution, observation access, action interface, human input, and resource allowances. Compare an evolving system with a fixed-pipeline control from the same starting point. Independent restarts and removal of retained libraries can isolate the value of accumulated experience~\citep{wang2026sear}.
\item \textbf{Record each revision.} Preserve the parent and candidate versions, feedback that motivated the change, changed artifact or component, validation results, and acceptance decision. Link each accepted update to the later trials that use it. Keep development and held-out evaluation tasks separate.
\item \textbf{Measure stage-specific costs and outcomes.} For collection, report usable trajectories per robot-hour, coverage of identified failures, and downstream learning value per sample. For training, report data and compute required to reach a capability target, experiment completion, and recovery from unsuccessful runs. For planning, score the spatial grounding of proposed actions~\citep{jung2026groundedplanbench}. For execution, report success, latency, model and tool calls, steps, and recovery cost. Count communication, resets, repairs, and human interventions in the complete-cycle budget.
\item \textbf{Check physical outcomes independently.} Keep time-stamped observations and actions alongside state checks, constraint violations, and irreversible outcomes. Record who decides completion and what evidence that decision uses; an agent's stopping claim and an independently verified outcome are distinct measurements~\citep{openrua2026revisiting,sharma2026roboarcade,wan2026nofreechecker}.
\item \textbf{Test retention, transfer, and attribution.} Compare parent and accepted versions on held-out tasks and repeated runs. Report uncertainty across trials, regressions on previously solved tasks, and transfer to changed scenes or embodiments. Ablate individual updates and their interactions. Finally, compare the benefit with the total collection, training, and execution cost across successive cycles.
\end{enumerate}

%% file: main.bbl
\begin{thebibliography}{937}
\providecommand{\natexlab}[1]{#1}
\providecommand{\url}[1]{\texttt{#1}}
\providecommand{\urlprefix}{URL }
\expandafter\ifx\csname urlstyle\endcsname\relax
  \providecommand{\doi}[1]{DOI:\discretionary{}{}{}#1}\else
  \providecommand{\doi}{DOI:\discretionary{}{}{}\begingroup \urlstyle{rm}\Url}\fi

\bibitem[{AbuSadeh et~al.(2026)AbuSadeh, Wei and Zhang}]{si260630109}
AbuSadeh M, Wei L and Zhang D (2026) {TacEvo: Self-Evolving Architecture Discovery for Robotic Tactile Perception via LLM-Driven Quality-Diversity Search}.
\newblock \urlprefix\url{https://arxiv.org/abs/2606.30109}.

\bibitem[{Adami et~al.(2026)Adami, Zudaire, Carli and Falco}]{activeperceptionreal2sim2026}
Adami A, Zudaire S, Carli R and Falco P (2026) {Real2Sim via Active Perception with Behavior Trees Automatically Generated by VLMs}.
\newblock arXiv:2601.08454.
\newblock \urlprefix\url{https://arxiv.org/abs/2601.08454}.

\bibitem[{Agarwal et~al.(2026)Agarwal, Shen, Chintalapudi, Cai, Kumar, Curtis, Manderson, Lozano-Pérez and Kaelbling}]{agarwal2026feijoa}
Agarwal A, Shen W, Chintalapudi S, Cai L, Kumar N, Curtis A, Manderson T, Lozano-Pérez T and Kaelbling LP (2026) {FEIJOA: Task-Driven Grounding and Execution for Mobile Manipulation}.
\newblock CoRL 2026 Workshop on Agentic Robotics.
\newblock \urlprefix\url{https://openreview.net/forum?id=Eq3xq6KZDC}.

\bibitem[{{agenticros}(n.d.)}]{agenticros2026}
{agenticros} (n.d.) {AgenticROS}.
\newblock Software repository.
\newblock \urlprefix\url{https://github.com/Corebrum/corebrum-synapse}.
\newblock Accessed: 2026-09-30.

\bibitem[{{AgiBot Research Team}(2026)}]{agibot2026go3nano}
{AgiBot Research Team} (2026) {Recursive Self-Improvement for Robotic Manipulation in RoboDojo}.
\newblock Technical report.
\newblock \urlprefix\url{https://github.com/StevenWang30/XPolicyLab/blob/e4178ee754413cd792d3a37e0411ea45c58761c3/policy/GO_3_Nano_preview/go3_agent_preview/docs/Appendix_Recursive_Self-Improvement_for_Robotic_Manipulation_in_RoboDojo.pdf}.

\bibitem[{Ahmad et~al.(2025)Ahmad, Ismail, Styrud, Stenmark and Krueger}]{xref250315202}
Ahmad F, Ismail H, Styrud J, Stenmark M and Krueger V (2025) {A Unified Framework for Real-Time Failure Handling in Robotics Using Vision-Language Models, Reactive Planner and Behavior Trees}.
\newblock \emph{arXiv preprint arXiv:2503.15202} \urlprefix\url{https://arxiv.org/abs/2503.15202}.

\bibitem[{Ahn et~al.(2024)Ahn, Dwibedi, Finn, Gonzalez~Arenas, Gopalakrishnan, Hausman, Ichter, Irpan, Joshi, Julian, Kirmani, Leal, Lee, Levine, Lu, Maddineni, Rao, Sadigh, Sanketi, Sermanet, Vuong, Welker, Xia, Xiao, Xu, Xu and Xu}]{autort2024}
Ahn M, Dwibedi D, Finn C, Gonzalez~Arenas M, Gopalakrishnan K, Hausman K, Ichter B, Irpan A, Joshi N, Julian R, Kirmani S, Leal I, Lee E, Levine S, Lu Y, Maddineni S, Rao K, Sadigh D, Sanketi P, Sermanet P, Vuong Q, Welker S, Xia F, Xiao T, Xu P, Xu S and Xu Z (2024) {AutoRT: Embodied Foundation Models for Large Scale Orchestration of Robotic Agents}.
\newblock arXiv:2401.12963.
\newblock \urlprefix\url{https://arxiv.org/abs/2401.12963}.
\newblock Accepted to ICRA 2024 VLMNM Workshop.

\bibitem[{Ahn et~al.(2025)Ahn, Choi, Lee, Park and Woo}]{ahn2025nesyro}
Ahn S, Choi W, Lee J, Park J and Woo H (2025) {Towards Reliable Code-as-Policies: A Neuro-Symbolic Framework for Embodied Task Planning}.
\newblock \emph{arXiv preprint arXiv:2510.21302} \urlprefix\url{https://arxiv.org/abs/2510.21302}.
\newblock Accepted to NeurIPS 2025.

\bibitem[{{AI4AI-for-Anything}(2026)}]{ai4ai2026researchboost}
{AI4AI-for-Anything} (2026) {ResearchBoost-Agent}: An autonomous agent that profiles, optimizes, and validates {AI} training code.
\newblock GitHub repository.
\newblock \urlprefix\url{https://github.com/DAGroup-PKU/ResearchBoost-Agent}.
\newblock Version 8af019a.

\bibitem[{Ali(2026)}]{ali2026jevarm}
Ali F (2026) {Jev vs LLM: Robot Arm Control}.
\newblock Software repository.
\newblock \urlprefix\url{https://github.com/FazalAAli/jev-robotics-demo}.
\newblock Accessed: 2026-10-25.

\bibitem[{{AlmondGod}(2026)}]{almondrobotautoresearch2026}
{AlmondGod} (2026) {robot-autoresearch}.
\newblock GitHub repository.
\newblock \urlprefix\url{https://github.com/AlmondGod/robotics-autoresearch}.

\bibitem[{{amazedsaint}(2026)}]{jevduck2026}
{amazedsaint} (2026) {Jevduck}.
\newblock \urlprefix\url{https://github.com/amazedsaint/jevduck}.

\bibitem[{Anonymous(2026)}]{tacharness2026}
Anonymous (2026) {TacHarness: Agentic Infrastructure for Continually Improving Visuotactile Robot Policies}.
\newblock Manuscript under review.
\newblock \urlprefix\url{https://openreview.net/forum?id=V8hQBmyydD}.
\newblock Anonymous manuscript under review.

\bibitem[{{Anthropic}(2026{\natexlab{a}})}]{anthropic2026fable52}
{Anthropic} (2026{\natexlab{a}}) {Claude Fable 5.2 and Mythos 5.2}.
\newblock \urlprefix\url{https://www.anthropic.com/claude-fable-and-mythos-5-2}.

\bibitem[{{Anthropic}(2026{\natexlab{b}})}]{anthropic2026mhs}
{Anthropic} (2026{\natexlab{b}}) {Previewing the Model Hardware Standard}.
\newblock \urlprefix\url{https://www.anthropic.com/news/model-hardware-standard-research-preview}.

\bibitem[{Anwar et~al.(2024)Anwar, Welsh, Biswas, Pouya and Chang}]{xref240913682}
Anwar A, Welsh J, Biswas J, Pouya S and Chang Y (2024) {ReMEmbR: Building and Reasoning Over Long-Horizon Spatio-Temporal Memory for Robot Navigation}.
\newblock \emph{arXiv preprint arXiv:2409.13682} \urlprefix\url{https://arxiv.org/abs/2409.13682}.

\bibitem[{Ao et~al.(2026)Ao, Jiang, Zhong, Liu, Gao, Zhao, Zhang, Zhu, Bai and Li}]{ao2026smart}
Ao J, Jiang S, Zhong Y, Liu Y, Gao Y, Zhao J, Zhang Y, Zhu S, Bai C and Li X (2026) {SMART: Zero-Shot Sim-to-Real Articulated Object Manipulation via Large-Scale Synthetic Pretraining}.
\newblock \doi{10.48550/arXiv.2610.07652}.
\newblock \urlprefix\url{https://arxiv.org/abs/2610.07652}.

\bibitem[{Ao et~al.(2024)Ao, Wu, Wu, Swikir and Haddadin}]{xref240910444}
Ao J, Wu F, Wu Y, Swikir A and Haddadin S (2024) {LLM-as-BT-Planner: Leveraging LLMs for Behavior Tree Generation in Robot Task Planning}.
\newblock \emph{arXiv preprint arXiv:2409.10444} \urlprefix\url{https://arxiv.org/abs/2409.10444}.
\newblock Accepted to ICRA 2025.

\bibitem[{{Apex Intelligence}(2026)}]{apex2026apex35b}
{Apex Intelligence} (2026) {Introducing Apex-35B-V1.0}.
\newblock Technical blog.
\newblock \urlprefix\url{https://apexin.com/apex-35b/}.

\bibitem[{{Artificial Analysis}(2026{\natexlab{a}})}]{artificialanalysis2026aaroboticssim}
{Artificial Analysis} (2026{\natexlab{a}}) {Can language models drive a humanoid robot? Introducing AA-Robotics-Sim v0.1}.
\newblock \urlprefix\url{https://artificialanalysis.ai/articles/aa-robotics-sim}.

\bibitem[{{Artificial Analysis}(2026{\natexlab{b}})}]{artificialanalysis2026robotics}
{Artificial Analysis} (2026{\natexlab{b}}) {Welcome to the Era of Embodied AI: Introducing AA-Robotics v0.1}.
\newblock Online research preview.
\newblock \urlprefix\url{https://artificialanalysis.ai/articles/aa-robotics}.
\newblock Published October 14, 2026; accessed 2026-10-19.

\bibitem[{Ashley et~al.(2026)Ashley, Przepióra, Chen, Abualsaud, Yesmagambet, Park, Feron and Schmidhuber}]{xref260305621}
Ashley DR, Przepióra J, Chen Y, Abualsaud A, Yesmagambet N, Park S, Feron E and Schmidhuber J (2026) {RACAS: Controlling Diverse Robots With a Single Agentic System}.
\newblock \emph{arXiv preprint arXiv:2603.05621} \urlprefix\url{https://arxiv.org/abs/2603.05621}.

\bibitem[{{automatika-robotics}(n.d.)}]{embodiedagents2026}
{automatika-robotics} (n.d.) {Embodied Agents}.
\newblock Software repository.
\newblock \urlprefix\url{https://github.com/automatika-robotics/embodied-agents}.
\newblock Accessed: 2026-09-30.

\bibitem[{{Autonomous AI}(2026)}]{autonomous2026lampos}
{Autonomous AI} (2026) {lampOS}: The open-source operating system for physical {AI}.
\newblock GitHub repository.
\newblock \urlprefix\url{https://github.com/autonomous-ai/lampOS}.
\newblock Version 1b5f7a7.

\bibitem[{{Awomo-PhysicalRSI Team} et~al.(2026){Awomo-PhysicalRSI Team}, Ma, Ma, Liu, Jia, Shan, Xu, Zhang, Xu, Guo, Yu, Zhang, Jin, Luo, Niu, Li, Feng, Zhang, Wang, Wang, Huang, Huang, Wang and Ma}]{physicalrsiteam2026awomo}
{Awomo-PhysicalRSI Team}, Ma D, Ma E, Liu H, Jia H, Shan H, Xu J, Zhang J, Xu J, Guo K, Yu K, Zhang L, Jin L, Luo M, Niu P, Li S, Feng S, Zhang T, Wang T, Wang X, Huang X, Huang Y, Wang Z and Ma Z (2026) {Awomo-SimDataEngine: Agentic Simulation-ReadyWorld Generation}.
\newblock \emph{arXiv preprint arXiv:2610.02274} \urlprefix\url{https://arxiv.org/abs/2610.02274}.

\bibitem[{Bagherieh et~al.(2026)Bagherieh, Balaji, Chen, Gonzalez, Heravi, Islam, Li, Regli, Ren, Švec, Yoon, Müller and Koltun}]{chen2026amp1}
Bagherieh O, Balaji NNA, Chen C, Gonzalez JT, Heravi N, Islam F, Li X, Regli JB, Ren K, Švec P, Yoon DJ, Müller M and Koltun V (2026) {On Agentic Manipulation Policies}.
\newblock \urlprefix\url{https://arxiv.org/abs/2611.09892}.

\bibitem[{Bai et~al.(2026{\natexlab{a}})Bai, Bai, Zhang, Li, Hu, Huang and Liu}]{bai2026skillrsi}
Bai J, Bai G, Zhang C, Li S, Hu M, Huang W and Liu S (2026{\natexlab{a}}) {Compile Once, Run Cheap: Frozen, Portable Skills for Embodied Agents, and What Recursive Self-Improvement Adds}.
\newblock \emph{arXiv preprint arXiv:2610.25114} \doi{10.48550/arXiv.2610.25114}.
\newblock \urlprefix\url{https://arxiv.org/abs/2610.25114}.

\bibitem[{Bai et~al.(2026{\natexlab{b}})Bai, Li, Huang and Karlsson}]{si260830760}
Bai Z, Li S, Huang T and Karlsson BF (2026{\natexlab{b}}) {PRACTICE: From Experience to Expertise in Self-Evolving Embodied Agents}.
\newblock \urlprefix\url{https://arxiv.org/abs/2608.30760}.

\bibitem[{Banwasi et~al.(2026)Banwasi, Muckelroy~III, Sundaresan, Zhao, Bohg and Ho}]{banwasi2026messymem}
Banwasi A, Muckelroy~III W, Sundaresan P, Zhao L, Bohg J and Ho C (2026) {MessyMem: Learning-from-Doing Memory for Mobile Manipulation}.
\newblock \emph{arXiv preprint arXiv:2609.15976} \urlprefix\url{https://arxiv.org/abs/2609.15976}.
\newblock Accepted to CoRL 2026.

\bibitem[{Bao et~al.(2026)Bao, Huang, Lin, Feng, Li, Zhao, Li, He, Chi and Zhang}]{roboharnevo2026}
Bao S, Huang F, Lin Y, Feng Y, Li G, Zhao C, Li Y, He J, Chi C and Zhang J (2026) {RoboHarn-Evo: Evolving Hierarchical Physical Knowledge for Self-Improving Robotic Manipulation}.
\newblock \emph{arXiv preprint arXiv:2609.37583} \urlprefix\url{https://arxiv.org/abs/2609.37583}.

\bibitem[{B{\"a}rmann et~al.(2024)B{\"a}rmann, Kartmann, Peller-Konrad, Niehues, Waibel and Asfour}]{audit230904316}
B{\"a}rmann L, Kartmann R, Peller-Konrad F, Niehues J, Waibel A and Asfour T (2024) {Incremental learning of humanoid robot behavior from natural interaction and large language models}.
\newblock \emph{Frontiers in Robotics and AI} 11: 1455375.
\newblock \doi{10.3389/frobt.2024.1455375}.
\newblock \urlprefix\url{https://doi.org/10.3389/frobt.2024.1455375}.

\bibitem[{Baumann et~al.(2026)Baumann, Boyle, Deng, Ghignone, Sun, Pollefeys, Benini and Magno}]{audit260627871}
Baumann N, Boyle L, Deng P, Ghignone E, Sun B, Pollefeys M, Benini L and Magno M (2026) {LocalNav: Distilling Frontier VLMs and Embodied RL for On-Device Object Goal Navigation}.
\newblock arXiv:2606.27871.
\newblock \urlprefix\url{https://arxiv.org/abs/2606.27871v1}.

\bibitem[{Ben~Amor et~al.(2025)Ben~Amor, Graesser, Iscen, D'Ambrosio, Abeyruwan, Bewley, Zhou, Kalirathinam, Mishra and Sanketi}]{benamor2025sasprompt}
Ben~Amor H, Graesser L, Iscen A, D'Ambrosio D, Abeyruwan S, Bewley A, Zhou Y, Kalirathinam K, Mishra S and Sanketi P (2025) {SAS-Prompt: Large Language Models as Numerical Optimizers for Robot Self-Improvement}.
\newblock \emph{arXiv preprint arXiv:2504.20459} \urlprefix\url{https://arxiv.org/abs/2504.20459}.
\newblock Accepted to ICRA 2025.

\bibitem[{Benaharon et~al.(2026)Benaharon, Zhang, Zheng, Zhang and Wei}]{benaharon2026coadaptive}
Benaharon L, Zhang C, Zheng W, Zhang W and Wei Y (2026) {Program-Level Adversarial Co-Adaptive Self-Play for Robotic Security}.
\newblock CoRL 2026 Workshop on Agentic Robotics.
\newblock \urlprefix\url{https://openreview.net/forum?id=LkW6BOlHMY}.

\bibitem[{Berman et~al.(2026)Berman, Ilie, Deng and Freeman}]{berman2026embody}
Berman S, Ilie M, Deng J and Freeman D (2026) {Claude plays robotics}.
\newblock Anthropic Research.
\newblock \urlprefix\url{https://www.anthropic.com/research/claude-plays-robotics}.
\newblock Accessed: 2026-10-25.

\bibitem[{Bhat et~al.(2025)Bhat, Kaypak, Krishnamurthy, Karri and Khorrami}]{xref240208546}
Bhat V, Kaypak AU, Krishnamurthy P, Karri R and Khorrami F (2025) {Grounding Large Language Models for Robot Task Planning Using Closed‐Loop State Feedback}.
\newblock \emph{Advanced Robotics Research} : e202500072\doi{10.1002/adrr.202500072}.
\newblock \urlprefix\url{https://doi.org/10.1002/adrr.202500072}.

\bibitem[{Bhat et~al.(2026)Bhat, Uy, Chen, Song, Zook, Yang, Birchfield, Blukis and Tremblay}]{audit260817129}
Bhat V, Uy MA, Chen S, Song CH, Zook A, Yang X, Birchfield S, Blukis V and Tremblay J (2026) {MG-VQA: Manipulation Grounded Visual Question Answering with VLMs}.
\newblock arXiv:2608.17129.
\newblock \urlprefix\url{https://arxiv.org/abs/2608.17129}.

\bibitem[{Bjorck et~al.(2026)Bjorck, Li, Man, Wang, Cheng, Liu, Wang, Yu, Badki, Birchfield, Blukis, Chebotar, Chen, Leng, Chou, Ding, Li, Luo, Su, Tremblay, Wang, Wen, Wu, Xie, Ye, Yin, Zentner, Gui, Wang, Zhu, Fan and Kautz}]{bjorck2026vesta}
Bjorck J, Li Z, Man Y, Wang J, Cheng AC, Liu S, Wang S, Yu Z, Badki A, Birchfield S, Blukis V, Chebotar Y, Chen S, Leng S, Chou YC, Ding T, Li B, Luo Z, Su H, Tremblay J, Wang T, Wen B, Wu J, Xie X, Ye H, Yin H, Zentner KR, Gui L, Wang YX, Zhu Y, Fan LJ and Kautz J (2026) {Vesta: A Generalist Embodied Reasoning Model}.
\newblock \urlprefix\url{https://arxiv.org/abs/2606.20905}.

\bibitem[{Black et~al.(2024)Black, Nakamoto, Atreya, Walke, Finn, Kumar and Levine}]{black2024susie}
Black K, Nakamoto M, Atreya P, Walke H, Finn C, Kumar A and Levine S (2024) Zero-shot robotic manipulation with pretrained image-editing diffusion models.
\newblock In: \emph{International Conference on Learning Representations}.
\newblock \urlprefix\url{https://openreview.net/forum?id=c0chJTSbci}.

\bibitem[{Boiko et~al.(2023)Boiko, MacKnight, Kline and Gomes}]{boiko2023coscientist}
Boiko DA, MacKnight R, Kline B and Gomes G (2023) Autonomous chemical research with large language models.
\newblock \emph{Nature} 624(7992): 570--578.
\newblock \doi{10.1038/s41586-023-06792-0}.
\newblock \urlprefix\url{https://doi.org/10.1038/s41586-023-06792-0}.

\bibitem[{Boyle et~al.(2025)Boyle, Baumann, Sivasothilingam, Magno and Benini}]{boyle2025robotxr1}
Boyle L, Baumann N, Sivasothilingam P, Magno M and Benini L (2025) {RobotxR1: Enabling Embodied Robotic Intelligence on Large Language Models through Closed-Loop Reinforcement Learning}.
\newblock \doi{10.48550/arXiv.2505.03238}.
\newblock \urlprefix\url{https://arxiv.org/abs/2505.03238}.

\bibitem[{{bryercowan}(2026)}]{hermesembodied2026}
{bryercowan} (2026) {Hermes Embodied: Self-Improving Robotics via Hermes Agent}.
\newblock GitHub repository.
\newblock \urlprefix\url{https://github.com/bryercowan/hermes-embodied}.

\bibitem[{Buchholz et~al.(2025{\natexlab{a}})Buchholz, Carlucho, Grimaldi and Petillot}]{audit250723735}
Buchholz M, Carlucho I, Grimaldi M and Petillot YR (2025{\natexlab{a}}) {Distributed AI Agents for Cognitive Underwater Robot Autonomy}.
\newblock \emph{arXiv preprint arXiv:2507.23735} \doi{10.48550/arXiv.2507.23735}.
\newblock \urlprefix\url{https://arxiv.org/abs/2507.23735}.

\bibitem[{Buchholz et~al.(2025{\natexlab{b}})Buchholz, Ellis, Abu~Hara, Carlucho and Petillot}]{auradiagnostics2025}
Buchholz M, Ellis N, Abu~Hara R, Carlucho I and Petillot YR (2025{\natexlab{b}}) {A Collaborative Reasoning Framework for Anomaly Diagnostics in Underwater Robotics}.
\newblock arXiv:2511.03075.
\newblock \urlprefix\url{https://arxiv.org/abs/2511.03075}.

\bibitem[{Burns et~al.(2024)Burns, Jain, Go, Xia, Stark, Schaal and Hausman}]{xref240406645}
Burns K, Jain A, Go K, Xia F, Stark M, Schaal S and Hausman K (2024) {GenCHiP: Generating Robot Policy Code for High-Precision and Contact-Rich Manipulation Tasks}.
\newblock \emph{arXiv preprint arXiv:2404.06645} \urlprefix\url{https://arxiv.org/abs/2404.06645}.

\bibitem[{Büchner et~al.(2026)Büchner, Röfer, Engelbracht, Welschehold, Bauer, Blum, Pollefeys and Valada}]{buchner2026momasg}
Büchner M, Röfer A, Engelbracht T, Welschehold T, Bauer Z, Blum H, Pollefeys M and Valada A (2026) {Articulated 3D Scene Graphs for Open-World Mobile Manipulation}.
\newblock arXiv:2602.16356.
\newblock \urlprefix\url{https://arxiv.org/abs/2602.16356}.

\bibitem[{Cai et~al.(2026{\natexlab{a}})Cai, Cai, Chen, Chen, Chen, Feng, Hou, Huang, Jiang, Ju, Li, Li, Li, Li, Li, Liu, Liu, Luo, Miao, Ni, Nie, Pan, Ren, Song, Wang, Wang, Wang, Wang, Wei, Wu, Xie, Yang, Ye, Yue, Zhang, Zhang, Zhao, Zhou and Zhou}]{audit260816885}
Cai X, Cai Y, Chen B, Chen J, Chen Z, Feng S, Hou T, Huang J, Jiang H, Ju R, Li D, Li M, Li S, Li X, Li Y, Liu Y, Liu Z, Luo J, Miao J, Ni R, Nie B, Pan M, Ren X, Song J, Wang J, Wang P, Wang S, Wang X, Wei D, Wu D, Xie P, Yang P, Ye H, Yue X, Zhang J, Zhang Q, Zhao X, Zhou P and Zhou Y (2026{\natexlab{a}}) {$\tau_0$-VLA: a Hierarchical Robot Foundation Model with World-Model-Guided Test-Time Computation}.
\newblock \emph{arXiv preprint arXiv:2608.16885} \doi{10.48550/arXiv.2608.16885}.
\newblock \urlprefix\url{https://arxiv.org/abs/2608.16885}.

\bibitem[{Cai et~al.(2026{\natexlab{b}})Cai, Zhu, Shang, Li, Pollefeys, Wu, Vedaldi and Zheng}]{cai2026artisan}
Cai X, Zhu R, Shang H, Li R, Pollefeys M, Wu S, Vedaldi A and Zheng C (2026{\natexlab{b}}) {Towards Physically Grounded Articulated 3D Assets: A Multi-Agent Pipeline and Benchmark}.
\newblock \emph{arXiv preprint arXiv:2610.13947} \doi{10.48550/arXiv.2610.13947}.
\newblock \urlprefix\url{https://arxiv.org/abs/2610.13947}.

\bibitem[{Cai et~al.(2026{\natexlab{c}})Cai, Wang, Han, Cheng, Liu and Wang}]{cai2026esi}
Cai Y, Wang Z, Han C, Cheng Y, Liu H and Wang Z (2026{\natexlab{c}}) {ESI: An Embodied Self-Improvement Flywheel for LLM-VLA Agents}.
\newblock \urlprefix\url{https://arxiv.org/abs/2611.18176}.

\bibitem[{Cardenas et~al.(2026)Cardenas, Arnett, Yeo, Sah and Kim}]{cardenas2026rosclaw}
Cardenas IS, Arnett MA, Yeo NC, Sah L and Kim JH (2026) {ROSClaw: An OpenClaw ROS 2 Framework for Agentic Robot Control and Interaction}.
\newblock \emph{arXiv preprint arXiv:2603.26997} \urlprefix\url{https://arxiv.org/abs/2603.26997}.

\bibitem[{Chai et~al.(2026)Chai, Shen, Yao, Xia, Zhao, Ma, Lin and Wang}]{si260618235}
Chai Q, Shen W, Yao N, Xia Y, Zhao K, Ma J, Lin G and Wang H (2026) {EvolveNav: Proactive Preflection and Self-Evolving Memory for Zero-Shot Object Goal Navigation}.
\newblock \urlprefix\url{https://arxiv.org/abs/2606.18235}.

\bibitem[{Chan et~al.(2026{\natexlab{a}})Chan, Padarath, Kwon, Greaves and Anderljung}]{chan2026measuringairda}
Chan A, Padarath R, Kwon J, Greaves H and Anderljung M (2026{\natexlab{a}}) Measuring {AI R\&D} automation.
\newblock \emph{arXiv preprint arXiv:2603.03992} \urlprefix\url{https://arxiv.org/abs/2603.03992}.

\bibitem[{Chan et~al.(2026{\natexlab{b}})Chan, Winter, Barto, Pachocki, Hinton, Horvitz, Bengio, Song, Clark, Greaves, Korinek, Hammond, Graepel, Bariach, Torr, McIlraith, Clune, Manning, Sastry, Davidson, Eth and Mindermann}]{chan2026intelligenceexplosion}
Chan A, Winter C, Barto A, Pachocki J, Hinton G, Horvitz E, Bengio Y, Song D, Clark J, Greaves H, Korinek A, Hammond S, Graepel T, Bariach B, Torr PHS, McIlraith SA, Clune J, Manning S, Sastry G, Davidson T, Eth D and Mindermann S (2026{\natexlab{b}}) What if automating {AI R\&D} triggers an intelligence explosion?
\newblock \emph{arXiv preprint arXiv:2609.36054} \urlprefix\url{https://arxiv.org/abs/2609.36054}.

\bibitem[{Chan et~al.(2026{\natexlab{c}})Chan, Zhang, Liu, Cui, Zeng, Ding, Ren, Zhang, Chen, Liu, Dai and Ju}]{si260518729}
Chan NT, Zhang Y, Liu Y, Cui R, Zeng F, Ding Z, Ren X, Zhang Z, Chen Q, Liu J, Dai Y and Ju X (2026{\natexlab{c}}) {Robo-Cortex: A Self-Evolving Embodied Agent via Dual-Grain Cognitive Memory and Autonomous Knowledge Induction}.
\newblock \urlprefix\url{https://arxiv.org/abs/2605.18729}.

\bibitem[{Chang et~al.(2026)Chang, Wu, Wu, Huang, Yang and Wang}]{chang2026scip}
Chang KP, Wu CH, Wu YS, Huang YH, Yang KC and Wang YCF (2026) {Self-Correcting Imaginative Planning for VLM Embodied Agents}.
\newblock CoRL 2026 Workshop on Continual Self-Improvement in Robotics.
\newblock \urlprefix\url{https://openreview.net/forum?id=aPS3938Wc2}.

\bibitem[{Chang et~al.(2024)Chang, Chhablani, Clegg, Cote, Desai, Hlavac, Karashchuk, Krantz, Mottaghi, Parashar, Patki, Prasad, Puig, Rai, Ramrakhya, Tran, Truong, Turner, Undersander and Yang}]{partnr}
Chang M, Chhablani G, Clegg A, Cote MD, Desai R, Hlavac M, Karashchuk V, Krantz J, Mottaghi R, Parashar P, Patki S, Prasad I, Puig X, Rai A, Ramrakhya R, Tran D, Truong J, Turner JM, Undersander E and Yang TY (2024) {PARTNR: A Benchmark for Planning and Reasoning in Embodied Multi-agent Tasks}.
\newblock arXiv:2411.00081.
\newblock \urlprefix\url{https://arxiv.org/abs/2411.00081}.

\bibitem[{Chen et~al.(2026{\natexlab{a}})Chen, Fang and Liu}]{chen2026astrarobomme}
Chen B, Fang H and Liu CK (2026{\natexlab{a}}) {Can Astra Solve RoboMME without Breaking the Bank?}
\newblock \emph{arXiv preprint arXiv:2611.06184} \doi{10.48550/arXiv.2611.06184}.
\newblock \urlprefix\url{https://arxiv.org/abs/2611.06184}.

\bibitem[{Chen et~al.(2022)Chen, Xia, Ichter, Rao, Gopalakrishnan, Ryoo, Stone and Kappler}]{chen2022nlmap}
Chen B, Xia F, Ichter B, Rao K, Gopalakrishnan K, Ryoo MS, Stone A and Kappler D (2022) {Open-vocabulary Queryable Scene Representations for Real World Planning}.
\newblock \emph{arXiv preprint arXiv:2209.09874} \urlprefix\url{https://arxiv.org/abs/2209.09874}.

\bibitem[{Chen et~al.(2026{\natexlab{b}})Chen, Jain, Goossen, Romig, Murray, Walker and Cakmak}]{chen2026architect}
Chen D, Jain AR, Goossen E, Romig E, Murray M, Walker N and Cakmak M (2026{\natexlab{b}}) {A Few Words Go a Long Way: Language Guided Robot Policy Synthesis}.
\newblock \emph{arXiv preprint arXiv:2607.23784} \urlprefix\url{https://arxiv.org/abs/2607.23784}.

\bibitem[{Chen et~al.(2025{\natexlab{a}})Chen, Xu, Chu, Zhou, Sun, Wu, Gao, Li, Yang and Ma}]{chen2025gendexhand}
Chen F, Xu Z, Chu T, Zhou X, Sun L, Wu Z, Gao S, Li Z, Yang Y and Ma Y (2025{\natexlab{a}}) {GenDexHand: Generative Simulation for Dexterous Hands}.
\newblock \doi{10.48550/arXiv.2511.01791}.
\newblock \urlprefix\url{https://arxiv.org/abs/2511.01791}.

\bibitem[{Chen et~al.(2026{\natexlab{c}})Chen, Zhou, He, Yao, Sun, Gao, Ding, Zhang, Zhou and Ma}]{chen2026embodiedplayground}
Chen F, Zhou P, He L, Yao Y, Sun L, Gao S, Ding M, Zhang K, Zhou K and Ma Y (2026{\natexlab{c}}) {Embodied Playground: A Long-Horizon Agentic Robotics Benchmark for Physical, Strategic, and Spatial Reasoning}.
\newblock \urlprefix\url{https://arxiv.org/abs/2610.27039}.

\bibitem[{Chen et~al.(2026{\natexlab{d}})Chen, Qu, Xia, Jing, Cheng, Ma, Yang, Zhou, Li, Suri, Sun, Liu, Chen, Zeng and Shi}]{diaggen2026}
Chen G, Qu Y, Xia Q, Jing P, Cheng Y, Ma B, Yang P, Zhou B, Li Z, Suri S, Sun G, Liu C, Chen PY, Zeng Z and Shi F (2026{\natexlab{d}}) {DiagGen: Agentic Generation of Deformable Assets with Sim-based Diagnostics for Robotic Simulation}.
\newblock \emph{arXiv preprint arXiv:2609.23103} \urlprefix\url{https://arxiv.org/abs/2609.23103}.

\bibitem[{Chen et~al.(2026{\natexlab{e}})Chen, Schott and Zhong}]{chen2026typego}
Chen G, Schott A and Zhong L (2026{\natexlab{e}}) {TypeGo: An OS Runtime for Embodied Agents}.
\newblock \doi{10.48550/arXiv.2607.05482}.
\newblock \urlprefix\url{https://arxiv.org/abs/2607.05482}.

\bibitem[{Chen et~al.(2026{\natexlab{f}})Chen, Xia, Peng, Zhang, Jing, Ma, Qian, Cheng, Jiao, Zhou, Qu, Ye, Zhang, Wang, Zeng, Chen, Yang, Zeng, Luo, Wang, Liu, Yuille, Shi, Zheng, Li, Jiang and Chen}]{agenticreal2sim2026}
Chen G, Xia Q, Peng J, Zhang H, Jing P, Ma B, Qian J, Cheng Y, Jiao Z, Zhou B, Qu Y, Ye L, Zhang K, Wang K, Zeng W, Chen Y, Yang P, Zeng Z, Luo S, Wang H, Liu C, Yuille A, Shi F, Zheng C, Li Y, Jiang C and Chen PY (2026{\natexlab{f}}) {Agentic Real2Sim: Physics-based World Modeling with Vision-Language Agents}.
\newblock \emph{arXiv preprint arXiv:2607.19190} \urlprefix\url{https://arxiv.org/abs/2607.19190}.

\bibitem[{Chen et~al.(2024{\natexlab{a}})Chen, Yang, Jia, Hu, Chen, Zhang, Wang and Pan}]{chen2024lasp}
Chen G, Yang L, Jia R, Hu Z, Chen Y, Zhang W, Wang W and Pan J (2024{\natexlab{a}}) {Language-Augmented Symbolic Planner for Open-World Task Planning}.
\newblock \emph{arXiv preprint arXiv:2407.09792} \urlprefix\url{https://arxiv.org/abs/2407.09792}.
\newblock Accepted to RSS 2024.

\bibitem[{Chen et~al.(2026{\natexlab{g}})Chen, Guo, Wang, Zhang, Huang, Zheng, Hou, Tie, Deng and Shao}]{chen2025goalvla}
Chen H, Guo J, Wang B, Zhang T, Huang X, Zheng B, Hou Y, Tie C, Deng J and Shao L (2026{\natexlab{g}}) Goal-vla: Image-generative vlms as object-centric world models empowering zero-shot robot manipulation.
\newblock In: \emph{2026 IEEE International Conference on Robotics and Automation (ICRA)}. pp. 10323--10330.
\newblock \doi{10.1109/ICRA57385.2026.11696379}.
\newblock \urlprefix\url{https://arxiv.org/abs/2506.23919}.

\bibitem[{Chen et~al.(2024{\natexlab{b}})Chen, Yao, Liu, Liu and Ichnowski}]{xref240903966}
Chen H, Yao Y, Liu R, Liu C and Ichnowski J (2024{\natexlab{b}}) {Automating Robot Failure Recovery Using Vision-Language Models With Optimized Prompts}.
\newblock \emph{arXiv preprint arXiv:2409.03966} \urlprefix\url{https://arxiv.org/abs/2409.03966}.

\bibitem[{Chen et~al.(2026{\natexlab{h}})Chen, Zhang, Schaefer, Chen, Chen, Cremers, Mees and Leutenegger}]{si260621406}
Chen H, Zhang A, Schaefer S, Chen K, Chen S, Cremers D, Mees O and Leutenegger S (2026{\natexlab{h}}) {Robot Self-Improvement via Human-Video Dynamics Models}.
\newblock \urlprefix\url{https://arxiv.org/abs/2606.21406}.

\bibitem[{Chen et~al.(2026{\natexlab{i}})Chen, Zou, Ma, Pu, Zhang, Chen and Yang}]{chen2026agps}
Chen H, Zou Z, Ma C, Pu Y, Zhang H, Chen Y and Yang Y (2026{\natexlab{i}}) {Accelerating Robotic Reinforcement Learning with Agent Guidance}.
\newblock \urlprefix\url{https://arxiv.org/abs/2602.11978}.

\bibitem[{Chen et~al.(2026{\natexlab{j}})Chen, Chen, Xu, Ge and Chen}]{chen2026roboevolve}
Chen HH, Chen S, Xu Y, Ge W and Chen YC (2026{\natexlab{j}}) Roboevolve: Co-evolving planner-simulator for robotic manipulation with limited data.
\newblock \urlprefix\url{https://arxiv.org/abs/2605.13775}.

\bibitem[{Chen et~al.(2024{\natexlab{c}})Chen, Lin, Xu, Chai, Liang and Wong}]{xref240107314}
Chen J, Lin B, Xu R, Chai Z, Liang X and Wong KYK (2024{\natexlab{c}}) {MapGPT: Map-Guided Prompting with Adaptive Path Planning for Vision-and-Language Navigation}.
\newblock \emph{arXiv preprint arXiv:2401.07314} \urlprefix\url{https://arxiv.org/abs/2401.07314}.
\newblock Accepted to ACL 2024.

\bibitem[{Chen et~al.(2024{\natexlab{d}})Chen, Mu, Yu, Wei, Wu, Yuan, Liang, Yang, Zhang, Shao, Qiao, Xu, Ding and Luo}]{grid58}
Chen J, Mu Y, Yu Q, Wei T, Wu S, Yuan Z, Liang Z, Yang C, Zhang K, Shao W, Qiao Y, Xu H, Ding M and Luo P (2024{\natexlab{d}}) {RoboScript: Code Generation for Free-Form Manipulation Tasks across Real and Simulation}.
\newblock arXiv:2402.14623.
\newblock \urlprefix\url{https://arxiv.org/abs/2402.14623}.

\bibitem[{Chen et~al.(2024{\natexlab{e}})Chen, Yu, Zhou, Xu, Mu, Hu, Shao, Wang, Li and Shao}]{grid41}
Chen J, Yu C, Zhou X, Xu T, Mu Y, Hu M, Shao W, Wang Y, Li G and Shao L (2024{\natexlab{e}}) {EMOS: Embodiment-aware Heterogeneous Multi-robot Operating System with LLM Agents}.
\newblock arXiv:2410.22662.
\newblock \urlprefix\url{https://arxiv.org/abs/2410.22662}.

\bibitem[{Chen et~al.(2026{\natexlab{k}})Chen, Zhang, Yao, Ni, Ji, Peng, Wang and Chen}]{chen2026aidsr}
Chen J, Zhang M, Yao L, Ni Y, Ji J, Peng J, Wang S and Chen T (2026{\natexlab{k}}) Bridging language and physics: Automated design of continuum robots with large language models.
\newblock In: \emph{Proceedings of Robotics: Science and Systems}, volume~22.
\newblock \doi{10.15607/RSS.2026.XXII.194}.
\newblock \urlprefix\url{https://www.roboticsproceedings.org/rss22/p194.html}.

\bibitem[{Chen et~al.(2026{\natexlab{l}})Chen, Xie, Fu, Yu, Pacini, Bajamahal, Kim, Drake, Kim, Xue, Francis, Juette, Schaldenbrand, Seker, Wickramarachchi, Yoo, Wang, Murali, Sundaralingam, Sastry, Zhu, Fan, Huang and Goldberg}]{chen2026gap}
Chen K, Xie S, Fu L, Yu J, Pacini W, Bajamahal S, Kim H, Drake J, Kim D, Xue H, Francis J, Juette C, Schaldenbrand P, Seker MY, Wickramarachchi R, Yoo U, Wang G, Murali A, Sundaralingam B, Sastry SS, Zhu Y, Fan LJ, Huang S and Goldberg K (2026{\natexlab{l}}) {GaP: A Graph-as-Policy Multi-Agent Self-Learning Harness For Variational Automation Tasks}.
\newblock \emph{arXiv preprint arXiv:2607.05369} \urlprefix\url{https://arxiv.org/abs/2607.05369}.
\newblock Accepted to CoRL 2026.

\bibitem[{Chen et~al.(2026{\natexlab{m}})Chen, Xie, Yoo, Zhu, Magner, Zhu, Kim, Heiden, Coumans, Macklin, Huang and Goldberg}]{chen2026robosimstudio}
Chen K, Xie S, Yoo U, Zhu X, Magner C, Zhu A, Kim D, Heiden E, Coumans E, Macklin M, Huang S and Goldberg K (2026{\natexlab{m}}) {RoboSimStudio: An Interactive Toolkit for Human and Agent Rapid Real2Sim of Robot Manipulation Tasks}.
\newblock \emph{arXiv preprint arXiv:2610.18835} \urlprefix\url{https://arxiv.org/abs/2610.18835}.

\bibitem[{Chen et~al.(2025{\natexlab{b}})Chen, Moorman and Gombolay}]{elemental2025}
Chen L, Moorman NM and Gombolay MC (2025{\natexlab{b}}) {ELEMENTAL: Interactive Learning from Demonstrations and Vision-Language Models for Reward Design in Robotics}.
\newblock In: \emph{Proceedings of the 42nd International Conference on Machine Learning}, \emph{Proceedings of Machine Learning Research}, volume 267. pp. 8700--8725.
\newblock \urlprefix\url{https://proceedings.mlr.press/v267/chen25at.html}.

\bibitem[{Chen et~al.(2026{\natexlab{n}})Chen, Liang, Yeh, Su, Chen, Sun and Hsu}]{chen2026adapt}
Chen PA, Liang YC, Yeh JF, Su HT, Chen YT, Sun M and Hsu W (2026{\natexlab{n}}) Adapt: Benchmarking commonsense planning under unspecified affordance constraints.
\newblock \urlprefix\url{https://arxiv.org/abs/2604.14902}.

\bibitem[{Chen et~al.(2025{\natexlab{c}})Chen, Sun, Wang, Lv, Zhang and Zeng}]{chen2025safemind}
Chen R, Sun Y, Wang J, Lv M, Zhang Q and Zeng Y (2025{\natexlab{c}}) {SafeMind: Benchmarking and Mitigating Safety Risks in Embodied LLM Agents}.
\newblock \urlprefix\url{https://arxiv.org/abs/2509.25885}.

\bibitem[{Chen et~al.(2026{\natexlab{o}})Chen, Hadfield, Zook, Uy, Song, Coumans, Yang, Ladhak, Qu, Birchfield, Tremblay and Blukis}]{grid23}
Chen S, Hadfield H, Zook A, Uy MA, Song CH, Coumans E, Yang X, Ladhak F, Qu Q, Birchfield S, Tremblay J and Blukis V (2026{\natexlab{o}}) {VoLo: A Physical Orchestrator for Open-Vocabulary Long-Horizon Manipulation}.
\newblock arXiv:2606.07723.
\newblock \urlprefix\url{https://arxiv.org/abs/2606.07723}.

\bibitem[{Chen et~al.(2026{\natexlab{p}})Chen, Huang, Wang, Huang, Li, Zhang, Li and Wu}]{chen2026behaviorspatial}
Chen S, Huang H, Wang Q, Huang W, Li M, Zhang R, Li FF and Wu J (2026{\natexlab{p}}) {BEHAVIOR-Spatial Benchmark: Evaluating Spatial Understanding for Robotics}.
\newblock arXiv:2610.34051.
\newblock \urlprefix\url{https://arxiv.org/abs/2610.34051}.

\bibitem[{Chen et~al.(2023{\natexlab{a}})Chen, Xiao and Hsu}]{grid54}
Chen S, Xiao A and Hsu D (2023{\natexlab{a}}) {LLM-State: Open World State Representation for Long-horizon Task Planning with Large Language Model}.
\newblock arXiv:2311.17406.
\newblock \urlprefix\url{https://arxiv.org/abs/2311.17406}.

\bibitem[{Chen et~al.(2026{\natexlab{q}})Chen, Zargarbashi, Cheng, An and Coros}]{chen2026kino}
Chen S, Zargarbashi F, Cheng J, An T and Coros S (2026{\natexlab{q}}) {KINO: A Keyframe Interface for VLM Planning and Whole-Body Control in Humanoid Loco-Manipulation}.
\newblock \urlprefix\url{https://arxiv.org/abs/2609.18869}.

\bibitem[{Chen(2026)}]{chen2026impossibletwins}
Chen T (2026) {Impossible Twins: Judge-Free Measurement of Verifier False Positives in Agentic Robot Controller Search}.
\newblock CoRL 2026 Workshop on Agentic Robotics.
\newblock \urlprefix\url{https://openreview.net/forum?id=uMkXmX3mAy}.

\bibitem[{Chen et~al.(2026{\natexlab{r}})Chen, Chen, Chen, Cai, Liu, Li, Liang, Lin, Ge, Gu, Deng, Guo, Nian, Xie, Chen, Su, Xu, Liu, Hu, Gao, Wang, Liang, Qin, Yang, Luo and Mu}]{robotwin2agent2025}
Chen T, Chen Z, Chen B, Cai Z, Liu Y, Li Z, Liang Q, Lin X, Ge Y, Gu Z, Deng W, Guo Y, Nian T, Xie X, Chen Q, Su K, Xu T, Liu G, Hu M, Gao Ha, Wang K, Liang Z, Qin Y, Yang X, Luo P and Mu Y (2026{\natexlab{r}}) {R}obo{T}win 2.0: A scalable data generator and benchmark with strong domain randomization for robust bimanual robotic manipulation.
\newblock In: \emph{Proceedings of the 43rd International Conference on Machine Learning}, \emph{Proceedings of Machine Learning Research}, volume 306. PMLR, pp. 13673--13699.
\newblock \urlprefix\url{https://proceedings.mlr.press/v306/chen26j.html}.

\bibitem[{Chen et~al.(2026{\natexlab{s}})Chen, Wang, Li, Qin, Shi, Li, Hu, Zhang, Wang, Chen, Wang, Wang, Yang, Xu, Wu, Mu, Yang, Dong and Luo}]{chen2026rmbench}
Chen T, Wang Y, Li M, Qin Y, Shi H, Li Z, Hu Y, Zhang Y, Wang K, Chen Y, Wang H, Wang J, Yang T, Xu R, Wu R, Mu Y, Yang Y, Dong H and Luo P (2026{\natexlab{s}}) {RMBench}: Memory-dependent robotic manipulation benchmark with insights into policy design.
\newblock arXiv:2603.01229.
\newblock \urlprefix\url{https://arxiv.org/abs/2603.01229}.

\bibitem[{Chen et~al.(2026{\natexlab{t}})Chen, Bhatia, Glossop, Mathihalli, Doshi, Tang, Driess, Pertsch and Levine}]{audit260213193}
Chen W, Bhatia JS, Glossop C, Mathihalli N, Doshi R, Tang A, Driess D, Pertsch K and Levine S (2026{\natexlab{t}}) {Steerable Vision-Language-Action Policies for Embodied Reasoning and Hierarchical Control}.
\newblock arXiv:2602.13193.
\newblock \urlprefix\url{https://arxiv.org/abs/2602.13193v1}.

\bibitem[{Chen et~al.(2025{\natexlab{d}})Chen, Li, Li, Wang, Li, Xue, Zhao and Liu}]{chen2025exoskeleton}
Chen W, Li G, Li M, Wang W, Li P, Xue X, Zhao X and Liu L (2025{\natexlab{d}}) {LLM-Enabled Incremental Learning Framework for Hand Exoskeleton Control}.
\newblock \emph{IEEE Transactions on Automation Science and Engineering} 22: 2617--2626.
\newblock \doi{10.1109/TASE.2024.3382679}.
\newblock \urlprefix\url{https://doi.org/10.1109/TASE.2024.3382679}.

\bibitem[{Chen et~al.(2024{\natexlab{f}})Chen, Arkin, Dawson, Zhang, Roy and Fan}]{chen2023autotamp}
Chen Y, Arkin J, Dawson C, Zhang Y, Roy N and Fan C (2024{\natexlab{f}}) {AutoTAMP: Autoregressive Task and Motion Planning with LLMs as Translators and Checkers}.
\newblock In: \emph{IEEE International Conference on Robotics and Automation}. pp. 6695--6702.
\newblock \doi{10.1109/icra57147.2024.10611163}.
\newblock \urlprefix\url{https://doi.org/10.1109/icra57147.2024.10611163}.

\bibitem[{Chen et~al.(2024{\natexlab{g}})Chen, Arkin, Zhang, Roy and Fan}]{xref230915943}
Chen Y, Arkin J, Zhang Y, Roy N and Fan C (2024{\natexlab{g}}) {Scalable Multi-Robot Collaboration with Large Language Models: Centralized or Decentralized Systems?}
\newblock In: \emph{IEEE International Conference on Robotics and Automation}. pp. 4311--4317.
\newblock \doi{10.1109/icra57147.2024.10610676}.
\newblock \urlprefix\url{https://doi.org/10.1109/icra57147.2024.10610676}.

\bibitem[{Chen et~al.(2026{\natexlab{u}})Chen, Bai, Cao, Zeng, Lin, Lin, Liang, Ma, Huang and Shou}]{showharness}
Chen Y, Bai Z, Cao Z, Zeng W, Lin KQ, Lin Y, Liang G, Ma KY, Huang Q and Shou MZ (2026{\natexlab{u}}) {Show-Harness: Just a VLM Agent Can Play Robots}.
\newblock arXiv:2609.10522.
\newblock \urlprefix\url{https://arxiv.org/abs/2609.10522}.

\bibitem[{Chen et~al.(2026{\natexlab{v}})Chen, Cao, Bai, Zeng, Ma, Lin, Liang, Liu, Sun, Lin and Shou}]{chen2026ruibench}
Chen Y, Cao Z, Bai Z, Zeng W, Ma KY, Lin Y, Liang G, Liu X, Sun K, Lin KQ and Shou MZ (2026{\natexlab{v}}) {RUIBench: What Makes a Good Robot-Use Interface?}
\newblock \urlprefix\url{https://arxiv.org/abs/2611.31115}.

\bibitem[{Chen et~al.(2026{\natexlab{w}})Chen, Cao, Ren, Yang, Li, Wang, Wang, Zhang, Shao, Zhao, Zhuang and Wu}]{chen2026roborouter}
Chen Y, Cao Z, Ren H, Yang C, Li W, Wang S, Wang Y, Zhang L, Shao Y, Zhao Z, Zhuang H and Wu Q (2026{\natexlab{w}}) {RoboRouter: Training-Free Policy Routing for Robotic Manipulation}.
\newblock \emph{arXiv preprint arXiv:2603.07892} \urlprefix\url{https://arxiv.org/abs/2603.07892}.

\bibitem[{Chen et~al.(2026{\natexlab{x}})Chen, Che, Huang, Fu, Wang, Cao, Liu, Yang, Su and Guo}]{chen2026harnessvln}
Chen Y, Che L, Huang Z, Fu W, Wang C, Cao X, Liu D, Yang Y, Su J and Guo LZ (2026{\natexlab{x}}) {HarnessVLN: Unifying Training-Free Embodied Navigation through an Agent Harness}.
\newblock \emph{arXiv preprint arXiv:2609.15195} \urlprefix\url{https://arxiv.org/abs/2609.15195}.

\bibitem[{Chen et~al.(2023{\natexlab{b}})Chen, Cui, Chen, Tan, Zhang, Zhao and Wang}]{chen2023robogpt}
Chen Y, Cui W, Chen Y, Tan M, Zhang X, Zhao D and Wang H (2023{\natexlab{b}}) {RoboGPT: an intelligent agent of making embodied long-term decisions for daily instruction tasks}.
\newblock \doi{10.48550/arXiv.2311.15649}.
\newblock \urlprefix\url{https://arxiv.org/abs/2311.15649}.

\bibitem[{Chen et~al.(2025{\natexlab{e}})Chen, Hao, Zhang and Fan}]{xref250301700}
Chen Y, Hao Y, Zhang Y and Fan C (2025{\natexlab{e}}) {Code-as-Symbolic-Planner: Foundation Model-Based Robot Planning via Symbolic Code Generation}.
\newblock In: \emph{IEEE/RSJ International Conference on Intelligent Robots and Systems}. pp. 19248--19254.
\newblock \doi{10.1109/iros60139.2025.11247174}.
\newblock \urlprefix\url{https://doi.org/10.1109/iros60139.2025.11247174}.

\bibitem[{Chen et~al.(2026{\natexlab{y}})Chen, Huai, Li, Wang, Zhang, Zhang, Chen, Gong, Jiang and Qiu}]{openeta}
Chen Y, Huai Z, Li S, Wang Y, Zhang H, Zhang Y, Chen H, Gong J, Jiang YG and Qiu X (2026{\natexlab{y}}) {ETA: A New Agentic Paradigm for Embodied Tasks}.
\newblock arXiv:2608.03924.
\newblock \urlprefix\url{https://arxiv.org/abs/2608.03924}.

\bibitem[{Chen et~al.(2026{\natexlab{z}})Chen, Li, Chen, Ma, Zhang, Du, Wei, Wang, Dong, Dou, Zhao, Zeng, Zhang and Zhou}]{chen2026physloop}
Chen Y, Li X, Chen H, Ma J, Zhang K, Du Z, Wei G, Wang F, Dong Z, Dou Z, Zhao L, Zeng Y, Zhang J and Zhou D (2026{\natexlab{z}}) {PhysLoop: Harnessing LLMs' Physics Awareness for Self-Improving Video Generation Models}.
\newblock \emph{arXiv preprint arXiv:2611.16748} \urlprefix\url{https://arxiv.org/abs/2611.16748}.

\bibitem[{Chen et~al.(2026{\natexlab{aa}})Chen, Liu, Wang, Jin, Zhu, Ruan and Han}]{chen2026recursia}
Chen Y, Liu F, Wang Z, Jin Z, Zhu X, Ruan L and Han W (2026{\natexlab{aa}}) {Learning to Compose: Recursive Self-Improvement of Shared Executable Stacks for Zero-Shot Long-Horizon Robot Manipulation}.
\newblock \emph{arXiv preprint arXiv:2610.24464} \doi{10.48550/arXiv.2610.24464}.
\newblock \urlprefix\url{https://arxiv.org/abs/2610.24464}.

\bibitem[{Chen et~al.(2026{\natexlab{ab}})Chen, Yang, Zhu, Huang, An, Xu, Yang, Liu, Gong, Cao, Bai, Zeng, Lin, Liang, Ma, Lin and Shou}]{chen2026papav}
Chen Y, Yang Z, Zhu J, Huang Q, An R, Xu P, Yang H, Liu R, Gong C, Cao Z, Bai Z, Zeng W, Lin Y, Liang G, Ma KY, Lin KQ and Shou MZ (2026{\natexlab{ab}}) Survey on multimodal embodied agents: A unified capability-centric perspective from computer-use to robot-use.
\newblock engrXiv preprint.
\newblock \doi{10.31224/8364}.
\newblock \urlprefix\url{https://engrxiv.org/preprint/view/8364}.

\bibitem[{Chen and Zhou(2026)}]{audit260722999}
Chen Y and Zhou K (2026) {WCM: World-Cognition Model for Generalizable Human-Robot Interaction}.
\newblock arXiv:2607.22999.
\newblock \urlprefix\url{https://arxiv.org/abs/2607.22999v1}.

\bibitem[{Chen et~al.(2026{\natexlab{ac}})Chen, Zhuang, Li, Wang, Zhao and Wu}]{chen2026manipevalagent}
Chen Y, Zhuang H, Li W, Wang S, Zhao X and Wu Q (2026{\natexlab{ac}}) {ManipEvalAgent: Promptable and Efficient Evaluation Framework for Robotic Manipulation Policies}.
\newblock In: \emph{International Conference on Learning Representations}. pp. 84574--84598.
\newblock \urlprefix\url{https://proceedings.iclr.cc/paper_files/paper/2026/file/8833c8aa10542d24d693bbaf6a4598f5-Paper-Conference.pdf}.

\bibitem[{Chen et~al.(2025{\natexlab{f}})Chen, Huo, Chen and Gao}]{chen2025robohorizon}
Chen Z, Huo J, Chen Y and Gao Y (2025{\natexlab{f}}) {RoboHorizon: An LLM-Assisted Multi-View World Model for Long-Horizon Robotic Manipulation}.
\newblock \doi{10.48550/arXiv.2501.06605}.
\newblock \urlprefix\url{https://arxiv.org/abs/2501.06605}.

\bibitem[{Chen et~al.(2026{\natexlab{ad}})Chen, Yin, Chen, Huo, Tian, Shi, Hou, Li and Gao}]{grid24}
Chen Z, Yin J, Chen Y, Huo J, Tian P, Shi J, Hou Y, Li Y and Gao Y (2026{\natexlab{ad}}) {DeCo: Task Decomposition and Skill Composition for Zero-Shot Generalization in Long-Horizon 3D Manipulation}.
\newblock \emph{IEEE Robotics and Automation Letters} 11(4): 5049--5056.
\newblock \doi{10.1109/lra.2026.3666363}.
\newblock \urlprefix\url{https://doi.org/10.1109/lra.2026.3666363}.

\bibitem[{Cheng et~al.(2024)Cheng, Ji, Yang, Gongye, Zou, Kautz, B{\i}y{\i}k, Yin, Liu and Wang}]{cheng2024navila}
Cheng AC, Ji Y, Yang Z, Gongye Z, Zou X, Kautz J, B{\i}y{\i}k E, Yin H, Liu S and Wang X (2024) {NaVILA: Legged Robot Vision-Language-Action Model for Navigation}.
\newblock \emph{arXiv preprint arXiv:2412.04453} \urlprefix\url{https://arxiv.org/abs/2412.04453}.

\bibitem[{Cheng et~al.(2026)Cheng, Liu, Bu, Bjorck, Yin, Luo, Kautz, Fan, Zhu and Liu}]{cheng2026asena}
Cheng AC, Liu I, Bu E, Bjorck J, Yin H, Luo Z, Kautz J, Fan L, Zhu Y and Liu S (2026) {ASENA: Self-evolving Agents for Embodied Navigation}.
\newblock arXiv:2609.39207.
\newblock \urlprefix\url{https://arxiv.org/abs/2609.39207}.

\bibitem[{Cheng et~al.(2024{\natexlab{ab}})Cheng, Nie and Swaminathan}]{cheng2024trace}
Cheng CA, Nie A and Swaminathan A (2024{\natexlab{ab}}) {Trace is the Next AutoDiff: Generative Optimization with Rich Feedback, Execution Traces, and LLMs}.
\newblock \urlprefix\url{https://arxiv.org/abs/2406.16218}.

\bibitem[{Cheng et~al.(2026{\natexlab{ab}})Cheng, Yi, Fang, Zhang, Feng, Li, Zhuang, Wang, Yang, Song, Xue, Wu, Gui, Wang and Wu}]{gptpolicy2026}
Cheng D, Yi T, Fang Y, Zhang X, Feng F, Li Y, Zhuang G, Wang R, Yang S, Song W, Xue W, Wu M, Gui J, Wang J and Wu T (2026{\natexlab{ab}}) {In-Context Robot Learning with VLM Agents}.
\newblock \emph{arXiv preprint arXiv:2609.19138} \urlprefix\url{https://arxiv.org/abs/2609.19138}.

\bibitem[{Cheng et~al.(2025)Cheng, Tu, Li, Dai, Hu, Hu, Li, Shi, Yu, Chen, Shi and Sun}]{audit250111858}
Cheng Z, Tu Y, Li R, Dai S, Hu J, Hu S, Li J, Shi Y, Yu T, Chen W, Shi L and Sun M (2025) {EmbodiedEval: Evaluate Multimodal LLMs as Embodied Agents}.
\newblock arXiv preprint arXiv:2501.11858.
\newblock \urlprefix\url{https://arxiv.org/abs/2501.11858}.

\bibitem[{Choi et~al.(2026)Choi, Wang, Su and Xu}]{scalinggenerativevlarl2026}
Choi A, Wang X, Su Z and Xu W (2026) {Scaling Sim-to-Real VLA Reinforcement Learning with Generative 3D Worlds}.
\newblock arXiv:2603.18532.
\newblock \urlprefix\url{https://arxiv.org/abs/2603.18532}.
\newblock Accepted to CoRL 2026.

\bibitem[{Choi et~al.(2024)Choi, Yoon, Ong, Kim and Jang}]{lotabench2024}
Choi JW, Yoon Y, Ong H, Kim J and Jang M (2024) {LoTa-Bench: Benchmarking Language-oriented Task Planners for Embodied Agents}.
\newblock \emph{arXiv preprint arXiv:2402.08178} \urlprefix\url{https://arxiv.org/abs/2402.08178}.
\newblock Accepted to ICLR 2024.

\bibitem[{Choi et~al.(2026{\natexlab{ab}})Choi, Lee, Park, Kim, Krishna, Fox and Yu}]{choi2026vlaeval}
Choi S, Lee Y, Park Y, Kim CD, Krishna R, Fox D and Yu Y (2026{\natexlab{ab}}) vla-eval: A unified evaluation harness for vision-language-action models.
\newblock \emph{arXiv preprint arXiv:2603.13966} \urlprefix\url{https://arxiv.org/abs/2603.13966}.

\bibitem[{Choi et~al.(2025)Choi, Ryu, Ock and Mehr}]{choi2025craft}
Choi S, Ryu K, Ock J and Mehr N (2025) {CRAFT: Coaching Reinforcement Learning Autonomously using Foundation Models for Multi-Robot Coordination Tasks}.
\newblock \urlprefix\url{https://arxiv.org/abs/2509.14380}.

\bibitem[{Chou et~al.(2026)Chou, Wang, Huang, Huang, Yang, Fan, Wang, Li, Kuang, Zheng, Bjorck, Liu, Su, Badki, Zhang, Yuille, Fan, Kautz and Yu}]{chou2026vflash}
Chou YC, Wang S, Huang J, Huang CP, Yang CA, Fan X, Wang J, Li Z, Kuang Y, Zheng R, Bjorck J, Liu S, Su H, Badki A, Zhang L, Yuille A, Fan L, Kautz J and Yu Z (2026) {VFlash: Fast, Accurate Visual Planning via Flash Primitive Decoding}.
\newblock \emph{arXiv preprint arXiv:2611.08449} \doi{10.48550/arXiv.2611.08449}.
\newblock \urlprefix\url{https://arxiv.org/abs/2611.08449}.

\bibitem[{Chu et~al.(2024)Chu, Zhao, Weber, Li, Lu and Wermter}]{grid61}
Chu K, Zhao X, Weber C, Li M, Lu W and Wermter S (2024) {Large Language Models for Orchestrating Bimanual Robots}.
\newblock In: \emph{2024 IEEE-RAS 23rd International Conference on Humanoid Robots (Humanoids)}. pp. 328--334.
\newblock \doi{10.1109/humanoids58906.2024.10769891}.
\newblock \urlprefix\url{https://doi.org/10.1109/humanoids58906.2024.10769891}.

\bibitem[{Chu et~al.(2026)Chu, Barr, Goues, O'Hearn, Harman, Sarro and Ye}]{audit261002459}
Chu Z, Barr ET, Goues CL, O'Hearn P, Harman M, Sarro F and Ye H (2026) {OpenRUA: Robot-Use Agents Are Zero-Shot Visuomotor Policies}.
\newblock arXiv:2610.02459.
\newblock \urlprefix\url{https://arxiv.org/abs/2610.02459v1}.

\bibitem[{Chung et~al.(2026)Chung, Ha, Kwak, Kwon, Lee, Lee and Lee}]{prompt2policy2026}
Chung W, Ha T, Kwak Y, Kwon T, Lee JG, Lee K and Lee S (2026) {Prompt-to-Policy: Agentic Engineering for Reinforcement Learning}.
\newblock Official project repository.
\newblock \urlprefix\url{https://github.com/krafton-ai/Prompt2Policy}.

\bibitem[{{\c{C}}i{\c{c}}ek et~al.(2026){\c{C}}i{\c{c}}ek, Er and Oguz}]{coral2026}
{\c{C}}i{\c{c}}ek B, Er MK and Oguz OS (2026) {CoRAL: Contact-Rich Adaptive LLM-based Control for Robotic Manipulation}.
\newblock arXiv:2605.02600.
\newblock \urlprefix\url{https://arxiv.org/abs/2605.02600}.
\newblock Accepted to RSS 2026.

\bibitem[{Community et~al.(2026)Community, Chen, Chen, Nian, Cai, Chen, Lin, Liang, Chen, Xiang, Su, Li, Tang, Qin, Chen, Zhu, Jiang, Wang, Li, Zhang, Wan, Chen, Su, Ye, Liu, Wang, Liang, Liu, Li, Wang, Chen, Bi, Huang, Tan, Cai, Mu, Guo, Wang, Zhu, Ke, Li, Tang, Li, Yang, Tu, Yang, Song, Ding, Zhang, Sun, Guo, Zhang, Chen, Cui, Zhang, Ma, Cai, Zhang, Yang, Ye, Chen, Liu, Su, Fang, Li, Cao, Wang, Chen, Luo and Ding}]{xpolicylab2026}
Community X, Chen T, Chen Y, Nian T, Cai Z, Chen G, Lin W, Liang Q, Chen Z, Xiang P, Su K, Li Z, Tang J, Qin Y, Chen Q, Zhu S, Jiang T, Wang Y, Li X, Zhang J, Wan W, Chen B, Su H, Ye K, Liu S, Wang K, Liang H, Liu Y, Li M, Wang Y, Chen B, Bi H, Huang S, Tan H, Cai J, Mu Y, Guo J, Wang X, Zhu Z, Ke W, Li H, Tang Y, Li X, Yang G, Tu Z, Yang S, Song W, Ding P, Zhang K, Sun Y, Guo J, Zhang T, Chen Y, Cui R, Zhang Z, Ma H, Cai J, Zhang H, Yang S, Ye J, Chen P, Liu S, Su X, Fang W, Li W, Cao Y, Wang C, Chen Q, Luo P and Ding W (2026) {XPolicyLab: A Unified Standard and Open Ecosystem for Robot Policy Evaluation and Deployment}.
\newblock \emph{arXiv preprint arXiv:2608.09892} \urlprefix\url{https://arxiv.org/abs/2608.09892}.

\bibitem[{Cui et~al.(2025)Cui, Wang, Li, Xie, Zou, Deng, Luo, Lu, Zhu and Dai}]{cui2023drivemlm}
Cui E, Wang W, Li Z, Xie J, Zou H, Deng H, Luo G, Lu L, Zhu X and Dai J (2025) {DriveMLM: aligning multi-modal large language models with behavioral planning states for autonomous driving}.
\newblock \emph{Visual Intelligence} 3(1): 22.
\newblock \doi{10.1007/s44267-025-00095-w}.
\newblock \urlprefix\url{https://doi.org/10.1007/s44267-025-00095-w}.

\bibitem[{Cui et~al.(2026)Cui, Zhao, Qin, Wang, Chen, Luo and Chen}]{cui2026embodisafety}
Cui E, Zhao S, Qin Y, Wang K, Chen J, Luo P and Chen T (2026) {EmbodiSafety: Benchmarking Risk Understanding and Response in Embodied AI}.
\newblock \urlprefix\url{https://arxiv.org/abs/2611.33054}.

\bibitem[{Cunningham et~al.(2026)Cunningham, Althoff, Halperin, Jabarian, Koh, Ramani, Trammell, Whitfill and Wu}]{cunningham2026economicsrsi}
Cunningham T, Althoff L, Halperin B, Jabarian B, Koh A, Ramani A, Trammell P, Whitfill P and Wu C (2026) {The Economics of Recursive Self-Improvement}.
\newblock \emph{arXiv preprint arXiv:2609.15802} \urlprefix\url{https://arxiv.org/abs/2609.15802}.

\bibitem[{Curtis et~al.(2025)Curtis, Kumar, Cao, Lozano-Pérez and Kaelbling}]{audit240605572}
Curtis A, Kumar N, Cao J, Lozano-Pérez T and Kaelbling LP (2025) Trust the proc3s: Solving long-horizon robotics problems with llms and constraint satisfaction.
\newblock In: \emph{Proceedings of The 8th Conference on Robot Learning}, \emph{Proceedings of Machine Learning Research}, volume 270. PMLR, pp. 1362--1383.
\newblock \urlprefix\url{https://proceedings.mlr.press/v270/curtis25a.html}.

\bibitem[{Dai et~al.(2026)Dai, Wang, Wang, Xie, Yang, Pan, Sun and Shu}]{dai2026evonavvln}
Dai G, Wang S, Wang Z, Xie GS, Yang Y, Pan J, Sun Q and Shu X (2026) {History to Future: Evolving Agent with Experience and Thought for Zero-shot Vision-and-Language Navigation}.
\newblock In: \emph{Proceedings of the IEEE/CVF Conference on Computer Vision and Pattern Recognition}. pp. 15177--15187.
\newblock \urlprefix\url{https://openaccess.thecvf.com/content/CVPR2026/html/Dai_History_to_Future_Evolving_Agent_with_Experience_and_Thought_for_CVPR_2026_paper.html}.

\bibitem[{Dai et~al.(2026{\natexlab{ab}})Dai, Mercat, Huang, Harikumar, Shah, Chai and Arora}]{dai2026mira}
Dai Y, Mercat J, Huang I, Harikumar S, Shah P, Chai J and Arora K (2026{\natexlab{ab}}) {When to Think, What to Remember: Mixed-Initiative Hierarchical Robot Control with Procedural Memory}.
\newblock CoRL 2026 Workshop on Memory for Robotics.
\newblock \urlprefix\url{https://corl2026-memory.github.io/papers/004.pdf}.

\bibitem[{Dalal et~al.(2024)Dalal, Chiruvolu, Chaplot and Salakhutdinov}]{dalal2024planseqlearn}
Dalal M, Chiruvolu T, Chaplot D and Salakhutdinov R (2024) {Plan-Seq-Learn: Language Model Guided RL for Solving Long Horizon Robotics Tasks}.
\newblock \doi{10.48550/arXiv.2405.01534}.
\newblock \urlprefix\url{https://arxiv.org/abs/2405.01534}.
\newblock Accepted to ICLR 2024.

\bibitem[{{DanMcInerney}(2026)}]{robotsworldjev2026}
{DanMcInerney} (2026) {Robots World}.
\newblock \urlprefix\url{https://github.com/DanMcInerney/robots-world}.

\bibitem[{Dao et~al.(2026)Dao, Ganai, Abukhadra, Sridhar, Azadani, Luo, Barrett, Wu, Finn and Pavone}]{dao2026direct}
Dao J, Ganai M, Abukhadra Y, Sridhar A, Azadani MN, Luo K, Barrett C, Wu J, Finn C and Pavone M (2026) {DIRECT: When and Where Should You Allocate Test-Time Compute in Embodied Planners?}
\newblock \urlprefix\url{https://arxiv.org/abs/2606.12402}.

\bibitem[{Darvish et~al.(2024)Darvish, Skreta, Zhao, Yoshikawa, Som, Bogdanovic, Cao, Hao, Xu, Aspuru-Guzik, Garg and Shkurti}]{audit240106949}
Darvish K, Skreta M, Zhao Y, Yoshikawa N, Som S, Bogdanovic M, Cao Y, Hao H, Xu H, Aspuru-Guzik A, Garg A and Shkurti F (2024) {ORGANA: A Robotic Assistant for Automated Chemistry Experimentation and Characterization}.
\newblock \emph{arXiv preprint arXiv:2401.06949} \doi{10.48550/arXiv.2401.06949}.
\newblock \urlprefix\url{https://arxiv.org/abs/2401.06949}.

\bibitem[{de~Bakker et~al.(2025)de~Bakker, Hejna, Lum, Celik, Taranovic, Blessing, Neumann, Bohg and Sadigh}]{debakker2025dexterityscaffolding}
de~Bakker V, Hejna J, Lum TGW, Celik O, Taranovic A, Blessing D, Neumann G, Bohg J and Sadigh D (2025) {Scaffolding Dexterous Manipulation with Vision-Language Models}.
\newblock \doi{10.48550/arXiv.2506.19212}.
\newblock \urlprefix\url{https://arxiv.org/abs/2506.19212}.

\bibitem[{{Demo2Skill Team}(2026)}]{demo2skill2026}
{Demo2Skill Team} (2026) {Demo2Skill: Equipping Robot-Use Agents with Dexterous Skills through Demonstration and Simulation}.
\newblock Project website.
\newblock \urlprefix\url{https://orangerangeangenge.github.io/Demo2Skill/}.

\bibitem[{Deng et~al.(2026)Deng, Gao, Lin, Liu, Wu and Wang}]{deng2026uniintervene}
Deng H, Gao Y, Lin Y, Liu H, Wu Z and Wang Z (2026) {UniIntervene: Agentic Intervention for Efficient Real-World Reinforcement Learning}.
\newblock \emph{arXiv preprint arXiv:2606.12372} \urlprefix\url{https://arxiv.org/abs/2606.12372}.

\bibitem[{Deng et~al.(2026{\natexlab{ab}})Deng, Liu, Zhang, Yan, Cao and Wang}]{dynaharness2026}
Deng H, Liu J, Zhang T, Yan L, Cao H and Wang Z (2026{\natexlab{ab}}) {DynaHarness: A Dynamic Physical Harness for Self-Evolving Robot Agents}.
\newblock \emph{arXiv preprint arXiv:2609.40306} \urlprefix\url{https://arxiv.org/abs/2609.40306}.

\bibitem[{Deng et~al.(2026{\natexlab{ac}})Deng, Lai, Li, Bai, Deng, Sun, Li and Yang}]{grid32}
Deng Y, Lai P, Li X, Bai C, Deng X, Sun C, Li X and Yang H (2026{\natexlab{ac}}) {SpaceVLN: A Zero-Shot Vision-and-Language Navigation Agent with Online Spatial Cognitive Memory and Reasoning}.
\newblock arXiv:2606.08992.
\newblock \urlprefix\url{https://arxiv.org/abs/2606.08992}.

\bibitem[{Di~Palo et~al.(2025)Di~Palo, Hasenclever, Humplik and Byravan}]{dipalo2025daag}
Di~Palo N, Hasenclever L, Humplik J and Byravan A (2025) {Diffusion Augmented Agents: A Framework for Efficient Exploration and Transfer Learning}.
\newblock In: \emph{Proceedings of The 3rd Conference on Lifelong Learning Agents}, \emph{Proceedings of Machine Learning Research}, volume 274. PMLR, pp. 268--284.
\newblock \urlprefix\url{https://proceedings.mlr.press/v274/palo25a.html}.

\bibitem[{Diao et~al.(2026)Diao, Chen, Xie, Lu, Peng, Zhong, Xu, Ye, Wu and Jiang}]{audit260939507}
Diao Z, Chen Y, Xie S, Lu T, Peng W, Zhong G, Xu H, Ye Z, Wu Z and Jiang YG (2026) {LIBERO-Agent: Evaluating General-Purpose Agents for Direct Embodied Manipulation}.
\newblock arXiv:2609.39507.
\newblock \urlprefix\url{https://arxiv.org/abs/2609.39507v1}.

\bibitem[{{dimensionalOS}(n.d.)}]{dimos2026}
{dimensionalOS} (n.d.) {Dimensional: Robotics Agent Framework}.
\newblock Software repository.
\newblock \urlprefix\url{https://github.com/dimensionalOS/dimos/tree/0f65afc34c6012e2d7924e620fb24fcb9f171510}.
\newblock Commit 0f65afc; accessed: 2026-10-14.

\bibitem[{{Dimweaker}(2026)}]{jevlibero2026}
{Dimweaker} (2026) {jev-libero}.
\newblock GitHub repository.
\newblock \urlprefix\url{https://github.com/Dimweaker/jev-libero}.
\newblock Accessed October 5, 2026.

\bibitem[{Ding et~al.(2026)Ding, Mi, Huang, Wang, Zhang, Hao, Chen, Li, Zheng, Guo, Wang, Li, Wu, Liu and Cao}]{zetta2026}
Ding X, Mi L, Huang M, Wang Z, Zhang C, Hao Z, Chen F, Li X, Zheng Y, Guo Y, Wang W, Li K, Wu H, Liu Y and Cao T (2026) {Zetta $\zeta$: An Efficient Closed-Loop Embodied Harness for Self-Evolving Physical Intelligence}.
\newblock \emph{arXiv preprint arXiv:2608.16590} \urlprefix\url{https://arxiv.org/abs/2608.16590}.

\bibitem[{Ding et~al.(2026{\natexlab{ab}})Ding, Yu, Dai, Qi, Sun, Wang, Yue and Wang}]{art2026}
Ding Y, Yu Y, Dai X, Qi X, Sun P, Wang X, Yue X and Wang J (2026{\natexlab{ab}}) {Evolve Vision-Language-Action Model into an Agent with On-the-Fly Tool-Use}.
\newblock In: \emph{2026 IEEE/CVF Conference on Computer Vision and Pattern Recognition- FINDINGS Track (CVPRF)}. pp. 1--12.
\newblock \doi{10.1109/cvprf71909.2026.00134}.
\newblock \urlprefix\url{https://doi.org/10.1109/cvprf71909.2026.00134}.

\bibitem[{Dong et~al.(2026)Dong, Qi, Zhang, Nambi, Ratliff, Vanbaar, Randle and Barreiros}]{dong2026agentguidedrl}
Dong W, Qi C, Zhang M, Nambi M, Ratliff LJ, Vanbaar J, Randle D and Barreiros J (2026) {Agent-Guided Reinforcement Learning for Robot Policy Improvement}.
\newblock CoRL 2026 Workshop on Continual Self-Improvement in Robotics.
\newblock \urlprefix\url{https://openreview.net/forum?id=Necql1OWWV}.

\bibitem[{Dong et~al.(2025)Dong, Zhu, Li, Mahon and Sun}]{dong2024zeroshotdriving}
Dong Z, Zhu Y, Li Y, Mahon K and Sun Y (2025) Generalizing end-to-end autonomous driving in real-world environments using zero-shot llms.
\newblock In: \emph{Proceedings of The 8th Conference on Robot Learning}, \emph{Proceedings of Machine Learning Research}, volume 270. PMLR, pp. 1231--1249.
\newblock \doi{10.48550/arXiv.2411.14256}.
\newblock \urlprefix\url{https://proceedings.mlr.press/v270/dong25a.html}.

\bibitem[{Dorbala et~al.(2024)Dorbala, Mullen~Jr. and Manocha}]{audit230303480}
Dorbala VS, Mullen~Jr JF and Manocha D (2024) {Can an Embodied Agent Find Your “Cat-shaped Mug”? LLM-Based Zero-Shot Object Navigation}.
\newblock \emph{IEEE Robotics and Automation Letters} 9(5): 4083--4090.
\newblock \doi{10.1109/lra.2023.3346800}.
\newblock \urlprefix\url{https://doi.org/10.1109/lra.2023.3346800}.

\bibitem[{Driess et~al.(2023)Driess, Xia, Sajjadi, Lynch, Chowdhery, Ichter, Wahid, Tompson, Vuong, Yu, Huang, Chebotar, Sermanet, Duckworth, Levine, Vanhoucke, Hausman, Toussaint, Greff, Zeng, Mordatch and Florence}]{audit230303378}
Driess D, Xia F, Sajjadi MSM, Lynch C, Chowdhery A, Ichter B, Wahid A, Tompson J, Vuong Q, Yu T, Huang W, Chebotar Y, Sermanet P, Duckworth D, Levine S, Vanhoucke V, Hausman K, Toussaint M, Greff K, Zeng A, Mordatch I and Florence P (2023) {P}a{LM}-e: An embodied multimodal language model.
\newblock In: \emph{Proceedings of the 40th International Conference on Machine Learning}, \emph{Proceedings of Machine Learning Research}, volume 202. PMLR, pp. 8469--8488.
\newblock \urlprefix\url{https://proceedings.mlr.press/v202/driess23a.html}.

\bibitem[{Du et~al.(2026)Du, Guo, Zhang, Wang, Dong, Gao, Han, Lu, Yu, Shen, Chang and Li}]{du2026sera}
Du T, Guo H, Zhang H, Wang X, Dong J, Gao C, Han M, Lu T, Yu M, Shen C, Chang X and Li Z (2026) {From VLM Planners to Robotic Agents: Self-Evolving Harnesses for History-Dependent Manipulation}.
\newblock \emph{arXiv preprint arXiv:2610.29560} \doi{10.48550/arXiv.2610.29560}.
\newblock \urlprefix\url{https://arxiv.org/abs/2610.29560}.

\bibitem[{Du et~al.(2026{\natexlab{ab}})Du, Kuang, Zhao, Wang, Guo, Xu, Li, He, Ren and Xiong}]{du2026agentrobocritic}
Du Y, Kuang Y, Zhao Q, Wang Y, Guo Z, Xu Y, Li X, He W, Ren L and Xiong C (2026{\natexlab{ab}}) {AgentRoboCritic: Understanding the Capability of Frontier Agents on Direct Robot Control}.
\newblock \emph{arXiv preprint arXiv:2611.26613} \urlprefix\url{https://arxiv.org/abs/2611.26613}.

\bibitem[{Duan et~al.(2024)Duan, Pumacay, Kumar, Wang, Tian, Yuan, Krishna, Fox, Mandlekar and Guo}]{aha2024}
Duan J, Pumacay W, Kumar N, Wang YR, Tian S, Yuan W, Krishna R, Fox D, Mandlekar A and Guo Y (2024) {AHA: A Vision-Language-Model for Detecting and Reasoning Over Failures in Robotic Manipulation}.
\newblock \emph{arXiv preprint arXiv:2410.00371} \urlprefix\url{https://arxiv.org/abs/2410.00371}.

\bibitem[{Duan et~al.(2025)Duan, Yuan, Pumacay, Wang, Ehsani, Fox and Krishna}]{xref240618915}
Duan J, Yuan W, Pumacay W, Wang YR, Ehsani K, Fox D and Krishna R (2025) Manipulate-anything: Automating real-world robots using vision-language models.
\newblock In: \emph{Proceedings of The 8th Conference on Robot Learning}, \emph{Proceedings of Machine Learning Research}, volume 270. PMLR, pp. 5326--5350.
\newblock \urlprefix\url{https://proceedings.mlr.press/v270/duan25a.html}.

\bibitem[{Duan et~al.(2024{\natexlab{ab}})Duan, Zhang and Xu}]{duan2024multimodaldrivingtokens}
Duan Y, Zhang Q and Xu R (2024{\natexlab{ab}}) {Prompting Multi-Modal Tokens to Enhance End-to-End Autonomous Driving Imitation Learning with LLMs}.
\newblock In: \emph{IEEE International Conference on Robotics and Automation}. pp. 6798--6805.
\newblock \doi{10.1109/icra57147.2024.10611614}.
\newblock \urlprefix\url{https://doi.org/10.1109/icra57147.2024.10611614}.

\bibitem[{Elmaaroufi et~al.(2026)Elmaaroufi, Svegliato, Kalade, Schelle, Seshia and Zaharia}]{rho2026}
Elmaaroufi K, Svegliato J, Kalade S, Schelle G, Seshia SA and Zaharia M (2026) {RHO: Your Coding Agent is Secretly a Roboticist}.
\newblock \emph{arXiv preprint arXiv:2606.16458} \urlprefix\url{https://arxiv.org/abs/2606.16458}.

\bibitem[{Elsayed et~al.(2026)Elsayed, Heakl, Shaker, Khan, Khan and {Claude Opus 5.5}}]{elsayed2026written}
Elsayed A, Heakl A, Shaker A, Khan F, Khan S and {Claude Opus 55} (2026) {Looped Weight-Space Programming: LLM-Written Transformers for Long Algorithmic Reasoning}.
\newblock \urlprefix\url{https://mbzuai-oryx.github.io/LoopedWSP/}.

\bibitem[{Espinola et~al.(2026)Espinola, Zhang and Xiang}]{espinola2026llmlibero}
Espinola M, Zhang Q and Xiang Y (2026) {From Reasoning to Action: Benchmarking Frontier AI Models on LIBERO-10 Robot Manipulation Tasks}.
\newblock \emph{arXiv preprint arXiv:2610.30051} \urlprefix\url{https://arxiv.org/abs/2610.30051}.

\bibitem[{Faldor et~al.(2024)Faldor, Zhang, Cully and Clune}]{si240515568}
Faldor M, Zhang J, Cully A and Clune J (2024) {OMNI-EPIC: Open-endedness via Models of human Notions of Interestingness with Environments Programmed in Code}.
\newblock \urlprefix\url{https://arxiv.org/abs/2405.15568}.

\bibitem[{Fan et~al.(2026)Fan, He, Jiang, Wu, Wang and Yue}]{fan2026captain}
Fan Y, He X, Jiang Y, Wu D, Wang J and Yue X (2026) {CaPtain}: A pipeline for post-training code-as-policy robot agents.
\newblock \doi{10.48550/arXiv.2610.40799}.
\newblock \urlprefix\url{https://arxiv.org/abs/2610.40799}.

\bibitem[{Fan et~al.(2026{\natexlab{ab}})Fan, Wang, Yuan, Zhang, Li, Xiao, Tang and Zhuang}]{fan2026embodiedharness}
Fan Z, Wang Z, Yuan Y, Zhang W, Li J, Xiao J, Tang S and Zhuang Y (2026{\natexlab{ab}}) {Embodied-Harness: Towards Recursive Self-Improvement for Robotic Manipulation}.
\newblock \urlprefix\url{https://embodied-harness.github.io/assets/embodied-harness.pdf}.

\bibitem[{Fang et~al.(2025)Fang, Zhang, Dong, Li, Wang, Zhang, Tian, Hu and Li}]{fang2025robix}
Fang H, Zhang M, Dong H, Li W, Wang Z, Zhang Q, Tian X, Hu Y and Li H (2025) {Robix: A Unified Model for Robot Interaction, Reasoning and Planning}.
\newblock \doi{10.48550/arXiv.2509.01106}.
\newblock \urlprefix\url{https://arxiv.org/abs/2509.01106}.

\bibitem[{Fang et~al.(2025{\natexlab{ab}})Fang, Sun, Ma, Lu and Yao}]{si250600276}
Fang J, Sun Y, Ma C, Lu Q and Yao L (2025{\natexlab{ab}}) {RoboMoRe: LLM-based Robot Co-design via Joint Optimization of Morphology and Reward}.
\newblock \urlprefix\url{https://arxiv.org/abs/2506.00276}.

\bibitem[{Fang et~al.(2025{\natexlab{ac}})Fang, Huang, Mao, Shone, Tenenbaum, Lozano-P{\'e}rez and Kaelbling}]{fang2025kalm}
Fang X, Huang BR, Mao J, Shone J, Tenenbaum JB, Lozano-P{\'e}rez T and Kaelbling LP (2025{\natexlab{ac}}) {KALM: Keypoint Abstraction Using Large Models for Object-Relative Imitation Learning}.
\newblock In: \emph{IEEE International Conference on Robotics and Automation (ICRA)}. pp. 8307--8314.
\newblock \doi{10.1109/ICRA55743.2025.11128681}.
\newblock \urlprefix\url{https://doi.org/10.1109/ICRA55743.2025.11128681}.

\bibitem[{Fang et~al.(2026)Fang, Yu, Chen, Li, Ma, Zhou, Xu, Va, Hu, Song, Du, Qian, Deng, Li, Wang, Li, Luo, Li, Wang, Deng and Li}]{audit260829896}
Fang Z, Yu Q, Chen Z, Li L, Ma H, Zhou K, Xu X, Va S, Hu Y, Song P, Du Q, Qian B, Deng Y, Li X, Wang Y, Li Z, Luo H, Li S, Wang Z, Deng W and Li X (2026) {EMERGE-Policy: A Robot Mind Emerges Beyond a Single Policy}.
\newblock arXiv:2608.29896.
\newblock \urlprefix\url{https://arxiv.org/abs/2608.29896v1}.

\bibitem[{Favaro and Clark(2026)}]{favaro2026whenai}
Favaro M and Clark J (2026) {When AI Builds Itself: Our Progress toward Recursive Self-Improvement, and Its Implications}.
\newblock The Anthropic Institute.
\newblock \urlprefix\url{https://www.anthropic.com/institute/recursive-self-improvement}.

\bibitem[{Fay et~al.(2026)Fay, Yi, Tolley and Wang}]{fay2026l2d}
Fay K, Yi S, Tolley MT and Wang X (2026) {Language to Design (L2D): Towards Self-Evolving Robots}.
\newblock CoRL 2026 Workshop on Continual Self-Improvement in Robotics.
\newblock \urlprefix\url{https://openreview.net/forum?id=fi5UmvIKPo}.

\bibitem[{{FBddcz}(2026)}]{embodiedjev}
{FBddcz} (2026) {EmbodiedJev}.
\newblock GitHub repository.
\newblock \urlprefix\url{https://github.com/FBddcz/embodied-jev}.
\newblock Accessed: September 29, 2026.

\bibitem[{Feng et~al.(2026)Feng, Chen, Liu, Pan, Xie, Cui, Zhou, Xiong, Zhang, Yin, Zhuang and Zhang}]{grid34}
Feng H, Chen S, Liu X, Pan M, Xie Y, Cui Y, Zhou Z, Xiong R, Zhang W, Yin J, Zhuang Y and Zhang X (2026) {Embodied-Navigator: Point, Think, Memorize, and Align for Efficient Navigation}.
\newblock arXiv:2608.17512.
\newblock \urlprefix\url{https://arxiv.org/abs/2608.17512}.

\bibitem[{Feng et~al.(2026{\natexlab{ab}})Feng, Liu, Gao, Li, Tang, Zhu, Shou, Chen, Yan, Zhang, Li, Wu, Zhang and Guo}]{feng2026metarobot}
Feng X, Liu Z, Gao W, Li X, Tang J, Zhu F, Shou Q, Chen H, Yan Z, Zhang H, Li J, Wu Z, Zhang J and Guo S (2026{\natexlab{ab}}) {META-Robot: Self-Cognition and Capability Scaling in Agentic Robots}.
\newblock \emph{arXiv preprint arXiv:2610.41868} \urlprefix\url{https://arxiv.org/abs/2610.41868}.

\bibitem[{Feng et~al.(2025)Feng, Han, Yang, Yue, Levine and Luo}]{audit250216707}
Feng Y, Han J, Yang Z, Yue X, Levine S and Luo J (2025) {Reflective Planning: Vision-Language Models for Multi-Stage Long-Horizon Robotic Manipulation}.
\newblock arXiv:2502.16707.
\newblock \urlprefix\url{https://arxiv.org/abs/2502.16707v1}.

\bibitem[{Feng et~al.(2026{\natexlab{ac}})Feng, Pan, Kang, Chen, Yang, Ye, Wang and Li}]{feng2026enactopt}
Feng Z, Pan Z, Kang Z, Chen Z, Yang S, Ye S, Wang H and Li H (2026{\natexlab{ac}}) {EnactOpt}: Diagnosis-guided harness evolution for embodied agents.
\newblock arXiv:2610.14977.
\newblock \urlprefix\url{https://arxiv.org/abs/2610.14977}.

\bibitem[{Firoozi et~al.(2025)Firoozi, Tucker, Tian, Majumdar, Sun, Liu, Zhu, Song, Kapoor, Hausman, Ichter, Driess, Wu, Lu and Schwager}]{firoozi2025foundation}
Firoozi R, Tucker J, Tian S, Majumdar A, Sun J, Liu W, Zhu Y, Song S, Kapoor A, Hausman K, Ichter B, Driess D, Wu J, Lu C and Schwager M (2025) {Foundation Models in Robotics: Applications, Challenges, and the Future}.
\newblock \emph{The International Journal of Robotics Research} 44(5): 701--739.
\newblock \doi{10.1177/02783649241281508}.
\newblock \urlprefix\url{https://doi.org/10.1177/02783649241281508}.

\bibitem[{{Foundation Model, Li Auto Inc.}(2026)}]{me2026vlm}
{Foundation Model, Li Auto Inc} (2026) {ME-VLM: A Unified VLM for Embodied Cognition and Agent Coordination}.
\newblock \urlprefix\url{https://arxiv.org/abs/2609.24526}.

\bibitem[{Fu et~al.(2024)Fu, Lei, Wen, Cai, Mao, Dou, Shi and Qiao}]{fu2024limsim}
Fu D, Lei W, Wen L, Cai P, Mao S, Dou M, Shi B and Qiao Y (2024) {LimSim++: A Closed-Loop Platform for Deploying Multimodal LLMs in Autonomous Driving}.
\newblock \doi{10.48550/arXiv.2402.01246}.
\newblock \urlprefix\url{https://arxiv.org/abs/2402.01246}.
\newblock Accepted to IEEE Intelligent Vehicles Symposium 2024.

\bibitem[{Fu et~al.(2023)Fu, Li, Wen, Dou, Cai, Shi and Qiao}]{fu2023drivelikeahuman}
Fu D, Li X, Wen L, Dou M, Cai P, Shi B and Qiao Y (2023) {Drive Like a Human: Rethinking Autonomous Driving with Large Language Models}.
\newblock \doi{10.48550/arXiv.2307.07162}.
\newblock \urlprefix\url{https://arxiv.org/abs/2307.07162}.

\bibitem[{Fu et~al.(2026)Fu, Nan, Sun, Li, Qian, Yang, Du, Barry, Kitani and Konidaris}]{xref260220119}
Fu J, Nan J, Sun L, Li H, Qian J, Yang B, Du Y, Barry JL, Kitani K and Konidaris G (2026) {NovaPlan: Zero-Shot Long-Horizon Manipulation via Closed-Loop Video Language Planning}.
\newblock \emph{arXiv preprint arXiv:2602.20119} \urlprefix\url{https://arxiv.org/abs/2602.20119}.
\newblock Accepted to CoRL 2026.

\bibitem[{Fu et~al.(2026{\natexlab{ab}})Fu, Yu, El-Refai, Kou, Xue, Huang, Xiao, Li, Shi, Wu, Sastry, Zhu, Goldberg and Fan}]{capx}
Fu L, Yu J, El-Refai K, Kou E, Xue H, Huang H, Xiao W, Li FF, Shi G, Wu J, Sastry SS, Zhu Y, Goldberg K and Fan L (2026{\natexlab{ab}}) {C}a{P}-x: A framework for benchmarking and improving coding agents for robot manipulation.
\newblock In: \emph{Proceedings of the 43rd International Conference on Machine Learning}, \emph{Proceedings of Machine Learning Research}, volume 306. PMLR, pp. 32155--32212.
\newblock \urlprefix\url{https://proceedings.mlr.press/v306/fu26u.html}.

\bibitem[{{gabec-slop}(2026)}]{autoresearchgym2026}
{gabec-slop} (2026) {Autoresearch Gym}.
\newblock GitHub repository.
\newblock \urlprefix\url{https://github.com/gabec-slop/autoresearch-gym}.

\bibitem[{Galanti et~al.(2026)Galanti, Shah and Dao}]{pigey}
Galanti L, Shah D and Dao T (2026) {Physical Agency: Better Robot Performance Through Better Interfaces}.
\newblock CoRL 2026 Workshop on Agentic Robotics.
\newblock \urlprefix\url{https://openreview.net/forum?id=SlmaqmaqW0}.

\bibitem[{{Galbot Team} et~al.(2026){Galbot Team}, Chen, Cheng, Deng, Ding, Dong, Gao, Jia, Li, Li, Lian, Liang, Lin, Liu, Liu, Liu, Ma, Qi, Su, Wang, Xu, Xu, Xu, Xu, Yan, Yan, Yi, Yu, Zhang, Zhang, Zhang, Zhang, Zheng and Zhu}]{su2026astra}
{Galbot Team}, Chen X, Cheng X, Deng Y, Ding L, Dong S, Gao X, Jia H, Li Z, Li Z, Lian Y, Liang S, Lin C, Liu D, Liu J, Liu Q, Ma Y, Qi Z, Su J, Wang H, Xu R, Xu T, Xu X, Xu Z, Yan M, Yan S, Yi L, Yu R, Zhang J, Zhang Y, Zhang Z, Zhang Z, Zheng Y and Zhu W (2026) {Systematically Exploring the Capabilities of GPT-6 Astra as Embodied Policies}.
\newblock \urlprefix\url{https://arxiv.org/abs/2609.38537}.

\bibitem[{Gao et~al.(2025)Gao, Liu, Chi, Huang, Fei, Hou, Zhang, Lin, Fang, Jiang and Shao}]{audit250617561}
Gao C, Liu Z, Chi Z, Huang J, Fei X, Hou Y, Zhang Y, Lin Y, Fang Z, Jiang Z and Shao L (2025) {VLA-OS: Structuring and Dissecting Planning Representations and Paradigms in Vision-Language-Action Models}.
\newblock arXiv preprint arXiv:2506.17561.
\newblock \urlprefix\url{https://arxiv.org/abs/2506.17561}.

\bibitem[{Gao et~al.(2025{\natexlab{ab}})Gao, Li, Shi, Li, Zhang, Figueroa and Jayaraman}]{audit250712644}
Gao GJ, Li T, Shi J, Li Y, Zhang Z, Figueroa N and Jayaraman D (2025{\natexlab{ab}}) {VLMgineer: Vision Language Models as Robotic Toolsmiths}.
\newblock \emph{arXiv preprint arXiv:2507.12644} \doi{10.48550/arXiv.2507.12644}.
\newblock \urlprefix\url{https://arxiv.org/abs/2507.12644}.

\bibitem[{Gao et~al.(2026)Gao, Zeng, Zhou, Ge, Shangguan, Wang, Wan, Zheng, Zhu, Li, Wu, Wang, He, Jia, Wu and He}]{gao2026hexashell}
Gao H, Zeng J, Zhou A, Ge S, Shangguan X, Wang Y, Wan Y, Zheng Z, Zhu J, Li X, Wu D, Wang Y, He L, Jia Z, Wu S and He X (2026) {HexaShell: A Universal Harness for Robot Coding Agents in Simulation and the Real World}.
\newblock \urlprefix\url{https://arxiv.org/abs/2610.23479}.

\bibitem[{Gao et~al.(2026{\natexlab{ab}})Gao, Zheng, Zhou, Zhu, Ge, Wang, Li, Zeng, Wu, Shangguan, Wan, He, Wang, Yu, Jia, Liu, Hooi, Dong, Zhang, Wu and He}]{gao2026embodiedlab}
Gao H, Zheng Z, Zhou A, Zhu J, Ge S, Wang Y, Li X, Zeng J, Wu D, Shangguan X, Wan Y, He L, Wang Y, Yu X, Jia Z, Liu Y, Hooi B, Dong Y, Zhang W, Wu S and He X (2026{\natexlab{ab}}) {EmbodiedLab: Benchmarking Embodied Agents That Act to Know}.
\newblock \urlprefix\url{https://arxiv.org/abs/2610.35385}.

\bibitem[{Gao et~al.(2026{\natexlab{ac}})Gao, Zhou, Zheng, Ge, Zhu, Wang, Zeng, Shangguan, Wu, He, Jia, Wu and He}]{physicalcoding2026}
Gao H, Zhou J, Zheng Z, Ge S, Zhu J, Wang Y, Zeng J, Shangguan X, Wu D, He L, Jia Z, Wu S and He X (2026{\natexlab{ac}}) {Self-Evolving Coding Agents: From Digital Programs to Physical-World Intelligence}.
\newblock \emph{arXiv preprint arXiv:2609.35432} \urlprefix\url{https://arxiv.org/abs/2609.35432}.

\bibitem[{Gao et~al.(2023)Gao, Sarkar, Xia, Xiao, Wu, Ichter, Majumdar and Sadigh}]{audit230902561}
Gao J, Sarkar B, Xia F, Xiao T, Wu J, Ichter B, Majumdar A and Sadigh D (2023) {Physically Grounded Vision-Language Models for Robotic Manipulation}.
\newblock arXiv preprint arXiv:2309.02561.
\newblock \urlprefix\url{https://arxiv.org/abs/2309.02561}.
\newblock Accepted to ICRA 2024.

\bibitem[{Gao et~al.(2025{\natexlab{ac}})Gao, Li, Lin, Liu, Ji, Pan, Xu, Li, Li, Ji, Wei, Tang, Wang, Shen, Xiao, Wu, Tang and Zhuang}]{gao2025arcadia}
Gao M, Li J, Lin Y, Liu X, Ji J, Pan X, Xu Z, Li X, Li M, Ji W, Wei R, Tang R, Wang Q, Shen K, Xiao J, Wu Q, Tang S and Zhuang Y (2025{\natexlab{ac}}) {Arcadia: Toward a Full-Lifecycle Framework for Embodied Lifelong Learning}.
\newblock \doi{10.48550/arXiv.2512.00076}.
\newblock \urlprefix\url{https://arxiv.org/abs/2512.00076}.

\bibitem[{Gao et~al.(2025{\natexlab{ad}})Gao, Chen, Yang, Chen, Tian, Li, Huang, Wang, Wang and Pang}]{gao2025genmanipllmdrivensimulationgeneralizable}
Gao N, Chen Y, Yang S, Chen X, Tian Y, Li H, Huang H, Wang H, Wang T and Pang J (2025{\natexlab{ad}}) {GENMANIP: LLM-driven Simulation for Generalizable Instruction-Following Manipulation}.
\newblock \urlprefix\url{https://arxiv.org/abs/2506.10966}.

\bibitem[{Gao(2026)}]{gao2026rsijev}
Gao S (2026) {RSI-Jev}: Typed-decision models trained by a self-improving loop of {AI} agents.
\newblock \urlprefix\url{https://github.com/Shanghua-Gao/RSI-Jev}.
\newblock Software and model release records, accessed October 12, 2026.

\bibitem[{Gao et~al.(2026{\natexlab{ad}})Gao, Zeng, Huang, Li, Xiong, Qin, Yu, Yao, Zhang, Lin, Liu, Zhao, Zhao, Guo, Luo, Shu, Ge, Hu and Itti}]{gao2026_3dcodeverse}
Gao Y, Zeng Z, Huang J, Li Y, Xiong H, Qin W, Yu X, Yao Y, Zhang J, Lin KY, Liu R, Zhao Y, Zhao Y, Guo M, Luo J, Shu L, Ge Y, Hu X and Itti L (2026{\natexlab{ad}}) {3DCodeVerse: Exploring 3D Worlds as Code}.
\newblock arXiv:2610.22341.
\newblock \urlprefix\url{https://arxiv.org/abs/2610.22341}.

\bibitem[{Garg et~al.(2024)Garg, Zhang, Arkin and Fan}]{xref240406413}
Garg K, Zhang S, Arkin J and Fan C (2024) {Foundation Models to the Rescue: Deadlock Resolution in Connected Multi-Robot Systems}.
\newblock \emph{arXiv preprint arXiv:2404.06413} \urlprefix\url{https://arxiv.org/abs/2404.06413}.

\bibitem[{Ge et~al.(2026)Ge, Zhou, Zeng, Wan, Wu, Zheng, Wang, Jia, Shangguan, Zhu, Liu, He, Wu, He and Gao}]{ge2026urai}
Ge S, Zhou A, Zeng J, Wan Y, Wu D, Zheng Z, Wang Y, Jia Z, Shangguan X, Zhu J, Liu Y, He L, Wu S, He X and Gao H (2026) {Make Code as Policy Great Again: Frontier Agents Write, Call, and Evolve Robot Tools}.
\newblock \emph{arXiv preprint arXiv:2609.39018} \urlprefix\url{https://arxiv.org/abs/2609.39018}.

\bibitem[{Ge et~al.(2026{\natexlab{ab}})Ge, Jia, Wu, Zhou, Wu and Jiang}]{si260603509}
Ge Z, Jia X, Wu C, Zhou Y, Wu Z and Jiang YG (2026{\natexlab{ab}}) {EvoMemNav: Efficient Self-Evolving Fine-Grained Memory for Zero-Shot Embodied Navigation}.
\newblock \urlprefix\url{https://arxiv.org/abs/2606.03509}.

\bibitem[{{Genesis Authors}(2024)}]{genesis2024}
{Genesis Authors} (2024) {Genesis: A Universal and Generative Physics Engine for Robotics and Beyond}.
\newblock Software repository.
\newblock \urlprefix\url{https://github.com/Genesis-Embodied-AI/genesis-world}.
\newblock Accessed: 2026-10-25.

\bibitem[{Ghasemzadeh~Kakroudi and Pieters(2026)}]{ghasemzadehkakroudi2026conversational}
Ghasemzadeh~Kakroudi A and Pieters R (2026) {A Conversational Framework for Human-Robot Collaborative Manipulation with Distributed Generative AI models}.
\newblock \urlprefix\url{https://arxiv.org/abs/2606.06061}.
\newblock Accepted to IEEE RO-MAN 2026.

\bibitem[{{GigaWorld Team} et~al.(2026){GigaWorld Team}, Ye, Ma, Wang, Ni, Ye, Huang, Li, Zhao, Yan, Li, Lu, Wang, Yu, Hu, Deng, Liu, Tian, Wang, Zhou, Xu, Chen, Wang, Zeng, Chang, Ye, Wu, Wu and Zhu}]{gigaworldpolicy052026}
{GigaWorld Team}, Ye A, Ma A, Wang B, Ni C, Ye F, Huang G, Li G, Zhao G, Yan H, Li H, Lu J, Wang K, Yu M, Hu Q, Deng Q, Liu S, Tian X, Wang X, Zhou X, Xu X, Chen X, Wang Y, Zeng Y, Chang Y, Ye Y, Wu Z, Wu Z and Zhu Z (2026) {GigaWorld-Policy-0.5: A Faster and Stronger WAM Empowered by AutoResearch}.
\newblock arXiv:2607.13960.
\newblock \urlprefix\url{https://arxiv.org/abs/2607.13960}.

\bibitem[{Glocker et~al.(2025)Glocker, Hönig, Hirschmanner and Vincze}]{xref250421716}
Glocker M, Hönig P, Hirschmanner M and Vincze M (2025) {LLM-Empowered Embodied Agent for Memory-Augmented Task Planning in Household Robotics}.
\newblock \emph{arXiv preprint arXiv:2504.21716} \urlprefix\url{https://arxiv.org/abs/2504.21716}.
\newblock Accepted to Austrian Robotics Workshop 2025.

\bibitem[{Goetting et~al.(2024)Goetting, Singh and Loquercio}]{audit241105755}
Goetting D, Singh HG and Loquercio A (2024) {End-to-End Navigation with Vision Language Models: Transforming Spatial Reasoning into Question-Answering}.
\newblock \emph{arXiv preprint arXiv:2411.05755} \doi{10.48550/arXiv.2411.05755}.
\newblock \urlprefix\url{https://arxiv.org/abs/2411.05755}.

\bibitem[{Goldfajn et~al.(2026)Goldfajn, Nakamura, Mahmud, Svegliato, Wray and Zilberstein}]{audit260923997}
Goldfajn DB, Nakamura M, Mahmud S, Svegliato J, Wray KH and Zilberstein S (2026) {RoboTalk: Learning Multi-Robot Communication and Coordination from Multimodal Demonstrations}.
\newblock arXiv:2609.23997.
\newblock \urlprefix\url{https://arxiv.org/abs/2609.23997v1}.

\bibitem[{Gong et~al.(2026)Gong, Zheng, Hou, Lee, Guo, Tan, Wen, Wang, He, Zhang and Mi}]{gong2026tinyastral}
Gong K, Zheng X, Hou G, Lee S, Guo M, Tan R, Wen Z, Wang Q, He X, Zhang M and Mi MB (2026) {TinyAstral: Transferring Self-Evolving Harnesses to Tiny Generalist Manipulation Agents}.
\newblock arXiv:2610.41216.
\newblock \urlprefix\url{https://arxiv.org/abs/2610.41216}.

\bibitem[{Gong et~al.(2026{\natexlab{ab}})Gong, Guo, Hu, Kong, Leng, Li, Li, Liu, Liu, Lu, Luo, Ming, Shen, Tao, Wang, Yin, Gu, Guan, Guo, Liu, Pang, Yang, Ye, Geng, Gu, Han, Jing, Pan, Sun, Yang, Zhang, Chen, He, Mei, Shi, Yang, Zhu, Cai, Ma, Su, Tang, Zheng, Chu, Wu, Tang and Xu}]{gong2026abotn1}
Gong R, Guo Y, Hu J, Kong J, Leng X, Li T, Li W, Liu F, Liu Z, Lu J, Luo M, Ming C, Shen Y, Tao J, Wang Z, Yin M, Gu M, Guan Z, Guo W, Liu G, Pang H, Yang M, Ye Z, Geng X, Gu Z, Han H, Jing D, Pan H, Sun M, Yang K, Zhang J, Chen Y, He Y, Mei W, Shi J, Yang X, Zhu Y, Cai Y, Ma J, Su S, Tang Z, Zheng L, Chu Z, Wu X, Tang W and Xu M (2026{\natexlab{ab}}) Abot-n1: Toward a general visual language navigation foundation model.
\newblock \emph{arXiv preprint arXiv:2607.10383} \urlprefix\url{https://arxiv.org/abs/2607.10383}.

\bibitem[{Gong et~al.(2025)Gong, Zhang, Shang, Minniti, Patel, Pepe, Yan, Gundogdu, Kapelyukh, Abbas, Yan, Patel, Herlant and Schmeckpeper}]{anytask2025}
Gong R, Zhang X, Shang J, Minniti MV, Patel J, Pepe V, Yan R, Gundogdu A, Kapelyukh I, Abbas A, Yan X, Patel H, Herlant L and Schmeckpeper K (2025) {AnyTask: an Automated Task and Data Generation Framework for Advancing Sim-to-Real Policy Learning}.
\newblock arXiv:2512.17853.
\newblock \urlprefix\url{https://arxiv.org/abs/2512.17853}.

\bibitem[{Gong et~al.(2026{\natexlab{ac}})Gong, Zheng, Huang, Chen, Tian, Yang, Liu and Yang}]{gong2026roboticsweightediting}
Gong Z, Zheng K, Huang Y, Chen W, Tian J, Yang J, Liu M and Yang Z (2026{\natexlab{ac}}) {We Asked Opus-5.5 to Edit $\pi_{0.5}$'s Weights. AutoResearch Made It Happen.}
\newblock Technical blog.
\newblock \urlprefix\url{https://cuzyoung.github.io/robotics-weight-editing/en/}.

\bibitem[{Gonz{\'{a}}lez-Santamarta et~al.(2023)Gonz{\'{a}}lez-Santamarta, Gonz{\'{a}}lez-Fern{\'{a}}ndez, Rodr{\'{i}}guez-Lera, Guerrero-Higueras and Matell{\'{a}}n-Olivera}]{gonzalezsantamarta2023merlin2}
Gonz{\'{a}}lez-Santamarta M{\'{A}}, Gonz{\'{a}}lez-Fern{\'{a}}ndez I, Rodr{\'{i}}guez-Lera FJ, Guerrero-Higueras {\'{A}}M and Matell{\'{a}}n-Olivera V (2023) {Integration of Large Language Models within Cognitive Architectures for Autonomous Robots}.
\newblock \doi{10.48550/arXiv.2309.14945}.
\newblock \urlprefix\url{https://arxiv.org/abs/2309.14945}.

\bibitem[{{Grid Robotics} et~al.(2026){Grid Robotics}, Lan, Li, Li, Lyu, Lu, Qi, Wang, Wang, Yang, Yang, Yang and Yu}]{grid2026agentic}
{Grid Robotics}, Lan Q, Li D, Li S, Lyu Y, Lu S, Qi Z, Wang S, Wang X, Yang M, Yang Z, Yang Z and Yu T (2026) {Envisioning Agentic Robotics: A New Era of Robotic Systems}.
\newblock \urlprefix\url{https://arxiv.org/abs/2611.18359}.

\bibitem[{{grmkris}(2026)}]{grid83}
{grmkris} (2026) {robo-harness}.
\newblock GitHub repository.
\newblock \urlprefix\url{https://github.com/grmkris/robo-harness}.
\newblock Accessed: September 29, 2026.

\bibitem[{Gu et~al.(2026)Gu, Chen, Cao, Shen, Sui, Zheng, Xiao, Lu, Xu, Shi, Shi, Zhang, Zhan, Tomizuka and Wu}]{gu2026real2worlds2real}
Gu J, Chen J, Cao X, Shen Y, Sui H, Zheng S, Xiao X, Lu H, Xu F, Shi L, Shi D, Zhang R, Zhan W, Tomizuka M and Wu R (2026) {Real2Worlds2Real: Scaling Self-Improving Agentic Real2Sim2Real for Robot Manipulation}.
\newblock \urlprefix\url{https://arxiv.org/abs/2611.31098}.

\bibitem[{Gu et~al.(2023)Gu, Kuwajerwala, Morin, Jatavallabhula, Sen, Agarwal, Rivera, Paul, Ellis, Chellappa, Gan, de~Melo, Tenenbaum, Torralba, Shkurti and Paull}]{gu2023conceptgraphs}
Gu Q, Kuwajerwala A, Morin S, Jatavallabhula KM, Sen B, Agarwal A, Rivera C, Paul W, Ellis K, Chellappa R, Gan C, de~Melo CM, Tenenbaum JB, Torralba A, Shkurti F and Paull L (2023) {ConceptGraphs: Open-Vocabulary 3D Scene Graphs for Perception and Planning}.
\newblock \emph{arXiv preprint arXiv:2309.16650} \urlprefix\url{https://arxiv.org/abs/2309.16650}.

\bibitem[{Gu et~al.(2024)Gu, Kondepudi, Gupta, Huang and Gopalan}]{gu2024colada}
Gu W, Kondepudi S, Gupta A, Huang L and Gopalan N (2024) {Continual Robot Skill and Task Learning via Dialogue}.
\newblock \doi{10.48550/arXiv.2409.03166}.
\newblock \urlprefix\url{https://arxiv.org/abs/2409.03166}.

\bibitem[{Gu et~al.(2026{\natexlab{ab}})Gu, Zhu, Lin, Zhu, Chen, Wang, Yuan, Yang, Zhao, Li, Su and Wang}]{gu2026harnesswam}
Gu Z, Zhu B, Lin T, Zhu G, Chen Y, Wang K, Yuan T, Yang F, Zhao C, Li Z, Su P and Wang J (2026{\natexlab{ab}}) {HarnessWAM: Bridging Prediction and Deliberation in World Action Models}.
\newblock \emph{arXiv preprint arXiv:2608.09516} \urlprefix\url{https://arxiv.org/abs/2608.09516}.

\bibitem[{Gumbsch et~al.(2026)Gumbsch, Barcellona, Sch{\"u}nemann, Karageorgis, Zadaianchuk, Wang, Zakharov, Despinoy, Aljundi and Gavves}]{demo2reward2026}
Gumbsch C, Barcellona L, Sch{\"u}nemann L, Karageorgis P, Zadaianchuk A, Wang Z, Zakharov S, Despinoy F, Aljundi R and Gavves E (2026) {From Demonstrations to Rewards: Test-Time Prompt Optimization for VLM Reward Models}.
\newblock arXiv:2606.00083.
\newblock \urlprefix\url{https://arxiv.org/abs/2606.00083}.

\bibitem[{Guo(2026)}]{guo2026physicalretarget}
Guo L (2026) {Agentic physical retargeting of human videos to Wuji hands}.
\newblock Demonstration report.
\newblock \urlprefix\url{https://x.com/Lingxiao234/status/2097717020540481630}.

\bibitem[{Guo(2026{\natexlab{ab}})}]{guo2026agenticreal2sim}
Guo L (2026{\natexlab{ab}}) {Agentic real-to-sim reconstruction from robot demonstrations}.
\newblock Demonstration report.
\newblock \urlprefix\url{https://x.com/Lingxiao234/status/2096992059731443923}.

\bibitem[{Guo et~al.(2026)Guo, Mo, Wang, Zhang, Wang, Deng, Zhang, Rao and Hu}]{audit260922966}
Guo MH, Mo ZH, Wang JJ, Zhang Y, Wang K, Deng YX, Zhang JP, Rao Y and Hu SM (2026) {Transferring the Intelligence of VLMs to Robotic Control}.
\newblock arXiv:2609.22966.
\newblock \urlprefix\url{https://arxiv.org/abs/2609.22966v1}.

\bibitem[{Guo et~al.(2026{\natexlab{ab}})Guo, Zhang, Mo, Wang, Wang, Chen, Song, Deng, Li, Chu, Zhang, ang Gao, Rao and Hu}]{guo2026nanoastra}
Guo MH, Zhang Y, Mo ZH, Wang HZ, Wang JJ, Chen XS, Song J, Deng YX, Li JN, Chu X, Zhang H, ang Gao H, Rao Y and Hu SM (2026{\natexlab{ab}}) {A Glimpse of Training VLMs as Foundation Models for Robotic Manipulation}.
\newblock \urlprefix\url{https://arxiv.org/abs/2611.29338}.

\bibitem[{Guo et~al.(2026{\natexlab{ac}})Guo, Mai, Xu, Zhang, Luu, Zhang, Miao, Ajoudani, Kingston, Qiu and She}]{sref260201662}
Guo P, Mai Z, Xu Z, Zhang K, Luu QK, Zhang H, Miao Z, Ajoudani A, Kingston Z, Qiu Q and She Y (2026{\natexlab{ac}}) {PLanAR: Planning-Language-Grounded Agentic Reasoning for Robot Manipulation}.
\newblock \urlprefix\url{https://arxiv.org/abs/2602.01662}.

\bibitem[{Guo et~al.(2024)Guo, Kingston and Kavraki}]{audit241022225}
Guo W, Kingston Z and Kavraki LE (2024) {CaStL: Constraints as Specifications through LLM Translation for Long-Horizon Task and Motion Planning}.
\newblock \emph{arXiv preprint arXiv:2410.22225} \doi{10.48550/arXiv.2410.22225}.
\newblock \urlprefix\url{https://arxiv.org/abs/2410.22225}.

\bibitem[{Guo et~al.(2024{\natexlab{ab}})Guo, Wang, Zha and Chen}]{grid46}
Guo Y, Wang YJ, Zha L and Chen J (2024{\natexlab{ab}}) {DoReMi: Grounding Language Model by Detecting and Recovering from Plan-Execution Misalignment}.
\newblock In: \emph{2024 IEEE/RSJ International Conference on Intelligent Robots and Systems (IROS)}. pp. 12124--12131.
\newblock \doi{10.1109/iros58592.2024.10802284}.
\newblock \urlprefix\url{https://doi.org/10.1109/iros58592.2024.10802284}.

\bibitem[{Gupta et~al.(2026)Gupta, Bae, Li and Isele}]{gupta2026scaleplan}
Gupta P, Bae S, Li J and Isele D (2026) {Scale-Plan}: Scalable language-enabled task planning for heterogeneous multi-robot teams.
\newblock arXiv preprint arXiv:2603.08814.
\newblock \urlprefix\url{https://arxiv.org/abs/2603.08814}.

\bibitem[{Ha et~al.(2023)Ha, Florence and Song}]{grid64}
Ha H, Florence P and Song S (2023) Scaling up and distilling down: Language-guided robot skill acquisition.
\newblock In: \emph{Proceedings of The 7th Conference on Robot Learning}, \emph{Proceedings of Machine Learning Research}, volume 229. PMLR, pp. 3766--3777.
\newblock \urlprefix\url{https://proceedings.mlr.press/v229/ha23a.html}.

\bibitem[{Ham et~al.(2026)Ham, Kim, Lee, Kim, Zhang, Wang and Liu}]{ham2026spinediagnosis}
Ham M, Kim D, Lee C, Kim MJ, Zhang Y, Wang J and Liu H (2026) {SPINE: Bridging the Cyber-Physical Gap with Agentic AI}.
\newblock \emph{arXiv preprint arXiv:2607.13049} \urlprefix\url{https://arxiv.org/abs/2607.13049}.

\bibitem[{Han et~al.(2024)Han, McInroe, Jelley, Albrecht, Bell and Storkey}]{llmpersonalize2024}
Han D, McInroe T, Jelley A, Albrecht SV, Bell P and Storkey A (2024) {LLM-Personalize: Aligning LLM Planners with Human Preferences via Reinforced Self-Training for Housekeeping Robots}.
\newblock \emph{arXiv preprint arXiv:2404.14285} \urlprefix\url{https://arxiv.org/abs/2404.14285}.

\bibitem[{Han et~al.(2024{\natexlab{ab}})Han, Zhu, Zhu, Wu and Zhu}]{han2024interpret}
Han M, Zhu Y, Zhu SC, Wu YN and Zhu Y (2024{\natexlab{ab}}) {InterPreT: Interactive Predicate Learning from Language Feedback for Generalizable Task Planning}.
\newblock \emph{arXiv preprint arXiv:2405.19758} \urlprefix\url{https://arxiv.org/abs/2405.19758}.
\newblock Accepted to RSS 2024.

\bibitem[{Han et~al.(2025)Han, Qiu, Liao, Huang, Gao, Yan and Liu}]{robocerebra2025}
Han S, Qiu B, Liao Y, Huang S, Gao C, Yan S and Liu S (2025) {RoboCerebra: A Large-scale Benchmark for Long-horizon Robotic Manipulation Evaluation}.
\newblock \emph{arXiv preprint arXiv:2506.06677} \urlprefix\url{https://arxiv.org/abs/2506.06677}.
\newblock Accepted to NeurIPS 2025.

\bibitem[{Han et~al.(2026)Han, Ravichandar, Narang, Xu and Yang}]{han2026comdex}
Han Y, Ravichandar H, Narang Y, Xu J and Yang W (2026) {ComDex: A Compositional Framework for Dexterity Learning from Offline Human Manipulation Videos}.
\newblock CoRL 2026 Workshop on Agentic Robotics.
\newblock \urlprefix\url{https://openreview.net/forum?id=l0BBYNeW4A}.

\bibitem[{Han et~al.(2026{\natexlab{ab}})Han, Guo, Zeng, Cui, Xin, Zhou, Zhou, Zhang and Li}]{han2026embodiedskillbench}
Han Z, Guo Z, Zeng C, Cui Z, Xin Y, Zhou M, Zhou D, Zhang W and Li Y (2026{\natexlab{ab}}) {Embodied Skill Bench: Benchmarking How Well Agents with Embodied Skills Perform on Physical Tasks}.
\newblock \urlprefix\url{https://arxiv.org/abs/2610.16238}.

\bibitem[{Han et~al.(2026{\natexlab{ac}})Han, Wang, Liu, Lyu, Shang, Dong, Liu and Han}]{han2026seqwalker}
Han Z, Wang X, Liu B, Lyu Q, Shang Z, Dong J, Liu L and Han Z (2026{\natexlab{ac}}) {SeqWalker: Sequential-Horizon Vision-and-Language Navigation with Hierarchical Planning}.
\newblock \doi{10.48550/arXiv.2601.04699}.
\newblock \urlprefix\url{https://arxiv.org/abs/2601.04699}.

\bibitem[{Hasan et~al.(2026)Hasan, Lee, Sarker and Iqbal}]{sref260411975}
Hasan S, Lee B, Sarker S and Iqbal T (2026) {M2HRI: An LLM-Driven Multimodal Multi-Agent Framework for Personalized Human-Robot Interaction}.
\newblock \urlprefix\url{https://arxiv.org/abs/2604.11975}.

\bibitem[{Hazra et~al.(2024)Hazra, Sygkounas, Persson, Loutfi and Zuidberg Dos~Martires}]{revolve2024}
Hazra R, Sygkounas A, Persson A, Loutfi A and Zuidberg Dos~Martires P (2024) {REvolve: Reward Evolution with Large Language Models using Human Feedback}.
\newblock arXiv:2406.01309.
\newblock \urlprefix\url{https://arxiv.org/abs/2406.01309}.
\newblock Accepted to ICLR 2025.

\bibitem[{He et~al.(2026)He, Chen, Ding, Shao, Shi, Li, Huo and Gao}]{he2026vdreamer}
He S, Chen Z, Ding H, Shao D, Shi J, Li C, Huo J and Gao Y (2026) {V-Dreamer: Automating Robotic Simulation and Trajectory Synthesis via Video Generation Priors}.
\newblock \emph{arXiv preprint arXiv:2603.18811} \urlprefix\url{https://arxiv.org/abs/2603.18811}.

\bibitem[{He et~al.(2026{\natexlab{ab}})He, Xie, Cao, Chen, Duan, Ge, Huo, Shao and Tian}]{audit260931770}
He S, Xie L, Cao Y, Chen P, Duan K, Ge J, Huo X, Shao J and Tian Q (2026{\natexlab{ab}}) {Robot Manipulation with GPT-6-Astra: Body Knowledge, Experience Reuse, Emergent Skills, and Sim2Real Transfer}.
\newblock arXiv:2609.31770.
\newblock \urlprefix\url{https://arxiv.org/abs/2609.31770v1}.

\bibitem[{He et~al.(2026{\natexlab{ac}})He, Li, Wang, Yu, Zhu, He, Liu, Zhang, Mao, Zhou, Shi, Xin, Li, Zheng, Zeng, Lu, Zhang, Guo, Zhang, Yu, Shi, Liu, Zhan and Xie}]{si260924271}
He W, Li H, Wang C, Yu Z, Zhu X, He M, Liu Z, Zhang X, Mao X, Zhou C, Shi J, Xin Y, Li J, Zheng J, Zeng S, Lu F, Zhang Z, Guo S, Zhang H, Yu P, Shi J, Liu Y, Zhan K and Xie Y (2026{\natexlab{ac}}) {ME-Brain-1.0: Memory, Cognition and Action for Evolving Embodied Intelligence}.
\newblock \urlprefix\url{https://arxiv.org/abs/2609.24271}.

\bibitem[{He et~al.(2026{\natexlab{ad}})He, Ma, Yu, Chen, Huang, Wu, Yang and Jiang}]{skill2real2026}
He X, Ma S, Yu C, Chen Y, Huang Y, Wu YN, Yang Y and Jiang C (2026{\natexlab{ad}}) {Skill2Real: Agentic Skill Learning for Zero-Shot Sim-to-Real Robot Manipulation}.
\newblock \urlprefix\url{https://arxiv.org/abs/2610.02788}.

\bibitem[{He et~al.(2026{\natexlab{ae}})He, Wei, Cen, Jiang, Li, Lu, Yang, Zhang and Zheng}]{audit260706990}
He YX, Wei L, Cen H, Jiang JJ, Li Z, Lu G, Yang Y, Zhang D and Zheng WS (2026{\natexlab{ae}}) {A Closed-Loop Multi-Agent Framework for Robust Multi-Robot Manipulation}.
\newblock arXiv:2607.06990.
\newblock \urlprefix\url{https://arxiv.org/abs/2607.06990v1}.
\newblock Accepted to RSS 2026.

\bibitem[{He et~al.(2026{\natexlab{af}})He, Li, Lan, Jin, Hou and Pan}]{he2026acap}
He Z, Li R, Lan Y, Jin C, Hou Z and Pan X (2026{\natexlab{af}}) {ACAP: Asynchronous Coding Agents for Efficient Robot Control}.
\newblock \urlprefix\url{https://arxiv.org/abs/2611.31205}.

\bibitem[{Hindsbo et~al.(2026)Hindsbo, Ehsani and Mishra}]{sref260602951}
Hindsbo N, Ehsani S and Mishra P (2026) {SCOPE: Real-Time Natural Language Camera Agent at the Edge: A Sim-to-Real Benchmark and Analysis of Open-Source Vision and Language Agents for PTZ Camera Tasks}.
\newblock In: \emph{Proceedings of the 21st ACM/IEEE International Conference on Human-Robot Interaction}. pp. 553--561.
\newblock \doi{10.1145/3757279.3785641}.
\newblock \urlprefix\url{https://doi.org/10.1145/3757279.3785641}.

\bibitem[{Holt et~al.(2026)Holt, Pabon, Alora, Frey and Pavone}]{audit260923841}
Holt FR, Pabon LA, Alora JI, Frey J and Pavone M (2026) {Structured World-State Reasoning for Agentic Robotic Search}.
\newblock arXiv:2609.23841.
\newblock \urlprefix\url{https://arxiv.org/abs/2609.23841v1}.

\bibitem[{Honerkamp et~al.(2024)Honerkamp, B{\"u}chner, Despinoy, Welschehold and Valada}]{grid38}
Honerkamp D, B{\"u}chner M, Despinoy F, Welschehold T and Valada A (2024) {Language-Grounded Dynamic Scene Graphs for Interactive Object Search With Mobile Manipulation}.
\newblock \emph{IEEE Robotics and Automation Letters} 9(10): 8298--8305.
\newblock \doi{10.1109/lra.2024.3441495}.
\newblock \urlprefix\url{https://doi.org/10.1109/lra.2024.3441495}.

\bibitem[{Hong et~al.(2026)Hong, Wei and Zhang}]{sref260829379}
Hong W, Wei L and Zhang D (2026) {Bridging Semantics and Physics with Constrained LLMs for Safe and Trustworthy Robotic Manipulation}.
\newblock \urlprefix\url{https://arxiv.org/abs/2608.29379}.

\bibitem[{Hong et~al.(2026{\natexlab{ab}})Hong, Huang, Li, Li, Guibas, Wu and Choi}]{si260221198}
Hong Y, Huang H, Li M, Li FF, Guibas L, Wu J and Choi Y (2026{\natexlab{ab}}) {Learning from Trials and Errors: Reflective Test-Time Planning for Embodied LLMs}.
\newblock \urlprefix\url{https://arxiv.org/abs/2602.21198}.

\bibitem[{Hong et~al.(2026{\natexlab{ac}})Hong, Liu, Yin, Li, Guibas, Li, Wu and Choi}]{esibench2026}
Hong Y, Liu J, Yin H, Li M, Guibas L, Li FF, Wu J and Choi Y (2026{\natexlab{ac}}) {ESI-Bench: Towards Embodied Spatial Intelligence that Closes the Perception-Action Loop}.
\newblock \emph{arXiv preprint arXiv:2605.18746} \urlprefix\url{https://arxiv.org/abs/2605.18746}.

\bibitem[{Hong et~al.(2025)Hong, Yu, Li and Choi}]{audit251102239}
Hong Y, Yu H, Li M and Choi C (2025) {LACY: A Vision-Language Model-based Language-Action Cycle for Self-Improving Robotic Manipulation}.
\newblock arXiv preprint arXiv:2511.02239.
\newblock \urlprefix\url{https://arxiv.org/abs/2511.02239}.
\newblock Accepted to ICRA 2026.

\bibitem[{Hu et~al.(2026)Hu, Gao, Sundaresan, Sharma, Bohg and Sadigh}]{via}
Hu H, Gao J, Sundaresan P, Sharma S, Bohg J and Sadigh D (2026) {VIA: Visual Interface Agent for Robot Control}.
\newblock \emph{arXiv preprint arXiv:2607.11119} \urlprefix\url{https://arxiv.org/abs/2607.11119}.

\bibitem[{Hu et~al.(2026{\natexlab{ab}})Hu, Huang, Liu, Wang, Suzuki, Brand and Koike-Akino}]{hu2026recovla}
Hu H, Huang CT, Liu J, Wang Y, Suzuki K, Brand M and Koike-Akino T (2026{\natexlab{ab}}) {ReCoVLA: VLM-Guided Reward Compilation for Failure Recovery in Vision-Language-Action Policies}.
\newblock \emph{arXiv preprint arXiv:2606.09630} \urlprefix\url{https://arxiv.org/abs/2606.09630}.

\bibitem[{Hu et~al.(2026{\natexlab{ac}})Hu, Mart{\'i}n-Mart{\'i}n, Stone, Duan, Jiang and Shi}]{simex2026}
Hu J, Mart{\'i}n-Mart{\'i}n R, Stone P, Duan R, Jiang Z and Shi G (2026{\natexlab{ac}}) {SimEX: Simulation-Integrated Robotics AutoResearch}.
\newblock \urlprefix\url{https://arxiv.org/abs/2609.38982}.

\bibitem[{Hu et~al.(2026{\natexlab{ad}})Hu, Shridhar, Lu, Shah, Chiang, Tan and Xie}]{audit260610267}
Hu J, Shridhar M, Lu C, Shah D, Chiang HTL, Tan J and Xie A (2026{\natexlab{ad}}) {What Matters in Orchestrating Robot Policies: A Systematic Study of Hierarchical VLA Agents}.
\newblock arXiv:2606.10267.
\newblock \urlprefix\url{https://arxiv.org/abs/2606.10267v1}.

\bibitem[{Hu et~al.(2026{\natexlab{ae}})Hu, Hu, Hong, Chen and Liu}]{hu2026coap}
Hu K, Hu S, Hong F, Chen Z and Liu Z (2026{\natexlab{ae}}) {Embodied Turing Machines: Stateful Code for Robot Recursive Self-Improvement}.
\newblock \emph{arXiv preprint arXiv:2610.12369} \urlprefix\url{https://arxiv.org/abs/2610.12369}.

\bibitem[{Hu et~al.(2026{\natexlab{af}})Hu, Abuduweili, Meng, Liu, Leboutet, Chen, Müller and Koltun}]{hu2026applehalo}
Hu S, Abuduweili A, Meng Y, Liu S, Leboutet Q, Chen C, Müller M and Koltun V (2026{\natexlab{af}}) {Perceptive Humanoid Motion from Simulation}.
\newblock \urlprefix\url{https://arxiv.org/abs/2611.06908}.

\bibitem[{Hu et~al.(2024)Hu, Fang, Fang, Deng, Chen and Fang}]{hu2024agentscodriver}
Hu S, Fang Z, Fang Z, Deng Y, Chen X and Fang Y (2024) {AgentsCoDriver: Large Language Model Empowered Collaborative Driving with Lifelong Learning}.
\newblock \doi{10.48550/arXiv.2404.06345}.
\newblock \urlprefix\url{https://arxiv.org/abs/2404.06345}.

\bibitem[{Hu et~al.(2026{\natexlab{ag}})Hu, Lu, Zeng, Weng, Zhang, Xie, Mou and Xu}]{hu2026ars}
Hu S, Lu W, Zeng L, Weng G, Zhang W, Xie K, Mou X and Xu Y (2026{\natexlab{ag}}) {ARS: Agentic Reward System for Robot Learning}.
\newblock \urlprefix\url{https://arxiv.org/abs/2609.34484}.

\bibitem[{Hu(2026)}]{hu2026dexgpt}
Hu X (2026) {DexGPT: GIF to Sharpa Wave}.
\newblock Software / demo.
\newblock \urlprefix\url{https://github.com/Hu-xiao-max/dexgpt}.

\bibitem[{Hu et~al.(2026{\natexlab{ah}})Hu, Chen, Cao and Detry}]{hu20262am}
Hu Y, Chen F, Cao X and Detry R (2026{\natexlab{ah}}) {2AM: Grounding Agent-Side Memory as Guidance for Steerable Action Models in Long-Horizon Manipulation}.
\newblock \emph{arXiv preprint arXiv:2609.11308} \urlprefix\url{https://arxiv.org/abs/2609.11308}.

\bibitem[{Hu et~al.(2023)Hu, Lin, Zhang, Yi and Gao}]{grid60}
Hu Y, Lin F, Zhang T, Yi L and Gao Y (2023) {Look Before You Leap: Unveiling the Power of GPT-4V in Robotic Vision-Language Planning}.
\newblock arXiv:2311.17842.
\newblock \urlprefix\url{https://arxiv.org/abs/2311.17842}.

\bibitem[{Hu et~al.(2024{\natexlab{ab}})Hu, Lucchetti, Schlesinger, Saxena, Freeman, Modak, Guha and Biswas}]{codebotler2026}
Hu Z, Lucchetti F, Schlesinger C, Saxena Y, Freeman A, Modak S, Guha A and Biswas J (2024{\natexlab{ab}}) {Deploying and Evaluating LLMs to Program Service Mobile Robots}.
\newblock \emph{IEEE Robotics and Automation Letters} 9(3): 2853--2860.
\newblock \doi{10.1109/lra.2024.3360020}.
\newblock \urlprefix\url{https://doi.org/10.1109/lra.2024.3360020}.

\bibitem[{Hua et~al.(2025)Hua, Liu, Macaluso, Lin, Zhang, Xu and Wang}]{hua2024gensim2}
Hua P, Liu M, Macaluso A, Lin Y, Zhang W, Xu H and Wang L (2025) Gensim2: Scaling robot data generation with multi-modal and reasoning llms.
\newblock In: \emph{Proceedings of The 8th Conference on Robot Learning}, \emph{Proceedings of Machine Learning Research}, volume 270. PMLR, pp. 5030--5066.
\newblock \urlprefix\url{https://proceedings.mlr.press/v270/hua25a.html}.

\bibitem[{Huang et~al.(2024)Huang, Chang, Lin, Liang, Zeng and Li}]{si241213492}
Huang C, Chang Y, Lin J, Liang J, Zeng R and Li J (2024) {Efficient Language-instructed Skill Acquisition via Reward-Policy Co-Evolution}.
\newblock \urlprefix\url{https://arxiv.org/abs/2412.13492}.
\newblock Accepted to AAAI 2025.

\bibitem[{Huang et~al.(2022)Huang, Mees, Zeng and Burgard}]{audit221005714}
Huang C, Mees O, Zeng A and Burgard W (2022) {Visual Language Maps for Robot Navigation}.
\newblock arXiv preprint arXiv:2210.05714.
\newblock \urlprefix\url{https://arxiv.org/abs/2210.05714}.
\newblock Accepted to ICRA 2023.

\bibitem[{Huang et~al.(2025)Huang, Wu, Chen, Wang and Yang}]{huang2025thinkact}
Huang CP, Wu YH, Chen MH, Wang YCF and Yang FE (2025) {ThinkAct: Vision-Language-Action Reasoning via Reinforced Visual Latent Planning}.
\newblock \emph{arXiv preprint arXiv:2507.16815} \urlprefix\url{https://arxiv.org/abs/2507.16815}.

\bibitem[{Huang et~al.(2026)Huang, Wang, Wu, Ran, Guo, Zhang, Men, Wang, Zhou, Yao and Wu}]{robocoliseum2026agent2act}
Huang D, Wang J, Wu J, Ran T, Guo L, Zhang S, Men C, Wang S, Zhou Y, Yao M and Wu M (2026) {Frontier LLMs Enter the World of Robotics: A Cross-Model Evaluation on RoboColiseum}.
\newblock \urlprefix\url{https://arxiv.org/abs/2610.27865}.

\bibitem[{Huang et~al.(2026{\natexlab{ab}})Huang, Zhao, Li, Zhang, Wei, Li, Yang, Zhou, Chen and Li}]{audit261000854}
Huang H, Zhao P, Li Z, Zhang Y, Wei Y, Li W, Yang H, Zhou K, Chen YC and Li Y (2026{\natexlab{ab}}) {Are Frontier VLM Agents Ready to Be Robot Generalists? An Empirical Study with the Embodied Agent Arena}.
\newblock arXiv:2610.00854.
\newblock \urlprefix\url{https://arxiv.org/abs/2610.00854v3}.

\bibitem[{Huang et~al.(2026{\natexlab{ac}})Huang, Chen, Liu, He, He, Huang, Zheng, Zhang and Zhong}]{huang2026embodiedclbench}
Huang J, Chen Y, Liu R, He K, He X, Huang R, Zheng Z, Zhang Z and Zhong Y (2026{\natexlab{ac}}) {EmbodiedCLBench: Evaluating Continual Learning for Self-Evolving Embodied Agents}.
\newblock arXiv:2610.13021.
\newblock \urlprefix\url{https://arxiv.org/abs/2610.13021}.

\bibitem[{Huang et~al.(2026{\natexlab{ad}})Huang, Hu, Li, Qi, Xiao, Wu, Ba, Zhang and Zhang}]{huang2026stageguard}
Huang J, Hu Y, Li Z, Qi R, Xiao Y, Wu Y, Ba T, Zhang Z and Zhang Y (2026{\natexlab{ad}}) {StageGuard: Learning Stage Transitions for Long-Horizon Robot Tasks via Agentic Distillation}.
\newblock \urlprefix\url{https://arxiv.org/abs/2609.20791}.

\bibitem[{Huang et~al.(2026{\natexlab{ae}})Huang, Li, Hu, Qi, Xiao, Coates, Cao, Zhang and Zhang}]{grid27}
Huang J, Li Z, Hu Y, Qi R, Xiao Y, Coates M, Cao T, Zhang Z and Zhang Y (2026{\natexlab{ae}}) {RoboHarness: Memory-Driven Orchestration of Heterogeneous Robot Policies for Long-Horizon Planning}.
\newblock arXiv:2607.18060.
\newblock \urlprefix\url{https://arxiv.org/abs/2607.18060}.
\newblock Accepted to ECCV 2026 Agent in the World Workshop.

\bibitem[{Huang et~al.(2026{\natexlab{af}})Huang, Li, Hu, Zhang, Coates, Quan and Zhang}]{si250921543}
Huang J, Li Z, Hu Y, Zhang Z, Coates M, Quan X and Zhang Y (2026{\natexlab{af}}) Self-{C}ri{T}each: {LLM} self-teaching and self-critiquing for improving robotic planning via automated domain generation.
\newblock In: \emph{Proceedings of the 43rd International Conference on Machine Learning}, \emph{Proceedings of Machine Learning Research}, volume 306. PMLR, pp. 46825--46850.
\newblock \urlprefix\url{https://proceedings.mlr.press/v306/huang26v.html}.

\bibitem[{Huang et~al.(2026{\natexlab{ag}})Huang, Zhang, Chen, Li, Wang, Lyu, Ye, Hong, Wang, Su, Xiong, Wu, Xie, Manderson, Ermon, Abbeel, Torralba, Song, Ye and Du}]{huang2026rax}
Huang R, Zhang X, Chen H, Li Y, Wang YJ, Lyu B, Ye H, Hong C, Wang Y, Su C, Xiong H, Wu Y, Xie E, Manderson T, Ermon S, Abbeel P, Torralba A, Song D, Ye W and Du Y (2026{\natexlab{ag}}) {Scaling Online Exploration for Robot Agents}.
\newblock \emph{arXiv preprint arXiv:2610.22496} \urlprefix\url{https://arxiv.org/abs/2610.22496}.

\bibitem[{Huang et~al.(2023)Huang, Jiang, Dong, Qiao, Gao and Li}]{grid57}
Huang S, Jiang Z, Dong H, Qiao Y, Gao P and Li H (2023) {Instruct2Act: Mapping Multi-modality Instructions to Robotic Actions with Large Language Model}.
\newblock arXiv:2305.11176.
\newblock \urlprefix\url{https://arxiv.org/abs/2305.11176}.

\bibitem[{Huang et~al.(2022{\natexlab{ab}})Huang, Abbeel, Pathak and Mordatch}]{huang2022zeroshotplanner}
Huang W, Abbeel P, Pathak D and Mordatch I (2022{\natexlab{ab}}) {Language Models as Zero-Shot Planners: Extracting Actionable Knowledge for Embodied Agents}.
\newblock \urlprefix\url{https://arxiv.org/abs/2201.07207}.

\bibitem[{Huang et~al.(2026{\natexlab{ah}})Huang, Bi, Yang, Si, Ni, Wang, Jiang, Zhang, Ruan, Zhang, Huang, Gong and Zhou}]{huang2026criteria}
Huang W, Bi J, Yang S, Si R, Ni R, Wang Q, Jiang S, Zhang Y, Ruan F, Zhang K, Huang Y, Gong S and Zhou D (2026{\natexlab{ah}}) {Learning the Task or Learning the Verifier? Rethinking Criteria Exposure in Embodied Self-Improvement}.
\newblock \urlprefix\url{https://arxiv.org/abs/2611.08318}.

\bibitem[{Huang et~al.(2025{\natexlab{ab}})Huang, Wang, Li, Zhang and Li}]{rekep2024}
Huang W, Wang C, Li Y, Zhang R and Li FF (2025{\natexlab{ab}}) Rekep: Spatio-temporal reasoning of relational keypoint constraints for robotic manipulation.
\newblock In: \emph{Proceedings of The 8th Conference on Robot Learning}, \emph{Proceedings of Machine Learning Research}, volume 270. PMLR, pp. 4573--4602.
\newblock \urlprefix\url{https://proceedings.mlr.press/v270/huang25g.html}.

\bibitem[{Huang et~al.(2023{\natexlab{ab}})Huang, Wang, Zhang, Li, Wu and Li}]{voxposer2023}
Huang W, Wang C, Zhang R, Li Y, Wu J and Li FF (2023{\natexlab{ab}}) Voxposer: Composable 3d value maps for robotic manipulation with language models.
\newblock In: \emph{Proceedings of The 7th Conference on Robot Learning}, \emph{Proceedings of Machine Learning Research}, volume 229. PMLR, pp. 540--562.
\newblock \urlprefix\url{https://proceedings.mlr.press/v229/huang23b.html}.

\bibitem[{Huang et~al.(2023{\natexlab{ac}})Huang, Xia, Shah, Driess, Zeng, Lu, Florence, Mordatch, Levine, Hausman and Ichter}]{audit230300855}
Huang W, Xia F, Shah D, Driess D, Zeng A, Lu Y, Florence P, Mordatch I, Levine S, Hausman K and Ichter B (2023{\natexlab{ac}}) {Grounded Decoding: Guiding Text Generation with Grounded Models for Embodied Agents}.
\newblock \emph{arXiv preprint arXiv:2303.00855} \doi{10.48550/arXiv.2303.00855}.
\newblock \urlprefix\url{https://arxiv.org/abs/2303.00855}.

\bibitem[{Huang et~al.(2023{\natexlab{ad}})Huang, Xia, Xiao, Chan, Liang, Florence, Zeng, Tompson, Mordatch, Chebotar, Sermanet, Jackson, Brown, Luu, Levine, Hausman and Ichter}]{innermonologue}
Huang W, Xia F, Xiao T, Chan H, Liang J, Florence P, Zeng A, Tompson J, Mordatch I, Chebotar Y, Sermanet P, Jackson T, Brown N, Luu L, Levine S, Hausman K and Ichter B (2023{\natexlab{ad}}) Inner monologue: Embodied reasoning through planning with language models.
\newblock In: \emph{Proceedings of The 6th Conference on Robot Learning}, \emph{Proceedings of Machine Learning Research}, volume 205. PMLR, pp. 1769--1782.
\newblock \urlprefix\url{https://proceedings.mlr.press/v205/huang23c.html}.

\bibitem[{Huang et~al.(2026{\natexlab{ai}})Huang, Chen and Yang}]{huang2026embodiedrsisurvey}
Huang Y, Chen Y and Yang Z (2026{\natexlab{ai}}) {Recursive Self Improvement for Embodied Agents---A Survey of Mechanisms Evaluation and Open Challenges}.
\newblock Preprints.org.
\newblock \urlprefix\url{https://www.preprints.org/manuscript/202610.2080/v1}.
\newblock Preprint, version 1.

\bibitem[{Huang et~al.(2026{\natexlab{aj}})Huang, Li, Saxena, Liang, Mishra, Ji, Zha, Wu, Kumar, Scherer, Xu and Silver}]{audit260425788}
Huang Y, Li B, Saxena V, Liang Y, Mishra UA, Ji L, Zha L, Wu J, Kumar N, Scherer S, Xu D and Silver T (2026{\natexlab{aj}}) {KinDER: A Physical Reasoning Benchmark for Robot Learning and Planning}.
\newblock arXiv preprint arXiv:2604.25788.
\newblock \urlprefix\url{https://arxiv.org/abs/2604.25788}.

\bibitem[{Huang et~al.(2026{\natexlab{ak}})Huang, Wu, Bu, Xiong, Jiang, Li, Ji, Xie, Huang, Wu, Jiang and Wang}]{sref260312939}
Huang Y, Wu J, Bu W, Xiong Z, Jiang G, Li Y, Ji K, Xie S, Huang Y, Wu C, Jiang J and Wang Z (2026{\natexlab{ak}}) {RoboStream: Weaving Spatio-Temporal Reasoning with Memory in Vision-Language Models for Robotics}.
\newblock \urlprefix\url{https://arxiv.org/abs/2603.12939}.
\newblock Accepted to ECCV 2026.

\bibitem[{Huang et~al.(2026{\natexlab{al}})Huang, Wu, Huang, Lin, Zhang, Xiong, Li, Xie, Wang, Liu, Huang, Zhang, Liu, Yang, Tang, Tang, Aktas, Wu, Wu, Ji, Xie, Liu, Bu, Li, Wang, Bai, Mao, Wang, Li, Yu, Adeli, Li, Li, Öztireli, Xu, Yu, Zhang, Zhang, Zhou, Zhu, Chi, Jiang and Wang}]{huang2026mind2actworld}
Huang Y, Wu J, Huang F, Lin J, Zhang W, Xiong Z, Li Y, Xie Z, Wang Z, Liu Z, Huang J, Zhang Y, Liu Y, Yang H, Tang C, Tang S, Aktas H, Wu A, Wu D, Ji K, Xie S, Liu H, Bu W, Li C, Wang X, Bai T, Mao Y, Wang J, Li M, Yu H, Adeli E, Li M, Li YL, Öztireli C, Xu H, Yu C, Zhang J, Zhang S, Zhou W, Zhu W, Chi C, Jiang J and Wang Z (2026{\natexlab{al}}) {Mind2Act World: A Diagnostic Benchmark for Embodied Mind, Physical Action, and Mind-Action Composition in Robotic Manipulation}.
\newblock \urlprefix\url{https://arxiv.org/abs/2611.26444}.

\bibitem[{{Huawei Noah EAI Team}(2026)}]{huaweinoah2026hiphysrsi}
{Huawei Noah EAI Team} (2026) {Scaling by Structure: What Forms Can Scaling Take in Embodied AI?}
\newblock Technical report.
\newblock \urlprefix\url{https://behavior1k-champion.com/assets/HiPhys-RSI_tech_report.pdf}.

\bibitem[{Huh et~al.(2026)Huh, Gu, Truong, Liu and Tevet}]{huh2026homebody}
Huh G, Gu C, Truong TE, Liu CK and Tevet G (2026) {HomeBody: A Humanoid That Explores, Remembers, and Acts on Its Own}.
\newblock Official project report.
\newblock \urlprefix\url{https://tml.stanford.edu/homebody/}.
\newblock Accessed October 16, 2026.

\bibitem[{Huo et~al.(2026)Huo, Liu, Liu, Qi, Sun, Gao, He, Yang, Chang, Xiong, Wei, Ma and Xu}]{huo2026abotclaw}
Huo D, Liu H, Liu G, Qi D, Sun Z, Gao M, He J, Yang Y, Chang X, Xiong F, Wei X, Ma Z and Xu M (2026) {ABot-Claw: A Foundation for Persistent, Cooperative, and Self-Evolving Robotic Agents}.
\newblock \emph{arXiv preprint arXiv:2604.10096} \urlprefix\url{https://arxiv.org/abs/2604.10096}.

\bibitem[{{hxwxss}(2026)}]{hxwxss2026glmskills}
{hxwxss} (2026) {GLM-5.3-Flash Skills: Embodied Data Engine}.
\newblock \url{https://github.com/hxwxss/glm-skills-embodied}.
\newblock \urlprefix\url{https://github.com/hxwxss/glm-skills-embodied}.
\newblock Software repository. Accessed September 28, 2026.

\bibitem[{Iacob et~al.(2026)Iacob, Jovanović, Shen, Burkhardt, Kurmanji, Tastan, Sani, Venanzi, Nie, Odonnat, Cao, Marino, Qiu, Antonova and Lane}]{iacob2026rqgm}
Iacob A, Jovanović A, Shen WF, Burkhardt D, Kurmanji M, Tastan N, Sani L, Venanzi NAE, Nie J, Odonnat A, Cao Z, Marino B, Qiu X, Antonova R and Lane ND (2026) The red queen g\"odel machine: Co-evolving agents and their evaluators.
\newblock \emph{arXiv preprint arXiv:2606.26294} \urlprefix\url{https://arxiv.org/abs/2606.26294}.

\bibitem[{Iannoli et~al.(2026)Iannoli, Zyrianoff, Trotta, Gigli and Di~Felice}]{audit261001569}
Iannoli A, Zyrianoff I, Trotta A, Gigli L and Di~Felice M (2026) {Managing Context and Communication in Distributed Agentic UAV Swarms}.
\newblock arXiv:2610.01569.
\newblock \urlprefix\url{https://arxiv.org/abs/2610.01569v1}.
\newblock Accepted to IEEE Consumer Communications \& Networking Conference 2027.

\bibitem[{Ichter et~al.(2023)Ichter, Brohan, Chebotar, Finn, Hausman, Herzog, Ho, Ibarz, Irpan, Jang, Julian, Kalashnikov, Levine, Lu, Parada, Rao, Sermanet, Toshev, Vanhoucke, Xia, Xiao, Xu, Yan, Brown, Ahn, Cortes, Sievers, Tan, Xu, Reyes, Rettinghouse, Quiambao, Pastor, Luu, Lee, Kuang, Jesmonth, Joshi, Jeffrey, Ruano, Hsu, Gopalakrishnan, David, Zeng and Fu}]{saycan2022}
Ichter B, Brohan A, Chebotar Y, Finn C, Hausman K, Herzog A, Ho D, Ibarz J, Irpan A, Jang E, Julian R, Kalashnikov D, Levine S, Lu Y, Parada C, Rao K, Sermanet P, Toshev AT, Vanhoucke V, Xia F, Xiao T, Xu P, Yan M, Brown N, Ahn M, Cortes O, Sievers N, Tan C, Xu S, Reyes D, Rettinghouse J, Quiambao J, Pastor P, Luu L, Lee KH, Kuang Y, Jesmonth S, Joshi NJ, Jeffrey K, Ruano RJ, Hsu J, Gopalakrishnan K, David B, Zeng A and Fu CK (2023) Do as i can, not as i say: Grounding language in robotic affordances.
\newblock In: \emph{Proceedings of The 6th Conference on Robot Learning}, \emph{Proceedings of Machine Learning Research}, volume 205. PMLR, pp. 287--318.
\newblock \urlprefix\url{https://proceedings.mlr.press/v205/ichter23a.html}.

\bibitem[{{Innate Inc.}(2026)}]{innate2026imitation}
{Innate Inc} (2026) {Imitating a recorded demonstration with GPT-6 Astra}.
\newblock Software / demo.
\newblock \urlprefix\url{https://github.com/innate-inc/innate-os/pull/817}.

\bibitem[{Ishibashi et~al.(2025)Ishibashi, Yano and Oyamada}]{ishibashi2025selfdeveloping}
Ishibashi Y, Yano T and Oyamada M (2025) {Can Large Language Models Invent Algorithms to Improve Themselves?: Algorithm Discovery for Recursive Self-Improvement through Reinforcement Learning}.
\newblock In: \emph{Proceedings of the 2025 Conference of the Nations of the Americas Chapter of the Association for Computational Linguistics: Human Language Technologies (Volume 1: Long Papers)}. Association for Computational Linguistics, pp. 10332--10363.
\newblock \doi{10.18653/v1/2025.naacl-long.519}.
\newblock \urlprefix\url{https://aclanthology.org/2025.naacl-long.519/}.

\bibitem[{Izzo et~al.(2026)Izzo, Battistini, Bardaro and Matteucci}]{izzo2026multimodalbt}
Izzo RA, Battistini C, Bardaro G and Matteucci M (2026) {Multimodal Behavior Tree Generation: A Small Vision-Language Model for Robot Task Planning}.
\newblock \urlprefix\url{https://arxiv.org/abs/2603.06084}.

\bibitem[{Jedlicki et~al.(2026)Jedlicki, Dieudonné and Yang}]{jedlicki2026occ4m}
Jedlicki JB, Dieudonné T and Yang H (2026) {OCC4M: Object-Centric 4D Memory for Spatiotemporal Reasoning in Long-Horizon Manipulation}.
\newblock CoRL 2026 Workshop on Memory for Robotics.
\newblock \urlprefix\url{https://arxiv.org/abs/2609.28798}.

\bibitem[{Jedlicki et~al.(2026{\natexlab{ab}})Jedlicki, Sethi and Yang}]{xref261017350}
Jedlicki JB, Sethi A and Yang H (2026{\natexlab{ab}}) {Can a Frontier VLM Serve as a Robot Policy for Memory-Dependent Manipulation?}
\newblock \emph{arXiv preprint arXiv:2610.17350} \urlprefix\url{https://arxiv.org/abs/2610.17350}.

\bibitem[{jellyheadandrew(2026)}]{autoresearchrobotics2026}
jellyheadandrew (2026) {autoresearch-robotics: Autonomous Robotics Research with Simulation Feedback}.
\newblock Software repository.
\newblock \urlprefix\url{https://github.com/jellyheadandrew/autoresearch-robotics/tree/ff30cc481af9343b9c96f448b5ed72b75bfcd562}.
\newblock Software repository; revision ff30cc481af9343b9c96f448b5ed72b75bfcd562; inspected 2026-10-06.

\bibitem[{Jena et~al.(2026)Jena, Wanaskar, Jain, Chadha, Sharma and Das}]{weightsskills2026}
Jena G, Wanaskar K, Jain V, Chadha A, Sharma V and Das A (2026) {Weights or Skills? A Survey of Robot-Learning Techniques: from Action-Predicting Weights to Robots that Write their Own Skills}.
\newblock \emph{arXiv preprint arXiv:2608.01851} \urlprefix\url{https://arxiv.org/abs/2608.01851}.

\bibitem[{Jeong et~al.(2026)Jeong, Swamy and Bajcsy}]{sref260612299}
Jeong HJ, Swamy G and Bajcsy A (2026) {Learning What to Say to Your VLA: Mostly Harmless Vision Language Action Model Steering}.
\newblock \urlprefix\url{https://arxiv.org/abs/2606.12299}.
\newblock Accepted to CoRL 2026.

\bibitem[{Ji et~al.(2026)Ji, Wang, Qian, Yu, Tian, Fei, Gong and Qiu}]{audit260703449}
Ji L, Wang S, Qian P, Yu X, Tian Y, Fei Z, Gong J and Qiu X (2026) {H}i{M}e: Hierarchical embodied memory for long-horizon vision-language-action control.
\newblock In: \emph{Proceedings of the 43rd International Conference on Machine Learning}, \emph{Proceedings of Machine Learning Research}, volume 306. PMLR, pp. 51611--51631.
\newblock \urlprefix\url{https://proceedings.mlr.press/v306/ji26k.html}.

\bibitem[{Jia et~al.(2026)Jia, Lin, Zhang, Zhang, Liu and Jiang}]{agp}
Jia M, Lin Y, Zhang X, Zhang Z, Liu X and Jiang M (2026) {Agent as Policy for Robotic Manipulation}.
\newblock arXiv:2609.12541.
\newblock \urlprefix\url{https://arxiv.org/abs/2609.12541}.

\bibitem[{Jiang et~al.(2024)Jiang, Chen, Liao, Zhang, Yin, Zhang, Huang, Liu and Wang}]{jiang2024senna}
Jiang B, Chen S, Liao B, Zhang X, Yin W, Zhang Q, Huang C, Liu W and Wang X (2024) {Senna: Bridging Large Vision-Language Models and End-to-End Autonomous Driving}.
\newblock \doi{10.48550/arXiv.2410.22313}.
\newblock \urlprefix\url{https://arxiv.org/abs/2410.22313}.

\bibitem[{Jiang et~al.(2026)Jiang, Zhang, Wang, Li, Wang, Ji, Lai, Ren, Li, Yang, Hu and Ma}]{si260913082}
Jiang B, Zhang F, Wang L, Li H, Wang Y, Ji Z, Lai J, Ren X, Li D, Yang Z, Hu J and Ma Q (2026) {Embodied-BenchForge: A Closed-Loop Agentic Workflow for Embodied Benchmark Construction}.
\newblock \urlprefix\url{https://arxiv.org/abs/2609.13082}.

\bibitem[{Jiang et~al.(2026{\natexlab{ab}})Jiang, Yang, Jiang, Bertasius, Szafir and Lioutikov}]{jiang2026halter}
Jiang J, Yang Y, Jiang X, Bertasius G, Szafir DJ and Lioutikov R (2026{\natexlab{ab}}) {From Rollout to Reset: A Graph-Based Harness for Autonomous Long-Horizon Manipulation Evaluation}.
\newblock \emph{arXiv preprint arXiv:2609.19413} \urlprefix\url{https://arxiv.org/abs/2609.19413}.

\bibitem[{Jiang et~al.(2024{\natexlab{ab}})Jiang, Cai, Cui, Li, Ren, Yu, Yang, Fu, Wen and Cai}]{jiang2024koma}
Jiang K, Cai X, Cui Z, Li A, Ren Y, Yu H, Yang H, Fu D, Wen L and Cai P (2024{\natexlab{ab}}) {KoMA: Knowledge-driven Multi-agent Framework for Autonomous Driving with Large Language Models}.
\newblock \doi{10.48550/arXiv.2407.14239}.
\newblock \urlprefix\url{https://arxiv.org/abs/2407.14239}.

\bibitem[{Jiang et~al.(2026{\natexlab{ac}})Jiang, Hu, Du, Li, Xue, Li and Soh}]{jiang2026appl}
Jiang P, Hu T, Du H, Li Y, Xue Z, Li X and Soh H (2026{\natexlab{ac}}) {Agent Priors-guided Policy Learning}.
\newblock \emph{arXiv preprint arXiv:2609.35690} \urlprefix\url{https://arxiv.org/abs/2609.35690}.

\bibitem[{Jiang et~al.(2025)Jiang, Yuan, Liu, Lu, Cui, Liu, Cheng, Gao, Xu and Zhao}]{jiang2025galaxeag0}
Jiang T, Yuan T, Liu Y, Lu C, Cui J, Liu X, Cheng S, Gao J, Xu H and Zhao H (2025) {Galaxea Open-World Dataset and G0 Dual-System VLA Model}.
\newblock \doi{10.48550/arXiv.2509.00576}.
\newblock \urlprefix\url{https://arxiv.org/abs/2509.00576}.

\bibitem[{Jiang et~al.(2026{\natexlab{ad}})Jiang, Gan, Wang, He, Zhao, Wu, Wang, Wang, He, Wang and Shao}]{audit260923578}
Jiang Y, Gan Z, Wang X, He T, Zhao C, Wu M, Wang X, Wang J, He J, Wang J and Shao Q (2026{\natexlab{ad}}) {AR-WAM: A Visual-Conditioned Agent-Ready World Action Model for Robotic Manipulation}.
\newblock arXiv:2609.23578.
\newblock \urlprefix\url{https://arxiv.org/abs/2609.23578v1}.

\bibitem[{Jin et~al.(2026)Jin, Tang, Luo, Huang, Sun, Yang, Xie and Wang}]{jin2026philia}
Jin W, Tang P, Luo K, Huang B, Sun B, Yang H, Xie S and Wang J (2026) {A Glimpse into Long-term Physical Coexistence with Intelligent Robots}.
\newblock \urlprefix\url{https://arxiv.org/abs/2607.11377}.

\bibitem[{Jin et~al.(2026{\natexlab{ab}})Jin, Guo, Jia, Deng, Li, Liu, Liao, Prasad, Franzius, Neumann and Chalvatzaki}]{jin2026nautilus}
Jin Y, Guo J, Jia X, Deng Y, Li Z, Liu H, Liao W, Prasad V, Franzius M, Neumann G and Chalvatzaki G (2026{\natexlab{ab}}) {Nautilus: From One Prompt to Plug-and-Play Robot Learning}.
\newblock \emph{arXiv preprint arXiv:2605.11665} \urlprefix\url{https://arxiv.org/abs/2605.11665}.

\bibitem[{Jin et~al.(2023)Jin, Yang, Yi, Shen, Peng, Liu, Qin, Li, Xie, Gao, Zhou and Gong}]{jin2023surrealdriver}
Jin Y, Yang R, Yi Z, Shen X, Peng H, Liu X, Qin J, Li J, Xie J, Gao P, Zhou G and Gong J (2023) {SurrealDriver: Designing LLM-powered Generative Driver Agent Framework based on Human Drivers' Driving-thinking Data}.
\newblock \doi{10.48550/arXiv.2309.13193}.
\newblock \urlprefix\url{https://arxiv.org/abs/2309.13193}.

\bibitem[{Jing et~al.(2025)Jing, Yang, Ao, Xiao, Jiang and Bai}]{humanoidgen2025}
Jing Z, Yang S, Ao J, Xiao T, Jiang YG and Bai C (2025) {HumanoidGen: Data Generation for Bimanual Dexterous Manipulation via LLM Reasoning}.
\newblock \urlprefix\url{https://arxiv.org/abs/2507.00833}.

\bibitem[{Joublin et~al.(2024)Joublin, Ceravola, Smirnov, Ocker, Deigmoeller, Belardinelli, Wang, Hasler, Tanneberg and Gienger}]{copal2023}
Joublin F, Ceravola A, Smirnov P, Ocker F, Deigmoeller J, Belardinelli A, Wang C, Hasler S, Tanneberg D and Gienger M (2024) {CoPAL: Corrective Planning of Robot Actions with Large Language Models}.
\newblock In: \emph{2024 IEEE International Conference on Robotics and Automation (ICRA)}. pp. 8664--8670.
\newblock \doi{10.1109/icra57147.2024.10610434}.
\newblock \urlprefix\url{https://doi.org/10.1109/icra57147.2024.10610434}.

\bibitem[{Ju et~al.(2026)Ju, Wang, Ding, Yang, Wu, Jiang, Zhang, Wen, Li, Wang, Li, Liu, Dai, Wang and Cao}]{embodiskill2026}
Ju R, Wang X, Ding X, Yang Y, Wu H, Jiang S, Zhang Q, Wen H, Li X, Wang W, Li K, Liu Y, Dai H, Wang W and Cao T (2026) {EmbodiSkill: Skill-Aware Reflection for Self-Evolving Embodied Agents}.
\newblock \emph{arXiv preprint arXiv:2605.10332} \urlprefix\url{https://arxiv.org/abs/2605.10332}.

\bibitem[{{juancamiloqhz}(2026)}]{roverlab2026}
{juancamiloqhz} (2026) {RoverLab}.
\newblock GitHub repository.
\newblock \urlprefix\url{https://github.com/juancamiloqhz/roverlab}.

\bibitem[{Jung et~al.(2026)Jung, Song, Kim, Tan, Gao, Lee and Kim}]{jung2026groundedplanbench}
Jung S, Song H, Kim DH, Tan R, Gao J, Lee YJ and Kim D (2026) {Spatially Grounded Long-Horizon Task Planning in the Wild}.
\newblock ICRA 2026 Workshop on Rigorous Perception.
\newblock \urlprefix\url{https://arxiv.org/abs/2603.13433}.

\bibitem[{Kachaev et~al.(2025)Kachaev, Spiridonov, Gorodetsky, Muravyev, Oskolkov, Narendra, Shakhuro, Makarov, Panov, Fedotova and Kovalev}]{kachaev2025kitchenr}
Kachaev N, Spiridonov A, Gorodetsky A, Muravyev K, Oskolkov N, Narendra A, Shakhuro V, Makarov D, Panov AI, Fedotova P and Kovalev AK (2025) {Mind and Motion Aligned: A Joint Evaluation IsaacSim Benchmark for Task Planning and Low-Level Policies in Mobile Manipulation}.
\newblock \urlprefix\url{https://arxiv.org/abs/2508.15663}.

\bibitem[{Kagaya et~al.(2025)Kagaya, Lakshmi, Lou, Yuan, Karlekar, Pranata, Murakami, Kinose and You}]{xref250924160}
Kagaya T, Lakshmi S, Lou Y, Yuan TJ, Karlekar J, Pranata S, Murakami N, Kinose A and You Y (2025) {Memory Transfer Planning: LLM-driven Context-Aware Code Adaptation for Robot Manipulation}.
\newblock \emph{arXiv preprint arXiv:2509.24160} \urlprefix\url{https://arxiv.org/abs/2509.24160}.

\bibitem[{Kang et~al.(2026)Kang, Ye, Liu, Mantri, Mao, Fleming, Regmi and Qin}]{self260509423}
Kang H, Ye X, Liu Y, Mantri SH, Mao L, Fleming J, Regmi D and Qin L (2026) {SimWorld Studio: Automatic Environment Generation with Evolving Coding Agent for Embodied Agent Learning}.
\newblock arXiv:2605.09423.
\newblock \urlprefix\url{https://arxiv.org/abs/2605.09423}.

\bibitem[{Kang et~al.(2026{\natexlab{ab}})Kang, Wang, Fan and Liu}]{audit260937359}
Kang Y, Wang Z, Fan Z and Liu B (2026{\natexlab{ab}}) {Encore: Few-Shot Agentic Discovery of Manipulation Strategies}.
\newblock arXiv:2609.37359.
\newblock \urlprefix\url{https://arxiv.org/abs/2609.37359v1}.

\bibitem[{Kannan et~al.(2024)Kannan, Venkatesh and Min}]{grid39}
Kannan SS, Venkatesh VLN and Min BC (2024) {SMART-LLM: Smart Multi-Agent Robot Task Planning using Large Language Models}.
\newblock In: \emph{2024 IEEE/RSJ International Conference on Intelligent Robots and Systems (IROS)}. pp. 12140--12147.
\newblock \doi{10.1109/iros58592.2024.10802322}.
\newblock \urlprefix\url{https://doi.org/10.1109/iros58592.2024.10802322}.

\bibitem[{Katara et~al.(2023)Katara, Xian and Fragkiadaki}]{gen2sim2023}
Katara P, Xian Z and Fragkiadaki K (2023) {Gen2Sim: Scaling up Robot Learning in Simulation with Generative Models}.
\newblock arXiv:2310.18308.
\newblock \urlprefix\url{https://arxiv.org/abs/2310.18308}.

\bibitem[{Katuwandeniya and Widhanapathirana(2025)}]{roshelpdesk2025}
Katuwandeniya K and Widhanapathirana SRJ (2025) {ROS Help Desk: GenAI Powered, User-Centric Framework for ROS Error Diagnosis and Debugging}.
\newblock \urlprefix\url{https://arxiv.org/abs/2507.07846}.

\bibitem[{Kawabe and Takano(2026)}]{audit260221670}
Kawabe T and Takano R (2026) {Hierarchical LLM-Based Multi-Agent Framework with Prompt Optimization for Multi-Robot Task Planning}.
\newblock arXiv:2602.21670.
\newblock \urlprefix\url{https://arxiv.org/abs/2602.21670v1}.
\newblock Accepted to ICRA 2026.

\bibitem[{Kawaharazuka et~al.(2026)Kawaharazuka, Obinata, Kanazawa, Jia and Okada}]{kawaharazuka2026efficient}
Kawaharazuka K, Obinata Y, Kanazawa N, Jia H and Okada K (2026) Efficient robot design with multi-objective black-box optimization and large language models.
\newblock \emph{IEEE Access} 14: 25404--25413.
\newblock \doi{10.1109/ACCESS.2026.3664844}.
\newblock \urlprefix\url{https://haraduka.github.io/urdf-llm-opt/}.

\bibitem[{Khan et~al.(2025)Khan, Andrev, Murtaza, Aguilera, Zhang, Ding, Hutchinson and Anwar}]{xref250315707}
Khan AA, Andrev M, Murtaza MA, Aguilera S, Zhang R, Ding J, Hutchinson S and Anwar A (2025) {Safety Aware Task Planning via Large Language Models in Robotics}.
\newblock \emph{arXiv preprint arXiv:2503.15707} \urlprefix\url{https://arxiv.org/abs/2503.15707}.

\bibitem[{Khan et~al.(2026)Khan, Yaqoot, Habel, Mustafa and Tsetserukou}]{si260603963}
Khan RA, Yaqoot Y, Habel AA, Mustafa MA and Tsetserukou D (2026) {AgenticRL: Agentic Reinforcement Learning with Self-Refinement for Complex UAV Navigation}.
\newblock \urlprefix\url{https://arxiv.org/abs/2606.03963}.

\bibitem[{Khandelwal and Gupta(2026)}]{khandelwal2026agentrl}
Khandelwal N and Gupta SS (2026) {Agent-Driven Autonomous Reinforcement Learning Research: Iterative Policy Improvement for Quadruped Locomotion}.
\newblock arXiv:2603.27416.
\newblock \urlprefix\url{https://arxiv.org/abs/2603.27416}.

\bibitem[{Khatana et~al.(2026)Khatana, Song, Hovakimyan and Voulgaris}]{khatana2026verifier}
Khatana V, Song Z, Hovakimyan N and Voulgaris P (2026) {Verifier-Mediated Multi-LLM Coordination for Multi-Robot Path Planning}.
\newblock arXiv:2610.12761.
\newblock \urlprefix\url{https://arxiv.org/abs/2610.12761}.

\bibitem[{Kim et~al.(2026)Kim, Trevithick, Wang and Ammanabrolu}]{audit261003615}
Kim B, Trevithick A, Wang R and Ammanabrolu P (2026) {Bridging Frontier Reasoning and Robot Execution: From Autonomous Demonstration Generation to Dense Language Supervision}.
\newblock arXiv:2610.03615.
\newblock \urlprefix\url{https://arxiv.org/abs/2610.03615v1}.

\bibitem[{Kim et~al.(2026{\natexlab{ab}})Kim, Lee, Kim, Kim and Kwak}]{kim2026rhd}
Kim S, Lee J, Kim M, Kim B and Kwak N (2026{\natexlab{ab}}) {Recursive Harness Distillation across Agents for Robot Manipulation}.
\newblock arXiv:2609.33378.
\newblock \urlprefix\url{https://arxiv.org/abs/2609.33378}.

\bibitem[{Kim et~al.(2025)Kim, Kim and Choi}]{grid53}
Kim T, Kim B and Choi J (2025) {Multi-Modal Grounded Planning and Efficient Replanning for Learning Embodied Agents with a Few Examples}.
\newblock \emph{Proceedings of the AAAI Conference on Artificial Intelligence} 39(4): 4329--4337.
\newblock \doi{10.1609/aaai.v39i4.32455}.
\newblock \urlprefix\url{https://doi.org/10.1609/aaai.v39i4.32455}.

\bibitem[{{Kinex contributors}(2026)}]{kinex2026}
{Kinex contributors} (2026) {Kinex: One Agent for Any Robot}.
\newblock Software project.
\newblock \urlprefix\url{https://kinex-agent.org/}.
\newblock Code: \url{https://github.com/stacx-lab/kinex}. Accessed: 2026-11-06.

\bibitem[{Kirk et~al.(2026)Kirk, Wray and Laird}]{kirk2026stars}
Kirk JR, Wray RE and Laird JE (2026) {STARS: extending interactive task learning with large language models}.
\newblock \emph{Frontiers in Computer Science} 8: 1848830.
\newblock \doi{10.3389/fcomp.2026.1848830}.
\newblock \urlprefix\url{https://www.frontiersin.org/journals/computer-science/articles/10.3389/fcomp.2026.1848830/full}.

\bibitem[{Knauer et~al.(2026)Knauer, Bustamante, Eiband, Albu-Schäffer, Stulp and Silvério}]{xref260303897}
Knauer M, Bustamante S, Eiband T, Albu-Schäffer A, Stulp F and Silvério J (2026) {IROSA: Interactive Robot Skill Adaptation Using Natural Language}.
\newblock \emph{IEEE Robotics and Automation Letters} 11(4): 5246--5253.
\newblock \doi{10.1109/lra.2026.3671560}.
\newblock \urlprefix\url{https://doi.org/10.1109/lra.2026.3671560}.

\bibitem[{Knauer et~al.(2026{\natexlab{ab}})Knauer, Gieraths, Mai, Bustamante, Albu-Sch{\"a}ffer, Stulp and Silv{\'e}rio}]{grid28}
Knauer M, Gieraths V, Mai T, Bustamante S, Albu-Sch{\"a}ffer A, Stulp F and Silv{\'e}rio J (2026{\natexlab{ab}}) {CLASP: Language-Driven Robot Skill Selection and Composition using Task-Parameterized Learning}.
\newblock arXiv:2606.08169.
\newblock \urlprefix\url{https://arxiv.org/abs/2606.08169}.

\bibitem[{{Knowin}(2026)}]{knowin2026knowinagent}
{Knowin} (2026) {GLOW: A Generative Learning Framework for General-Purpose Embodied Intelligence}.
\newblock \urlprefix\url{https://www.knowinai.com/en/tech.html}.

\bibitem[{Kozlov et~al.(2026)Kozlov, Maliukov, Marchenko, Plotnikov, Cabrera and Tsetserukou}]{audit260931418}
Kozlov T, Maliukov D, Marchenko A, Plotnikov D, Cabrera MA and Tsetserukou D (2026) {CognitiveReality: Robot-Agnostic Semantic Gaussian Mapping with an LLM Agent for Immersive Collaborative VR Teleoperation}.
\newblock arXiv:2609.31418.
\newblock \urlprefix\url{https://arxiv.org/abs/2609.31418v1}.

\bibitem[{Kumar(2026)}]{kumar2026aor}
Kumar V (2026) {Act-Observe-Rewrite: Multimodal Coding Agents as In-Context Policy Learners for Robot Manipulation}.
\newblock \emph{arXiv preprint arXiv:2603.04466} \urlprefix\url{https://arxiv.org/abs/2603.04466}.

\bibitem[{Kuramshin et~al.(2026)Kuramshin, Aslan, Neary and Berseth}]{kuramshin2026tread}
Kuramshin A, Aslan {\"{O}}, Neary C and Berseth G (2026) {Task Robustness via Re-Labelling Vision-Action Robot Data}.
\newblock \doi{10.48550/arXiv.2606.10918}.
\newblock \urlprefix\url{https://arxiv.org/abs/2606.10918}.

\bibitem[{Kuroki et~al.(2026)Kuroki, Kawaharazuka and Tang}]{kuroki2026closedloop}
Kuroki S, Kawaharazuka K and Tang Y (2026) {Learning and Transferring Closed-Loop Robot Software}.
\newblock \urlprefix\url{https://arxiv.org/abs/2609.19906}.

\bibitem[{Kwon et~al.(2025)Kwon, Choi, Kim, Kim, Moon, Kwak, Huang and Yeo}]{kwon2025memento}
Kwon T, Choi D, Kim H, Kim S, Moon S, Kwak Bw, Huang KH and Yeo J (2025) {Embodied Agents Meet Personalization: Investigating Challenges and Solutions Through the Lens of Memory Utilization}.
\newblock \urlprefix\url{https://arxiv.org/abs/2505.16348}.

\bibitem[{Kwon et~al.(2024)Kwon, Di~Palo and Johns}]{audit231011604}
Kwon T, Di~Palo N and Johns E (2024) {Language Models as Zero-Shot Trajectory Generators}.
\newblock \emph{IEEE Robotics and Automation Letters} 9(7): 6728--6735.
\newblock \doi{10.1109/lra.2024.3410155}.
\newblock \urlprefix\url{https://doi.org/10.1109/lra.2024.3410155}.

\bibitem[{Lai et~al.(2026)Lai, Zhong, Li, Wang, Liu, Tang, Li, Xiao, Yang, Zhao, Mai and Fan}]{lai2026xharness}
Lai Y, Zhong J, Li YJ, Wang L, Liu C, Tang Y, Li M, Xiao X, Yang X, Zhao L, Mai Z and Fan R (2026) {Toward Agentic Robotics: A Large-Scale Study of Harness Design for Robot-Use Agents}.
\newblock \emph{arXiv preprint arXiv:2611.34778} \urlprefix\url{https://arxiv.org/abs/2611.34778}.

\bibitem[{Lang et~al.(2026)Lang, Lu, Sang, Luo, Chen, Li, Yu, Shi and Wang}]{embodirsi2026}
Lang H, Lu H, Sang S, Luo H, Chen G, Li Q, Yu J, Shi Y and Wang J (2026) {EmbodiRSI: Recursive Self-Improvement for Data-Efficient Robot Adaptation}.
\newblock \emph{arXiv preprint arXiv:2609.38905} \urlprefix\url{https://arxiv.org/abs/2609.38905}.

\bibitem[{Lange et~al.(2025)Lange, Yildiz, Arief, Khattak, Kochenderfer and Georgakis}]{lange2025arna}
Lange B, Yildiz A, Arief M, Khattak S, Kochenderfer M and Georgakis G (2025) {General-Purpose Robotic Navigation via LVLM-Orchestrated Perception, Reasoning, and Acting}.
\newblock \emph{arXiv preprint arXiv:2506.17462} \urlprefix\url{https://arxiv.org/abs/2506.17462}.

\bibitem[{Le et~al.(2025)Le, Xie, Liang, Wang, Yang, Ma, Vedder, Krishna, Jayaraman and Eaton}]{le2024articulateanything}
Le L, Xie J, Liang W, Wang HJ, Yang Y, Ma YJ, Vedder K, Krishna A, Jayaraman D and Eaton E (2025) {Articulate-Anything: Automatic Modeling of Articulated Objects via a Vision-Language Foundation Model}.
\newblock In: \emph{International Conference on Learning Representations}. pp. 17578--17602.
\newblock \urlprefix\url{https://proceedings.iclr.cc/paper_files/paper/2025/hash/2c8047bf3ed8ef6905351608d641f02f-Abstract-Conference.html}.

\bibitem[{Lee et~al.(2026)Lee, Gillet, Morency, Breazeal and Park}]{audit260204157}
Lee DW, Gillet S, Morency LP, Breazeal C and Park HW (2026) {A Modern System Recipe for Situated Embodied Human-Robot Conversation with Real-Time Multimodal LLMs and Tool-Calling}.
\newblock arXiv:2602.04157.
\newblock \urlprefix\url{https://arxiv.org/abs/2602.04157v1}.

\bibitem[{Lee et~al.(2026{\natexlab{ab}})Lee, Jung, Cha, Lee, Wang, Choe and Park}]{lee2026beyondcse}
Lee H, Jung H, Cha E, Lee D, Wang YCF, Choe J and Park J (2026{\natexlab{ab}}) {Beyond the Current Scene: Event-Referential Grasping with Active View Selection}.
\newblock CoRL 2026 Workshop on Memory for Robotics.
\newblock \urlprefix\url{https://arxiv.org/abs/2609.39375}.

\bibitem[{Lee et~al.(2026{\natexlab{ac}})Lee, Subramanian, Abbatematteo, Veerabadran, Matias, Ridgeway and Kamra}]{lee2026rda}
Lee H, Subramanian A, Abbatematteo B, Veerabadran V, Matias P, Ridgeway K and Kamra N (2026{\natexlab{ac}}) {RDA: Reward Design Agent for Reinforcement Learning}.
\newblock \emph{arXiv preprint arXiv:2606.01672} \urlprefix\url{https://arxiv.org/abs/2606.01672}.
\newblock Accepted to Reinforcement Learning Conference 2026.

\bibitem[{Lee et~al.(2026{\natexlab{ad}})Lee, Xu, Seely, Lee, Sojoudi, Zaharia and Tang}]{lee2026mass}
Lee H, Xu J, Seely J, Lee D, Sojoudi S, Zaharia M and Tang Y (2026{\natexlab{ad}}) {Recursive Self-Improvement through Multi-Agent Self-Supervision}.
\newblock \emph{arXiv preprint arXiv:2610.12176} \doi{10.48550/arXiv.2610.12176}.
\newblock \urlprefix\url{https://arxiv.org/abs/2610.12176}.

\bibitem[{Lee et~al.(2026{\natexlab{ae}})Lee, Kim, Jeong, Kam, Kim and Lee}]{lee2026emho}
Lee HJ, Kim J, Jeong J, Kam TE, Kim D and Lee YJ (2026{\natexlab{ae}}) {EMHO: EMbodied Agent Harness Optimization via Experience Traces}.
\newblock \urlprefix\url{https://arxiv.org/abs/2610.08432}.

\bibitem[{Lee et~al.(2026{\natexlab{af}})Lee, Baek, Kim, Jeon, Kim, Kim and Kwak}]{lee2026robotuse}
Lee J, Baek I, Kim S, Jeon S, Kim M, Kim B and Kwak N (2026{\natexlab{af}}) {RobotUse: Allocating Computation, Context, and Decisions}.
\newblock \urlprefix\url{https://arxiv.org/abs/2610.04929}.

\bibitem[{Lee et~al.(2026{\natexlab{ag}})Lee, Seo, Cho, Choe, Wang, Saravanan, Huang, Huang, Scherer, Shi, Kim, Lee and Lee}]{lee2026flybycodeembodiedcodingagents}
Lee J, Seo J, Cho S, Choe G, Wang Y, Saravanan B, Huang JB, Huang F, Scherer S, Shi G, Kim HJ, Lee S and Lee D (2026{\natexlab{ag}}) Fly-by-code: Embodied coding agents for aerial manipulation with active visual and physical feedback.
\newblock \urlprefix\url{https://arxiv.org/abs/2610.12549}.

\bibitem[{Lee et~al.(2025)Lee, Hong and Kim}]{lee2025graspcorrect}
Lee S, Hong Y and Kim KI (2025) {GraspCorrect: Robotic Grasp Correction via Vision-Language Model-Guided Feedback}.
\newblock \doi{10.48550/arXiv.2503.15035}.
\newblock \urlprefix\url{https://arxiv.org/abs/2503.15035}.

\bibitem[{Leet et~al.(2026)Leet, Menon, Machcha, Koenig, Chan and Chooi}]{leet2026bitter}
Leet C, Menon A, Machcha S, Koenig S, Chan TK and Chooi J (2026) {The Bitter Lesson Comes to Robotics}.
\newblock \emph{arXiv preprint arXiv:2610.25287} \doi{10.48550/arXiv.2610.25287}.
\newblock \urlprefix\url{https://arxiv.org/abs/2610.25287}.

\bibitem[{Leet et~al.(2026{\natexlab{ab}})Leet, Menon, Machcha, Zou, Patel, Singh, Singh, Vamsi, Ahuja, Yamani, Anand, Alle, Zhang, Chan and Chooi}]{inspectrobots2026}
Leet C, Menon A, Machcha S, Zou S, Patel A, Singh AK, Singh AK, Vamsi G, Ahuja J, Yamani SA, Anand T, Alle V, Zhang ZJ, Chan TK and Chooi J (2026{\natexlab{ab}}) {Inspect Robots: Evaluating the Capabilities and Safety of Embodied AI}.
\newblock \emph{arXiv preprint arXiv:2610.06306} \doi{10.48550/arXiv.2610.06306}.
\newblock \urlprefix\url{https://arxiv.org/abs/2610.06306}.

\bibitem[{Lehman et~al.(2022)Lehman, Gordon, Jain, Ndousse, Yeh and Stanley}]{audit220608896}
Lehman J, Gordon J, Jain S, Ndousse K, Yeh C and Stanley KO (2022) {Evolution through Large Models}.
\newblock arXiv preprint arXiv:2206.08896.
\newblock \urlprefix\url{https://arxiv.org/abs/2206.08896}.

\bibitem[{Lei et~al.(2026)Lei, Song, Zhang, Pei, Chen, Yan, Zhao, Ding, Zhang, Huang, Wang, Wang and Li}]{audit260510921}
Lei H, Song W, Zhang H, Pei J, Chen J, Yan H, Zhao H, Ding P, Zhang Z, Huang L, Wang D, Wang Y and Li H (2026) {RoboMemArena: A Comprehensive and Challenging Robotic Memory Benchmark}.
\newblock arXiv:2605.10921.
\newblock \urlprefix\url{https://arxiv.org/abs/2605.10921v1}.

\bibitem[{Lei et~al.(2026{\natexlab{ab}})Lei, Li, Yap, Zhang, Zhong, Gui, Liu and Deng}]{audit260704162}
Lei IT, Li Q, Yap YJ, Zhang Y, Zhong R, Gui H, Liu X and Deng Z (2026{\natexlab{ab}}) {ACE: Agentic Control for Embodied Manipulation via Zero-shot Workflow Reasoning}.
\newblock arXiv:2607.04162.
\newblock \urlprefix\url{https://arxiv.org/abs/2607.04162v1}.

\bibitem[{Lei et~al.(2025)Lei, Cai, Yang, Wu, Ren, Cui, Tan, Hong, Hu, Zhu, Jiang, Wang, Tan, Wan, Xue, Li, Li, Cui, Zhao and Han}]{xref250801415}
Lei M, Cai H, Yang Y, Wu Y, Ren J, Cui Z, Tan L, Hong J, Hu G, Zhu S, Jiang S, Wang G, Tan J, Wan Z, Xue Z, Li Z, Li Z, Cui S, Zhao Y and Han Y (2025) {RoboMemory: A Brain-inspired Multi-memory Agentic Framework for Interactive Environmental Learning in Physical Embodied Systems}.
\newblock \emph{arXiv preprint arXiv:2508.01415} \urlprefix\url{https://arxiv.org/abs/2508.01415}.
\newblock Accepted to CoRL 2026.

\bibitem[{Lei et~al.(2025{\natexlab{ab}})Lei, Wang, Zhao, Mai, Zhao, Guo, Li, Cui, Han and Ren}]{grid45}
Lei M, Wang G, Zhao Y, Mai Z, Zhao Q, Guo Y, Li Z, Cui S, Han Y and Ren J (2025{\natexlab{ab}}) {CLEA: Closed-Loop Embodied Agent for Enhancing Task Execution in Dynamic Environments}.
\newblock In: \emph{2025 IEEE/RSJ International Conference on Intelligent Robots and Systems (IROS)}. pp. 21048--21054.
\newblock \doi{10.1109/iros60139.2025.11246638}.
\newblock \urlprefix\url{https://doi.org/10.1109/iros60139.2025.11246638}.

\bibitem[{Lei et~al.(2026{\natexlab{ac}})Lei, Liu, Xiong, Xiong, Ding, Zhang, Li and Chen}]{vlaastools2026}
Lei Z, Liu C, Xiong Y, Xiong M, Ding Y, Zhang Z, Li W and Chen S (2026{\natexlab{ac}}) {Towards Long-horizon Embodied Agents with Tool-Aligned Vision-Language-Action Models}.
\newblock \emph{arXiv preprint arXiv:2605.13119} \urlprefix\url{https://arxiv.org/abs/2605.13119}.

\bibitem[{Lei et~al.(2026{\natexlab{ad}})Lei, Liu, Zhang, Liu, Cai, Chen, Wu, Wang, Li, Wen, Zhao, Zhang, Lian and Chen}]{lei2026robocoach}
Lei Z, Liu G, Zhang Y, Liu Q, Cai Y, Chen S, Wu J, Wang Y, Li W, Wen C, Zhao B, Zhang S, Lian W and Chen S (2026{\natexlab{ad}}) {From Digital to Physical: Digital Agents as Autonomous Coaches for Physical Intelligence}.
\newblock \emph{arXiv preprint arXiv:2601.21570} \doi{10.48550/arXiv.2601.21570}.
\newblock \urlprefix\url{https://arxiv.org/abs/2601.21570}.

\bibitem[{Lei et~al.(2026{\natexlab{ae}})Lei, Zhou, Chen, Zhang, Huang, Yin, Yuan, Wu, Li and Chen}]{audit260830396}
Lei Z, Zhou G, Chen XH, Zhang J, Huang Y, Yin H, Yuan H, Wu Q, Li W and Chen S (2026{\natexlab{ae}}) {Scaffolding Foundation Models into Physical-World Agents Pushes the Frontier of Long-Horizon Navigation}.
\newblock arXiv:2608.30396.
\newblock \urlprefix\url{https://arxiv.org/abs/2608.30396v1}.

\bibitem[{Li et~al.(2025)Li, He, Xu, Yuan, Xu, Zang, Hu, Yue, Jiang, Hu, Karlsson, Tang and Lu}]{li2025propriomllm}
Li B, He S, Xu H, Yuan H, Xu X, Zang Y, Hu L, Yue J, Jiang Z, Hu P, Karlsson BF, Tang Y and Lu Z (2025) {Towards Proprioception-Aware Embodied Planning for Dual-Arm Humanoid Robots}.
\newblock \emph{arXiv preprint arXiv:2510.07882} \urlprefix\url{https://arxiv.org/abs/2510.07882}.

\bibitem[{Li et~al.(2026)Li, Song, Wu, Yao, Zhang and Zhang}]{motormind2026}
Li B, Song S, Wu Y, Yao J, Zhang T and Zhang H (2026) {MotorMind: Scaffolding General Vision Language Models for Zero-Shot Robot Manipulation}.
\newblock \emph{arXiv preprint arXiv:2609.38078} \urlprefix\url{https://arxiv.org/abs/2609.38078}.
\newblock Project: https://motor-mind.github.io/.

\bibitem[{Li et~al.(2025{\natexlab{ab}})Li, Wu, Yan, Liu, Li, Zeng and Zha}]{li2025reflectiveadaptation}
Li B, Wu D, Yan Z, Liu X, Li L, Zeng Z and Zha H (2025{\natexlab{ab}}) {Reflection-Based Task Adaptation for Self-Improving VLA}.
\newblock \emph{arXiv preprint arXiv:2510.12710} \urlprefix\url{https://arxiv.org/abs/2510.12710}.

\bibitem[{Li et~al.(2025{\natexlab{ac}})Li, Wu, Abbeel and Malik}]{li2025itp}
Li B, Wu P, Abbeel P and Malik J (2025{\natexlab{ac}}) Interactive task planning with language models.
\newblock \emph{Transactions on Machine Learning Research} \urlprefix\url{https://openreview.net/forum?id=VmfWywWuYQ}.

\bibitem[{Li et~al.(2026{\natexlab{ab}})Li, Zhang and Liu}]{arstag2026}
Li B, Zhang Y and Liu C (2026{\natexlab{ab}}) {ARSTAG: An Agentic Real2Sim2Real System for Task-Specific Robot Data Generation}.
\newblock \emph{arXiv preprint arXiv:2609.24563} \urlprefix\url{https://arxiv.org/abs/2609.24563}.

\bibitem[{Li et~al.(2026{\natexlab{ac}})Li, Zhao, Li, Gao, Liu, Li, Wang, Si, Pan, Ouyang and Zhou}]{robobridge2026}
Li C, Zhao Z, Li R, Gao Y, Liu K, Li J, Wang D, Si T, Pan M, Ouyang W and Zhou D (2026{\natexlab{ac}}) {RoboBridge: A Self-Evolving Embodied Agent Framework for Sim-to-Real Transfer}.
\newblock \urlprefix\url{https://arxiv.org/abs/2610.02717}.

\bibitem[{Li and Zhu(2026)}]{li2026sharppen}
Li C and Zhu W (2026) {Autonomous construction and training of a Sharpa pen-spinning task}.
\newblock Demonstration report.
\newblock \urlprefix\url{https://x.com/walterzhu8/status/2100212420840989112}.

\bibitem[{Li et~al.(2024)Li, Chen and Lin}]{xref240308833}
Li D, Chen W and Lin X (2024) {TINA: Think, Interaction, and Action Framework for Zero-Shot Vision Language Navigation}.
\newblock \emph{arXiv preprint arXiv:2403.08833} \urlprefix\url{https://arxiv.org/abs/2403.08833}.

\bibitem[{Li et~al.(2026{\natexlab{ad}})Li, Lei, Wang, Liu, Yang, Wang, Liu, Zheng and Fan}]{audit260313528}
Li D, Lei J, Wang H, Liu L, Yang Y, Wang Z, Liu B, Zheng M and Fan Z (2026{\natexlab{ad}}) {Learning Actionable Manipulation Recovery via Counterfactual Failure Synthesis}.
\newblock arXiv preprint arXiv:2603.13528.
\newblock \urlprefix\url{https://arxiv.org/abs/2603.13528}.

\bibitem[{Li et~al.(2025{\natexlab{ad}})Li, Peng, Li, Qiao, Zheng, Sun, Qin, Li, Luan, Wu, Zhan, Sun, Xu, Li, Shen and He}]{li2025atomicskilllibrary}
Li D, Peng B, Li C, Qiao N, Zheng Q, Sun L, Qin Y, Li B, Luan Y, Wu B, Zhan Y, Sun M, Xu T, Li L, Shen H and He X (2025{\natexlab{ad}}) {An Atomic Skill Library Construction Method for Data-Efficient Embodied Manipulation}.
\newblock \doi{10.48550/arXiv.2501.15068}.
\newblock \urlprefix\url{https://arxiv.org/abs/2501.15068}.

\bibitem[{Li et~al.(2026{\natexlab{ae}})Li, Han, Li, Li, Wang, Ding, Zhang and Xu}]{audit260302772}
Li G, Han R, Li C, Li H, Wang S, Ding W, Zhang H and Xu C (2026{\natexlab{ae}}) {Self-Evolutionary Replanning for Failure-Aware Motion Planning}.
\newblock arXiv:2603.02772.
\newblock \urlprefix\url{https://arxiv.org/abs/2603.02772}.

\bibitem[{Li et~al.(2026{\natexlab{af}})Li, Wang, Sun, Yuan, Liu, Zong, Hu, Gao, Wang and Yang}]{li2026joyinside}
Li G, Wang L, Sun J, Yuan J, Liu H, Zong H, Hu Y, Gao T, Wang J and Yang X (2026{\natexlab{af}}) {JoyInside-VLM-Agent: Product-Conditioned Multimodal Policy Learning for Embodied Agents}.
\newblock \urlprefix\url{https://arxiv.org/abs/2610.12957}.

\bibitem[{Li et~al.(2026{\natexlab{ag}})Li, Dong, Jiang, Zhou and Ma}]{xref260307997}
Li H, Dong X, Jiang H, Zhou Y and Ma X (2026{\natexlab{ag}}) {CMMR-VLN: Vision-and-Language Navigation via Continual Multimodal Memory Retrieval}.
\newblock \emph{arXiv preprint arXiv:2603.07997} \urlprefix\url{https://arxiv.org/abs/2603.07997}.

\bibitem[{Li et~al.(2026{\natexlab{ah}})Li, You, Su and Guibas}]{audit260220323}
Li H, You Y, Su H and Guibas L (2026{\natexlab{ah}}) {PhysMem: Scaling Test-Time Memory for Embodied Physical Reasoning}.
\newblock arXiv:2602.20323.
\newblock \urlprefix\url{https://arxiv.org/abs/2602.20323v1}.

\bibitem[{Li et~al.(2024{\natexlab{ab}})Li, Chen, Wu, Zheng, Xu and Jia}]{li2024robocoder}
Li J, Chen P, Wu S, Zheng C, Xu H and Jia J (2024{\natexlab{ab}}) {RoboCoder: Robotic Learning from Basic Skills to General Tasks with Large Language Models}.
\newblock \emph{arXiv preprint arXiv:2406.03757} \urlprefix\url{https://arxiv.org/abs/2406.03757}.

\bibitem[{Li et~al.(2026{\natexlab{ai}})Li, Li, Xu, Liu, Wang, Deng and Qin}]{li2026manipscope}
Li J, Li Y, Xu C, Liu Q, Wang H, Deng Z and Qin J (2026{\natexlab{ai}}) {ManipScope: A Diagnostic Benchmark for Instruction Following and Generalization in Robotic Manipulation}.
\newblock \urlprefix\url{https://embodied-ai-research.github.io/ManipScope_webpage/}.

\bibitem[{Li et~al.(2026{\natexlab{aj}})Li, Wang, Li, Zhang, Ye, Liu, Liang and Lian}]{hero2026}
Li J, Wang Y, Li H, Zhang G, Ye Y, Liu Q, Liang H and Lian W (2026{\natexlab{aj}}) {Practice Makes Policies: Bootstrapping and Consolidating Robotic Capabilities from Zero Human Demonstrations}.
\newblock \urlprefix\url{https://arxiv.org/abs/2607.26809}.

\bibitem[{Li et~al.(2026{\natexlab{ak}})Li, Luo, Li, Tian, Liu, Jian, Yang, Xi, Zhang, Luo, Cheng and Lin}]{li2026worldtestsuite}
Li K, Luo M, Li Y, Tian Y, Liu X, Jian X, Yang Y, Xi Z, Zhang Y, Luo Z, Cheng M and Lin H (2026{\natexlab{ak}}) {When the World Is the Test Suite: Agentic Intelligence for the Physical World}.
\newblock \urlprefix\url{https://arxiv.org/abs/2611.22291}.

\bibitem[{Li et~al.(2026{\natexlab{al}})Li, Zhao, Yang, Xu, Wang, Liew, Qu, Shen and Pan}]{li2026spotter}
Li L, Zhao Q, Yang Y, Xu F, Wang Z, Liew AWC, Qu C, Shen HT and Pan S (2026{\natexlab{al}}) Spotter: Let the embodied model lead, and the {VLM} reflect for it.
\newblock \emph{arXiv preprint arXiv:2609.36808} \urlprefix\url{https://arxiv.org/abs/2609.36808}.

\bibitem[{Li et~al.(2024{\natexlab{ac}})Li, Zhao, Wang, Wang, Zhou, Srivastava, Gokmen, Lee, Li, Zhang, Liu, Liang, Li, Mao and Wu}]{eai2024}
Li M, Zhao S, Wang Q, Wang K, Zhou Y, Srivastava S, Gokmen C, Lee T, Li LE, Zhang R, Liu W, Liang P, Li FF, Mao J and Wu J (2024{\natexlab{ac}}) {Embodied Agent Interface: Benchmarking LLMs for Embodied Decision Making}.
\newblock \emph{arXiv preprint arXiv:2410.07166} \urlprefix\url{https://arxiv.org/abs/2410.07166}.
\newblock Accepted to NeurIPS 2024.

\bibitem[{Li et~al.(2026{\natexlab{am}})Li, Zhuang, Gao, Dong, Li, Jiang, Dou, Xi, Zhou, Huang, Li, Gong, Ma, Gui, Wu, Zhang, Huang, Jiang and Qiu}]{li2026fromw1}
Li P, Zhuang Z, Gao Y, Dong Y, Li S, Jiang C, Dou S, Xi Z, Zhou E, Huang J, Li H, Gong J, Ma X, Gui T, Wu Z, Zhang Q, Huang X, Jiang YG and Qiu X (2026{\natexlab{am}}) {FRoM-W1: Towards General Humanoid Whole-Body Control with Language Instructions}.
\newblock \urlprefix\url{https://arxiv.org/abs/2601.12799}.

\bibitem[{Li et~al.(2026{\natexlab{an}})Li, Zhou, Zhu, Chen, Wang, Hu, Yu, Jiang, Su, Ma, Shen, Yang, Ren, Yao, Wang and Mu}]{roboclaw}
Li R, Zhou Y, Zhu Y, Chen K, Wang S, Hu K, Yu M, Jiang B, Su Z, Ma J, Shen Y, Yang Y, Ren G, Yao M, Wang W and Mu Y (2026{\natexlab{an}}) {RoboClaw: An Agentic Framework for Scalable Long-Horizon Robotic Tasks}.
\newblock In: \emph{Computer Vision – ECCV 2026}, Lecture Notes in Computer Science. pp. 304--320.
\newblock \doi{10.1007/978-3-032-36987-1_17}.
\newblock \urlprefix\url{https://doi.org/10.1007/978-3-032-36987-1_17}.

\bibitem[{Li et~al.(2026{\natexlab{ao}})Li, Dong, Ma, Chen, Zhao and Zhou}]{xref260628397}
Li S, Dong X, Ma X, Chen J, Zhao H and Zhou Y (2026{\natexlab{ao}}) {CLOSER-VLN: Closed-Loop Self-Verified Retrieval-Augmented Reasoning for Aerial Vision-Language Navigation}.
\newblock \emph{arXiv preprint arXiv:2606.28397} \urlprefix\url{https://arxiv.org/abs/2606.28397}.

\bibitem[{Li et~al.(2026{\natexlab{ap}})Li, Gu, Liu, Hu, Li, Li, Tang, Wu, Shugurov, Ma, Zollhoefer, An, Mittal, Zhao, Krishna, Li, Liu and Guo}]{humanclaw}
Li S, Gu J, Liu S, Hu K, Li Z, Li L, Tang C, Wu PC, Shugurov I, Ma L, Zollhoefer M, An S, Mittal A, Zhao A, Krishna R, Li M, Liu Z and Guo C (2026{\natexlab{ap}}) {HumanCLAW: Can Vision-Language Models Act Through a Body?}
\newblock arXiv:2607.27180.
\newblock \urlprefix\url{https://arxiv.org/abs/2607.27180}.

\bibitem[{Li et~al.(2026{\natexlab{aq}})Li, Lang, Yan, Yu, Liu, Du, Zheng, Wang and Liu}]{audit260908444}
Li S, Lang T, Yan A, Yu J, Liu F, Du Y, Zheng Y, Wang X and Liu P (2026{\natexlab{aq}}) {Safe Task Planning with Long-Term Graph Memory for Embodied Agents}.
\newblock arXiv:2609.08444.
\newblock \urlprefix\url{https://arxiv.org/abs/2609.08444v1}.
\newblock Accepted to CoRL 2026.

\bibitem[{Li et~al.(2026{\natexlab{ar}})Li, Chi, Zheng, Liu, Wang, Gao, Jiang, Hong, Xiao, Yang and Na}]{li2026embodiedevo}
Li W, Chi Y, Zheng Y, Liu C, Wang X, Gao M, Jiang Z, Hong D, Xiao C, Yang K and Na Q (2026{\natexlab{ar}}) {EmbodiedEvo: Self-Evolving Coding Agent For Physical Intelligence}.
\newblock \emph{arXiv preprint arXiv:2610.18867} \urlprefix\url{https://arxiv.org/abs/2610.18867}.

\bibitem[{Li et~al.(2026{\natexlab{as}})Li, Su, Niu, Cao, Xu, Qu, Fan, You and Xu}]{li2026sentinelvla}
Li W, Su X, Niu D, Cao Y, Xu H, Qu Z, Fan L, You S and Xu C (2026{\natexlab{as}}) Sentinel-{VLA}: A metacognitive {VLA} model with active status monitoring for dynamic reasoning and error recovery.
\newblock In: \emph{Proceedings of the 43rd International Conference on Machine Learning}, \emph{Proceedings of Machine Learning Research}, volume 306. PMLR, pp. 69972--69983.
\newblock \urlprefix\url{https://proceedings.mlr.press/v306/li26fm.html}.

\bibitem[{Li(2026)}]{li2026robouse}
Li X (2026) {Robo Use}: agent as a policy for embodied agents.
\newblock Software, version 0.6.1.
\newblock \urlprefix\url{https://robouse.ai/}.

\bibitem[{Li et~al.(2026{\natexlab{at}})Li, Lyu, Wu, Liu, Liu and Liu}]{audit260402318}
Li X, Lyu F, Wu H, Liu M, Liu JN and Liu G (2026{\natexlab{at}}) {Stop Wandering: Efficient Vision-Language Navigation via Metacognitive Reasoning}.
\newblock arXiv:2604.02318.
\newblock \urlprefix\url{https://arxiv.org/abs/2604.02318v1}.

\bibitem[{Li et~al.(2024{\natexlab{ad}})Li, Zhang, Geng, Geng, Long, Shen, Zhang, Liu and Dong}]{li2024manipllm}
Li X, Zhang M, Geng Y, Geng H, Long Y, Shen Y, Zhang R, Liu J and Dong H (2024{\natexlab{ad}}) {ManipLLM}: Embodied multimodal large language model for object-centric robotic manipulation.
\newblock In: \emph{Proceedings of the IEEE/CVF Conference on Computer Vision and Pattern Recognition}. pp. 18061--18070.
\newblock \urlprefix\url{https://openaccess.thecvf.com/content/CVPR2024/html/Li_ManipLLM_Embodied_Multimodal_Large_Language_Model_for_Object-Centric_Robotic_Manipulation_CVPR_2024_paper.html}.

\bibitem[{Li et~al.(2025{\natexlab{ae}})Li, Deng, Zhang, Jang, Memmel, Yu, Garrett, Ramos, Fox, Li, Gupta and Goyal}]{li2025hamster}
Li Y, Deng Y, Zhang J, Jang J, Memmel M, Yu R, Garrett CR, Ramos F, Fox D, Li A, Gupta A and Goyal A (2025{\natexlab{ae}}) {HAMSTER: Hierarchical Action Models For Open-World Robot Manipulation}.
\newblock \emph{arXiv preprint arXiv:2502.05485} \urlprefix\url{https://arxiv.org/abs/2502.05485}.

\bibitem[{Li et~al.(2026{\natexlab{au}})Li, Feng, Sun, Chen, Ren, Yang, Xu, Wu, Cheng and Xu}]{li2026agenticrsr}
Li Y, Feng Y, Sun S, Chen J, Ren H, Yang B, Xu W, Wu Q, Cheng H and Xu R (2026{\natexlab{au}}) {Agentic RSR: Real-to-Sim-to-Real through Scene Reconstruction and Execution-Grounded Robot Policies}.
\newblock \emph{arXiv preprint arXiv:2610.10479} \urlprefix\url{https://arxiv.org/abs/2610.10479}.

\bibitem[{Li et~al.(2026{\natexlab{av}})Li, Li, Zheng, Cai, Qin and Yang}]{audit260610577}
Li Y, Li C, Zheng H, Cai J, Qin T and Yang M (2026{\natexlab{av}}) {AgenticNav: Zero-Shot Vision-and-Language Navigation as a Tool-Calling Harness}.
\newblock arXiv:2606.10577.
\newblock \urlprefix\url{https://arxiv.org/abs/2606.10577}.

\bibitem[{Li et~al.(2025{\natexlab{af}})Li, Yan, Macaluso, Ji, Zou and Wang}]{li2025lmm3dp}
Li Y, Yan G, Macaluso A, Ji M, Zou X and Wang X (2025{\natexlab{af}}) {Integrating LMM Planners and 3D Skill Policies for Generalizable Manipulation}.
\newblock \doi{10.48550/arXiv.2501.18733}.
\newblock \urlprefix\url{https://arxiv.org/abs/2501.18733}.

\bibitem[{Li et~al.(2026{\natexlab{aw}})Li, You, Xiong, Chen, Zhao, Wu, Zhang and Liu}]{audit260910895}
Li Y, You J, Xiong M, Chen Y, Zhao Z, Wu D, Zhang D and Liu B (2026{\natexlab{aw}}) {ReactHuman: A Physics-Grounded Benchmark for Human-Like Reactive Decision-Making in Embodied Multimodal LLMs}.
\newblock \emph{arXiv preprint arXiv:2609.10895} \doi{10.48550/arXiv.2609.10895}.
\newblock \urlprefix\url{https://arxiv.org/abs/2609.10895}.

\bibitem[{Li et~al.(2026{\natexlab{ax}})Li, Zhou, Chen, Ding, Liu, Han, Zhan, Wang, Zhang and Sheng}]{audit260924124}
Li Y, Zhou E, Chen R, Ding Y, Liu M, Han Y, Zhan J, Wang L, Zhang S and Sheng L (2026{\natexlab{ax}}) {ActiveArena: Benchmarking and Understanding Active Perception in Robotic Manipulation}.
\newblock \emph{arXiv preprint arXiv:2609.24124} \doi{10.48550/arXiv.2609.24124}.
\newblock \urlprefix\url{https://arxiv.org/abs/2609.24124}.

\bibitem[{Li et~al.(2026{\natexlab{ay}})Li, Jin, Liu, Liu, Prasad, D'Eramo and Chalvatzaki}]{li2026harbor}
Li Z, Jin Y, Liu X, Liu P, Prasad V, D'Eramo C and Chalvatzaki G (2026{\natexlab{ay}}) {HARBOR: A Harness Framework for Agentic Robot Reinforcement Learning}.
\newblock \emph{arXiv preprint arXiv:2606.08610} \urlprefix\url{https://arxiv.org/abs/2606.08610}.

\bibitem[{Li et~al.(2026{\natexlab{az}})Li, Li, Zhang, Li and Zhou}]{sref260118492}
Li Z, Li S, Zhang Z, Li B and Zhou S (2026{\natexlab{az}}) {DV-VLN: Dual Verification for Reliable LLM-Based Vision-and-Language Navigation}.
\newblock \urlprefix\url{https://arxiv.org/abs/2601.18492}.

\bibitem[{Li et~al.(2026{\natexlab{ba}})Li, Mao, Ren, Huang, Yang, Huang, Chen and Lv}]{li2026continualdrive}
Li Z, Mao Z, Ren Y, Huang H, Yang H, Huang Q, Chen Z and Lv C (2026{\natexlab{ba}}) {ContinualDrive: A Deployment-Anchored, Multi-Axis Benchmark for Continual Updates of Learned Driving Planners}.
\newblock \urlprefix\url{https://lzrbit.github.io/continualdrive/}.

\bibitem[{Li et~al.(2026{\natexlab{bb}})Li, Ozdil, Zandonati, Cai, Le, Lozano-Pérez, Kaelbling, Du and Ye}]{li2026gema}
Li Z, Ozdil PG, Zandonati B, Cai R, Le H, Lozano-Pérez T, Kaelbling LP, Du Y and Ye W (2026{\natexlab{bb}}) {GEMA: Generative Memory for Robotic Manipulation}.
\newblock CoRL 2026 Workshop on Memory for Robotics.
\newblock \urlprefix\url{https://corl2026-memory.github.io/papers/030.pdf}.

\bibitem[{Li et~al.(2026{\natexlab{bc}})Li, Yu, Zhang, Niu, Qi, Zheng, Chen, Liu, Zhang, Wang and Yi}]{li2026physrsieval}
Li Z, Yu J, Zhang J, Niu L, Qi Z, Zheng J, Chen X, Liu D, Zhang Y, Wang H and Yi L (2026{\natexlab{bc}}) {Where is Physical RSI capability from? An Observational Scaling Law of LLM-as-policy}.
\newblock \emph{arXiv preprint arXiv:2610.26848} \doi{10.48550/arXiv.2610.26848}.
\newblock \urlprefix\url{https://arxiv.org/abs/2610.26848}.

\bibitem[{Li et~al.(2026{\natexlab{bd}})Li, Zhang, Ren, Huang, Lin, Wang, Gao, Guo, Karray, Liu, Russell and Gong}]{si260932698}
Li Z, Zhang H, Ren Z, Huang T, Lin R, Wang H, Gao Z, Guo Y, Karray F, Liu T, Russell C and Gong M (2026{\natexlab{bd}}) {SEES: A Self-Evolving Embodied System via Failure-Guided VLA Policy Adaptation}.
\newblock \urlprefix\url{https://arxiv.org/abs/2609.32698}.

\bibitem[{Li et~al.(2026{\natexlab{be}})Li, Zhang, Huang, Zhou, Li, Wang, Chang, Wei, Zhang, Chen, Zhou, Li and Cheng}]{racap2026}
Li Z, Zhang Y, Huang H, Zhou B, Li W, Wang C, Chang Y, Wei Y, Zhang T, Chen YC, Zhou K, Li Y and Cheng J (2026{\natexlab{be}}) {RACaP: Agentic Reasoning, Acting, and Coding as Policies for Evolvable Robot Learning}.
\newblock \emph{arXiv preprint arXiv:2609.29394} \urlprefix\url{https://arxiv.org/abs/2609.29394}.

\bibitem[{Li et~al.(2026{\natexlab{bf}})Li, Zhang, Li, Huang, Wang, Deng, Wei, Zhang, Xie, Zhou, Chang, Zhou, Chen, Cheng and Li}]{li2026roboharnessk1}
Li Z, Zhang Y, Li W, Huang H, Wang C, Deng S, Wei Y, Zhang T, Xie B, Zhou B, Chang Y, Zhou K, Chen YC, Cheng J and Li Y (2026{\natexlab{bf}}) {Robo-Harness K1: Harnessing Robot-Use Agents via Perception Augmentation}.
\newblock \emph{arXiv preprint arXiv:2609.29389} \urlprefix\url{https://arxiv.org/abs/2609.29389}.

\bibitem[{Liang et~al.(2026)Liang, Chen, Li, Huang, Sun, Yin, Lai, Feng, Jia and Zhang}]{liang2026a2meas}
Liang C, Chen K, Li F, Huang S, Sun Y, Yin Z, Lai J, Feng H, Jia C and Zhang Z (2026) {Toward the Era of Agentic Embodied Intelligence}.
\newblock \urlprefix\url{https://arxiv.org/abs/2611.09124}.

\bibitem[{Liang et~al.(2023)Liang, Huang, Xia, Xu, Hausman, Ichter, Florence and Zeng}]{grid66}
Liang J, Huang W, Xia F, Xu P, Hausman K, Ichter B, Florence P and Zeng A (2023) {Code as Policies: Language Model Programs for Embodied Control}.
\newblock In: \emph{2023 IEEE International Conference on Robotics and Automation (ICRA)}. pp. 9493--9500.
\newblock \doi{10.1109/icra48891.2023.10160591}.
\newblock \urlprefix\url{https://doi.org/10.1109/icra48891.2023.10160591}.

\bibitem[{Liang et~al.(2026{\natexlab{bb}})Liang, Liao, Zhang, Hou, Cai, Deng, He, Huang, Tan, Wang, Yu, Tian, Sartoretti, Luo, Mu, Wu, Li and Lv}]{liang2026robofoundry}
Liang J, Liao S, Zhang S, Hou D, Cai Y, Deng X, He C, Huang W, Tan R, Wang Z, Yu L, Tian X, Sartoretti G, Luo J, Mu Y, Wu W, Li W and Lv C (2026{\natexlab{bb}}) {RoboFoundry: System-as-Policy Evolution for Self-Learning Embodied Agents}.
\newblock \urlprefix\url{https://arxiv.org/abs/2609.32862}.

\bibitem[{Liang et~al.(2024)Liang, Xia, Yu, Zeng, Gonzalez~Arenas, Attarian, Bauza, Bennice, Bewley, Dostmohamed, Fu, Gileadi, Giustina, Gopalakrishnan, Hasenclever, Humplik, Hsu, Joshi, Jyenis, Kew, Kirmani, Lee, Lee, Michaely, Moore, Oslund, Rao, Ren, Tabanpour, Vuong, Wahid, Xiao, Xu, Zhuang, Xu, Frey, Caluwaerts, Zhang, Ichter, Tompson, Takayama, Vanhoucke, Shafran, Mataric, Sadigh, Heess, Rao, Stewart, Tan and Parada}]{liang2024lmpc}
Liang J, Xia F, Yu W, Zeng A, Gonzalez~Arenas M, Attarian M, Bauza M, Bennice M, Bewley A, Dostmohamed A, Fu CK, Gileadi N, Giustina M, Gopalakrishnan K, Hasenclever L, Humplik J, Hsu J, Joshi N, Jyenis B, Kew C, Kirmani S, Lee TWE, Lee KH, Michaely AH, Moore J, Oslund K, Rao D, Ren A, Tabanpour B, Vuong Q, Wahid A, Xiao T, Xu Y, Zhuang V, Xu P, Frey E, Caluwaerts K, Zhang T, Ichter B, Tompson J, Takayama L, Vanhoucke V, Shafran I, Mataric M, Sadigh D, Heess N, Rao K, Stewart N, Tan J and Parada C (2024) {Learning to Learn Faster from Human Feedback with Language Model Predictive Control}.
\newblock \emph{arXiv preprint arXiv:2402.11450} \urlprefix\url{https://arxiv.org/abs/2402.11450}.

\bibitem[{Liang et~al.(2026{\natexlab{bc}})Liang, Zhong, Pan, Zhou, Liu, Li, Li, Chen, Zhang and Zhang}]{liang2026embodiedmemory}
Liang L, Zhong X, Pan M, Zhou X, Liu X, Li Q, Li P, Chen J, Zhang X and Zhang W (2026{\natexlab{bc}}) {EmbodiedMemory-Bench: Benchmarking Embodied Memory for Long-Horizon Embodied Tasks}.
\newblock \doi{10.48550/arXiv.2609.28236}.
\newblock \urlprefix\url{https://arxiv.org/abs/2609.28236}.

\bibitem[{Liang et~al.(2024{\natexlab{bb}})Liang, Wang, Wang, Bastani, Jayaraman and Ma}]{liang2024eurekaverse}
Liang W, Wang S, Wang HJ, Bastani O, Jayaraman D and Ma YJ (2024{\natexlab{bb}}) {Eurekaverse: Environment Curriculum Generation via Large Language Models}.
\newblock \emph{arXiv preprint arXiv:2411.01775} \urlprefix\url{https://arxiv.org/abs/2411.01775}.

\bibitem[{Liang et~al.(2026{\natexlab{bd}})Liang, Wang, Wang, Bastani, Ma and Jayaraman}]{liang2026tether}
Liang W, Wang S, Wang HJ, Bastani O, Ma YJ and Jayaraman D (2026{\natexlab{bd}}) {Tether: Autonomous Functional Play with Correspondence-Driven Trajectory Warping}.
\newblock \emph{arXiv preprint arXiv:2603.03278} \urlprefix\url{https://arxiv.org/abs/2603.03278}.
\newblock Accepted to ICLR 2026.

\bibitem[{Liao et~al.(2026)Liao, Lv, Lew, Zhang, Liang, Li, Cao, Wu and Sartoretti}]{audit260114681}
Liao S, Lv X, Lew J, Zhang S, Liang J, Li P, Cao Y, Wu W and Sartoretti G (2026) {FARE: Fast-Slow Agentic Robotic Exploration}.
\newblock arXiv:2601.14681.
\newblock \urlprefix\url{https://arxiv.org/abs/2601.14681v1}.

\bibitem[{{lilyzhng}(2026)}]{skillclaw2026}
{lilyzhng} (2026) {SkillClaw: Collectively Self-Improving Robot Agents}.
\newblock GitHub repository.
\newblock \urlprefix\url{https://github.com/lilyzhng/SkillClaw}.

\bibitem[{Lim and Clites(2026)}]{lim2026lithe}
Lim HK and Clites TR (2026) {LITHE: Bridging Best-Effort Python and Real-Time C++ for Hot-Swapping Robotic Control Laws on Commodity Linux}.
\newblock \emph{arXiv preprint arXiv:2603.07442} \urlprefix\url{https://arxiv.org/abs/2603.07442}.

\bibitem[{Lim et~al.(2026)Lim, Sun, Bai, Xu, Yao, Yang and Li}]{lim2026prism}
Lim YK, Sun P, Bai Z, Xu X, Yao A, Yang X and Li S (2026) {PRISM: Planning and Reasoning with Intent in Simulated Embodied Environments}.
\newblock \urlprefix\url{https://arxiv.org/abs/2605.11534}.

\bibitem[{Lima et~al.(2026)Lima, Vinci, Günther, Renz, Sung, Stock, Brust, Niecksch, Yi, Igelbrink, Kisliuk, Atzmueller and Hertzberg}]{audit260213081}
Lima O, Vinci M, Günther M, Renz M, Sung A, Stock S, Brust J, Niecksch L, Yi Z, Igelbrink F, Kisliuk B, Atzmueller M and Hertzberg J (2026) {Agentic AI for Robot Control: Flexible but Still Fragile}.
\newblock \emph{Proceedings of the AAAI Symposium Series} 8(1): 465--473.
\newblock \doi{10.1609/aaaiss.v8i1.42578}.
\newblock \urlprefix\url{https://doi.org/10.1609/aaaiss.v8i1.42578}.

\bibitem[{Lin et~al.(2025)Lin, Yuan, Wang, Qiu, Wang, Guo, Wang, Narang, Fox and Gan}]{lin2025robotsmithgenerativerobotictool}
Lin C, Yuan H, Wang Y, Qiu X, Wang TH, Guo M, Wang B, Narang Y, Fox D and Gan C (2025) {RobotSmith: Generative Robotic Tool Design for Acquisition of Complex Manipulation Skills}.
\newblock \urlprefix\url{https://arxiv.org/abs/2506.14763}.

\bibitem[{Lin et~al.(2026)Lin, Zhang, Yu, Chen, Griffiths, Choi, Held and Gan}]{robowits2026}
Lin C, Zhang H, Yu F, Chen Z, Griffiths TL, Choi Y, Held D and Gan C (2026) {RoboWits: Unexpected Challenges for Robotic Creative Problem Solving}.
\newblock \emph{arXiv preprint arXiv:2605.30326} \urlprefix\url{https://arxiv.org/abs/2605.30326}.

\bibitem[{Lin et~al.(2026{\natexlab{bb}})Lin, Chen, Chang, Chen, Chang, Huang, Yang, Chen, Chen, Wang and Sun}]{lin2026physcap}
Lin CY, Chen JW, Chang HE, Chen HA, Chang SH, Huang CP, Yang FE, Chen MH, Chen YT, Wang YCF and Sun SH (2026{\natexlab{bb}}) {PhysCaP: Grounding Code-as-Policy Agent with Physics-Informed Exploration}.
\newblock CoRL 2026 Workshop on Agentic Robotics.
\newblock \urlprefix\url{https://arxiv.org/abs/2608.21031}.
\newblock ArXiv:2608.21031. \href{https://physcap.github.io/}{Project page} and \href{https://github.com/physcap/physcap}{code}.

\bibitem[{Lin et~al.(2023)Lin, Agia, Migimatsu, Pavone and Bohg}]{grid67}
Lin K, Agia C, Migimatsu T, Pavone M and Bohg J (2023) {Text2Motion: from natural language instructions to feasible plans}.
\newblock \emph{Autonomous Robots} 47(8): 1345--1365.
\newblock \doi{10.1007/s10514-023-10131-7}.
\newblock \urlprefix\url{https://doi.org/10.1007/s10514-023-10131-7}.

\bibitem[{Lin et~al.(2026{\natexlab{bc}})Lin, Zhu, Zhang, Liu, Li, Zeng, Wang and Li}]{lin2026miracompanion}
Lin L, Zhu Y, Zhang F, Liu Y, Li B, Zeng X, Wang J and Li Y (2026{\natexlab{bc}}) {MIRA: Real-Time Full-Duplex Human-Robot Interaction for Embodied Companions}.
\newblock \urlprefix\url{https://arxiv.org/abs/2609.24547}.

\bibitem[{Lin et~al.(2026{\natexlab{bd}})Lin, Du, Wang, Zhong and Zhao}]{lin2026fastharness}
Lin T, Du Y, Wang H, Zhong Z and Zhao B (2026{\natexlab{bd}}) Fastharness: Experience-driven review scheduling for vision-language-action policies.
\newblock \emph{arXiv preprint arXiv:2610.31427} \urlprefix\url{https://arxiv.org/abs/2610.31427}.

\bibitem[{Lin et~al.(2026{\natexlab{be}})Lin, Deng, Yuan, Yang, Xue and Wang}]{lin2026uniintervenepp}
Lin Y, Deng H, Yuan Z, Yang Z, Xue Y and Wang Z (2026{\natexlab{be}}) {UniIntervene++: An Adaptive Intervention Agent for Efficient Real-World Reinforcement Learning}.
\newblock \emph{arXiv preprint arXiv:2610.03620} \urlprefix\url{https://arxiv.org/abs/2610.03620}.

\bibitem[{Lin et~al.(2026{\natexlab{bf}})Lin, Liu, Qi, Xue, Song, Qin and Liu}]{lin2026acdc}
Lin Y, Liu SZ, Qi R, Xue GZ, Song X, Qin C and Liu HHT (2026{\natexlab{bf}}) {Agentic Aerial Cinematography: From Dialogue Cues to Cinematic Trajectories}.
\newblock CoRL 2026 Workshop on Agentic Robotics.
\newblock \urlprefix\url{https://openreview.net/forum?id=AVqVpgtKYF}.

\bibitem[{Lin et~al.(2026{\natexlab{bg}})Lin, Jiang, Cen, Wu, Zhang and Zheng}]{grid63}
Lin Z, Jiang JJ, Cen H, Wu XM, Zhang D and Zheng WS (2026{\natexlab{bg}}) {HATS: A Human-Agent Teleoperation System for Multi-Arm Data Collection}.
\newblock arXiv:2606.16491.
\newblock \urlprefix\url{https://arxiv.org/abs/2606.16491}.

\bibitem[{Lin et~al.(2025{\natexlab{bb}})Lin, Tang, Ye and Zhang}]{xref250302698}
Lin Z, Tang C, Ye H and Zhang H (2025{\natexlab{bb}}) {FlowPlan: Zero-Shot Task Planning with LLM Flow Engineering for Robotic Instruction Following}.
\newblock \emph{arXiv preprint arXiv:2503.02698} \urlprefix\url{https://arxiv.org/abs/2503.02698}.

\bibitem[{Lin et~al.(2026{\natexlab{bh}})Lin, Xu, Bian, Wang and Gui}]{audit261002196}
Lin Z, Xu S, Bian L, Wang YX and Gui LY (2026{\natexlab{bh}}) {InterEvolve: Test-Time Evolution of Reward Programs for Humanoid Loco-Manipulation}.
\newblock arXiv:2610.02196.
\newblock \urlprefix\url{https://arxiv.org/abs/2610.02196v1}.

\bibitem[{Liu et~al.(2026)Liu, Liu, Ding, Jiang, Chen, Tang, Leng, Zhang and Cheng}]{audit260729600}
Liu A, Liu B, Ding H, Jiang Y, Chen Y, Tang F, Leng C, Zhang H and Cheng J (2026) {HAM-VLN: Harnessing Hierarchical Agentic Memory for Zero-Shot Vision-and-Language Navigation}.
\newblock arXiv:2607.29600.
\newblock \urlprefix\url{https://arxiv.org/abs/2607.29600v1}.

\bibitem[{Liu et~al.(2025)Liu, Liu, Wang, Yang and Chen}]{liu2025colmdriver}
Liu C, Liu G, Wang Z, Yang J and Chen S (2025) {CoLMDriver: LLM-based Negotiation Benefits Cooperative Autonomous Driving}.
\newblock \doi{10.48550/arXiv.2503.08683}.
\newblock \urlprefix\url{https://arxiv.org/abs/2503.08683}.

\bibitem[{Liu et~al.(2026{\natexlab{bb}})Liu, Zhang, Lin, Wang, Wang, Wang and Qi}]{audit261003476}
Liu C, Zhang Y, Lin J, Wang J, Wang B, Wang Z and Qi X (2026{\natexlab{bb}}) {MobiAgent: Dual-Loop Recursive Policy Self-Improvement for Long-Horizon Mobile Manipulation}.
\newblock arXiv:2610.03476.
\newblock \urlprefix\url{https://arxiv.org/abs/2610.03476v1}.
\newblock Accepted to CoRL 2026.

\bibitem[{Liu et~al.(2024)Liu, Fang, Abbeel and Levine}]{liu2024moka}
Liu F, Fang K, Abbeel P and Levine S (2024) {MOKA: Open-World Robotic Manipulation through Mark-Based Visual Prompting}.
\newblock \emph{arXiv preprint arXiv:2403.03174} \urlprefix\url{https://arxiv.org/abs/2403.03174}.

\bibitem[{Liu et~al.(2026{\natexlab{bc}})Liu, Shen, Cheng, Mi, Tang, Liu, Xiang, Li, Li and Tang}]{roboicl2026}
Liu F, Shen Y, Cheng A, Mi W, Tang C, Liu C, Xiang Y, Li T, Li YL and Tang Y (2026{\natexlab{bc}}) {RoboICL: Embodied In-Context Learning with GPT-6 Astra}.
\newblock \emph{arXiv preprint arXiv:2609.34261} \urlprefix\url{https://arxiv.org/abs/2609.34261}.

\bibitem[{Liu et~al.(2023)Liu, Chen, Zhu, Swaminathan, Kolobov and Cheng}]{liu2023olaf}
Liu H, Chen A, Zhu Y, Swaminathan A, Kolobov A and Cheng CA (2023) {Interactive Robot Learning from Verbal Correction}.
\newblock \emph{arXiv preprint arXiv:2310.17555} \urlprefix\url{https://arxiv.org/abs/2310.17555}.

\bibitem[{Liu et~al.(2025{\natexlab{bb}})Liu, Guo, Mai, Cao, Li and Ma}]{liu2025robodexvlm}
Liu H, Guo S, Mai P, Cao J, Li H and Ma J (2025{\natexlab{bb}}) {RoboDexVLM: Visual Language Model-Enabled Task Planning and Motion Control for Dexterous Robot Manipulation}.
\newblock \doi{10.48550/arXiv.2503.01616}.
\newblock \urlprefix\url{https://arxiv.org/abs/2503.01616}.

\bibitem[{Liu et~al.(2026{\natexlab{bd}})Liu, Li, Ding, Wen, Ren, Ma, Jian, Wang, Yu, Li and Song}]{revolvecollection2026}
Liu H, Li Q, Ding Y, Wen J, Ren K, Ma Y, Jian T, Wang Z, Yu Z, Li X and Song Z (2026{\natexlab{bd}}) {REVOLVE: An Automated Closed-Loop Framework for Evolving Robot Manipulation with Minimal Human Intervention}.
\newblock arXiv:2609.14633.
\newblock \urlprefix\url{https://arxiv.org/abs/2609.14633}.

\bibitem[{Liu et~al.(2026{\natexlab{be}})Liu, Li, Yao, Shi, Zhou, Huang, Huang and Mao}]{grid26}
Liu H, Li X, Yao S, Shi P, Zhou T, Huang JB, Huang F and Mao J (2026{\natexlab{be}}) {Guava: Distilling Frontier VLMs into a Compact Agent through a Robotic Manipulation Harness}.
\newblock arXiv:2606.18363.
\newblock \urlprefix\url{https://arxiv.org/abs/2606.18363}.

\bibitem[{Liu et~al.(2026{\natexlab{bf}})Liu, Xue, Zhou, Yang, Deng, Liang, Xie and Wang}]{liu2026unimanip}
Liu H, Xue Y, Zhou Y, Yang L, Deng H, Liang Y, Xie L and Wang Z (2026{\natexlab{bf}}) {UniManip}: Open-world robotic manipulation via agentic operational graphs.
\newblock arXiv:2602.13086.
\newblock \urlprefix\url{https://arxiv.org/abs/2602.13086}.

\bibitem[{Liu et~al.(2026{\natexlab{bg}})Liu, Cheng, Bjorck, Yu, Yin, Kautz, Fan, Zhu and Liu}]{recova2026}
Liu I, Cheng AC, Bjorck J, Yu Z, Yin H, Kautz J, Fan L, Zhu Y and Liu S (2026{\natexlab{bg}}) {Recova: Agent-Guided Failure Recovery for Autonomous Robotic Manipulation}.
\newblock \urlprefix\url{https://arxiv.org/abs/2610.01178}.

\bibitem[{Liu et~al.(2026{\natexlab{bh}})Liu, Chen, Liu, Zhang, Chen, Xie, Yao, Xu, Cui and Zhang}]{liu2026robocoach}
Liu J, Chen Y, Liu Y, Zhang J, Chen R, Xie S, Yao G, Xu M, Cui S and Zhang C (2026{\natexlab{bh}}) Robocoach: World models as active coaches for compositional robot skills.
\newblock \emph{arXiv preprint arXiv:2609.39685} \urlprefix\url{https://arxiv.org/abs/2609.39685}.

\bibitem[{Liu et~al.(2026{\natexlab{bi}})Liu, Wang, Wang, Hong, Gao, Li, Chi, Zheng, Yang, Wang, He, Zuo, Chen, Chen, Qian, Chen, Zhou, Yao, Tang, Zhou, Zhang, Xiao, Altman and ang Gao}]{liu2026robouniverse}
Liu J, Wang Y, Wang S, Hong D, Gao M, Li W, Chi Y, Zheng Y, Yang K, Wang X, He B, Zuo Y, Chen B, Chen Y, Qian C, Chen T, Zhou H, Yao H, Tang H, Zhou H, Zhang YQ, Xiao C, Altman T and ang Gao H (2026{\natexlab{bi}}) {RoboUniverse-10K: Toward the Scaling Era of Agentic Robotics via Real-to-Sim Environments}.
\newblock \urlprefix\url{https://arxiv.org/abs/2611.31002}.

\bibitem[{Liu et~al.(2026{\natexlab{bj}})Liu, Yao, Gu and Xu}]{audit260927526}
Liu J, Yao S, Gu J and Xu L (2026{\natexlab{bj}}) {NavProbe: Evidence-Grounded Reasoning with Active Memory Retrieval for Zero-Shot Navigation}.
\newblock arXiv:2609.27526.
\newblock \urlprefix\url{https://arxiv.org/abs/2609.27526v1}.

\bibitem[{Liu et~al.(2023{\natexlab{bb}})Liu, Yang, Idrees, Liang, Schornstein, Tellex and Shah}]{audit230211649}
Liu JX, Yang Z, Idrees I, Liang S, Schornstein B, Tellex S and Shah A (2023{\natexlab{bb}}) Grounding complex natural language commands for temporal tasks in unseen environments.
\newblock In: \emph{Proceedings of The 7th Conference on Robot Learning}, \emph{Proceedings of Machine Learning Research}, volume 229. PMLR, pp. 1084--1110.
\newblock \doi{10.48550/arXiv.2302.11649}.
\newblock \urlprefix\url{https://proceedings.mlr.press/v229/liu23d.html}.

\bibitem[{Liu et~al.(2024{\natexlab{bb}})Liu, Tang, Wang, Wang, Li and Zhao}]{xref240915146}
Liu K, Tang Z, Wang D, Wang Z, Li X and Zhao B (2024{\natexlab{bb}}) {COHERENT: Collaboration of Heterogeneous Multi-Robot System with Large Language Models}.
\newblock \emph{arXiv preprint arXiv:2409.15146} \urlprefix\url{https://arxiv.org/abs/2409.15146}.
\newblock Accepted to ICRA 2025.

\bibitem[{Liu et~al.(2026{\natexlab{bk}})Liu, Han, Yi, Yu, Wang, Du, Yuan, Yuan, Feng, Liu, Zhang and Yu}]{audit260604172}
Liu L, Han Y, Yi P, Yu W, Wang H, Du H, Yuan E, Yuan Z, Feng R, Liu M, Zhang Q and Yu J (2026{\natexlab{bk}}) {Affordance2Action: Task-Conditioned Scene-level Affordance Grounding for Real-Time Manipulation}.
\newblock arXiv preprint arXiv:2606.04172.
\newblock \urlprefix\url{https://arxiv.org/abs/2606.04172}.

\bibitem[{Liu et~al.(2026{\natexlab{bl}})Liu, Corcodel and Zhao}]{liu2026elevate}
Liu S, Corcodel R and Zhao D (2026{\natexlab{bl}}) {ELEVATE: Agentic Controller Acquisition from Task Failure for Quadrupedal Manipulation}.
\newblock arXiv:2610.38388.
\newblock \urlprefix\url{https://arxiv.org/abs/2610.38388}.

\bibitem[{Liu et~al.(2026{\natexlab{bm}})Liu, Yang, Huang, Wang, Han and Chen}]{liu2026agenticlonglive}
Liu S, Yang S, Huang W, Wang L, Han S and Chen Y (2026{\natexlab{bm}}) {Agentic LongLive: Asynchronous Agentic Memory for Long Video Generation}.
\newblock \emph{arXiv preprint arXiv:2611.34907} \urlprefix\url{https://arxiv.org/abs/2611.34907}.

\bibitem[{Liu et~al.(2026{\natexlab{bn}})Liu, Wang, Zhang, Wang, Wei, Xiang, Wang, Ye, Dai, Liu, Xu, Chen, Sun, Shen, He and Yang}]{audit260213591}
Liu W, Wang Y, Zhang X, Wang Y, Wei J, Xiang Y, Wang Y, Ye M, Dai E, Liu Z, Xu Y, Chen X, Sun H, Shen J, He J and Yang T (2026{\natexlab{bn}}) {AgentRob: From Virtual Forum Agents to Hijacked Physical Robots}.
\newblock arXiv:2602.13591.
\newblock \urlprefix\url{https://arxiv.org/abs/2602.13591v1}.

\bibitem[{Liu et~al.(2026{\natexlab{bo}})Liu, Cui, Yao, Cao, Ma, Zhang and Zhang}]{liu2026scene2demo}
Liu X, Cui S, Yao G, Cao Z, Ma J, Zhang M and Zhang C (2026{\natexlab{bo}}) {Scene2Demo: Self-Evolving Embodied Data Generation via Object-Action Graph}.
\newblock \emph{arXiv preprint arXiv:2602.12065} \urlprefix\url{https://arxiv.org/abs/2602.12065}.

\bibitem[{Liu et~al.(2026{\natexlab{bp}})Liu, Sadikoglu, Chatterjee and Senanayake}]{audit260822657}
Liu X, Sadikoglu E, Chatterjee R and Senanayake R (2026{\natexlab{bp}}) {Physical Agentic AI: An Architecture for Orchestrating a Robot Crew with LLMs}.
\newblock arXiv:2608.22657.
\newblock \urlprefix\url{https://arxiv.org/abs/2608.22657v1}.

\bibitem[{Liu et~al.(2026{\natexlab{bq}})Liu, Xu, Bao, Li, Bao, Chen and Shang}]{liu2026agenticroborl}
Liu X, Xu B, Bao W, Li M, Bao W, Chen C and Shang Y (2026{\natexlab{bq}}) {AgenticRoboRL: Bringing Coding-Agent Post-Training to Agentic Robotics}.
\newblock \emph{arXiv preprint arXiv:2610.28589} \urlprefix\url{https://arxiv.org/abs/2610.28589}.

\bibitem[{Liu et~al.(2026{\natexlab{br}})Liu, Chen, Song, Pu, Mo, Bai, Chen, Sun, Zhong, Shen and Lin}]{phyagentos2026}
Liu Y, Chen W, Song X, Pu T, Mo S, Bai Y, Chen Z, Sun Q, Zhong L, Shen Y and Lin L (2026{\natexlab{br}}) {PhyAgentOS: A Self-Evolving Operating System for Embodied Agents with Decoupled Cognitive Planning and Physical Execution}.
\newblock \emph{arXiv preprint arXiv:2607.16636} \urlprefix\url{https://arxiv.org/abs/2607.16636}.

\bibitem[{Liu et~al.(2026{\natexlab{bs}})Liu, Chen and Zheng}]{liu2026geometricabstraction}
Liu Y, Chen Y and Zheng M (2026{\natexlab{bs}}) {Do Agentic Policies Need to Look at Every Step? Geometric Abstraction for Reasoning and Action}.
\newblock CoRL 2026 Workshop on Agentic Robotics.
\newblock \urlprefix\url{https://openreview.net/forum?id=CQFstcjx1Z}.

\bibitem[{Liu et~al.(2026{\natexlab{bt}})Liu, Mao, Hsu, Kaelbling and Lozano-P{\'e}rez}]{rapid2026}
Liu Y, Mao J, Hsu D, Kaelbling LP and Lozano-P{\'e}rez T (2026{\natexlab{bt}}) {RAPID: Robot Agentic Programming from Demonstrations}.
\newblock \emph{arXiv preprint arXiv:2609.30249} \urlprefix\url{https://arxiv.org/abs/2609.30249}.

\bibitem[{Liu et~al.(2026{\natexlab{bu}})Liu, Wang and Ye}]{vcage2026}
Liu Y, Wang Ab and Ye N (2026{\natexlab{bu}}) {V-CAGE: Vision-Closed-Loop Agentic Generation Engine for Robotic Manipulation}.
\newblock arXiv:2604.09036.
\newblock \urlprefix\url{https://arxiv.org/abs/2604.09036}.

\bibitem[{Liu et~al.(2026{\natexlab{bv}})Liu, Wang, Wu, He, Zhan, Ma, Chen, Ge, Tang, Gu, Zhang, Wang, Niu, Dong, Zhu, Darrell, Fan, Han and Cai}]{liu2026robotaembed}
Liu Y, Wang J, Wu S, He W, Zhan G, Ma X, Chen J, Ge W, Tang H, Gu Y, Zhang Z, Wang Z, Niu D, Dong Z, Zhu Y, Darrell T, Fan LJ, Han S and Cai H (2026{\natexlab{bv}}) {Robota-Embed: Unified VLA Embeddings for Robot In-Context Learning and Evolution}.
\newblock \emph{arXiv preprint arXiv:2610.19495} \urlprefix\url{https://arxiv.org/abs/2610.19495}.

\bibitem[{Liu et~al.(2023{\natexlab{bc}})Liu, Bahety and Song}]{grid47}
Liu Z, Bahety A and Song S (2023{\natexlab{bc}}) Reflect: Summarizing robot experiences for failure explanation and correction.
\newblock In: \emph{Proceedings of The 7th Conference on Robot Learning}, \emph{Proceedings of Machine Learning Research}, volume 229. PMLR, pp. 3468--3484.
\newblock \urlprefix\url{https://proceedings.mlr.press/v229/liu23g.html}.

\bibitem[{Liu et~al.(2026{\natexlab{bw}})Liu, Fernandez-Ayala, Wang, Qin, Wang, Dimarogonas and Wang}]{liu2026industrialsdi}
Liu Z, Fernandez-Ayala VN, Wang T, Qin Q, Wang XV, Dimarogonas DV and Wang L (2026{\natexlab{bw}}) {Agentic Neuro-Symbolic Planning and Commissioning for Human-in-the-Loop Industrial Robotics with Digital Twins}.
\newblock \doi{10.48550/arXiv.2606.08214}.
\newblock \urlprefix\url{https://arxiv.org/abs/2606.08214}.

\bibitem[{Liu et~al.(2026{\natexlab{bx}})Liu, Ning, Hu, Xie, Li, Tang, Wang, Yang, Wang, Liu and Pu}]{grid25}
Liu Z, Ning X, Hu Z, Xie X, Li W, Tang Z, Wang C, Yang Z, Wang H, Liu Y and Pu Z (2026{\natexlab{bx}}) {Goal2Skill: Long-Horizon Manipulation with Adaptive Planning and Reflection}.
\newblock arXiv:2604.13942.
\newblock \urlprefix\url{https://arxiv.org/abs/2604.13942}.

\bibitem[{Liu et~al.(2026{\natexlab{by}})Liu, Niu, Zhang, Shao, Wang, Zheng, Sharma, Lu, Liu, Zhang, Herzig, Bai, Goldberg, Xu, Zhu, Wang, Fan and Darrell}]{liu2026aw0}
Liu Z, Niu D, Zhang B, Shao B, Wang J, Zheng R, Sharma Y, Lu R, Liu J, Zhang R, Herzig R, Bai Y, Goldberg K, Xu D, Zhu Y, Wang G, Fan LJ and Darrell T (2026{\natexlab{by}}) {Agentic World Action Models}.
\newblock \emph{arXiv preprint arXiv:2610.30136} \urlprefix\url{https://arxiv.org/abs/2610.30136}.

\bibitem[{Long et~al.(2024)Long, Shi, Liu and Li}]{long2024vlmmpc}
Long K, Shi H, Liu J and Li X (2024) {VLM-MPC: Vision Language Foundation Model (VLM)-Guided Model Predictive Controller (MPC) for Autonomous Driving}.
\newblock \doi{10.48550/arXiv.2408.04821}.
\newblock \urlprefix\url{https://arxiv.org/abs/2408.04821}.

\bibitem[{Long et~al.(2023)Long, Li, Cai and Dong}]{xref230911382}
Long Y, Li X, Cai W and Dong H (2023) {Discuss Before Moving: Visual Language Navigation via Multi-expert Discussions}.
\newblock \emph{arXiv preprint arXiv:2309.11382} \urlprefix\url{https://arxiv.org/abs/2309.11382}.

\bibitem[{Lou et~al.(2026)Lou, Yang, Wang, Lv, Liu, Hou, Liu, Shi and Wu}]{audit260918520}
Lou J, Yang Y, Wang H, Lv X, Liu X, Hou D, Liu X, Shi R and Wu W (2026) {AeroWeaver: An Embodied-Agent Harness for Weaving Aerial Skills into Distributed, Adaptive Swarm Execution}.
\newblock arXiv:2609.18520.
\newblock \urlprefix\url{https://arxiv.org/abs/2609.18520v1}.

\bibitem[{Lou et~al.(2026{\natexlab{bb}})Lou, Zeng, Wang and Si}]{lou2026knowbody}
Lou Z, Zeng Y, Wang Y and Si C (2026{\natexlab{bb}}) {Know Your Body: A Harness for Direct and Self-Improving Robot Control with VLMs}.
\newblock arXiv:2609.28530.
\newblock \urlprefix\url{https://arxiv.org/abs/2609.28530}.

\bibitem[{Lu et~al.(2026)Lu, Lu, Lange, Yamada, Hu, Foerster, Ha and Clune}]{lu2026aiscientist}
Lu C, Lu C, Lange RT, Yamada Y, Hu S, Foerster J, Ha D and Clune J (2026) Towards end-to-end automation of {AI} research.
\newblock \emph{Nature} 651(8107): 914--919.
\newblock \doi{10.1038/s41586-026-10265-5}.
\newblock \urlprefix\url{https://www.nature.com/articles/s41586-026-10265-5}.

\bibitem[{Lu et~al.(2026{\natexlab{bb}})Lu, Shen, Yu, Xiao, Liu, Zhang, Wu, Chen, Ye, Wang, Wang and Liu}]{lu2026annotateanything}
Lu H, Shen M, Yu S, Xiao Y, Liu S, Zhang J, Wu S, Chen Y, Ye G, Wang J, Wang Z and Liu H (2026{\natexlab{bb}}) {AnnotateAnything: Automatic Annotation of 3D Assets for Robot Manipulation}.
\newblock \emph{arXiv preprint arXiv:2606.17446} \urlprefix\url{https://arxiv.org/abs/2606.17446}.

\bibitem[{Lu et~al.(2026{\natexlab{bc}})Lu, Xiao and Hsu}]{lu2026migu}
Lu M, Xiao A and Hsu D (2026{\natexlab{bc}}) Migu: Multimodal instruction grounding under uncertainty for manipulation planning.
\newblock \emph{arXiv preprint arXiv:2609.24995} .

\bibitem[{Lu et~al.(2026{\natexlab{bd}})Lu, Wu, Kou, Lin, Fu, Xiao, Mandlekar, Xu, Shi, Goldberg, Chen, Chowdhury, Kautz, Zhu, Fan and Wang}]{self260700272}
Lu R, Wu Y, Kou E, Lin H, Fu L, Xiao W, Mandlekar A, Xu Y, Shi G, Goldberg K, Chen A, Chowdhury M, Kautz J, Zhu Y, Fan LJ and Wang G (2026{\natexlab{bd}}) {ASPIRE: Agentic /Skills Discovery for Robotics}.
\newblock arXiv:2607.00272.
\newblock \urlprefix\url{https://arxiv.org/abs/2607.00272}.

\bibitem[{Lu et~al.(2025)Lu, Chen, Hu, Zhou, Zhang, Liu, Sheng and Shao}]{lu2025isbench}
Lu X, Chen Z, Hu X, Zhou Y, Zhang W, Liu D, Sheng L and Shao J (2025) Is-bench: Evaluating interactive safety of vlm-driven embodied agents in daily household tasks.
\newblock \urlprefix\url{https://arxiv.org/abs/2506.16402}.

\bibitem[{Luo et~al.(2025)Luo, Zeng, Jia, Du and Sun}]{si250606658}
Luo C, Zeng Z, Jia M, Du Y and Sun C (2025) {Self-Improving Loops for Visual Robotic Planning}.
\newblock \urlprefix\url{https://arxiv.org/abs/2506.06658}.
\newblock Accepted to ICLR 2026.

\bibitem[{Luo et~al.(2026)Luo, Qin, Cao, Huang, Yang, Chen, Lu, Gui, Shi and Wang}]{luo2026instagent}
Luo H, Qin J, Cao J, Huang Y, Yang G, Chen X, Lu H, Gui T, Shi Y and Wang J (2026) {InstAgent: A Recursively Self-Improving Agent Operating System for Heterogeneous Robot Collaboration}.
\newblock \emph{arXiv preprint arXiv:2610.23098} \doi{10.48550/arXiv.2610.23098}.
\newblock \urlprefix\url{https://arxiv.org/abs/2610.23098}.

\bibitem[{Luo et~al.(2026{\natexlab{bb}})Luo, Bai, Xu, Li, Li, Qian, Lu, Zhang, Zou, Luo, Chen and Lin}]{luo2026physicalrsi}
Luo M, Bai Z, Xu S, Li Y, Li K, Qian M, Lu W, Zhang R, Zou Y, Luo Z, Chen W and Lin H (2026{\natexlab{bb}}) {Toward Recursive Self-Improvement in Physical AI}.
\newblock \emph{arXiv preprint arXiv:2610.29677} \urlprefix\url{https://arxiv.org/abs/2610.29677}.

\bibitem[{Luo et~al.(2026{\natexlab{bc}})Luo, Guo, Qin, Guo, Zhao, Ma and Yang}]{luo2026robolineage}
Luo Q, Guo W, Qin Z, Guo N, Zhao Y, Ma Y and Yang Y (2026{\natexlab{bc}}) {RoboLineage: Agent-Native Data Lifecycle Governance Across Robot Policy Iterations}.
\newblock \emph{arXiv preprint arXiv:2606.22142} \urlprefix\url{https://arxiv.org/abs/2606.22142}.
\newblock Accepted to CoRL 2026.

\bibitem[{Luo et~al.(2026{\natexlab{bd}})Luo, Liu, Cao, Tancik, Buch, Stearns and Xian}]{luo2026scenerig}
Luo R, Liu Y, Cao D, Tancik M, Buch S, Stearns C and Xian W (2026{\natexlab{bd}}) {SceneRig}: An agentic system for simulation-ready 3d scene reconstruction from single images.
\newblock arXiv:2610.28585.
\newblock \urlprefix\url{https://arxiv.org/abs/2610.28585}.
\newblock Accepted to CoRL 2026 Agentic Robotics Workshop.

\bibitem[{Luo et~al.(2026{\natexlab{be}})Luo, He, Wu, Pan, Ren, Guo, Yang, Zuo, Kou, Yang, Lai, Ding, Yu, Zhang, Qiu, Li and Ding}]{luo2026handson}
Luo T, He Y, Wu H, Pan T, Ren D, Guo H, Yang J, Zuo Y, Kou Z, Yang K, Lai Z, Ding Y, Yu J, Zhang Y, Qiu Z, Li H and Ding N (2026{\natexlab{be}}) {HandsOn}: A large-scale simulation benchmark for generalizable dexterous manipulation.
\newblock \urlprefix\url{https://arxiv.org/abs/2610.26765}.

\bibitem[{Luo et~al.(2025{\natexlab{bb}})Luo, Yang, Zhang and Zheng}]{xref250109307}
Luo Z, Yang Y, Zhang Y and Zheng F (2025{\natexlab{bb}}) {RoboReflect: A Robotic Reflective Reasoning Framework for Grasping Ambiguous-Condition Objects}.
\newblock \emph{arXiv preprint arXiv:2501.09307} \urlprefix\url{https://arxiv.org/abs/2501.09307}.

\bibitem[{Lykov et~al.(2025)Lykov, Sam, Nguyen, Kozlovskiy, Mahmoud, Serpiva, Cabrera, Konenkov and Tsetserukou}]{lykov2025physicalagent}
Lykov A, Sam J, Nguyen HK, Kozlovskiy V, Mahmoud Y, Serpiva V, Cabrera MA, Konenkov M and Tsetserukou D (2025) {PhysicalAgent: Towards General Cognitive Robotics with Foundation World Models}.
\newblock \doi{10.48550/arXiv.2509.13903}.
\newblock \urlprefix\url{https://arxiv.org/abs/2509.13903}.

\bibitem[{{lykycy123}(2026)}]{robojev2026}
{lykycy123} (2026) {RoboJEV}.
\newblock GitHub repository.
\newblock \urlprefix\url{https://github.com/lykycy123/RoboJEV}.
\newblock Accessed October 5, 2026.

\bibitem[{Lyu et~al.(2026)Lyu, Wu, Li, Hu, Si, Miao, Yang, Wang, Xiao, Hu, Sun and Hao}]{xref260314807}
Lyu K, Wu K, Li P, Hu X, Si Q, Miao C, Yang N, Wang Z, Xiao L, Hu L, Sun J and Hao C (2026) {HiMemVLN: Enhancing Reliability of Open-Source Zero-Shot Vision-and-Language Navigation with Hierarchical Memory System}.
\newblock \emph{arXiv preprint arXiv:2603.14807} \urlprefix\url{https://arxiv.org/abs/2603.14807}.

\bibitem[{Lyu et~al.(2026{\natexlab{bb}})Lyu, Jiang, Li, Lu and Zhang}]{lyu2026codeactionbench}
Lyu Y, Jiang X, Li W, Lu S and Zhang G (2026{\natexlab{bb}}) {CodeActionBench: Evaluating Agentic Code-as-Policy for Embodied Manipulation}.
\newblock \emph{arXiv preprint arXiv:2609.33807} \urlprefix\url{https://arxiv.org/abs/2609.33807}.

\bibitem[{Ma et~al.(2026)Ma, Lu, Yang, Liu, Xu, Byrne, Havoutis, Trigoni and Markham}]{ma2026cyclevla}
Ma C, Lu K, Yang G, Liu J, Xu S, Byrne B, Havoutis I, Trigoni N and Markham A (2026) {CycleVLA: Proactive Self-Correcting Vision-Language-Action Models via Subtask Backtracking and Minimum Bayes Risk Decoding}.
\newblock \emph{arXiv preprint arXiv:2601.02295} \urlprefix\url{https://arxiv.org/abs/2601.02295}.

\bibitem[{Ma et~al.(2026{\natexlab{bb}})Ma, Guo, Tang, Zhou and Rong}]{ma2026atpvpdsg}
Ma G, Guo J, Tang X, Zhou H and Rong Y (2026{\natexlab{bb}}) Adaptive task planning for long-horizon robotic manipulation based on video priors and dynamic scene graphs.
\newblock \emph{Sensors} 26(17): 5595.
\newblock \doi{10.3390/s26175595}.
\newblock \urlprefix\url{https://doi.org/10.3390/s26175595}.

\bibitem[{Ma et~al.(2026{\natexlab{bc}})Ma, Gao, Qiang, Dai and Li}]{ma2026rlebench}
Ma H, Gao C, Qiang R, Dai B and Li N (2026{\natexlab{bc}}) {RLE-Bench: A Qualifying Exam for Coding Agents as Robot Learning Engineers}.
\newblock \urlprefix\url{https://arxiv.org/abs/2609.34210}.

\bibitem[{Ma et~al.(2026{\natexlab{bd}})Ma, Hu, Tang, Hu, Liao, Wu and Liu}]{xref260416331}
Ma X, Hu L, Tang W, Hu Z, Liao Z, Wu Z and Liu Y (2026{\natexlab{bd}}) {BrainMem: Brain-Inspired Evolving Memory for Embodied Agent Task Planning}.
\newblock \emph{arXiv preprint arXiv:2604.16331} \urlprefix\url{https://arxiv.org/abs/2604.16331}.

\bibitem[{Ma et~al.(2023)Ma, Cui, Cao, Ye, Liu, Lu, Abdelraouf, Gupta, Han, Bera, Rehg and Wang}]{ma2023lampilot}
Ma Y, Cui C, Cao X, Ye W, Liu P, Lu J, Abdelraouf A, Gupta R, Han K, Bera A, Rehg JM and Wang Z (2023) {LaMPilot: An Open Benchmark Dataset for Autonomous Driving with Language Model Programs}.
\newblock \doi{10.48550/arXiv.2312.04372}.
\newblock \urlprefix\url{https://arxiv.org/abs/2312.04372}.
\newblock Accepted to CVPR 2024.

\bibitem[{Ma et~al.(2024)Ma, Liang, Wang, Zhu, Fan, Bastani and Jayaraman}]{ma2024dreureka}
Ma Y, Liang W, Wang HJ, Zhu Y, Fan L, Bastani O and Jayaraman D (2024) {DrEureka: Language Model Guided Sim-To-Real Transfer}.
\newblock In: \emph{Robotics: Science and Systems XX}.
\newblock \doi{10.15607/rss.2024.xx.094}.
\newblock \urlprefix\url{https://doi.org/10.15607/rss.2024.xx.094}.

\bibitem[{Ma et~al.(2023{\natexlab{bb}})Ma, Liang, Wang, Huang, Bastani, Jayaraman, Zhu, Fan and Anandkumar}]{ma2023eureka}
Ma YJ, Liang W, Wang G, Huang DA, Bastani O, Jayaraman D, Zhu Y, Fan L and Anandkumar A (2023{\natexlab{bb}}) {Eureka: Human-Level Reward Design via Coding Large Language Models}.
\newblock \emph{arXiv preprint arXiv:2310.12931} \urlprefix\url{https://arxiv.org/abs/2310.12931}.
\newblock Accepted to ICLR 2024.

\bibitem[{{MachEmbodied}(2026)}]{machembodied2026mebrainrsi}
{MachEmbodied} (2026) {ME-Brain-RSI}: A dual-harness framework for self-evolving embodied agents.
\newblock Technical blog.
\newblock \urlprefix\url{https://machembodied.com/ME-Brain/ME-Brain-RSI/index.html}.
\newblock Published October 27, 2026; accessed October 28, 2026.

\bibitem[{{mahajanparth}(2026)}]{smartrobotjev2026}
{mahajanparth} (2026) {JEV Smart Robot Control}.
\newblock \urlprefix\url{https://github.com/mahajanparth/JEV_SMARTROBOTCONTROL}.

\bibitem[{Mahmud et~al.(2026)Mahmud, Dhurubo, Kirttonia, Ujjal and Haque}]{audit260913335}
Mahmud KA, Dhurubo NK, Kirttonia T, Ujjal SH and Haque MA (2026) {Bridging Thought and Action: Taming Long-Horizon Instability in Open-Source LLM Agents with a MetaTool-Enhanced ROS Framework}.
\newblock arXiv:2609.13335.
\newblock \urlprefix\url{https://arxiv.org/abs/2609.13335v1}.

\bibitem[{Mandi et~al.(2024)Mandi, Jain and Song}]{roco}
Mandi Z, Jain S and Song S (2024) {RoCo: Dialectic Multi-Robot Collaboration with Large Language Models}.
\newblock In: \emph{2024 IEEE International Conference on Robotics and Automation (ICRA)}. pp. 286--299.
\newblock \doi{10.1109/icra57147.2024.10610855}.
\newblock \urlprefix\url{https://doi.org/10.1109/icra57147.2024.10610855}.

\bibitem[{{Manifold AI}(2026)}]{manifold2026r2si}
{Manifold AI} (2026) {WorldScape R\textsuperscript{2}SI}.
\newblock Project.
\newblock \urlprefix\url{https://r2si.worldscape.io/}.

\bibitem[{Mao et~al.(2023)Mao, Ye, Qian, Pavone and Wang}]{mao2023agentdriver}
Mao J, Ye J, Qian Y, Pavone M and Wang Y (2023) {A Language Agent for Autonomous Driving}.
\newblock \doi{10.48550/arXiv.2311.10813}.
\newblock \urlprefix\url{https://arxiv.org/abs/2311.10813}.

\bibitem[{Mao et~al.(2026)Mao, Cao, He, Xu, Yan, Jin, Zhang, Tang, Jin, Liu, Shao, Huang, Nan, Han, Xi, Xu, Liang, Zhu, Wu, Wang, Huang, Zou, Zhu, Zhou and Huang}]{mao2026robotagentbench}
Mao L, Cao J, He L, Xu W, Yan Y, Jin H, Zhang Z, Tang J, Jin S, Liu Z, Shao Y, Huang J, Nan F, Han R, Xi Z, Xu Z, Liang S, Zhu S, Wu W, Wang X, Huang Y, Zou X, Zhu G, Zhou K and Huang B (2026) {RobotAgentBench: Towards Open-world Robot Autonomy}.
\newblock \doi{10.48550/arXiv.2610.34649}.
\newblock \urlprefix\url{https://arxiv.org/abs/2610.34649}.

\bibitem[{Mei et~al.(2024)Mei, Zhu, Zhang and Gan}]{xref240721762}
Mei A, Zhu GN, Zhang H and Gan Z (2024) {ReplanVLM: Replanning Robotic Tasks with Visual Language Models}.
\newblock \emph{arXiv preprint arXiv:2407.21762} \urlprefix\url{https://arxiv.org/abs/2407.21762}.

\bibitem[{Mei et~al.(2026)Mei, Xu, Jin, Dai, Cheng, Lv, Fang, Tang, Chen, Zhao, Shen and He}]{sref260902653}
Mei M, Xu H, Jin S, Dai Z, Cheng Q, Lv Z, Fang H, Tang S, Chen G, Zhao X, Shen H and He Z (2026) {HINT: Human-Intent Inception for Long-Horizon Robot Manipulation}.
\newblock \urlprefix\url{https://arxiv.org/abs/2609.02653}.

\bibitem[{Memmel(2026)}]{memmel2026sphere}
Memmel M (2026) {Human sphere-handling video to a Sharpa robot trajectory}.
\newblock Demonstration report.
\newblock \urlprefix\url{https://x.com/memmelma/status/2113601346280804846}.

\bibitem[{Memmel et~al.(2026)Memmel, Yin, Rutia, Wadhwa, Li, Angeloff, Goyal, Zhang, Fox, Ramos, Li and Gupta}]{memmel2026dna}
Memmel M, Yin S, Rutia U, Wadhwa A, Li E, Angeloff G, Goyal A, Zhang J, Fox D, Ramos F, Li A and Gupta A (2026) {D\&A: Decoupling Description and Assessment Enables VLMs as Zero-Shot Robotic Reward Models}.
\newblock arXiv:2611.03622.
\newblock \urlprefix\url{https://arxiv.org/abs/2611.03622}.

\bibitem[{Meng et~al.(2025)Meng, Sun, Fest, Yao, Li, Bing and Knoll}]{meng2025lyra}
Meng Y, Sun Z, Fest M, Yao X, Li X, Bing Z and Knoll A (2025) {Robotic Ultra-Long-Horizon Manipulation Skills via Human-guided Lifelong Code Generation}.
\newblock \emph{arXiv preprint arXiv:2509.18597} \urlprefix\url{https://arxiv.org/abs/2509.18597}.

\bibitem[{Meng et~al.(2025{\natexlab{bb}})Meng, Yao, Ye, Zhou, Zhang, Sun, Li, Bing and Knoll}]{xref250321969}
Meng Y, Yao X, Ye H, Zhou Y, Zhang S, Sun Z, Li X, Bing Z and Knoll A (2025{\natexlab{bb}}) {Robotic Long-Horizon Manipulation with Progressive In-Context Code Generation and Episodic Feedback}.
\newblock \emph{arXiv preprint arXiv:2503.21969} \urlprefix\url{https://arxiv.org/abs/2503.21969}.
\newblock Accepted to IROS 2026.

\bibitem[{Meng et~al.(2026)Meng, Liu, Ma, Li, Liu, Nie, Wei, Chen, Xu, Yuan and Zhang}]{meng2026drivechoreo}
Meng Z, Liu Y, Ma C, Li Z, Liu J, Nie W, Wei B, Chen S, Xu W, Yuan J and Zhang M (2026) {OmniDrive: An LLM-Choreographed Multi-Agent World Model with Unified Latent Co-Compression for Multi-View Driving Video Generation}.
\newblock \doi{10.48550/arXiv.2606.17536}.
\newblock \urlprefix\url{https://arxiv.org/abs/2606.17536}.

\bibitem[{Merler et~al.(2026)Merler, Li, Roy, Liang, Wang, Huang and Silver}]{audit260930233}
Merler M, Li B, Roy J, Liang Y, Wang Q, Huang Y and Silver T (2026) {Coding Agents for Generalized Task and Motion Planning Problems}.
\newblock arXiv:2609.30233.
\newblock \urlprefix\url{https://arxiv.org/abs/2609.30233v1}.

\bibitem[{{metrox-eth}(2026)}]{mossjev2026}
{metrox-eth} (2026) {MOSS $\times$ Jev}.
\newblock \urlprefix\url{https://github.com/metrox-eth/moss-jev}.

\bibitem[{Mi et~al.(2026)Mi, Ma, Yao, Gao, Chen, Li, Zhu, Li, Yin, Wang, Mu, Pang and Wang}]{real2026}
Mi B, Ma M, Yao Y, Gao X, Chen J, Li Y, Zhu Z, Li G, Yin Z, Wang T, Mu Y, Pang J and Wang H (2026) {Exploratory, Communicative, and Deployable: Vision-Driven Embodied Agents for Open-World Mobile Manipulation}.
\newblock In: \emph{Computer Vision – ECCV 2026}, Lecture Notes in Computer Science. pp. 627--645.
\newblock \doi{10.1007/978-3-032-37422-6_36}.
\newblock \urlprefix\url{https://doi.org/10.1007/978-3-032-37422-6_36}.

\bibitem[{Miceli-Barone et~al.(2023)Miceli-Barone, Lascarides and Innes}]{micelibarone2023dialoguebasedscenarios}
Miceli-Barone AV, Lascarides A and Innes C (2023) {Dialogue-based generation of self-driving simulation scenarios using Large Language Models}.
\newblock \doi{10.48550/arXiv.2310.17372}.
\newblock \urlprefix\url{https://arxiv.org/abs/2310.17372}.

\bibitem[{Mishra et~al.(2026)Mishra, Degirmenci, Chen, Xu, Hermans and Garrett}]{mishra2026tampagent}
Mishra UA, Degirmenci A, Chen Y, Xu D, Hermans T and Garrett C (2026) {TAMPAgent: Agentic Belief-Space Planning for Mobile Manipulation}.
\newblock arXiv:2611.03584.
\newblock \urlprefix\url{https://arxiv.org/abs/2611.03584}.

\bibitem[{Mon-Williams et~al.(2025)Mon-Williams, Li, Long, Du and Lucas}]{monwilliams2025ellmer}
Mon-Williams R, Li G, Long R, Du W and Lucas CG (2025) {Embodied large language models enable robots to complete complex tasks in unpredictable environments}.
\newblock \emph{Nature Machine Intelligence} 7(4): 592--601.
\newblock \doi{10.1038/s42256-025-01005-x}.
\newblock \urlprefix\url{https://doi.org/10.1038/s42256-025-01005-x}.

\bibitem[{Mower et~al.(2026)Mower, Wan, Yu, Grosnit, Gonzalez-Billandon, Zimmer, Wang, Zhang, Zhao, Zhai, Liu, Palenicek, Tateo, Cadena, Hutter, Peters, Tian, Zhuang, Shao, Quan, Hao, Wang and Bou-Ammar}]{grid42}
Mower CE, Wan Y, Yu H, Grosnit A, Gonzalez-Billandon J, Zimmer M, Wang J, Zhang X, Zhao Y, Zhai A, Liu P, Palenicek D, Tateo D, Cadena C, Hutter M, Peters J, Tian G, Zhuang Y, Shao K, Quan X, Hao J, Wang J and Bou-Ammar H (2026) {A robot operating system framework for using large language models in embodied AI}.
\newblock \emph{Nature Machine Intelligence} 8(3): 313--325.
\newblock \doi{10.1038/s42256-026-01186-z}.
\newblock \urlprefix\url{https://doi.org/10.1038/s42256-026-01186-z}.

\bibitem[{Mu et~al.(2024)Mu, Chen, Zhang, Chen, Yu, Ge, Chen, Liang, Hu, Tao, Sun, Yu, Yang, Shao, Wang, Dai, Qiao, Ding and Luo}]{mu2024robocodex}
Mu Y, Chen J, Zhang QL, Chen S, Yu Q, Ge C, Chen R, Liang Z, Hu M, Tao C, Sun P, Yu H, Yang C, Shao W, Wang W, Dai J, Qiao Y, Ding M and Luo P (2024) {RoboCodeX: Multimodal Code Generation for Robotic Behavior Synthesis}.
\newblock In: \emph{Proceedings of the 41st International Conference on Machine Learning}, \emph{Proceedings of Machine Learning Research}, volume 235. pp. 36434--36454.
\newblock \urlprefix\url{https://proceedings.mlr.press/v235/mu24a.html}.

\bibitem[{Mu et~al.(2025)Mu, Chen, Chen, Peng, Lan, Gao, Liang, Yu, Zou, Xu, Lin, Xie, Ding and Luo}]{mu2025robotwin}
Mu Y, Chen T, Chen Z, Peng S, Lan Z, Gao Z, Liang Z, Yu Q, Zou Y, Xu M, Lin L, Xie Z, Ding M and Luo P (2025) {RoboTwin}: Dual-arm robot benchmark with generative digital twins.
\newblock \emph{arXiv preprint arXiv:2504.13059} \urlprefix\url{https://arxiv.org/abs/2504.13059}.
\newblock Accepted to CVPR 2025.

\bibitem[{Munn et~al.(2026)Munn, Tidd, Bohm, Gallagher and Howard}]{audit260201515}
Munn H, Tidd B, Bohm P, Gallagher M and Howard D (2026) {RAPT: Model-Predictive Out-of-Distribution Detection and Failure Diagnosis for Sim-to-Real Humanoid Deployment}.
\newblock \emph{arXiv preprint arXiv:2602.01515} \doi{10.48550/arXiv.2602.01515}.
\newblock \urlprefix\url{https://arxiv.org/abs/2602.01515}.

\bibitem[{Nam et~al.(2026)Nam, Cho, Kim and Choi}]{nam2026ontoplan}
Nam H, Cho W, Kim J and Choi J (2026) {OntoPlan: An Ontology-Grounded Scene Representation and Agentic Framework for Scalable Robot Task Planning}.
\newblock \doi{10.48550/arXiv.2610.07649}.
\newblock \urlprefix\url{https://arxiv.org/abs/2610.07649}.

\bibitem[{Naouali et~al.(2026)Naouali, Wu, Wong, Puthran and Younis}]{xref260816978}
Naouali D, Wu M, Wong C, Puthran A and Younis OG (2026) {VLCP: Vision Language Control Policy Closed-Loop Code Replanning for Robot Manipulation}.
\newblock \emph{arXiv preprint arXiv:2608.16978} \urlprefix\url{https://arxiv.org/abs/2608.16978}.

\bibitem[{Nasiriany et~al.(2024)Nasiriany, Xia, Yu, Xiao, Liang, Dasgupta, Xie, Driess, Wahid, Xu, Vuong, Zhang, Lee, Lee, Xu, Kirmani, Zhu, Zeng, Hausman, Heess, Finn, Levine and Ichter}]{nasiriany2024pivot}
Nasiriany S, Xia F, Yu W, Xiao T, Liang J, Dasgupta I, Xie A, Driess D, Wahid A, Xu Z, Vuong Q, Zhang T, Lee TWE, Lee KH, Xu P, Kirmani S, Zhu Y, Zeng A, Hausman K, Heess N, Finn C, Levine S and Ichter B (2024) {PIVOT}: Iterative visual prompting elicits actionable knowledge for {VLM}s.
\newblock In: \emph{Proceedings of the 41st International Conference on Machine Learning}, \emph{Proceedings of Machine Learning Research}, volume 235. PMLR, pp. 37321--37341.
\newblock \urlprefix\url{https://proceedings.mlr.press/v235/nasiriany24a.html}.

\bibitem[{Navaratnarajah et~al.(2026)Navaratnarajah, Kim, Ruthardt, Bhimwal, Yamada, Blei, Burgard and Asano}]{audit260722014}
Navaratnarajah S, Kim T, Ruthardt J, Bhimwal I, Yamada R, Blei Y, Burgard W and Asano YM (2026) {Zero-Shot Mission-Level Evaluation for Aerial MLLM Agents}.
\newblock arXiv:2607.22014.
\newblock \urlprefix\url{https://arxiv.org/abs/2607.22014v1}.

\bibitem[{Nayak et~al.(2026)Nayak, Parameswaran, Peri and Ramanan}]{nayak2026mimicagent}
Nayak LK, Parameswaran NP, Peri N and Ramanan D (2026) {MimicAgent: Quadruped Skills via Prompt-to-Trajectory Generation}.
\newblock arXiv:2609.24145.
\newblock \urlprefix\url{https://arxiv.org/abs/2609.24145}.

\bibitem[{Nayak et~al.(2024)Nayak, Orozco, Have, Thirumalai, Zhang, Chen, Kapoor, Robinson, Gopalakrishnan, Harrison, Ichter, Mahajan and Balakrishnan}]{grid40}
Nayak S, Orozco AM, Have MT, Thirumalai V, Zhang J, Chen D, Kapoor A, Robinson E, Gopalakrishnan K, Harrison J, Ichter B, Mahajan A and Balakrishnan H (2024) {LLaMAR: Long-Horizon Planning for Multi-Agent Robots in Partially Observable Environments}.
\newblock arXiv:2407.10031.
\newblock \urlprefix\url{https://arxiv.org/abs/2407.10031}.

\bibitem[{Nazzari et~al.(2026)Nazzari, Cerisara, Tonnis, Zakir, Labarile, Zhu, Dorigo and Heinrich}]{audit260916346}
Nazzari A, Cerisara N, Tonnis D, Zakir R, Labarile L, Zhu W, Dorigo M and Heinrich MK (2026) {Auto-HSI: Personalized human control of a robot swarm on demand by using LLMs for online automatic code generation}.
\newblock arXiv:2609.16346.
\newblock \urlprefix\url{https://arxiv.org/abs/2609.16346v1}.

\bibitem[{{Nebius}(2026)}]{nebiusphysicalai2026}
{Nebius} (2026) {Nebius Physical AI}.
\newblock Software repository.
\newblock \urlprefix\url{https://github.com/nebius/nebius-physical-ai}.
\newblock Version 823fcf5.

\bibitem[{Ng et~al.(2026)Ng, Zhang, Wu, Lui, Zhao, Gao and Ren}]{ng2026endorsi}
Ng CK, Zhang Y, Wu T, Lui SH, Zhao H, Gao H and Ren H (2026) {EndoRSI: Physical-Evidence-Driven Recursive Improvement of Endoscopic Robot Policies}.
\newblock \emph{arXiv preprint arXiv:2610.13376} \urlprefix\url{https://arxiv.org/abs/2610.13376}.

\bibitem[{Nguyen et~al.(2026)Nguyen, Raha, Le, Huh, Niyato and Hong}]{nguyen2026selfevolvinghumanoids}
Nguyen LX, Raha AD, Le HQ, Huh EN, Niyato D and Hong CS (2026) {Self-Evolving AI for Humanoids: Mechanisms, Safety, and Evaluation of Post-Deployment Self-Improvement}.
\newblock \emph{arXiv preprint arXiv:2609.13236} \urlprefix\url{https://arxiv.org/abs/2609.13236}.

\bibitem[{Nguyen et~al.(2025)Nguyen, Wang, Hong, Aasi, Silva, Rosman, Karaman and Rus}]{nguyen2025regengenerativerobotsimulation}
Nguyen P, Wang TH, Hong ZW, Aasi E, Silva A, Rosman G, Karaman S and Rus D (2025) {ReGen: Generative Robot Simulation via Inverse Design}.
\newblock \urlprefix\url{https://arxiv.org/abs/2511.04769}.

\bibitem[{Nguyen et~al.(2024)Nguyen, Wang, Hong, Karaman and Rus}]{nguyen2024texttodrive}
Nguyen P, Wang TH, Hong ZW, Karaman S and Rus D (2024) {Text-to-Drive: Diverse Driving Behavior Synthesis via Large Language Models}.
\newblock \doi{10.48550/arXiv.2406.04300}.
\newblock \urlprefix\url{https://arxiv.org/abs/2406.04300}.

\bibitem[{Ni et~al.(2026)Ni, Lu and Cai}]{core2026}
Ni H, Lu T and Cai Y (2026) {C}o{R}e: Combined rewards with vision-language model feedback for preference-aligned reinforcement learning.
\newblock In: \emph{Proceedings of the 43rd International Conference on Machine Learning}, \emph{Proceedings of Machine Learning Research}, volume 306. PMLR, pp. 92975--92999.
\newblock \urlprefix\url{https://proceedings.mlr.press/v306/ni26d.html}.

\bibitem[{Nie et~al.(2026)Nie, Liu and Wang}]{nie2026teachandgrow}
Nie C, Liu Z and Wang H (2026) {Teach and Grow: An Agent-Centered Architecture for General Robot Learning}.
\newblock \emph{arXiv preprint arXiv:2608.17209} \urlprefix\url{https://arxiv.org/abs/2608.17209}.

\bibitem[{Nie et~al.(2025)Nie, Guo, Duan, Zhang and Chen}]{xref250302247}
Nie D, Guo X, Duan Y, Zhang R and Chen L (2025) {WMNav: Integrating Vision-Language Models into World Models for Object Goal Navigation}.
\newblock In: \emph{IEEE/RSJ International Conference on Intelligent Robots and Systems}. pp. 2392--2399.
\newblock \doi{10.1109/iros60139.2025.11246684}.
\newblock \urlprefix\url{https://doi.org/10.1109/iros60139.2025.11246684}.

\bibitem[{{NVIDIA}(2026)}]{nvidia2026agentskills}
{NVIDIA} (2026) {Agent Skills for NVIDIA Products}.
\newblock Software repository.
\newblock \urlprefix\url{https://github.com/NVIDIA/skills}.

\bibitem[{{NVIDIA}(2026{\natexlab{bb}})}]{isaacsim}
{NVIDIA} (2026{\natexlab{bb}}) {Isaac Sim}.
\newblock NVIDIA Developer Documentation.
\newblock \urlprefix\url{https://developer.nvidia.com/isaac-sim}.
\newblock Accessed: 2026-10-25.

\bibitem[{{NVIDIA}(2026{\natexlab{bc}})}]{nvidia2026v2dingestion}
{NVIDIA} (2026{\natexlab{bc}}) {Video to Data: Video Ingestion Agent}.
\newblock NVIDIA Isaac project documentation.
\newblock \urlprefix\url{https://nvidia-isaac.github.io/video_to_data/}.

\bibitem[{{NVIDIA et al.}(2026)}]{nvidia2026gr00t2}
{NVIDIA et al} (2026) {GR00T 2: Scaling Robot Foundation Models for Generalization, Dexterity, and Adaptation}.
\newblock Technical report, NVIDIA.
\newblock \urlprefix\url{https://research.nvidia.com/labs/gear/gr00t-2/assets/gr00t2_whitepaper.pdf}.

\bibitem[{{NVIDIA Research}(2026)}]{nvlabs2026aspirecode}
{NVIDIA Research} (2026) {ASPIRE}: Public robot skill discovery implementation.
\newblock Source code and configuration documentation.
\newblock \urlprefix\url{https://github.com/NVlabs/ASPIRE/tree/fa00f0ef2bc1d91d594b9aa502c22133c9f538f7}.
\newblock Revision fa00f0e, accessed October 19, 2026.

\bibitem[{Oikawa et~al.(2026)Oikawa, Endo, Uzawa, Zhang, Anjaria, Pinto, Yang and Tsuda}]{oikawa2026wetrobo}
Oikawa Y, Endo K, Uzawa T, Zhang Y, Anjaria M, Pinto L, Yang S and Tsuda K (2026) {WetRobo: A Reproducible Robot Kit for Coding Agents in Biological Laboratories}.
\newblock \emph{arXiv preprint arXiv:2609.18435} \urlprefix\url{https://arxiv.org/abs/2609.18435}.

\bibitem[{Oloo et~al.(2026)Oloo, Altaweel, Hayamizu, Liu, Ding, Amiri, Yang, Kaminski, Esselink, Paxton, Zhang and Zhang}]{audit260426988}
Oloo A, Altaweel Z, Hayamizu Y, Liu P, Ding Y, Amiri S, Yang H, Kaminski A, Esselink C, Paxton C, Zhang X and Zhang S (2026) {Robot Planning and Situation Handling with Active Perception}.
\newblock arXiv:2604.26988.
\newblock \urlprefix\url{https://arxiv.org/abs/2604.26988v1}.

\bibitem[{{omnilink-tech}(n.d.)}]{omnisim2026}
{omnilink-tech} (n.d.) {OmniSim}.
\newblock Software repository.
\newblock \urlprefix\url{https://github.com/omnilink-tech/omnisim}.
\newblock Accessed: 2026-10-26.

\bibitem[{{OpenRoboKit}(2026)}]{openrobokit2026roboskills}
{OpenRoboKit} (2026) {RoboSkills}: A minimal set of agent workflows and tools for robotics.
\newblock Software repository.
\newblock \urlprefix\url{https://github.com/openrobokit/roboskills}.

\bibitem[{{openroboto-ai}(2026)}]{openroboto2026}
{openroboto-ai} (2026) {jev-robot-control}.
\newblock GitHub repository.
\newblock \urlprefix\url{https://github.com/openroboto-ai/jev-robot-control}.
\newblock Accessed October 5, 2026.

\bibitem[{{OpenRUA Team}(2026)}]{openrua2026revisiting}
{OpenRUA Team} (2026) {Behind the RoboDojo Boom: Revisiting Embodied Evaluation of Robot-Use Agents}.
\newblock \urlprefix\url{https://openrua.ai/revisiting-rua}.

\bibitem[{Ord(2026)}]{ord2026dynamics}
Ord T (2026) {The Dynamics of Intelligence Explosions}.
\newblock \emph{arXiv preprint arXiv:2608.14426} \urlprefix\url{https://arxiv.org/abs/2608.14426}.

\bibitem[{Ouyang et~al.(2024)Ouyang, Li, Li, Li, Yu, Sreenath and Wu}]{ouyang2024ouyangetal}
Ouyang Y, Li J, Li Y, Li Z, Yu C, Sreenath K and Wu Y (2024) {Long-horizon Locomotion and Manipulation on a Quadrupedal Robot with Large Language Models}.
\newblock \doi{10.48550/arXiv.2404.05291}.
\newblock \urlprefix\url{https://arxiv.org/abs/2404.05291}.

\bibitem[{Pan et~al.(2025)Pan, Zhang, Wu, Zhao, Gao and Dong}]{omnimanip2025}
Pan M, Zhang J, Wu T, Zhao Y, Gao W and Dong H (2025) {OmniManip: Towards General Robotic Manipulation via Object-Centric Interaction Primitives as Spatial Constraints}.
\newblock arXiv:2501.03841.
\newblock \urlprefix\url{https://arxiv.org/abs/2501.03841}.

\bibitem[{Pang et~al.(2024)Pang, Wang and Li}]{pang2024lgdrl}
Pang H, Wang Z and Li G (2024) {Large Language Model guided Deep Reinforcement Learning for Decision Making in Autonomous Driving}.
\newblock \doi{10.48550/arXiv.2412.18511}.
\newblock \urlprefix\url{https://arxiv.org/abs/2412.18511}.

\bibitem[{Parakh et~al.(2023)Parakh, Fong, Simeonov, Chen, Gupta and Agrawal}]{parakh2023halp}
Parakh M, Fong A, Simeonov A, Chen T, Gupta A and Agrawal P (2023) {Lifelong Robot Learning with Human Assisted Language Planners}.
\newblock \emph{arXiv preprint arXiv:2309.14321} \urlprefix\url{https://arxiv.org/abs/2309.14321}.

\bibitem[{Patel et~al.(2025)Patel, Yin, Huang, Garg, Nayyeri, Li, Lazebnik and Li}]{iker2025}
Patel S, Yin X, Huang W, Garg S, Nayyeri H, Li FF, Lazebnik S and Li Y (2025) {A Real-to-Sim-to-Real Approach to Robotic Manipulation with VLM-Generated Iterative Keypoint Rewards}.
\newblock arXiv:2502.08643.
\newblock \urlprefix\url{https://arxiv.org/abs/2502.08643}.
\newblock Accepted to ICRA 2025.

\bibitem[{Peng et~al.(2024)Peng, Sucholutsky, Li, Sumers, Griffiths, Andreas and Shah}]{peng2024lga}
Peng A, Sucholutsky I, Li BZ, Sumers TR, Griffiths TL, Andreas J and Shah JA (2024) {Learning with Language-Guided State Abstractions}.
\newblock In: \emph{International Conference on Learning Representations}. pp. 38711--38744.
\newblock \urlprefix\url{https://arxiv.org/abs/2402.18759}.

\bibitem[{Peng et~al.(2026)Peng, Yu, Feng, Yang, Cai, Xiong, Yang, Zheng, Cao, Wei, Pang, Shen and Wang}]{sref260705377}
Peng J, Yu X, Feng D, Yang Y, Cai W, Xiong J, Yang G, Zheng J, Cao J, Wei X, Pang J, Shen Y and Wang T (2026) {Cortex: A Bidirectionally Aligned Embodied Agent Framework for Long-horizon Manipulation}.
\newblock \urlprefix\url{https://arxiv.org/abs/2607.05377}.

\bibitem[{Peng et~al.(2026{\natexlab{bb}})Peng, He, Li, Ma and Zhou}]{peng2026slowbrain}
Peng Z, He H, Li Q, Ma Y and Zhou B (2026{\natexlab{bb}}) {Slow Brain, Fast Planner: Latency-Resilient VLM-Augmented Urban Navigation}.
\newblock \emph{arXiv preprint arXiv:2606.20458} \urlprefix\url{https://arxiv.org/abs/2606.20458}.

\bibitem[{Peng et~al.(2025)Peng, Wang, Han, Zheng and Ma}]{peng2025learningflow}
Peng Z, Wang Y, Han X, Zheng L and Ma J (2025) {LearningFlow: Automated Policy Learning Workflow for Urban Driving with Large Language Models}.
\newblock \doi{10.48550/arXiv.2501.05057}.
\newblock \urlprefix\url{https://arxiv.org/abs/2501.05057}.

\bibitem[{Perez et~al.(2023)Perez, Proux, Roux and Niemaz}]{perez2023larg2}
Perez J, Proux D, Roux C and Niemaz M (2023) {LARG, Language-based Automatic Reward and Goal Generation}.
\newblock \urlprefix\url{https://arxiv.org/abs/2306.10985}.

\bibitem[{Periasami et~al.(2026)Periasami, Luo, Li and Cheng}]{periasami2026frames}
Periasami AV, Luo X, Li H and Cheng X (2026) {FRAMES: Failure Recovery And Monitoring of Embodied Skills for Humanoid Loco-Manipulation}.
\newblock \doi{10.48550/arXiv.2609.22538}.
\newblock \urlprefix\url{https://arxiv.org/abs/2609.22538}.

\bibitem[{Petkov et~al.(2026)Petkov, Albanese, Nikolov, Van~Gool, Tremblay and Paudel}]{petkov2026ladder}
Petkov P, Albanese G, Nikolov N, Van~Gool L, Tremblay J and Paudel DP (2026) {LADDER}: A long-horizon robot manipulation benchmark.
\newblock arXiv:2610.18053.
\newblock \urlprefix\url{https://arxiv.org/abs/2610.18053}.

\bibitem[{Pfaff et~al.(2026)Pfaff, Cohn, Zakharov, Cory and Tedrake}]{scenesmith2026}
Pfaff N, Cohn T, Zakharov S, Cory R and Tedrake R (2026) {S}cene{S}mith: Agentic generation of simulation-ready indoor scenes.
\newblock In: \emph{Proceedings of the 43rd International Conference on Machine Learning}, \emph{Proceedings of Machine Learning Research}, volume 306. PMLR, pp. 98499--98567.
\newblock \urlprefix\url{https://proceedings.mlr.press/v306/pfaff26a.html}.

\bibitem[{{PhyAgentOS Team}(2026)}]{phyagentos2026core}
{PhyAgentOS Team} (2026) {PhyAgentOS-core: Public Runtime and Robot Configuration Documentation}.
\newblock Software repository.
\newblock \urlprefix\url{https://github.com/PhyAgentOS/PhyAgentOS-core/tree/74546aba45733cf102211e2f13dbf2c80ff6df8b}.
\newblock Revision 74546ab; accessed 2026-10-19. Concrete robot Skill runtimes and assets are separate dependencies.

\bibitem[{PhysicalRSI()}]{physicalrsi2026}
PhysicalRSI (2026) {PhysicalRSI 1.0: Recursive Self-Harness for Scaling Embodied Skills}.
\newblock Project website.
\newblock \urlprefix\url{https://mmlab.hk/research/PhysicalRSI}.
\newblock Accessed September 30, 2026.

\bibitem[{Pikus et~al.(2026)Pikus, Saunders, Mironenco, Mohammadi, Mathew, Zhu, Makkar, Huang, Li, Narasimhaswamy, Cen, Xu, Buch and Neff}]{pikus2026pretrainingknowswhatsimulation}
Pikus B, Saunders D, Mironenco M, Mohammadi S, Mathew M, Zhu S, Makkar D, Huang X, Li Q, Narasimhaswamy S, Cen S, Xu D, Buch S and Neff T (2026) {Pretraining Knows What, Simulation Teaches How: Compositional Generalization for Agentic Robotics}.
\newblock \urlprefix\url{https://arxiv.org/abs/2610.35817}.

\bibitem[{Pomichter et~al.(2026)Pomichter, Dasyam and Ventura}]{canjevnav2026}
Pomichter S, Dasyam R and Ventura H (2026) {Can Jev Nav?}
\newblock \urlprefix\url{https://research.dimensional.org/system-one-navigation/}.

\bibitem[{Puigjaner and Alexis(2026)}]{audit260926360}
Puigjaner AG and Alexis K (2026) {Hierarchical Floorplan-Guided Vision-Language Exploration for Embodied Question Answering}.
\newblock arXiv:2609.26360.
\newblock \urlprefix\url{https://arxiv.org/abs/2609.26360v1}.

\bibitem[{Qi et~al.(2025)Qi, Zhang, Ding, Dong, Yu, Li, Xu, Li, He, Fan, Zhang, He, Gu, Jin, Ma, Zhang, Wang and Yi}]{audit250213143}
Qi Z, Zhang W, Ding Y, Dong R, Yu X, Li J, Xu L, Li B, He X, Fan G, Zhang J, He J, Gu J, Jin X, Ma K, Zhang Z, Wang H and Yi L (2025) {SoFar: Language-Grounded Orientation Bridges Spatial Reasoning and Object Manipulation}.
\newblock arXiv:2502.13143.
\newblock \urlprefix\url{https://arxiv.org/abs/2502.13143v1}.
\newblock Accepted to NeurIPS 2025.

\bibitem[{Qiao et~al.(2024)Qiao, Lyu, Wang, Wang, Li, Zhang, Tan and Wu}]{xref240918794}
Qiao Y, Lyu W, Wang H, Wang Z, Li Z, Zhang Y, Tan M and Wu Q (2024) {Open-Nav: Exploring Zero-Shot Vision-and-Language Navigation in Continuous Environment with Open-Source LLMs}.
\newblock \emph{arXiv preprint arXiv:2409.18794} \urlprefix\url{https://arxiv.org/abs/2409.18794}.
\newblock Accepted to ICRA 2025.

\bibitem[{Qin et~al.(2026)Qin, Wu, Feng, Wang, Yu, Sun, Fan, He, Chen, Lian, Ji, Ren, Hu, Lin, Jiang and Yue}]{qin2026towardsagenticrobots}
Qin C, Wu D, Feng K, Wang J, Yu X, Sun P, Fan Y, He X, Chen B, Lian W, Ji R, Ren W, Hu X, Lin D, Jiang YG and Yue X (2026) {Towards Agentic Robots: Model, Data, Environment, and Harness}.
\newblock \urlprefix\url{https://arxiv.org/abs/2610.20442}.

\bibitem[{Qin et~al.(2026{\natexlab{bb}})Qin, Deng, Liang, Song, Lin, Jiang, Fu, Sun, Lei, Gong, Chen, Zhao, Bai, Ding, Wang and Li}]{qin2026embodiedsmithscalingembodieddata}
Qin Y, Deng Y, Liang M, Song W, Lin Z, Jiang Z, Fu J, Sun Q, Lei H, Gong X, Chen J, Zhao H, Bai S, Ding P, Wang P and Li H (2026{\natexlab{bb}}) Embodiedsmith: Scaling embodied data through recursive self-improvement flywheel in simulation.
\newblock \urlprefix\url{https://arxiv.org/abs/2610.07969}.

\bibitem[{Qin et~al.(2025)Qin, Kang, Song, Yin, Liu, Liu, Zhang and Bai}]{audit250316408}
Qin Y, Kang L, Song X, Yin Z, Liu X, Liu X, Zhang R and Bai L (2025) {RoboFactory: Exploring Embodied Agent Collaboration with Compositional Constraints}.
\newblock arXiv:2503.16408.
\newblock \urlprefix\url{https://arxiv.org/abs/2503.16408v1}.

\bibitem[{Qiu and Cygan(2026)}]{qiu2026debate2create}
Qiu K and Cygan M (2026) {Debate2Create}: Robot co-design via multi-agent {LLM} debate.
\newblock In: \emph{Proceedings of the 43rd International Conference on Machine Learning}, \emph{Proceedings of Machine Learning Research}, volume 306. pp. 101918--101945.
\newblock \urlprefix\url{https://proceedings.mlr.press/v306/qiu26c.html}.

\bibitem[{Qiu et~al.(2024)Qiu, Pa{\l}ucki, Ciebiera, Fija{\l}kowski, Cygan and Kuci{\'n}ski}]{si240708626}
Qiu K, Pa{\l}ucki W, Ciebiera K, Fija{\l}kowski P, Cygan M and Kuci{\'n}ski {\L} (2024) {RoboMorph: Evolving Robot Morphology using Large Language Models}.
\newblock \urlprefix\url{https://arxiv.org/abs/2407.08626}.

\bibitem[{Qiu et~al.(2025)Qiu, Yang, Wang, Chen, Wang, Wang, Xian and Gan}]{qiu2025articulateanymeshopenvocabulary3d}
Qiu X, Yang J, Wang Y, Chen Z, Wang Y, Wang TH, Xian Z and Gan C (2025) Articulate anymesh: Open-vocabulary 3d articulated objects modeling.
\newblock \urlprefix\url{https://arxiv.org/abs/2502.02590}.

\bibitem[{{Qoo Tech}(2026)}]{qoo2026q0preview}
{Qoo Tech} (2026) {Q0-Preview}.
\newblock Benchmark submission.
\newblock \urlprefix\url{https://robodojo-benchmark.com/leaderboard/rollouts/Q0_Preview?bench=sim}.

\bibitem[{Qu et~al.(2026)Qu, Chen and Zhang}]{bincequ2026roboharness}
Qu B, Chen W and Zhang B (2026) {RoboHarness: A Simple Harness Outperforms VLA and World Action Models}.
\newblock \emph{arXiv preprint arXiv:2610.14018} \doi{10.48550/arXiv.2610.14018}.
\newblock \urlprefix\url{https://arxiv.org/abs/2610.14018}.

\bibitem[{Qu et~al.(2025)Qu, Lan, Zurbr{\"u}gg, Chen, Mower, Bou-Ammar and Hutter}]{qu2025pragmabot}
Qu K, Lan G, Zurbr{\"u}gg R, Chen C, Mower CE, Bou-Ammar H and Hutter M (2025) {A Pragmatist Robot: Learning to Plan Tasks by Experiencing the Real World}.
\newblock \emph{arXiv preprint arXiv:2507.16713} \urlprefix\url{https://arxiv.org/abs/2507.16713}.
\newblock Accepted to IEEE Robotics and Automation Letters.

\bibitem[{Rachwa{\l} et~al.(2025)Rachwa{\l}, Majek, Boczek, D{\k{a}}browski, Liberadzki, D{\k{a}}browski and Ganzha}]{rai}
Rachwa{\l} K, Majek M, Boczek B, D{\k{a}}browski K, Liberadzki P, D{\k{a}}browski A and Ganzha M (2025) {RAI: Flexible Agent Framework for Embodied AI}.
\newblock In: \emph{Highlights in Practical Applications of Agents, Multi-Agent Systems and Computational Social Science. The PAAMS Collection}, Communications in Computer and Information Science. pp. 195--206.
\newblock \doi{10.1007/978-3-032-05925-3_16}.
\newblock \urlprefix\url{https://doi.org/10.1007/978-3-032-05925-3_16}.

\bibitem[{{raghavpillai}(2026)}]{px4jevroom2026}
{raghavpillai} (2026) {PX4 Jev Room Search}.
\newblock \urlprefix\url{https://github.com/raghavpillai/jev-drone}.

\bibitem[{Rajvanshi et~al.(2023)Rajvanshi, Sikka, Lin, Lee, Chiu and Velasquez}]{xref230904077}
Rajvanshi A, Sikka K, Lin X, Lee B, Chiu HP and Velasquez A (2023) {SayNav: Grounding Large Language Models for Dynamic Planning to Navigation in New Environments}.
\newblock \emph{arXiv preprint arXiv:2309.04077} \urlprefix\url{https://arxiv.org/abs/2309.04077}.

\bibitem[{Ramabadran et~al.(2026)Ramabadran, Mahns and Gessler}]{audit260938948}
Ramabadran A, Mahns S and Gessler T (2026) {DrivingBench: Can Vision-Language Models Drive a Toyota Corolla?}
\newblock arXiv:2609.38948.
\newblock \urlprefix\url{https://arxiv.org/abs/2609.38948v1}.

\bibitem[{Ramrakhya et~al.(2025)Ramrakhya, Chang, Puig, Desai, Kira and Mottaghi}]{ramrakhya2025autoask}
Ramrakhya R, Chang M, Puig X, Desai R, Kira Z and Mottaghi R (2025) {Grounding Multimodal LLMs to Embodied Agents that Ask for Help with Reinforcement Learning}.
\newblock \urlprefix\url{https://arxiv.org/abs/2504.00907}.

\bibitem[{Rana et~al.(2023)Rana, Haviland, Garg, Abou-Chakra, Reid and Suenderhauf}]{sayplan2023}
Rana K, Haviland J, Garg S, Abou-Chakra J, Reid I and Suenderhauf N (2023) Sayplan: Grounding large language models using 3d scene graphs for scalable robot task planning.
\newblock In: \emph{Proceedings of The 7th Conference on Robot Learning}, \emph{Proceedings of Machine Learning Research}, volume 229. PMLR, pp. 23--72.
\newblock \urlprefix\url{https://proceedings.mlr.press/v229/rana23a.html}.

\bibitem[{Ranawaka et~al.(2026)Ranawaka, Wong, Pai, Chu, Dai, Moghani, Yin, Jiang, Durbano, Huynh, Fang, Xu, Zhang, Li, Fan, Wen, Mandlekar and Zhu}]{simfoundry2026}
Ranawaka N, Wong J, Pai WL, Chu WT, Dai T, Moghani M, Yin H, Jiang Y, Durbano W, Huynh B, Fang Y, Xu D, Zhang R, Li FF, Fan L, Wen B, Mandlekar A and Zhu Y (2026) {SimFoundry: Modular and Automated Scene Generation for Policy Learning and Evaluation}.
\newblock arXiv:2606.28276.
\newblock \urlprefix\url{https://arxiv.org/abs/2606.28276}.

\bibitem[{Rank et~al.(2026)Rank, Bhatnagar, Prabhu, Eisenberg, Nguyen, Bethge and Andriushchenko}]{rank2026posttrainbench}
Rank B, Bhatnagar H, Prabhu A, Eisenberg S, Nguyen K, Bethge M and Andriushchenko M (2026) {P}ost{T}rain{B}ench: Can {LLM} agents automate {LLM} post-training?
\newblock In: \emph{Proceedings of the 43rd International Conference on Machine Learning}, \emph{Proceedings of Machine Learning Research}, volume 306. PMLR, pp. 103808--103823.
\newblock \urlprefix\url{https://proceedings.mlr.press/v306/rank26a.html}.

\bibitem[{Rao and Advani(2026)}]{si260629871}
Rao A and Advani NK (2026) {AI Training Manager: Bounded Closed-Loop Control of Adaptive Training Recipes}.
\newblock \urlprefix\url{https://arxiv.org/abs/2606.29871}.

\bibitem[{Rao(2026)}]{rao2026argon}
Rao K (2026) {Gemini 4 Argon for Robotics}.
\newblock Google Developers Blog.
\newblock \urlprefix\url{https://developers.googleblog.com/gemini-4-argon-for-robotics/}.

\bibitem[{Ravichandran et~al.(2026)Ravichandran, Diller, Cladera, Murali, Pappas and Kumar}]{audit260930428}
Ravichandran Z, Diller J, Cladera F, Murali V, Pappas GJ and Kumar V (2026) {Actively Resolving Contextual Uncertainty for Underspecified Tasks in Natural Language}.
\newblock arXiv:2609.30428.
\newblock \urlprefix\url{https://arxiv.org/abs/2609.30428v1}.
\newblock Accepted to ISRR 2026.

\bibitem[{Ravichandran et~al.(2025)Ravichandran, Murali, Tzes, Pappas and Kumar}]{grid37}
Ravichandran Z, Murali V, Tzes M, Pappas GJ and Kumar V (2025) {SPINE: Online Semantic Planning for Missions with Incomplete Natural Language Specifications in Unstructured Environments}.
\newblock In: \emph{2025 IEEE International Conference on Robotics and Automation (ICRA)}. pp. 13714--13721.
\newblock \doi{10.1109/icra55743.2025.11128238}.
\newblock \urlprefix\url{https://doi.org/10.1109/icra55743.2025.11128238}.

\bibitem[{Ravichandran et~al.(2025{\natexlab{bb}})Ravichandran, Robey, Kumar, Pappas and Hassani}]{ravichandran2025roboguard}
Ravichandran Z, Robey A, Kumar V, Pappas GJ and Hassani H (2025{\natexlab{bb}}) {Safety Guardrails for LLM-Enabled Robots}.
\newblock \emph{arXiv preprint arXiv:2503.07885} \urlprefix\url{https://arxiv.org/abs/2503.07885}.

\bibitem[{Ravichandran et~al.(2026{\natexlab{bb}})Ravichandran, Snyder, Robey, Hassani, Kumar and Pappas}]{sref260219983}
Ravichandran Z, Snyder D, Robey A, Hassani H, Kumar V and Pappas GJ (2026{\natexlab{bb}}) {Contextual Safety Reasoning and Grounding for Open-World Robots}.
\newblock \urlprefix\url{https://arxiv.org/abs/2602.19983}.

\bibitem[{Ren et~al.(2023)Ren, Dixit, Bodrova, Singh, Tu, Brown, Xu, Takayama, Xia, Varley, Xu, Sadigh, Zeng and Majumdar}]{grid49}
Ren AZ, Dixit A, Bodrova A, Singh S, Tu S, Brown N, Xu P, Takayama L, Xia F, Varley J, Xu Z, Sadigh D, Zeng A and Majumdar A (2023) Robots that ask for help: Uncertainty alignment for large language model planners.
\newblock In: \emph{Proceedings of The 7th Conference on Robot Learning}, \emph{Proceedings of Machine Learning Research}, volume 229. PMLR, pp. 661--682.
\newblock \urlprefix\url{https://proceedings.mlr.press/v229/ren23a.html}.

\bibitem[{Ren et~al.(2026)Ren, Yao, Long, Huang, Wang, Chen, Ding, Wang, Luo, Ivanovic, Pavone, Tomizuka, Luo, Chen and Tian}]{ren2026atlas}
Ren J, Yao K, Long J, Huang T, Wang H, Chen Y, Ding W, Wang Y, Luo W, Ivanovic B, Pavone M, Tomizuka M, Luo P, Chen Y and Tian R (2026) {ATLAS}: Agentic harness for whole-body intelligence from human video.
\newblock \urlprefix\url{https://atlas-humanoid.github.io/}.
\newblock Early release.

\bibitem[{Ren et~al.(2026{\natexlab{bb}})Ren, Xu, Song, Xu, Dai, Yu and Lu}]{real2gym2026}
Ren K, Xu Y, Song K, Xu L, Dai B, Yu M and Lu T (2026{\natexlab{bb}}) {Real2Gym: Building Gyms from Videos, Bringing Skills to Robots}.
\newblock \emph{arXiv preprint arXiv:2609.37089} \urlprefix\url{https://arxiv.org/abs/2609.37089}.

\bibitem[{Ren et~al.(2024)Ren, Li, Luo, Song, Chen, Liufu, Yang, Zheng, Xu, Huang, Ding, Xie, Zhang, Fu, Liu, Lin, Zheng and Liang}]{ren2024infiniteworld}
Ren P, Li M, Luo Z, Song X, Chen Z, Liufu W, Yang Y, Zheng H, Xu R, Huang Z, Ding T, Xie L, Zhang K, Fu C, Liu Y, Lin L, Zheng F and Liang X (2024) {InfiniteWorld: A Unified Scalable Simulation Framework for General Visual-Language Robot Interaction}.
\newblock \doi{10.48550/arXiv.2412.05789}.
\newblock \urlprefix\url{https://arxiv.org/abs/2412.05789}.

\bibitem[{Renhang et~al.(2026)Renhang, Majumder, Pala and Poria}]{audit261010388}
Renhang L, Majumder N, Pala TD and Poria S (2026) {RoboQuest: Generalist Physical Agents that Search, Inspect and Test}.
\newblock arXiv:2610.10388.
\newblock \urlprefix\url{https://arxiv.org/abs/2610.10388v1}.

\bibitem[{Rivera et~al.(2024)Rivera, Byrd, Paul, Feldman, Booker, Holmes, Handelman, Kemp, Badger, Schmidt, Jatavallabhula, de~Melo, Seenivasan, Unberath and Chellappa}]{audit241006108}
Rivera C, Byrd G, Paul W, Feldman T, Booker M, Holmes E, Handelman D, Kemp B, Badger A, Schmidt A, Jatavallabhula KM, de~Melo CM, Seenivasan L, Unberath M and Chellappa R (2024) {ConceptAgent: LLM-Driven Precondition Grounding and Tree Search for Robust Task Planning and Execution}.
\newblock \emph{arXiv preprint arXiv:2410.06108} \doi{10.48550/arXiv.2410.06108}.
\newblock \urlprefix\url{https://arxiv.org/abs/2410.06108}.

\bibitem[{Robey et~al.(2024)Robey, Ravichandran, Kumar, Hassani and Pappas}]{robey2024robopair}
Robey A, Ravichandran Z, Kumar V, Hassani H and Pappas GJ (2024) Jailbreaking {LLM}-controlled robots.
\newblock \emph{arXiv preprint arXiv:2410.13691} \doi{10.48550/arXiv.2410.13691}.
\newblock \urlprefix\url{https://arxiv.org/abs/2410.13691}.

\bibitem[{{RoboChallenge Team}(2026)}]{robochallengezero2026}
{RoboChallenge Team} (2026) {RoboChallenge Zero: A Continually Evolving Benchmark for Zero-Shot Evaluation on Real Robots}.
\newblock Technical report, RoboChallenge.
\newblock \urlprefix\url{https://robochallenge.ai/docs/robochallenge_techreport_ZERO.pdf}.

\bibitem[{{robocurve}(2026)}]{roboharm2026}
{robocurve} (2026) {RoboHarm}.
\newblock Software repository.
\newblock \urlprefix\url{https://github.com/robocurve/roboharm}.
\newblock Accessed: 2026-11-06.

\bibitem[{{RobotKitAI}(2026)}]{piperjev2026}
{RobotKitAI} (2026) {PiPER Astra/Jev}.
\newblock \urlprefix\url{https://github.com/RobotKitAI/piper-astra-jev}.

\bibitem[{{robotmcp}(n.d.)}]{rosmcp2026}
{robotmcp} (n.d.) {ROS MCP Server}.
\newblock Software repository.
\newblock \urlprefix\url{https://github.com/robotmcp/ros-mcp-server}.
\newblock Accessed: 2026-11-11.

\bibitem[{{rokbenko}(2026)}]{grid79}
{rokbenko} (2026) {quackd}.
\newblock GitHub repository.
\newblock \urlprefix\url{https://github.com/rokbenko/quackd}.
\newblock Accessed: September 29, 2026.

\bibitem[{{RomanSlack}(2026)}]{jevdrone2026}
{RomanSlack} (2026) {jev-drone}.
\newblock GitHub repository.
\newblock \urlprefix\url{https://github.com/RomanSlack/jev-drone}.
\newblock Accessed October 5, 2026.

\bibitem[{Royce et~al.(2025)Royce, Kaufmann, Becktor, Moon, Carpenter, Pak, Towler, Thakker and Khattak}]{rosapaper2024}
Royce R, Kaufmann M, Becktor J, Moon S, Carpenter K, Pak K, Towler A, Thakker R and Khattak S (2025) {Enabling Novel Mission Operations and Interactions with ROSA: The Robot Operating System Agent}.
\newblock In: \emph{2025 IEEE Aerospace Conference}. pp. 1--16.
\newblock \doi{10.1109/aero63441.2025.11068426}.
\newblock \urlprefix\url{https://doi.org/10.1109/aero63441.2025.11068426}.

\bibitem[{Ruan et~al.(2026)Ruan, Zhang, Lv, Huang, Zhang, Si, Yang, Wang, Ni, Duan, Bi, Fan, Li, Fu, Zhang, Wang, Xie and Zhou}]{ruan2026pyruadex}
Ruan F, Zhang K, Lv G, Huang W, Zhang Y, Si R, Yang S, Wang L, Ni R, Duan W, Bi J, Fan Y, Li H, Fu Z, Zhang J, Wang D, Xie E and Zhou D (2026) {PyRUA-Dex: Learning Programmatic Policies for Dexterous Bimanual Manipulation with Coding Agents}.
\newblock \urlprefix\url{https://arxiv.org/abs/2611.33144}.

\bibitem[{Rudra et~al.(2026)Rudra, Zhang, Manas, Prasad and Chalvatzaki}]{rudra2026suto}
Rudra S, Zhang Y, Manas K, Prasad V and Chalvatzaki G (2026) {SUTO: Failure-Localized Reasoning for Robot Task Planning and Execution}.
\newblock CoRL 2026 Workshop on Agentic Robotics.
\newblock \urlprefix\url{https://openreview.net/forum?id=BCdDYO0jo9}.

\bibitem[{Rui et~al.(2026)Rui, Zuo, Wang, Jia, Ma, Zhou and Li}]{rui2026semantichandoff}
Rui K, Zuo Y, Wang J, Jia H, Ma J, Zhou W and Li M (2026) {Diagnosing Semantic Handoff Failures in Agent-Orchestrated Vision-Language-Action Skill Composition}.
\newblock \emph{arXiv preprint arXiv:2607.06256} \urlprefix\url{https://arxiv.org/abs/2607.06256}.
\newblock Presented at RSS 2026 SemRob Workshop.

\bibitem[{Rui et~al.(2026{\natexlab{bb}})Rui, Wang, Huang and Gao}]{si260938178}
Rui Z, Wang R, Huang H and Gao Y (2026{\natexlab{bb}}) {Skill-Space Shooting for Autonomous Robot Policy Improvement}.
\newblock \urlprefix\url{https://arxiv.org/abs/2609.38178}.

\bibitem[{Ryu et~al.(2024)Ryu, Liao, Li, Delgosha, Sreenath and Mehr}]{curricullm2024}
Ryu K, Liao Q, Li Z, Delgosha P, Sreenath K and Mehr N (2024) {CurricuLLM: Automatic Task Curricula Design for Learning Complex Robot Skills using Large Language Models}.
\newblock arXiv:2409.18382.
\newblock \urlprefix\url{https://arxiv.org/abs/2409.18382}.
\newblock Accepted to ICRA 2025.

\bibitem[{Salimpour et~al.(2025)Salimpour, Fu, Rachwa{\l}, Bertrand, O'Sullivan, Jakob, Keramat, Militano, Toffetti, Edelman and Queralta}]{embodiedsurvey2025}
Salimpour S, Fu L, Rachwa{\l} K, Bertrand P, O'Sullivan K, Jakob R, Keramat F, Militano L, Toffetti G, Edelman H and Queralta JP (2025) {Towards Embodied Agentic AI: Review and Classification of LLM- and VLM-Driven Robot Autonomy and Interaction}.
\newblock \emph{arXiv preprint arXiv:2508.05294} \urlprefix\url{https://arxiv.org/abs/2508.05294}.

\bibitem[{Santos et~al.(2026)Santos, Khadraoui, Farhat, Yous, Teffahi and Hacid}]{grid43}
Santos VGd, Khadraoui I, Farhat I, Yous H, Teffahi S and Hacid H (2026) {ALRM: Agentic LLM for Robotic Manipulation}.
\newblock arXiv:2601.19510.
\newblock \urlprefix\url{https://arxiv.org/abs/2601.19510}.

\bibitem[{Sato et~al.(2026)Sato, Iwasawa, Tang and Kuroki}]{sail2026}
Sato M, Iwasawa Y, Tang Y and Kuroki S (2026) {SAIL: Test-Time Scaling for In-Context Imitation Learning with VLM}.
\newblock \emph{arXiv preprint arXiv:2603.08269} \urlprefix\url{https://arxiv.org/abs/2603.08269}.
\newblock Accepted to IROS 2026; project: https://pub.sakana.ai/sail/.

\bibitem[{Schakkal et~al.(2025)Schakkal, Zandonati, Yang and Azizan}]{schakkal2025hierarchical}
Schakkal A, Zandonati B, Yang Z and Azizan N (2025) {Hierarchical Vision-Language Planning for Multi-Step Humanoid Manipulation}.
\newblock \emph{arXiv preprint arXiv:2506.22827} \urlprefix\url{https://arxiv.org/abs/2506.22827}.
\newblock Accepted to RSS 2025 Workshop on Robot Planning in the Era of Foundation Models.

\bibitem[{Schroeder et~al.(2025)Schroeder, Biza, Weng, Luo and Glass}]{schroeder2025rover}
Schroeder P, Biza O, Weng T, Luo H and Glass J (2025) {ROVER: Recursive Reasoning Over Videos with Vision-Language Models for Embodied Tasks}.
\newblock \emph{arXiv preprint arXiv:2508.01943} \urlprefix\url{https://arxiv.org/abs/2508.01943}.

\bibitem[{{SE3 Labs}(2026)}]{se3labs2026agentic}
{SE3 Labs} (2026) Where should a robot's intelligence live? code-as-policy versus agentic control.
\newblock SE3 Labs, Dispatch 001.
\newblock \urlprefix\url{https://se3labs.ai/blog/agentic-vs-code-as-policy/}.
\newblock Published 20 September 2026.

\bibitem[{Seker et~al.(2026)Seker, Aggarwal, Wickramarachchi, Francis and Kroemer}]{audit260909808}
Seker MY, Aggarwal S, Wickramarachchi R, Francis J and Kroemer O (2026) {GTA-2: A Multi-VLM Framework for Synthesizing Robot Manipulation Skills via Grounded Task Axes}.
\newblock arXiv:2609.09808.
\newblock \urlprefix\url{https://arxiv.org/abs/2609.09808}.

\bibitem[{{SenseNova} et~al.(2026){SenseNova}, {Joint Research Team}, Cai, Cao, Chen, Chua, Dai, Deng, Deng, Diao, Ding, Fan, Fu, Gong, Gu, He, Hou, Li, Li, Li, Li, Liao, Lin, Lin, Liu, Liu, Liu, Liu, Liu, Loy, Lu, Luo, Ma, Pan, Pan, Qian, Shang, Shao, Su, Tao, Tao, Wang, Wang, Wang, Wang, Wen, Wu, Wu, Wu, Wu, Wu, Xiao, Xu, Yamashita, Yang, Yang, Yin, Yin, Zhao, Zhong, Zhuo and Zuo}]{sensenova2026embodied}
{SenseNova}, {Joint Research Team}, Cai Z, Cao Z, Chen T, Chua HC, Dai L, Deng H, Deng H, Diao H, Ding C, Fan W, Fu X, Gong Z, Gu C, He Z, Hou Z, Li B, Li Q, Li Q, Li X, Liao Z, Lin D, Lin Z, Liu B, Liu C, Liu P, Liu T, Liu Z, Loy CC, Lu L, Luo J, Ma Y, Pan L, Pan X, Qian O, Shang Z, Shao Y, Su D, Tao D, Tao Q, Wang Q, Wang R, Wang X, Wang Z, Wen B, Wu D, Wu J, Wu S, Wu X, Wu Z, Xiao H, Xu W, Yamashita A, Yang L, Yang Z, Yin B, Yin W, Zhao H, Zhong T, Zhuo L and Zuo J (2026) {Has GPT-6-Astra Achieved Embodied Intelligence?}
\newblock Technical Report arXiv:2610.22471, SenseNova.
\newblock \urlprefix\url{https://arxiv.org/abs/2610.22471}.
\newblock Technical report. \href{https://huggingface.co/blog/sensenova/roborsi}{Action-review experiments}.

\bibitem[{Sermanet et~al.(2025)Sermanet, Majumdar, Irpan, Kalashnikov and Sindhwani}]{sermanet2025asimov}
Sermanet P, Majumdar A, Irpan A, Kalashnikov D and Sindhwani V (2025) Generating robot constitutions \& benchmarks for semantic safety.
\newblock In: \emph{Proceedings of the 9th Conference on Robot Learning}, \emph{Proceedings of Machine Learning Research}, volume 305. PMLR, pp. 4767--4823.
\newblock \urlprefix\url{https://proceedings.mlr.press/v305/sermanet25a.html}.

\bibitem[{Sha et~al.(2023)Sha, Mu, Jiang, Chen, Xu, Luo, Li, Tomizuka, Zhan and Ding}]{sha2023languagempc}
Sha H, Mu Y, Jiang Y, Chen L, Xu C, Luo P, Li SE, Tomizuka M, Zhan W and Ding M (2023) {LanguageMPC: Large Language Models as Decision Makers for Autonomous Driving}.
\newblock \doi{10.48550/arXiv.2310.03026}.
\newblock \urlprefix\url{https://arxiv.org/abs/2310.03026}.

\bibitem[{Shah et~al.(2025)Shah, Yu, Zhu, Zhu and Mart{\'i}n-Mart{\'i}n}]{bumble}
Shah R, Yu A, Zhu Y, Zhu Y and Mart{\'i}n-Mart{\'i}n R (2025) {BUMBLE: Unifying Reasoning and Acting with Vision-Language Models for Building-wide Mobile Manipulation}.
\newblock In: \emph{2025 IEEE International Conference on Robotics and Automation (ICRA)}. pp. 13337--13345.
\newblock \doi{10.1109/icra55743.2025.11128444}.
\newblock \urlprefix\url{https://doi.org/10.1109/icra55743.2025.11128444}.

\bibitem[{Shaji et~al.(2026)Shaji, Huppertz, Mitrevski and Houben}]{audit260303148}
Shaji S, Huppertz F, Mitrevski A and Houben S (2026) {From Language to Action: Can LLM-Based Agents Be Used for Embodied Robot Cognition?}
\newblock arXiv:2603.03148.
\newblock \urlprefix\url{https://arxiv.org/abs/2603.03148v1}.
\newblock Accepted to ICRA 2026.

\bibitem[{Shan et~al.(2026)Shan, Hu, Zhou, Dong, Hou, Bai, Ma, Dong and Yang}]{shan2026contactharness}
Shan S, Hu Y, Zhou C, Dong Y, Hou B, Bai J, Ma B, Dong J and Yang J (2026) {Contact Harness: Enabling VLM Agents to Perform Contact-Rich Manipulation}.
\newblock \urlprefix\url{https://arxiv.org/abs/2611.29377}.

\bibitem[{Sharma et~al.(2026)Sharma, Hu, Kamat, Sadigh and Bohg}]{sharma2026roboarcade}
Sharma S, Hu H, Kamat M, Sadigh D and Bohg J (2026) {RoboARCade: Benchmarking Agentic Robot Control}.
\newblock \emph{arXiv preprint arXiv:2611.08209} \urlprefix\url{https://arxiv.org/abs/2611.08209}.

\bibitem[{Sharrock et~al.(2026)Sharrock, Aronsson, Petersson, Backlund, Wennström, Petersson, Nordström and Carlsson}]{sharrock2026dronebench}
Sharrock C, Aronsson E, Petersson L, Backlund A, Wennström A, Petersson H, Nordström K and Carlsson R (2026) {Drone-Bench: Tracking Simple Drone Surveillance Capabilities of Frontier Models}.
\newblock \urlprefix\url{https://andonlabs.com/docs/Drone_Bench.pdf}.

\bibitem[{Shen et~al.(2026)Shen, Cheng, Zhao, Zhu, Yi, Zhao, Yang, Shi and Lin}]{shen2026openarmhandoffs}
Shen Y, Cheng C, Zhao Z, Zhu J, Yi Z, Zhao Q, Yang J, Shi Y and Lin CT (2026) {Representation Handoffs for OpenArm-Based Laboratory Mobile Manipulation}.
\newblock RSS 2026 Workshop From Perception to Action.
\newblock \urlprefix\url{https://openreview.net/forum?id=aecUusZZ4H}.

\bibitem[{Shen et~al.(2026{\natexlab{bb}})Shen, You, Liu, Zheng, Zha, Yamazaki, Zhang, Huang, Sun, Chen, He, Liu, Chang, Fragkiadaki, Shah, Schwager, Henderson, Abraham and Xu}]{embodiedswe2026}
Shen Z, You H, Liu Y, Zheng Z, Zha L, Yamazaki K, Zhang M, Huang S, Sun J, Chen Q, He L, Liu K, Chang H, Fragkiadaki K, Shah D, Schwager M, Henderson P, Abraham I and Xu C (2026{\natexlab{bb}}) {EmbodiedSWE: Coding Agents for Long Horizon Dexterous Robotics}.
\newblock \emph{arXiv preprint arXiv:2609.27308} \urlprefix\url{https://arxiv.org/abs/2609.27308}.

\bibitem[{Sheng et~al.(2026)Sheng, Wang, Li, Dai, Liu and Chen}]{sheng2026evolife}
Sheng K, Wang L, Li J, Dai H, Liu C and Chen Q (2026) {Ask to Evolve: A Unified Harness for Lifelong Embodied Navigation with Hybrid Action Space and Persistent Memory}.
\newblock arXiv:2610.39446.
\newblock \urlprefix\url{https://arxiv.org/abs/2610.39446}.

\bibitem[{Shentu et~al.(2024)Shentu, Wu, Rajeswaran and Abbeel}]{shentu2024fromllmstoactions}
Shentu Y, Wu P, Rajeswaran A and Abbeel P (2024) {From LLMs to Actions: Latent Codes as Bridges in Hierarchical Robot Control}.
\newblock \doi{10.48550/arXiv.2405.04798}.
\newblock \urlprefix\url{https://arxiv.org/abs/2405.04798}.

\bibitem[{Shi et~al.(2025)Shi, Wu, Kumar and Sukhatme}]{shi2025pipllm}
Shi G, Wu Y, Kumar V and Sukhatme GS (2025) {PIP-LLM: Integrating PDDL-Integer Programming with LLMs for Coordinating Multi-Robot Teams Using Natural Language}.
\newblock arXiv:2510.22784.
\newblock \urlprefix\url{https://arxiv.org/abs/2510.22784}.

\bibitem[{Shi et~al.(2026)Shi, Li, Ding, Deng, Wang and Nie}]{audit260939915}
Shi H, Li Z, Ding M, Deng X, Wang Y and Nie L (2026) {NavHarness: Adaptive Goals for Agentic Vision-Language Navigation}.
\newblock arXiv:2609.39915.
\newblock \urlprefix\url{https://arxiv.org/abs/2609.39915v1}.

\bibitem[{Shi et~al.(2025{\natexlab{bb}})Shi, Yang, Chao, Wan, Shao, Lei, Qian, Le, Chaudhari, Daniilidis, Wen and Jayaraman}]{shi2025maestro}
Shi J, Yang R, Chao K, Wan SB, Shao Y, Lei J, Qian J, Le L, Chaudhari P, Daniilidis K, Wen C and Jayaraman D (2025{\natexlab{bb}}) {MAESTRO: Orchestrating Robotics Modules with Coding Agents for Zero-Shot Generalist Robots}.
\newblock \emph{arXiv preprint arXiv:2511.00917} \urlprefix\url{https://arxiv.org/abs/2511.00917}.
\newblock Accepted to CoRL 2026.

\bibitem[{Shi et~al.(2025{\natexlab{bc}})Shi, Ichter, Equi, Ke, Pertsch, Vuong, Tanner, Walling, Wang, Fusai, Li-Bell, Driess, Groom, Levine and Finn}]{grid22}
Shi LX, Ichter B, Equi MR, Ke L, Pertsch K, Vuong Q, Tanner J, Walling A, Wang H, Fusai N, Li-Bell A, Driess D, Groom L, Levine S and Finn C (2025{\natexlab{bc}}) Hi robot: Open-ended instruction following with hierarchical vision-language-action models.
\newblock In: \emph{Proceedings of the 42nd International Conference on Machine Learning}, \emph{Proceedings of Machine Learning Research}, volume 267. PMLR, pp. 54919--54933.
\newblock \urlprefix\url{https://proceedings.mlr.press/v267/shi25d.html}.

\bibitem[{Shi et~al.(2026{\natexlab{bb}})Shi, Liu, Hou, Chen, Zhang, Hu, Liu and Zhang}]{shi2026atomicwam}
Shi S, Liu Z, Hou Z, Chen Y, Zhang T, Hu K, Liu Y and Zhang J (2026{\natexlab{bb}}) {Atomic-WAM: Execution-Grounded Atomic-Motion Reasoning for World Action Models}.
\newblock \emph{arXiv preprint arXiv:2610.20157} \urlprefix\url{https://arxiv.org/abs/2610.20157}.

\bibitem[{Shin et~al.(2024)Shin, jeon, Kim, Kang and Zhang}]{shin2024socraticplanner}
Shin S, jeon S, Kim J, Kang GC and Zhang BT (2024) {Socratic Planner: Self-QA-Based Zero-Shot Planning for Embodied Instruction Following}.
\newblock \doi{10.48550/arXiv.2404.15190}.
\newblock \urlprefix\url{https://arxiv.org/abs/2404.15190}.
\newblock Accepted to ICRA 2025.

\bibitem[{Shirai et~al.(2023)Shirai, Beltran-Hernandez, Hamaya, Hashimoto, Tanaka, Kawaharazuka, Tanaka, Ushiku and Mori}]{shirai2023vilain}
Shirai K, Beltran-Hernandez CC, Hamaya M, Hashimoto A, Tanaka S, Kawaharazuka K, Tanaka K, Ushiku Y and Mori S (2023) {Vision-Language Interpreter for Robot Task Planning}.
\newblock \doi{10.48550/arXiv.2311.00967}.
\newblock \urlprefix\url{https://arxiv.org/abs/2311.00967}.
\newblock Accepted to ICRA 2024.

\bibitem[{Si et~al.(2026)Si, Bi, Yang, Ni, Huang, Wang, Jiang, Zhang, Ruan, Zhang, Wang, Li, Feng, Dong and Zhou}]{si2026fewer}
Si R, Bi J, Yang S, Ni R, Huang W, Wang Q, Jiang S, Zhang Y, Ruan F, Zhang K, Wang D, Li X, Feng H, Dong Z and Zhou D (2026) {Fewer Tokens, Better Action: GPT-6 Astra Robot Agents with 14\% Higher Success Rate but 65\% Fewer Tokens}.
\newblock \emph{arXiv preprint arXiv:2610.01939} \urlprefix\url{https://arxiv.org/abs/2610.01939}.

\bibitem[{Singh et~al.(2024)Singh, Das, Han, Nakov and Laptev}]{xref241117636}
Singh H, Das RJ, Han M, Nakov P and Laptev I (2024) {MALMM: Multi-Agent Large Language Models for Zero-Shot Robotics Manipulation}.
\newblock \emph{arXiv preprint arXiv:2411.17636} \urlprefix\url{https://arxiv.org/abs/2411.17636}.

\bibitem[{Singh et~al.(2023)Singh, Blukis, Mousavian, Goyal, Xu, Tremblay, Fox, Thomason and Garg}]{grid65}
Singh I, Blukis V, Mousavian A, Goyal A, Xu D, Tremblay J, Fox D, Thomason J and Garg A (2023) {ProgPrompt: Generating Situated Robot Task Plans using Large Language Models}.
\newblock In: \emph{2023 IEEE International Conference on Robotics and Automation (ICRA)}. pp. 11523--11530.
\newblock \doi{10.1109/icra48891.2023.10161317}.
\newblock \urlprefix\url{https://doi.org/10.1109/icra48891.2023.10161317}.

\bibitem[{Skreta et~al.(2024)Skreta, Zhou, Yuan, Darvish, Aspuru-Guzik and Garg}]{xref240104157}
Skreta M, Zhou Z, Yuan JL, Darvish K, Aspuru-Guzik A and Garg A (2024) {RePLan: Robotic Replanning with Perception and Language Models}.
\newblock \emph{arXiv preprint arXiv:2401.04157} \urlprefix\url{https://arxiv.org/abs/2401.04157}.

\bibitem[{Song et~al.(2023)Song, Wu, Washington, Sadler, Chao and Su}]{grid56}
Song CH, Wu J, Washington C, Sadler BM, Chao WL and Su Y (2023) {LLM-Planner: Few-Shot Grounded Planning for Embodied Agents with Large Language Models}.
\newblock In: \emph{2023 IEEE/CVF International Conference on Computer Vision (ICCV)}. pp. 2986--2997.
\newblock \doi{10.1109/iccv51070.2023.00280}.
\newblock \urlprefix\url{https://doi.org/10.1109/iccv51070.2023.00280}.

\bibitem[{Song et~al.(2023{\natexlab{bb}})Song, Zhou, Liu, Fang, Shu and Ma}]{selfrefinedreward2023}
Song J, Zhou Z, Liu J, Fang C, Shu Z and Ma L (2023{\natexlab{bb}}) {Self-Refined Large Language Model as Automated Reward Function Designer for Deep Reinforcement Learning in Robotics}.
\newblock arXiv:2309.06687.
\newblock \urlprefix\url{https://arxiv.org/abs/2309.06687}.

\bibitem[{Song et~al.(2025)Song, Chen, Ma, Jin, Zhao, Ding, Xiong and Pan}]{song2025collabot}
Song K, Chen G, Ma S, Jin N, Zhao G, Ding M, Xiong Z and Pan J (2025) {CollaBot: Vision-Language Guided Simultaneous Collaborative Manipulation}.
\newblock \urlprefix\url{https://arxiv.org/abs/2508.03526}.

\bibitem[{Song et~al.(2026)Song, Liang, Fu, Wang, Yang and Liu}]{song2026embodiedrsiactivecontinualrobot}
Song P, Liang Z, Fu K, Wang M, Yang J and Liu S (2026) Embodiedrsi: Active continual robot learning through hypothesis-guided co-evolution.
\newblock \urlprefix\url{https://arxiv.org/abs/2610.10498}.

\bibitem[{Song et~al.(2025{\natexlab{bb}})Song, Xie, Li, Li, Wang and Qi}]{xref260206967}
Song S, Xie X, Li Z, Li Y, Wang S and Qi B (2025{\natexlab{bb}}) {Leveraging Adaptive Group Negotiation for Heterogeneous Multi-Robot Collaboration with Large Language Models}.
\newblock \emph{arXiv preprint arXiv:2602.06967} \urlprefix\url{https://arxiv.org/abs/2602.06967}.

\bibitem[{Song et~al.(2025{\natexlab{bc}})Song, Chen, Li, He, Zhao, Cui, Ding, Su, Tang, Wang, Cheng, Ge, Zheng, Liu, Wang and Li}]{song2025rationalvla}
Song W, Chen J, Li W, He X, Zhao H, Cui C, Ding P, Su S, Tang F, Wang D, Cheng X, Ge Z, Zheng X, Liu Z, Wang H and Li H (2025{\natexlab{bc}}) {RationalVLA: A Rational Vision-Language-Action Model with Dual System}.
\newblock \doi{10.48550/arXiv.2506.10826}.
\newblock \urlprefix\url{https://arxiv.org/abs/2506.10826}.

\bibitem[{Song et~al.(2024)Song, Chen, Liu, Chen, Li and Lin}]{song2024navgen}
Song X, Chen W, Liu Y, Chen W, Li G and Lin L (2024) {Towards Long-Horizon Vision-Language Navigation: Platform, Benchmark and Method}.
\newblock \doi{10.48550/arXiv.2412.09082}.
\newblock \urlprefix\url{https://arxiv.org/abs/2412.09082}.
\newblock Accepted to CVPR 2025.

\bibitem[{Song et~al.(2026{\natexlab{bb}})Song, Mittal and Jain}]{song2026glide}
Song Y, Mittal A and Jain U (2026{\natexlab{bb}}) {Learning Beyond What Humans Can Demonstrate}.
\newblock \emph{arXiv preprint arXiv:2609.24996} \doi{10.48550/arXiv.2609.24996}.
\newblock \urlprefix\url{https://arxiv.org/abs/2609.24996}.
\newblock Accepted to CoRL 2026.

\bibitem[{Soni et~al.(2024)Soni, Venkataraman, Chandra, Fischmeister, Liang, Dai and Yang}]{soni2024videoagent}
Soni A, Venkataraman S, Chandra A, Fischmeister S, Liang P, Dai B and Yang S (2024) {VideoAgent: Self-Improving Video Generation}.
\newblock \doi{10.48550/arXiv.2410.10076}.
\newblock \urlprefix\url{https://arxiv.org/abs/2410.10076}.

\bibitem[{Sridhar et~al.(2026)Sridhar, Pan, Sharma and Finn}]{sridhar2026memer}
Sridhar A, Pan J, Sharma S and Finn C (2026) {MemER: Scaling Up Memory for Robot Control via Experience Retrieval}.
\newblock In: \emph{International Conference on Learning Representations}.
\newblock \urlprefix\url{https://openreview.net/forum?id=1dH4ARGdwD}.

\bibitem[{Srikanth et~al.(2026)Srikanth, Zhao, Xu, Wu and Jiang}]{srikanth2026aide2}
Srikanth D, Zhao B, Xu D, Wu Y and Jiang Z (2026) {Recursive self-improvement of AI research agents}.
\newblock \emph{arXiv preprint arXiv:2609.26457} \doi{10.48550/arXiv.2609.26457}.
\newblock \urlprefix\url{https://arxiv.org/abs/2609.26457}.

\bibitem[{Stamatopoulou et~al.(2024)Stamatopoulou, Li and Kanoulas}]{sds2024}
Stamatopoulou M, Li J and Kanoulas D (2024) {SDS -- See it, Do it, Sorted: Quadruped Skill Synthesis from Single Video Demonstration}.
\newblock arXiv:2410.11571.
\newblock \urlprefix\url{https://arxiv.org/abs/2410.11571}.

\bibitem[{{strands-labs}(n.d.)}]{strandsrobots2026}
{strands-labs} (n.d.) {Strands Robots}.
\newblock Software repository.
\newblock \urlprefix\url{https://github.com/strands-labs/robots}.
\newblock Accessed: 2026-10-26.

\bibitem[{Strobel et~al.(2024)Strobel, Dorigo and Fritz}]{xref241011387}
Strobel V, Dorigo M and Fritz M (2024) {LLM2Swarm: Robot Swarms that Responsively Reason, Plan, and Collaborate through LLMs}.
\newblock \emph{arXiv preprint arXiv:2410.11387} \urlprefix\url{https://arxiv.org/abs/2410.11387}.
\newblock Accepted to NeurIPS 2024 Workshop on Open-World Agents.

\bibitem[{Su et~al.(2026)Su, Xu, van Kaick, Xu and Hu}]{audit260302669}
Su X, Xu J, van Kaick O, Xu K and Hu R (2026) {IMR-LLM: Industrial Multi-Robot Task Planning and Program Generation using Large Language Models}.
\newblock arXiv:2603.02669.
\newblock \urlprefix\url{https://arxiv.org/abs/2603.02669v1}.

\bibitem[{Sun et~al.(2026)Sun, Li and Tian}]{sun2026pi05callastra}
Sun J, Li C and Tian Y (2026) {Make $\pi_{0.5}$ Learn to Call Astra for Reliable Robot Manipulation}.
\newblock \urlprefix\url{https://arxiv.org/abs/2610.14117}.

\bibitem[{Sun et~al.(2025)Sun, Zhang, Han, Zhao, Yong, He, Wang, Cao, Guo and Xu}]{sun2025trinityhumanoid}
Sun J, Zhang Q, Han G, Zhao W, Yong Z, He Y, Wang J, Cao J, Guo Y and Xu R (2025) {Trinity: A Modular Humanoid Robot AI System}.
\newblock \doi{10.48550/arXiv.2503.08338}.
\newblock \urlprefix\url{https://arxiv.org/abs/2503.08338}.

\bibitem[{Sun et~al.(2023)Sun, Jha, Hori, Jain, Corcodel, Zhu, Tomizuka and Romeres}]{sun2023interactiveplanning}
Sun L, Jha DK, Hori C, Jain S, Corcodel R, Zhu X, Tomizuka M and Romeres D (2023) {Interactive Planning Using Large Language Models for Partially Observable Robotics Tasks}.
\newblock \doi{10.48550/arXiv.2312.06876}.
\newblock \urlprefix\url{https://arxiv.org/abs/2312.06876}.

\bibitem[{Sun et~al.(2024)Sun, Li, Zhao, Huang and Xu}]{sun2024llmlocomotion}
Sun S, Li C, Zhao Z, Huang H and Xu W (2024) {Leveraging large language models for comprehensive locomotion control in humanoid robots design}.
\newblock \emph{Biomimetic Intelligence and Robotics} 4(4): 100187.
\newblock \doi{10.1016/j.birob.2024.100187}.
\newblock \urlprefix\url{https://doi.org/10.1016/j.birob.2024.100187}.

\bibitem[{Sun et~al.(2024{\natexlab{bb}})Sun, Liu, Lyu, Yang, Zhang and Li}]{card2024}
Sun S, Liu R, Lyu J, Yang JW, Zhang L and Li X (2024{\natexlab{bb}}) {A Large Language Model-Driven Reward Design Framework via Dynamic Feedback for Reinforcement Learning}.
\newblock arXiv:2410.14660.
\newblock \urlprefix\url{https://arxiv.org/abs/2410.14660}.

\bibitem[{Sun et~al.(2024{\natexlab{bc}})Sun, Hou, Wang, Yu, Liu, Yang, Liang, Gan and Han}]{sun2024dadue}
Sun W, Hou S, Wang Z, Yu B, Liu S, Yang X, Liang S, Gan Y and Han Y (2024{\natexlab{bc}}) {DaDu-E: Rethinking the Role of Large Language Model in Robotic Computing Pipeline}.
\newblock \doi{10.48550/arXiv.2412.01663}.
\newblock \urlprefix\url{https://arxiv.org/abs/2412.01663}.

\bibitem[{Suresh and Atkeson(2026)}]{suresh2026needforspeed}
Suresh K and Atkeson C (2026) {The need for speed}.
\newblock Technical blog.
\newblock \urlprefix\url{https://krishnasuresh.org/blog/2026/robot-whips/}.

\bibitem[{Sygkounas et~al.(2026)Sygkounas, Aregbede, Loutfi and Persson}]{sygkounas2026memento}
Sygkounas A, Aregbede V, Loutfi A and Persson A (2026) {MEMENTO: Memory-Guided Memetic Code-as-Policy Evolution}.
\newblock \emph{arXiv preprint arXiv:2607.22832} \urlprefix\url{https://arxiv.org/abs/2607.22832}.

\bibitem[{Takehi et~al.(2026)Takehi, Teoh, Zhang and Khattab}]{takehi2026weightspace}
Takehi R, Teoh J, Zhang A and Khattab O (2026) {Sparks of Weight-Space Programming: LLMs can natively read \& write neural network weights}.
\newblock Technical blog.
\newblock \urlprefix\url{https://rikiyat.github.io/blog/weight-space-programming/}.

\bibitem[{Talwar et~al.(2026)Talwar, Nijs, Verleysen and wyffels}]{audit260926499}
Talwar R, Nijs E, Verleysen A and wyffels F (2026) {Generalizing Manipulation Skills with a Local Coding Agent}.
\newblock arXiv:2609.26499.
\newblock \urlprefix\url{https://arxiv.org/abs/2609.26499v1}.

\bibitem[{Tan et~al.(2025)Tan, Chi, Chen, Ji, Zhao, Hao, Lyu, Cao, Zhao, Lyu, Zhou, Chen, Fu, Peng, Guo, Liang, Chen, Lyu, He, Ao, Lin, Wang, Wang and Zhang}]{tan2025roboosnext}
Tan H, Chi C, Chen X, Ji Y, Zhao Z, Hao X, Lyu Y, Cao M, Zhao J, Lyu H, Zhou E, Chen N, Fu Y, Peng C, Guo W, Liang D, Chen Z, Lyu M, He C, Ao Y, Lin Y, Wang P, Wang Z and Zhang S (2025) {RoboOS-NeXT: A Unified Memory-based Framework for Lifelong, Scalable, and Robust Multi-Robot Collaboration}.
\newblock \doi{10.48550/arXiv.2510.26536}.
\newblock \urlprefix\url{https://arxiv.org/abs/2510.26536}.

\bibitem[{Tan et~al.(2025{\natexlab{bb}})Tan, Hao, Chi, Lin, Lyu, Cao, Liang, Chen, Lyu, Peng, He, Ao, Lin, Wang, Wang and Zhang}]{audit250503673}
Tan H, Hao X, Chi C, Lin M, Lyu Y, Cao M, Liang D, Chen Z, Lyu M, Peng C, He C, Ao Y, Lin Y, Wang P, Wang Z and Zhang S (2025{\natexlab{bb}}) {RoboOS: A Hierarchical Embodied Framework for Cross-Embodiment and Multi-Agent Collaboration}.
\newblock arXiv:2505.03673.
\newblock \urlprefix\url{https://arxiv.org/abs/2505.03673}.

\bibitem[{Tan et~al.(2026)Tan, Wang, Wang and Yuan}]{tan2026forca}
Tan X, Wang Y, Wang Z and Yuan W (2026) {FORCA: A Force-Aware Coding Agent for Contact-Rich Robot Manipulation}.
\newblock \urlprefix\url{https://arxiv.org/abs/2610.31841}.

\bibitem[{Tanahashi et~al.(2023)Tanahashi, Inoue, Yamaguchi, Yaginuma, Shiotsuka, Shimatani, Iwamasa, Inoue, Yamaguchi, Igari, Horinouchi, Tokuhiro, Tokuchi and Aoki}]{tanahashi2023tanahashietal}
Tanahashi K, Inoue Y, Yamaguchi Y, Yaginuma H, Shiotsuka D, Shimatani H, Iwamasa K, Inoue Y, Yamaguchi T, Igari K, Horinouchi T, Tokuhiro K, Tokuchi Y and Aoki S (2023) {Evaluation of Large Language Models for Decision Making in Autonomous Driving}.
\newblock \doi{10.48550/arXiv.2312.06351}.
\newblock \urlprefix\url{https://arxiv.org/abs/2312.06351}.
\newblock Accepted to 2023 Symposium on Machine Learning for Autonomous Driving.

\bibitem[{Tanaka(2026)}]{tanaka2026taczero}
Tanaka K (2026) {TacZero: Training-Free Peg Insertion Using a General-Purpose Vision-Language Model with Tactile Feedback}.
\newblock arXiv:2610.07621.
\newblock \urlprefix\url{https://arxiv.org/abs/2610.07621}.

\bibitem[{Tang et~al.(2026)Tang, Jia, Chen, Li, Huang, Lv and Ji}]{tang2026vlmdewm}
Tang G, Jia Q, Chen G, Li T, Huang Z, Lv Z and Ji N (2026) {VLM-DEWM: Dynamic External World Model for Verifiable and Resilient Vision-Language Planning in Manufacturing}.
\newblock \urlprefix\url{https://arxiv.org/abs/2602.15549}.

\bibitem[{Tang et~al.(2026{\natexlab{bb}})Tang, Peng, Ding, Shen and Chen}]{si260929204}
Tang J, Peng J, Ding Z, Shen Y and Chen T (2026{\natexlab{bb}}) {AdaHVLA: Adaptive Harnesses for Long-Horizon Vision-Language-Action Execution}.
\newblock \urlprefix\url{https://arxiv.org/abs/2609.29204}.

\bibitem[{Tang et~al.(2026{\natexlab{bc}})Tang, Zhang, Guo, Zhao, Zhao, Li, Li, Zhao, Bian, Liu and Han}]{pear2026}
Tang J, Zhang Z, Guo J, Zhao F, Zhao Y, Li Z, Li X, Zhao Y, Bian S, Liu Z and Han S (2026{\natexlab{bc}}) {Training Reasoning Models in Simulation to Natively Use Real Robots}.
\newblock \urlprefix\url{https://arxiv.org/abs/2611.18562}.

\bibitem[{Tang et~al.(2026{\natexlab{bd}})Tang, Zhang, Yang, Yan, Zhou, Mao, Han, Cao, Xu, He, Fu, Feng and Huang}]{tang2026video2world}
Tang J, Zhang Z, Yang J, Yan Y, Zhou K, Mao L, Han R, Cao J, Xu W, He L, Fu M, Feng F and Huang B (2026{\natexlab{bd}}) {Video2World: Benchmarking Coding Agents for Interactive World Modeling from Embodied Videos}.
\newblock \emph{arXiv preprint arXiv:2610.04432} \urlprefix\url{https://arxiv.org/abs/2610.04432}.

\bibitem[{Tang et~al.(2025)Tang, Zhang, Wen, Zhang, Bian, xia zhou, Yan, Zhan, Jia, Wu, Lin and Liang}]{tang2025robopearlseditablevideosimulation}
Tang T, Zhang L, Wen Y, Zhang K, Bian JW, xia zhou, Yan T, Zhan K, Jia P, Wu H, Lin L and Liang X (2025) {RoboPearls: Editable Video Simulation for Robot Manipulation}.
\newblock \urlprefix\url{https://arxiv.org/abs/2506.22756}.
\newblock Accepted to ICCV 2025.

\bibitem[{Tang et~al.(2026{\natexlab{be}})Tang, Wang, Zhang, Wang, Chen, Zhang, Ma, Jiang, Guo, Li, Jiao, Cui, Yin, Liang, Wan, Lin, Zhao, Zhang, Zhou, Zhang and You}]{tang2026trinity}
Tang W, Wang R, Zhang J, Wang H, Chen H, Zhang Z, Ma Y, Jiang Z, Guo Z, Li H, Jiao Z, Cui Z, Yin X, Liang S, Wan Z, Lin H, Zhao W, Zhang K, Zhou P, Zhang O and You Y (2026{\natexlab{be}}) {The Trinity}: Capability-driven recursive self-improvement for multimodal agents.
\newblock \doi{10.48550/arXiv.2610.36077}.
\newblock \urlprefix\url{https://arxiv.org/abs/2610.36077}.

\bibitem[{Tao et~al.(2025)Tao, Yang, Ding and Erickson}]{tao2025lams}
Tao Y, Yang J, Ding D and Erickson Z (2025) {LAMS: LLM-Driven Automatic Mode Switching for Assistive Teleoperation}.
\newblock \doi{10.48550/arXiv.2501.08558}.
\newblock \urlprefix\url{https://arxiv.org/abs/2501.08558}.

\bibitem[{Team et~al.(2025)Team, Abdolmaleki, Abeyruwan, Ainslie, Alayrac, Gonzalez~Arenas, Balakrishna, Batchelor, Bewley, Bingham, Bloesch, Bousmalis, Brakel, Brohan, Buschmann, Byravan, Cabi, Caluwaerts, Casarini, Chan, Chang, Chappellet-Volpini, Chen, Chen, Chiang, Choromanski, Collister, D'Ambrosio, Dasari, Davchev, Dave, Devin, Di~Palo, Ding, Doersch, Dostmohamed, Du, Dwibedi, Egambaram, Elabd, Erez, Fang, Fantacci, Fong, Frey, Fu, Gao, Giustina, Gopalakrishnan, Graesser, Groth, Gupta, Hafner, Hansen, Hasenclever, Haves, Heess, Hernaez, Hofer, Hsu, Huang, Huang, Iscen, Jacob, Jain, Jesmonth, Jindal, Julian, Kalashnikov, Karagozler, Karp, Kecman, Kew, Kim, Kim, Kim, Kipf, Kirmani, Konyushkova, Ku, Kuang, Lampe, Laurens, Le, Leal, Lee, Lee, Lever, Liang, Lin, Liu, Long, Lu, Maddineni, Majumdar, Maninis, Marmon, Martinez, Michaely, Milonopoulos, Moore, Moreno, Neunert, Nori, Ortiz, Oslund, Parada, Parisotto, Paryag, Pooley, Power, Quaglino, Qureshi, Raju, Ran, Rao, Rao, Reid, Rendleman, Reymann, Rivas,
  Romano, Rubanova, Sampedro, Sanketi, Shah, Sharma, Shea, Shridhar, Shu, Sindhwani, Singh, Soricut, Sterneck, Storz, Surdulescu, Tan, Tompson, Tunyasuvunakool, Varley, Vesom, Vezzani, Villalonga, Vinyals, Wagner, Wahid, Welker, Wohlhart, Wu, Wulfmeier, Xia, Xiao, Xie, Xie, Xu, Xu, Xu, Xu, Yan, Yang, Yang, Yang, Yu, Yu, Yuan, Yuan, Zhang, Zhang, Zhang, Zhou, Zhou and Zhou}]{geminiroboticsteam2025geminirobotics15}
Team GR, Abdolmaleki A, Abeyruwan S, Ainslie J, Alayrac JB, Gonzalez~Arenas M, Balakrishna A, Batchelor N, Bewley A, Bingham J, Bloesch M, Bousmalis K, Brakel P, Brohan A, Buschmann T, Byravan A, Cabi S, Caluwaerts K, Casarini F, Chan C, Chang O, Chappellet-Volpini L, Chen JE, Chen X, Chiang HTL, Choromanski K, Collister A, D'Ambrosio DB, Dasari S, Davchev T, Dave MK, Devin C, Di~Palo N, Ding T, Doersch C, Dostmohamed A, Du Y, Dwibedi D, Egambaram ST, Elabd M, Erez T, Fang X, Fantacci C, Fong C, Frey E, Fu C, Gao R, Giustina M, Gopalakrishnan K, Graesser L, Groth O, Gupta A, Hafner R, Hansen S, Hasenclever L, Haves S, Heess N, Hernaez B, Hofer A, Hsu J, Huang L, Huang SH, Iscen A, Jacob MG, Jain D, Jesmonth S, Jindal A, Julian R, Kalashnikov D, Karagozler ME, Karp S, Kecman M, Kew JC, Kim D, Kim F, Kim J, Kipf T, Kirmani S, Konyushkova K, Ku LY, Kuang Y, Lampe T, Laurens A, Le TA, Leal I, Lee AX, Lee TWE, Lever G, Liang J, Lin LH, Liu F, Long S, Lu C, Maddineni S, Majumdar A, Maninis KK, Marmon A, Martinez S,
  Michaely AH, Milonopoulos N, Moore J, Moreno R, Neunert M, Nori F, Ortiz J, Oslund K, Parada C, Parisotto E, Paryag A, Pooley A, Power T, Quaglino A, Qureshi H, Raju RV, Ran H, Rao D, Rao K, Reid I, Rendleman D, Reymann K, Rivas M, Romano F, Rubanova Y, Sampedro PP, Sanketi PR, Shah D, Sharma M, Shea K, Shridhar M, Shu C, Sindhwani V, Singh S, Soricut R, Sterneck R, Storz I, Surdulescu R, Tan J, Tompson J, Tunyasuvunakool S, Varley J, Vesom G, Vezzani G, Villalonga MB, Vinyals O, Wagner R, Wahid A, Welker S, Wohlhart P, Wu C, Wulfmeier M, Xia F, Xiao T, Xie A, Xie J, Xu P, Xu S, Xu Y, Xu Z, Yan J, Yang S, Yang S, Yang Y, Yu HH, Yu W, Yuan W, Yuan Y, Zhang J, Zhang T, Zhang Z, Zhou A, Zhou G and Zhou Y (2025) {Gemini Robotics 1.5: Pushing the Frontier of Generalist Robots with Advanced Embodied Reasoning, Thinking, and Motion Transfer}.
\newblock \emph{arXiv preprint arXiv:2510.03342} \urlprefix\url{https://arxiv.org/abs/2510.03342}.

\bibitem[{Team et~al.(2025{\natexlab{bb}})Team, Abeyruwan, Ainslie, Alayrac, Gonzalez~Arenas, Armstrong, Balakrishna, Baruch, Bauza, Blokzijl, Bohez, Bousmalis, Brohan, Buschmann, Byravan, Cabi, Caluwaerts, Casarini, Chang, Chen, Chen, Chiang, Choromanski, D'Ambrosio, Dasari, Davchev, Devin, Di~Palo, Ding, Dostmohamed, Driess, Du, Dwibedi, Elabd, Fantacci, Fong, Frey, Fu, Giustina, Gopalakrishnan, Graesser, Hasenclever, Heess, Hernaez, Herzog, Hofer, Humplik, Iscen, Jacob, Jain, Julian, Kalashnikov, Karagozler, Karp, Kew, Kirkland, Kirmani, Kuang, Lampe, Laurens, Leal, Lee, Lee, Liang, Lin, Maddineni, Majumdar, Michaely, Moreno, Neunert, Nori, Parada, Parisotto, Pastor, Pooley, Rao, Reymann, Sadigh, Saliceti, Sanketi, Sermanet, Shah, Sharma, Shea, Shu, Sindhwani, Singh, Soricut, Springenberg, Sterneck, Surdulescu, Tan, Tompson, Vanhoucke, Varley, Vesom, Vezzani, Vinyals, Wahid, Welker, Wohlhart, Xia, Xiao, Xie, Xie, Xu, Xu, Xu, Xu, Yang, Yao, Yaroshenko, Yu, Yuan, Zhang, Zhang, Zhou and Zhou}]{audit250320020}
Team GR, Abeyruwan S, Ainslie J, Alayrac JB, Gonzalez~Arenas M, Armstrong T, Balakrishna A, Baruch R, Bauza M, Blokzijl M, Bohez S, Bousmalis K, Brohan A, Buschmann T, Byravan A, Cabi S, Caluwaerts K, Casarini F, Chang O, Chen JE, Chen X, Chiang HTL, Choromanski K, D'Ambrosio D, Dasari S, Davchev T, Devin C, Di~Palo N, Ding T, Dostmohamed A, Driess D, Du Y, Dwibedi D, Elabd M, Fantacci C, Fong C, Frey E, Fu C, Giustina M, Gopalakrishnan K, Graesser L, Hasenclever L, Heess N, Hernaez B, Herzog A, Hofer RA, Humplik J, Iscen A, Jacob MG, Jain D, Julian R, Kalashnikov D, Karagozler ME, Karp S, Kew C, Kirkland J, Kirmani S, Kuang Y, Lampe T, Laurens A, Leal I, Lee AX, Lee TWE, Liang J, Lin Y, Maddineni S, Majumdar A, Michaely AH, Moreno R, Neunert M, Nori F, Parada C, Parisotto E, Pastor P, Pooley A, Rao K, Reymann K, Sadigh D, Saliceti S, Sanketi P, Sermanet P, Shah D, Sharma M, Shea K, Shu C, Sindhwani V, Singh S, Soricut R, Springenberg JT, Sterneck R, Surdulescu R, Tan J, Tompson J, Vanhoucke V, Varley J,
  Vesom G, Vezzani G, Vinyals O, Wahid A, Welker S, Wohlhart P, Xia F, Xiao T, Xie A, Xie J, Xu P, Xu S, Xu Y, Xu Z, Yang Y, Yao R, Yaroshenko S, Yu W, Yuan W, Zhang J, Zhang T, Zhou A and Zhou Y (2025{\natexlab{bb}}) {Gemini Robotics: Bringing AI into the Physical World}.
\newblock arXiv preprint arXiv:2503.20020.
\newblock \urlprefix\url{https://arxiv.org/abs/2503.20020}.

\bibitem[{Tedrake and {The Drake Development Team}(2019)}]{drake2019}
Tedrake R and {The Drake Development Team} (2019) {Drake: Model-Based Design and Verification for Robotics}.
\newblock \urlprefix\url{https://drake.mit.edu}.

\bibitem[{{Tencent Robotics X Team and Futian Laboratory, Shenzhen}(2025)}]{tencentroboticsxte2025tairos}
{Tencent Robotics X Team and Futian Laboratory, Shenzhen} (2025) {TAIROS: An Embodied AI Platform for Robotics Applications}.
\newblock \urlprefix\url{https://cdn.tairos.tencent-cloud.com/downloads/tools/pdf/tairos.pdf}.

\bibitem[{Teo(2026)}]{teo2026heuristic}
Teo EH (2026) {From False Positives to Failure Topologies: Heuristic Learning for Auditable Dexterous Grasping Development}.
\newblock Zenodo preprint.
\newblock \doi{10.5281/zenodo.20922238}.
\newblock \urlprefix\url{https://doi.org/10.5281/zenodo.20922238}.

\bibitem[{Tessler et~al.(2026)Tessler, Jiang, Otto, Shi, Iqbal, Chechik and Peng}]{tessler2026metatron}
Tessler C, Jiang Y, Otto A, Shi Y, Iqbal U, Chechik G and Peng XB (2026) {METATRON: A Promptable Whole-Body Execution Layer for Humanoids}.
\newblock \urlprefix\url{https://research.nvidia.com/labs/par/projects/metatron/assets/metatron-paper.pdf}.

\bibitem[{{Text2MuJoCo contributors}(2026)}]{text2mujoco2026}
{Text2MuJoCo contributors} (2026) {Text2MuJoCo}.
\newblock Software repository.
\newblock \urlprefix\url{https://github.com/ShawnJoeng/Text2Mujoco}.
\newblock Version d76a2a6.

\bibitem[{{ThinkFlowLab}(2026)}]{system1agents2026}
{ThinkFlowLab} (2026) {System1-Agents}.
\newblock GitHub repository.
\newblock \urlprefix\url{https://github.com/ThinkFlowLab/system1-agents}.

\bibitem[{Tian et~al.(2026)Tian, Liu, Xu, Lu, Guan, Han, Yang, Gu, Qian, Zhang, Zhu, Ye, Yang, Wang, Hu, Li, Zhang, Su, Ji, Wang, Feng, Liu, Zhang, Wu, Tang, Gu, Cai, Zheng, Ma, Yin, Chu, Tang and Xu}]{tian2026abotagentos}
Tian J, Liu S, Xu Y, Lu J, Guan Z, Han H, Yang D, Gu M, Qian Y, Zhang T, Zhu Y, Ye Z, Yang M, Wang F, Hu X, Li X, Zhang W, Su S, Ji Y, Wang J, Feng Z, Liu J, Zhang Z, Wu X, Tang Z, Gu Z, Cai Y, Zheng L, Ma J, Yin M, Chu Z, Tang W and Xu M (2026) {ABot-AgentOS: A General Robotic Agent OS with Lifelong Multi-modal Memory}.
\newblock \urlprefix\url{https://arxiv.org/abs/2607.10350}.

\bibitem[{Tian et~al.(2026{\natexlab{bb}})Tian, Zhang, Liu, Liu, Liu and Wang}]{tian2026physevo}
Tian W, Zhang Z, Liu Z, Liu F, Liu Q and Wang L (2026{\natexlab{bb}}) {PhysEvo: Astra Can Act, Let It}.
\newblock \emph{arXiv preprint arXiv:2610.08995} \urlprefix\url{https://arxiv.org/abs/2610.08995}.

\bibitem[{Tian et~al.(2025)Tian, Gu, Li, Liu, Wang, Zhao, Zhan, Jia, Lang and Zhao}]{tian2024drivevlmdual}
Tian X, Gu J, Li B, Liu Y, Wang Y, Zhao Z, Zhan K, Jia P, Lang X and Zhao H (2025) Drivevlm: The convergence of autonomous driving and large vision-language models.
\newblock In: \emph{Proceedings of The 8th Conference on Robot Learning}, \emph{Proceedings of Machine Learning Research}, volume 270. PMLR, pp. 4698--4726.
\newblock \doi{10.48550/arXiv.2402.12289}.
\newblock \urlprefix\url{https://proceedings.mlr.press/v270/tian25c.html}.

\bibitem[{Tomar(2026)}]{tomar2026askable}
Tomar T (2026) {jev-askable-arm}.
\newblock Software repository.
\newblock \urlprefix\url{https://github.com/TarunTomar122/jev-askable-arm}.
\newblock Accessed: 2026-10-25.

\bibitem[{Torne et~al.(2026)Torne, Pertsch, Walke, Vedder, Nair, Ichter, Ren, Wang, Tang, Stachowicz, Dhabalia, Equi, Vuong, Springenberg, Levine, Finn and Driess}]{torne2026mem}
Torne M, Pertsch K, Walke H, Vedder K, Nair S, Ichter B, Ren AZ, Wang H, Tang J, Stachowicz K, Dhabalia K, Equi M, Vuong Q, Springenberg JT, Levine S, Finn C and Driess D (2026) {MEM: Multi-Scale Embodied Memory for Vision Language Action Models}.
\newblock \doi{10.48550/arXiv.2603.03596}.
\newblock \urlprefix\url{https://arxiv.org/abs/2603.03596}.

\bibitem[{Torshizi et~al.(2026)Torshizi, Singh, Mathur, Habib, Du and Tokekar}]{step2026}
Torshizi K, Singh A, Mathur S, Habib K, Du L and Tokekar P (2026) {STeP: Signal Temporal Logic for Precise Specifications for Action Generation with Vision Language Models}.
\newblock arXiv:2607.18580.
\newblock \urlprefix\url{https://arxiv.org/abs/2607.18580}.

\bibitem[{Tsui et~al.(2026)Tsui, Fang and Hwu}]{audit260120334}
Tsui BY, Fang AY and Hwu TJ (2026) {Demonstration-Free Robotic Control via LLM Agents}.
\newblock arXiv:2601.20334.
\newblock \urlprefix\url{https://arxiv.org/abs/2601.20334v1}.
\newblock Accepted to IROS 2026.

\bibitem[{Turcato et~al.(2025)Turcato, Iovino, Synodinos, Dalla~Libera, Carli and Falco}]{turcato2025archie}
Turcato N, Iovino M, Synodinos A, Dalla~Libera A, Carli R and Falco P (2025) {Towards Autonomous Reinforcement Learning for Real-World Robotic Manipulation with Large Language Models}.
\newblock \doi{10.48550/arXiv.2503.04280}.
\newblock \urlprefix\url{https://arxiv.org/abs/2503.04280}.

\bibitem[{Tziafas and Kasaei(2024)}]{self240618746}
Tziafas G and Kasaei H (2024) {Lifelong Robot Library Learning: Bootstrapping Composable and Generalizable Skills for Embodied Control with Language Models}.
\newblock In: \emph{2024 IEEE International Conference on Robotics and Automation (ICRA)}. pp. 515--522.
\newblock \doi{10.1109/icra57147.2024.10611448}.
\newblock \urlprefix\url{https://doi.org/10.1109/icra57147.2024.10611448}.

\bibitem[{Uysal et~al.(2026)Uysal, Kehrer and Panichella}]{audit260600104}
Uysal E, Kehrer T and Panichella S (2026) {PEACE: A Planner-Executor Agent with Constraint Enforcement for UAVs}.
\newblock arXiv:2606.00104.
\newblock \urlprefix\url{https://arxiv.org/abs/2606.00104v1}.
\newblock Accepted to ICRA 2026 Workshop on Semantics for Reliable Robot Autonomy.

\bibitem[{Vemprala et~al.(2024)Vemprala, Bonatti, Bucker and Kapoor}]{vemprala2024chatgpt}
Vemprala SH, Bonatti R, Bucker A and Kapoor A (2024) {ChatGPT for Robotics: Design Principles and Model Abilities}.
\newblock \emph{IEEE Access} 12: 55682--55696.
\newblock \doi{10.1109/ACCESS.2024.3387941}.
\newblock \urlprefix\url{https://doi.org/10.1109/ACCESS.2024.3387941}.

\bibitem[{Venuto et~al.(2024)Venuto, Islam, Klissarov, Precup, Yang and Anand}]{venuto2024codeasreward}
Venuto D, Islam MSN, Klissarov M, Precup D, Yang S and Anand A (2024) {Code as Reward: Empowering Reinforcement Learning with VLMs}.
\newblock In: \emph{Proceedings of the 41st International Conference on Machine Learning}, \emph{Proceedings of Machine Learning Research}, volume 235. pp. 49368--49387.
\newblock \urlprefix\url{https://proceedings.mlr.press/v235/venuto24a.html}.

\bibitem[{Vu et~al.(2026)Vu, Yao, Liu, Aryan and Aralikatti}]{audit260912285}
Vu LG, Yao C, Liu Y, Aryan F and Aralikatti RC (2026) {AnchorVLN: Geometry-Anchored Vision-Language Grounding Reasoning for Open-Vocabulary Navigation}.
\newblock arXiv:2609.12285.
\newblock \urlprefix\url{https://arxiv.org/abs/2609.12285v1}.

\bibitem[{{Waddle Team}(2026)}]{waddle2026introducing}
{Waddle Team} (2026) {Introducing Waddle: Agents that Control Robots}.
\newblock Waddle Labs Blog.
\newblock \urlprefix\url{https://waddlelabs.ai/research/introducing-waddle}.

\bibitem[{Wake and Wagle(2026)}]{wake2026sharedkvbt}
Wake N and Wagle J (2026) {SharedKV-BT: Node-Local Typed Decisions for Behavior-Tree Agents}.
\newblock \doi{10.48550/arXiv.2610.07327}.
\newblock \urlprefix\url{https://arxiv.org/abs/2610.07327}.

\bibitem[{Wan et~al.(2026)Wan, Yue, Liu, Chu, Wang, Chen, Jiang, Zhu, Dong and Zhu}]{wan2026nofreechecker}
Wan Y, Yue X, Liu Z, Chu Z, Wang S, Chen Y, Jiang X, Zhu X, Dong Y and Zhu L (2026) {No Free Checker: A Survey of Verifiers for Robot Policies}.
\newblock \emph{arXiv preprint arXiv:2609.09250} \urlprefix\url{https://arxiv.org/abs/2609.09250}.

\bibitem[{Wang et~al.(2026)Wang, Cui, Zhang, Zheng, Guo and Yu}]{audit260120577}
Wang B, Cui H, Zhang B, Zheng X, Guo B and Yu Z (2026) {MeCo: Enhancing LLM-Empowered Multi-Robot Collaboration via Similar Task Memoization}.
\newblock arXiv:2601.20577.
\newblock \urlprefix\url{https://arxiv.org/abs/2601.20577}.

\bibitem[{Wang et~al.(2026{\natexlab{bb}})Wang, Zhang, Yang, Gu, Wang, Wang, Ze, Chang, Zhao, Shao, Huang, Liu, Zhang, Xue, Guan, Pu, Guo, Wang and Zhu}]{wang2026zero2skill}
Wang B, Zhang Z, Yang Z, Gu P, Wang S, Wang X, Ze X, Chang Y, Zhao G, Shao J, Huang G, Liu H, Zhang Y, Xue W, Guan C, Pu C, Guo Y, Wang X and Zhu Z (2026{\natexlab{bb}}) {Zero2Skill: Bootstrapping Robot Skills through Autonomous Data Collection, Training, and Deployment}.
\newblock \emph{arXiv preprint arXiv:2607.14047} \urlprefix\url{https://arxiv.org/abs/2607.14047}.

\bibitem[{Wang et~al.(2026{\natexlab{bc}})Wang, Ye, Ying, Jing, Ma, Chen, Wu, Wysocki, Duan and Sheil}]{audit260905985}
Wang G, Ye P, Ying Q, Jing Y, Ma Y, Chen H, Wu H, Wysocki O, Duan M and Sheil B (2026{\natexlab{bc}}) {A Brain-inspired Hierarchical Framework for Zero-Shot Robot Task Reasoning and Execution}.
\newblock arXiv:2609.05985.
\newblock \urlprefix\url{https://arxiv.org/abs/2609.05985v1}.

\bibitem[{Wang et~al.(2023)Wang, Gonzalez-Pumariega, Sharma and Choudhury}]{audit230516744}
Wang H, Gonzalez-Pumariega G, Sharma Y and Choudhury S (2023) {Demo2Code: From Summarizing Demonstrations to Synthesizing Code via Extended Chain-of-Thought}.
\newblock \emph{arXiv preprint arXiv:2305.16744} \doi{10.48550/arXiv.2305.16744}.
\newblock \urlprefix\url{https://arxiv.org/abs/2305.16744}.

\bibitem[{Wang et~al.(2024)Wang, Kedia, Ren, Abdullah, Bhardwaj, Chao, Chen, Chin, Dan, Fan, Gonzalez-Pumariega, Kompella, Pace, Sharma, Sun, Sunkara and Choudhury}]{audit240218796}
Wang H, Kedia K, Ren J, Abdullah R, Bhardwaj A, Chao A, Chen KY, Chin N, Dan P, Fan X, Gonzalez-Pumariega G, Kompella A, Pace MA, Sharma Y, Sun X, Sunkara N and Choudhury S (2024) {MOSAIC: Modular Foundation Models for Assistive and Interactive Cooking}.
\newblock arXiv:2402.18796.
\newblock \urlprefix\url{https://arxiv.org/abs/2402.18796}.
\newblock Accepted to CoRL 2024.

\bibitem[{Wang et~al.(2026{\natexlab{bd}})Wang, Wang, Anilkumar, Zou, Serban and Negrut}]{sref260514398}
Wang H, Wang J, Anilkumar A, Zou B, Serban R and Negrut D (2026{\natexlab{bd}}) {ChronoAgentic: A Code-based Multi-Agent World Simulator for Physically Grounded Simulation Construction}.
\newblock \urlprefix\url{https://arxiv.org/abs/2605.14398}.

\bibitem[{Wang(2026)}]{si260931760}
Wang J (2026) {What Stops Recursive Self-Improvement in Robotics? Lessons from 123 Rounds of Agentic Skill Discovery}.
\newblock \urlprefix\url{https://arxiv.org/abs/2609.31760}.

\bibitem[{Wang et~al.(2026{\natexlab{be}})Wang, Jiang, Shen and Soh}]{wang2026agenticroboticssurvey}
Wang J, Jiang Y, Shen Z and Soh H (2026{\natexlab{be}}) {A Survey of Agentic Robotics: Toward Continual Self-Improvement}.
\newblock Preprints.org.
\newblock \doi{10.20944/preprints202610.2747.v1}.
\newblock \urlprefix\url{https://www.preprints.org/manuscript/202610.2747/v1}.
\newblock Version 1, October 2026.

\bibitem[{Wang et~al.(2026{\natexlab{bf}})Wang, Pechuk, Xu, Wu, Kim, Millan and Erickson}]{wang2026genphri}
Wang J, Pechuk N, Xu X, Wu T, Kim N, Millan J and Erickson Z (2026{\natexlab{bf}}) {GenPHRI: Agentic Generative Simulation for Physical Human-Robot Interaction}.
\newblock \emph{arXiv preprint arXiv:2604.08664} \urlprefix\url{https://arxiv.org/abs/2604.08664}.

\bibitem[{Wang et~al.(2026{\natexlab{bg}})Wang, Yu, Zhu, Yang and Chen}]{audit261012457}
Wang J, Yu Y, Zhu B, Yang Z and Chen J (2026{\natexlab{bg}}) {SpatialHarness: Test-Time Spatial Scaffolding for Fine Robotic Manipulation}.
\newblock arXiv:2610.12457.
\newblock \urlprefix\url{https://arxiv.org/abs/2610.12457v1}.

\bibitem[{Wang et~al.(2026{\natexlab{bh}})Wang, Zhao, He, Peng and Zhang}]{si260413533}
Wang J, Zhao B, He Y, Peng J and Zhang X (2026{\natexlab{bh}}) {Evolvable Embodied Agent for Robotic Manipulation via Long Short-Term Reflection and Optimization}.
\newblock \urlprefix\url{https://arxiv.org/abs/2604.13533}.
\newblock Accepted to IJCNN 2026.

\bibitem[{Wang et~al.(2026{\natexlab{bi}})Wang, Jie, Yan, Heng and Li}]{si260927612}
Wang K, Jie H, Yan Y, Heng Z and Li Z (2026{\natexlab{bi}}) {RegenHarness: A Robot Agent Harness with Evidence-Gated Recursive Self-Improvement}.
\newblock \urlprefix\url{https://arxiv.org/abs/2609.27612}.

\bibitem[{Wang et~al.(2025)Wang, Lu, Liu, Jiang, Li, Zhang, Zheng, Yu, Chen and Shen}]{wang2025odyssey}
Wang K, Lu L, Liu M, Jiang J, Li Z, Zhang B, Zheng W, Yu X, Chen H and Shen C (2025) {ODYSSEY: Open-World Quadrupeds Exploration and Manipulation for Long-Horizon Tasks}.
\newblock \doi{10.48550/arXiv.2508.08240}.
\newblock \urlprefix\url{https://arxiv.org/abs/2508.08240}.

\bibitem[{Wang et~al.(2026{\natexlab{bj}})Wang, Wang, Chen, Zhang, Gui, Wang, Chen, Xiao, Huang, Zhang, Li, Wang, Kim, Sun, Deng, Hou, Zhu, Liu, Xing, Yao, Fan, Liu, Song, Abbeel, Li, Wu, Zhu, Shi, Zhang and Li}]{embodiedfirstexam2026}
Wang K, Wang Q, Chen C, Zhang P, Gui C, Wang F, Chen C, Xiao W, Huang X, Zhang T, Li L, Wang Z, Kim M, Sun Y, Deng M, Hou J, Zhu Q, Liu H, Xing EP, Yao H, Fan Z, Liu Y, Song D, Abbeel P, Li FF, Wu J, Zhu Y, Shi G, Zhang R and Li M (2026{\natexlab{bj}}) {Embodied First Exam}.
\newblock \emph{arXiv preprint arXiv:2611.12177} \doi{10.48550/arXiv.2611.12177}.
\newblock \urlprefix\url{https://arxiv.org/abs/2611.12177}.
\newblock Companion failure analysis: \url{https://www.embodied-first-exam.ai/blogs/failure-analysis/}.

\bibitem[{Wang et~al.(2024{\natexlab{bb}})Wang, Ling, Yuan, Shridhar, Bao, Qin, Wang, Xu and Wang}]{gensim2023}
Wang L, Ling Y, Yuan Z, Shridhar M, Bao C, Qin Y, Wang B, Xu H and Wang X (2024{\natexlab{bb}}) {GenSim: Generating Robotic Simulation Tasks via Large Language Models}.
\newblock In: \emph{International Conference on Learning Representations}.
\newblock \urlprefix\url{https://arxiv.org/abs/2310.01361}.

\bibitem[{Wang et~al.(2025{\natexlab{bb}})Wang, Ying, Yang, Zou, Yin, Li, Yang, Yang, Liu and Liu}]{sref251221220}
Wang L, Ying Z, Yang X, Zou Q, Yin Z, Li T, Yang J, Yang Y, Liu A and Liu X (2025{\natexlab{bb}}) {RoboSafe: Safeguarding Embodied Agents via Executable Safety Logic}.
\newblock \urlprefix\url{https://arxiv.org/abs/2512.21220}.

\bibitem[{Wang et~al.(2026{\natexlab{bk}})Wang, Osterberg, Tian, Shorinwa, Bowkett, Pastor, Wu and Schwager}]{wang2026insight}
Wang M, Osterberg L, Tian S, Shorinwa O, Bowkett J, Pastor D, Wu J and Schwager M (2026{\natexlab{bk}}) {InSight: Self-Guided Skill Acquisition via Steerable VLAs}.
\newblock \emph{arXiv preprint arXiv:2606.24884} \urlprefix\url{https://arxiv.org/abs/2606.24884}.

\bibitem[{Wang and Srivastava(2026)}]{wang2026homememory}
Wang O and Srivastava M (2026) {How Should a Home Robot Update Its Memories When Human Routines Change?}
\newblock CoRL 2026 Workshop on Memory for Robotics.
\newblock \urlprefix\url{https://corl2026-memory.github.io/papers/046.pdf}.

\bibitem[{Wang et~al.(2026{\natexlab{bl}})Wang, Chang, Ma, Yang, Feng, Liu, Yang, Cao, Wang, Su and Wang}]{wang2026adarsi}
Wang P, Chang Y, Ma Z, Yang X, Feng S, Liu D, Yang Y, Cao X, Wang C, Su J and Wang D (2026{\natexlab{bl}}) {AdaRSI: On-Demand Recursive Self-Improvement for Embodied Agents}.
\newblock \emph{arXiv preprint arXiv:2610.20474} \urlprefix\url{https://arxiv.org/abs/2610.20474}.

\bibitem[{Wang et~al.(2026{\natexlab{bm}})Wang, Hu, Liu, Yang, Wang and Wen}]{wang2026atlasva}
Wang P, Hu Y, Liu X, Yang J, Wang H and Wen Z (2026{\natexlab{bm}}) {AtlasVA: Self-Evolving Visual Skill Memory for Teacher-Free VLM Agents}.
\newblock \doi{10.48550/arXiv.2605.17933}.
\newblock \urlprefix\url{https://arxiv.org/abs/2605.17933}.

\bibitem[{Wang et~al.(2026{\natexlab{bn}})Wang, Ma, Chang, Luo, Zhang, Wang, Yang, Feng, Yang and Li}]{grid62}
Wang P, Ma Z, Chang Y, Luo X, Zhang Y, Wang Z, Yang X, Feng S, Yang Y and Li D (2026{\natexlab{bn}}) {Self-Evolving Embodied Agents via Skill-Harness Evolution}.
\newblock arXiv:2608.11350.
\newblock \urlprefix\url{https://arxiv.org/abs/2608.11350}.

\bibitem[{Wang et~al.(2026{\natexlab{bo}})Wang, Bi, Jiang, Si, Huang, Chen, Li, Li, Li and Zhou}]{wang2026pyruareal}
Wang Q, Bi J, Jiang S, Si R, Huang W, Chen K, Li J, Li L, Li X and Zhou D (2026{\natexlab{bo}}) {PyRUA-REAL: Robot Exploration and Adaptive Learning through RSI in the Real World}.
\newblock \urlprefix\url{https://arxiv.org/abs/2611.06768}.

\bibitem[{Wang et~al.(2026{\natexlab{bp}})Wang, Wang, Li, Ban, Chen, Ge, Qin, Li and Zhu}]{thea2026}
Wang Q, Wang T, Li C, Ban S, Chen Y, Ge Y, Qin J, Li C and Zhu W (2026{\natexlab{bp}}) {Towards the Harness of Embodied Agents}.
\newblock \emph{arXiv preprint arXiv:2608.11246} \urlprefix\url{https://arxiv.org/abs/2608.11246}.

\bibitem[{Wang et~al.(2026{\natexlab{bq}})Wang, Hsu, Mehta, Kim, Dou and Pajic}]{wang2026gavel}
Wang R, Hsu HL, Mehta S, Kim J, Dou Z and Pajic M (2026{\natexlab{bq}}) {GAVEL: Graph World Models for Verified and Efficient Long-Horizon LLM Task Planning}.
\newblock \urlprefix\url{https://arxiv.org/abs/2609.19315}.

\bibitem[{Wang et~al.(2026{\natexlab{br}})Wang, Yao, Zhou, Deng, Su and Jia}]{wang2026demofoundry}
Wang R, Yao X, Zhou X, Deng Y, Su Y and Jia K (2026{\natexlab{br}}) {DemoFoundry: Toward Scalable Photorealistic Robot Data with Coding Agents and Generative Rendering}.
\newblock \emph{arXiv preprint arXiv:2611.34671} \urlprefix\url{https://arxiv.org/abs/2611.34671}.

\bibitem[{Wang et~al.(2024{\natexlab{bc}})Wang, Han, Jiao, Zhang, Wu, Zhu and Liu}]{grid50}
Wang S, Han M, Jiao Z, Zhang Z, Wu YN, Zhu SC and Liu H (2024{\natexlab{bc}}) {LLM$^3$: Large Language Model-based Task and Motion Planning with Motion Failure Reasoning}.
\newblock In: \emph{2024 IEEE/RSJ International Conference on Intelligent Robots and Systems (IROS)}. pp. 12086--12092.
\newblock \doi{10.1109/iros58592.2024.10801328}.
\newblock \urlprefix\url{https://doi.org/10.1109/iros58592.2024.10801328}.

\bibitem[{Wang et~al.(2025{\natexlab{bc}})Wang, Zhao, Cai and Sun}]{wang2025imaginenav}
Wang T, Zhao X, Cai W and Sun C (2025{\natexlab{bc}}) {ImagineNav++: Prompting Vision-Language Models as Embodied Navigator through Scene Imagination}.
\newblock \doi{10.48550/arXiv.2512.17435}.
\newblock \urlprefix\url{https://arxiv.org/abs/2512.17435}.

\bibitem[{Wang et~al.(2026{\natexlab{bs}})Wang, Li, Lei, Martinez, Gao, Jiang, Jiang, Wu, Terzopoulos and Gong}]{wang2026sunagenticrobotpolicy}
Wang W, Li Z, Lei Y, Martinez D, Gao X, Jiang Y, Jiang C, Wu Y, Terzopoulos D and Gong R (2026{\natexlab{bs}}) Sun: Agentic robot policy learning with persistent task programs.
\newblock \urlprefix\url{https://arxiv.org/abs/2608.31167}.

\bibitem[{Wang et~al.(2025{\natexlab{bd}})Wang, Dong, Rangasrinivasan, Nwogu, Setlur and Govindaraju}]{xref250306791}
Wang X, Dong L, Rangasrinivasan S, Nwogu I, Setlur S and Govindaraju V (2025{\natexlab{bd}}) {AutoMisty: A Multi-Agent LLM Framework for Automated Code Generation in the Misty Social Robot}.
\newblock \emph{arXiv preprint arXiv:2503.06791} \urlprefix\url{https://arxiv.org/abs/2503.06791}.
\newblock Accepted to IROS 2025.

\bibitem[{Wang et~al.(2026{\natexlab{bt}})Wang, Li, Wang, Zhang, Cui, Tang, Buch and Xu}]{wang2026sear}
Wang X, Li B, Wang H, Zhang L, Cui AY, Tang Z, Buch S and Xu D (2026{\natexlab{bt}}) {SEAR: Self-Evolving Agentic Robotics}.
\newblock \urlprefix\url{https://arxiv.org/abs/2610.18792}.

\bibitem[{Wang et~al.(2026{\natexlab{bu}})Wang, Liu, Ding, Choi, Huang, Zhao, Li, Jiang, Yu, Liu, Xu and Su}]{wang2026embodiedgenv2}
Wang X, Liu L, Ding T, Choi A, Huang C, Zhao M, Li Z, Jiang J, Yu C, Liu S, Xu W and Su Z (2026{\natexlab{bu}}) {EmbodiedGen V2: An Agentic, Simulation-Ready 3D World Engine for Embodied AI}.
\newblock \emph{arXiv preprint arXiv:2607.07459} \urlprefix\url{https://arxiv.org/abs/2607.07459}.

\bibitem[{Wang et~al.(2026{\natexlab{bv}})Wang, Wu, Zhang, Wang, Shao, Luan, Li, Li, Gu, Zhou, Shi, Ni, Lu, Meng, Li, Wu, Zhao, Guo, Li, Tang, Zhang, Xie and Qu}]{si260929166}
Wang X, Wu W, Zhang M, Wang Z, Shao K, Luan J, Li Y, Li Q, Gu S, Zhou H, Shi S, Ni F, Lu S, Meng W, Li K, Wu J, Zhao K, Guo S, Li G, Tang Y, Zhang Z, Xie Y and Qu H (2026{\natexlab{bv}}) {HarnessPAI: An Evolving Harness for Physical AI}.
\newblock \urlprefix\url{https://arxiv.org/abs/2609.29166}.

\bibitem[{Wang(2026{\natexlab{cb}})}]{wang2026vlaweightediting}
Wang Y (2026{\natexlab{cb}}) {How Far Can an LLM Go by Editing VLA Weights?}
\newblock Technical blog.
\newblock \urlprefix\url{https://no-518.github.io/no-518-blog/blog/vla-weight-space-programming/}.

\bibitem[{Wang et~al.(2026{\natexlab{bw}})Wang, Huai, Zhang, Wu, Chen, Zhang, Li, Gong, Lin, Chen and Qiu}]{wang2026tactileeta}
Wang Y, Huai Z, Zhang Y, Wu J, Chen Y, Zhang H, Li S, Gong J, Lin Z, Chen Z and Qiu X (2026{\natexlab{bw}}) {Embodied Task Agents Can Perform Fine Manipulation through Tactile In-Context Learning}.
\newblock \urlprefix\url{https://arxiv.org/abs/2610.27613}.

\bibitem[{Wang et~al.(2023{\natexlab{bb}})Wang, Jiao, Zhan, Lang, Huang, Wang, Yang and Zhu}]{wang2023safetyawaredriving}
Wang Y, Jiao R, Zhan SS, Lang C, Huang C, Wang Z, Yang Z and Zhu Q (2023{\natexlab{bb}}) {Empowering Autonomous Driving with Large Language Models: A Safety Perspective}.
\newblock \doi{10.48550/arXiv.2312.00812}.
\newblock \urlprefix\url{https://arxiv.org/abs/2312.00812}.
\newblock Accepted to ICLR 2024 LLMAgents Workshop.

\bibitem[{Wang et~al.(2026{\natexlab{bx}})Wang, Li, Li, Wu and Huang}]{wang2026dexagent}
Wang Y, Li Y, Li FF, Wu J and Huang H (2026{\natexlab{bx}}) {DexAgent: An Agentic Human2Sim2Robot Framework for Dexterous Manipulation with Self-Evolving Tool Library}.
\newblock \emph{arXiv preprint arXiv:2609.35318} \urlprefix\url{https://arxiv.org/abs/2609.35318}.

\bibitem[{Wang et~al.(2024{\natexlab{bd}})Wang, Xian, Chen, Wang, Wang, Fragkiadaki, Erickson, Held and Gan}]{wang2023robogen}
Wang Y, Xian Z, Chen F, Wang TH, Wang Y, Fragkiadaki K, Erickson Z, Held D and Gan C (2024{\natexlab{bd}}) {R}obo{G}en: Towards unleashing infinite data for automated robot learning via generative simulation.
\newblock In: \emph{Proceedings of the 41st International Conference on Machine Learning}, \emph{Proceedings of Machine Learning Research}, volume 235. PMLR, pp. 51936--51983.
\newblock \urlprefix\url{https://proceedings.mlr.press/v235/wang24cc.html}.

\bibitem[{Wang et~al.(2026{\natexlab{by}})Wang, Xu, Li and Huang}]{si260525832}
Wang Y, Xu X, Li Y and Huang X (2026{\natexlab{by}}) {When Search Becomes Memory: Accelerating Robot Design Discovery with Self-Evolving Skills}.
\newblock \urlprefix\url{https://arxiv.org/abs/2605.25832}.
\newblock Accepted to EMNLP 2026.

\bibitem[{Wang et~al.(2026{\natexlab{bz}})Wang, Yang, Guo, Qiu, Wang, Matusik, Tenenbaum and Gan}]{physcensis2026}
Wang Y, Yang H, Guo M, Qiu X, Wang TH, Matusik W, Tenenbaum JB and Gan C (2026{\natexlab{bz}}) {PhyScensis: Physics-Augmented LLM Agents for Complex Physical Scene Arrangement}.
\newblock arXiv:2602.14968.
\newblock \urlprefix\url{https://arxiv.org/abs/2602.14968}.
\newblock Accepted to ICLR 2026.

\bibitem[{Wang et~al.(2026{\natexlab{ca}})Wang, Zhu, Hu, Zhang, Wang, Yeung, Zhao and Sun}]{wang2026kpi}
Wang Y, Zhu H, Hu X, Zhang H, Wang Z, Yeung YF, Zhao D and Sun L (2026{\natexlab{ca}}) {KPI}: A promptable kernel for physical interaction on humanoids.
\newblock \emph{arXiv preprint arXiv:2609.36151} \urlprefix\url{https://arxiv.org/abs/2609.36151}.

\bibitem[{Wang et~al.(2026{\natexlab{cb}})Wang, Jiang, Deng, Xue, Ye, Duan, Haghtalab, Sastry, Abbeel and Qi}]{rpg2026}
Wang YJ, Jiang H, Deng S, Xue H, Ye W, Duan R, Haghtalab N, Sastry SS, Abbeel P and Qi H (2026{\natexlab{cb}}) {Reconstruct, Practice, Go Real: Guided Self-Improvement for Embodied Agents}.
\newblock \emph{arXiv preprint arXiv:2610.02204} \urlprefix\url{https://arxiv.org/abs/2610.02204}.

\bibitem[{Wang et~al.(2023{\natexlab{bc}})Wang, Zhang, Chen and Sreenath}]{audit230909969}
Wang YJ, Zhang B, Chen J and Sreenath K (2023{\natexlab{bc}}) {Prompt a Robot to Walk with Large Language Models}.
\newblock \emph{arXiv preprint arXiv:2309.09969} \doi{10.48550/arXiv.2309.09969}.
\newblock \urlprefix\url{https://arxiv.org/abs/2309.09969}.

\bibitem[{Wang et~al.(2026{\natexlab{cc}})Wang, Ung, Gubarev, Tan, Srinivasa and Fox}]{roboplayground2026}
Wang YR, Ung C, Gubarev E, Tan C, Srinivasa S and Fox D (2026{\natexlab{cc}}) {RoboPlayground: Democratizing Robotic Evaluation through Structured Physical Domains}.
\newblock arXiv:2604.05226.
\newblock \urlprefix\url{https://arxiv.org/abs/2604.05226}.

\bibitem[{Wang et~al.(2026{\natexlab{cd}})Wang, Dang, Shi, Gao, Yin, Tam, Liu, Wang, Lu, Zhang, Cong and Yao}]{wang2026justrua}
Wang Z, Dang Y, Shi X, Gao M, Yin C, Tam S, Liu H, Wang Y, Lu Y, Zhang S, Cong X and Yao Y (2026{\natexlab{cd}}) {JustRUA: A Controlled Study of GPT-6 Robot-Use Agents}.
\newblock \urlprefix\url{https://arxiv.org/abs/2611.16009}.

\bibitem[{Wang et~al.(2026{\natexlab{ce}})Wang, Lin, Huang, Lin, Zhang and Wang}]{wang2026gpttactile}
Wang Z, Lin Y, Huang R, Lin L, Zhang D and Wang G (2026{\natexlab{ce}}) {GPT-6-Astra Understands Touch and Outperform Vision--Tactile--Language--Action Models}.
\newblock \doi{10.48550/arXiv.2610.28457}.
\newblock \urlprefix\url{https://arxiv.org/abs/2610.28457}.

\bibitem[{Wang et~al.(2026{\natexlab{cf}})Wang, Liu, Chen, Yang, Ye, Chen, Lu, Liu, Yu and Jia}]{wang2026carryon}
Wang Z, Liu Y, Chen Y, Yang S, Ye J, Chen P, Lu C, Liu S, Yu B and Jia J (2026{\natexlab{cf}}) {CarryOn}: A dual-loop harness for personalized and reliable robot assistance.
\newblock arXiv:2610.21453.
\newblock \urlprefix\url{https://arxiv.org/abs/2610.21453}.

\bibitem[{Wang et~al.(2024{\natexlab{be}})Wang, Shen and Stadie}]{grid52}
Wang Z, Shen R and Stadie B (2024{\natexlab{be}}) {Wonderful Team: Zero-Shot Physical Task Planning with Visual LLMs}.
\newblock arXiv:2407.19094.
\newblock \urlprefix\url{https://arxiv.org/abs/2407.19094}.

\bibitem[{Wang et~al.(2024{\natexlab{bf}})Wang, Yu, Zhao, Sun, Hou, Liang, Hu, Han and Gan}]{wang2024karma}
Wang Z, Yu B, Zhao J, Sun W, Hou S, Liang S, Hu X, Han Y and Gan Y (2024{\natexlab{bf}}) {KARMA: Augmenting Embodied AI Agents with Long-and-short Term Memory Systems}.
\newblock \urlprefix\url{https://arxiv.org/abs/2409.14908}.

\bibitem[{Wei et~al.(2026)Wei, Huang, Ma, Zhang and Cui}]{wei2026fate}
Wei B, Huang B, Ma J, Zhang Z and Cui S (2026) {FATE: Closed-Loop Feasibility-Aware Task Generation with Active Repair for Physically Grounded Robotic Curricula}.
\newblock \emph{arXiv preprint arXiv:2603.01505} \urlprefix\url{https://arxiv.org/abs/2603.01505}.

\bibitem[{Wei et~al.(2025)Wei, Wan, Peng, Yu, Yang, Feng, Cai, Zhu, Wang, Pang and Liu}]{wei2025dualvln}
Wei M, Wan C, Peng J, Yu X, Yang Y, Feng D, Cai W, Zhu C, Wang T, Pang J and Liu X (2025) {Ground Slow, Move Fast: A Dual-System Foundation Model for Generalizable Vision-and-Language Navigation}.
\newblock \emph{arXiv preprint arXiv:2512.08186} \doi{10.48550/arXiv.2512.08186}.
\newblock \urlprefix\url{https://arxiv.org/abs/2512.08186}.

\bibitem[{Wei et~al.(2026{\natexlab{cb}})Wei, Wang, Li, He, Gan, Peng, Guo, Zhou, Dong, Shen, Guo, Yu and Lu}]{wei2026roboatom}
Wei X, Wang X, Li X, He Y, Gan W, Peng J, Guo W, Zhou J, Dong J, Shen C, Guo H, Yu M and Lu T (2026{\natexlab{cb}}) {RoboAtom}: Large-scale atomic-action supervision for robot learning.
\newblock \doi{10.48550/arXiv.2610.40207}.
\newblock \urlprefix\url{https://arxiv.org/abs/2610.40207}.

\bibitem[{Wei et~al.(2026{\natexlab{cc}})Wei, Fang, Yang, Wu, Wang and Qi}]{wei2026scoper}
Wei Y, Fang Y, Yang S, Wu T, Wang T and Qi X (2026{\natexlab{cc}}) {Towards a Systematic Study of Embodied System 2 over Full Task Execution Cycles}.
\newblock \urlprefix\url{https://arxiv.org/abs/2610.17657}.

\bibitem[{Wei et~al.(2024)Wei, Wang, Lu, Xu, Liu, Zhao, Chen and Wang}]{wei2024chatsim}
Wei Y, Wang Z, Lu Y, Xu C, Liu C, Zhao H, Chen S and Wang Y (2024) {Editable Scene Simulation for Autonomous Driving via Collaborative LLM-Agents}.
\newblock \doi{10.48550/arXiv.2402.05746}.
\newblock \urlprefix\url{https://arxiv.org/abs/2402.05746}.
\newblock Accepted to CVPR 2024.

\bibitem[{Wei et~al.(2026{\natexlab{cd}})Wei, Tan, Wang and Zhou}]{wei2026scaleselection}
Wei Z, Tan F, Wang J and Zhou H (2026{\natexlab{cd}}) {Scale and Selection: What Makes Automatic Harness Evolution Work for Visual-Interface Robot Agents}.
\newblock \emph{arXiv preprint arXiv:2609.39304} \urlprefix\url{https://arxiv.org/abs/2609.39304}.

\bibitem[{Wen et~al.(2026)Wen, Qiu, Benton, Kirchner and Leike}]{wen2026automatedw2s}
Wen J, Qiu L, Benton J, Kirchner JH and Leike J (2026) {Automated Weak-to-Strong Researcher}.
\newblock Anthropic Alignment Science Blog.
\newblock \urlprefix\url{https://alignment.anthropic.com/2026/automated-w2s-researcher/}.

\bibitem[{Wen et~al.(2023)Wen, Fu, Li, Cai, Ma, Cai, Dou, Shi, He and Qiao}]{wen2023dilu}
Wen L, Fu D, Li X, Cai X, Ma T, Cai P, Dou M, Shi B, He L and Qiao Y (2023) {DiLu: A Knowledge-Driven Approach to Autonomous Driving with Large Language Models}.
\newblock \doi{10.48550/arXiv.2309.16292}.
\newblock \urlprefix\url{https://arxiv.org/abs/2309.16292}.
\newblock Accepted to ICLR 2024.

\bibitem[{Wen and Wang(2026)}]{wen2026dworld}
Wen X and Wang B (2026) {D-World: An Explicit World-Language-Action Model with Recursive Self-Improvement under Dual Control}.
\newblock \emph{arXiv preprint arXiv:2610.17114} \doi{10.48550/arXiv.2610.17114}.
\newblock \urlprefix\url{https://arxiv.org/abs/2610.17114}.

\bibitem[{Wen et~al.(2026{\natexlab{cb}})Wen, Chen, Cao, Chen, Zou, Yu, Kuang, Xue, Lv, Wen and Lu}]{noematrix2026roborsi}
Wen Z, Chen Y, Cao Y, Chen W, Zou Y, Yu W, Kuang F, Xue H, Lv J, Wen C and Lu C (2026{\natexlab{cb}}) {RoboRSI: Stable, efficient, and reusable robot self-evolution in complex real-world environments}.
\newblock \urlprefix\url{https://arxiv.org/abs/2610.12424}.

\bibitem[{Wijk et~al.(2025)Wijk, Lin, Becker, Jawhar, Parikh, Broadley, Chan, Chen, Clymer, Dhyani, Ericheva, Garcia, Goodrich, Jurkovic, Kinniment, Lajko, Nix, Koba~Sato, Saunders, Taran, West and Barnes}]{wijk2024rebench}
Wijk H, Lin TR, Becker J, Jawhar S, Parikh N, Broadley T, Chan L, Chen M, Clymer JM, Dhyani J, Ericheva E, Garcia K, Goodrich B, Jurkovic N, Kinniment M, Lajko A, Nix S, Koba~Sato LJ, Saunders W, Taran M, West B and Barnes E (2025) {RE}-bench: Evaluating frontier {AI} r\&d capabilities of language model agents against human experts.
\newblock In: \emph{Proceedings of the 42nd International Conference on Machine Learning}, \emph{Proceedings of Machine Learning Research}, volume 267. PMLR, pp. 66772--66832.
\newblock \urlprefix\url{https://proceedings.mlr.press/v267/wijk25a.html}.

\bibitem[{Woods and Seifi(2026)}]{woods2026skillcomposer}
Woods J and Seifi H (2026) {SkillComposer: Learning Reusable Skills for Natural-Language Robot Programming}.
\newblock \emph{arXiv preprint arXiv:2608.14944} \urlprefix\url{https://arxiv.org/abs/2608.14944}.

\bibitem[{Wu et~al.(2026)Wu, Han, Luo, Wang, He and Wan}]{si260414399}
Wu A, Han H, Luo X, Wang R, He S and Wan X (2026) {SpaceMind: A Modular and Self-Evolving Embodied Vision-Language Agent Framework for Autonomous On-orbit Servicing}.
\newblock \urlprefix\url{https://arxiv.org/abs/2604.14399}.

\bibitem[{Wu et~al.(2026{\natexlab{cb}})Wu, Ye, Chen, Chen, Zhou, Ji, Liu and Du}]{wu2026memoryharnesses}
Wu F, Ye W, Chen H, Chen H, Zhou J, Ji Y, Liu Y and Du Y (2026{\natexlab{cb}}) {How Hard Are Current Robot Memory Benchmarks for Agentic Harnesses?}
\newblock CoRL 2026 Workshop on Agentic Robotics.
\newblock \urlprefix\url{https://openreview.net/forum?id=z5FsJD37Dj}.

\bibitem[{Wu et~al.(2026{\natexlab{cc}})Wu, Kumar and Berenson}]{wu2026topology}
Wu H, Kumar A and Berenson D (2026{\natexlab{cc}}) {Topology-Informed Visual Prompting for Hierarchical Vision-Language-Action Control}.
\newblock CoRL 2026 Workshop on Agentic Robotics.
\newblock \urlprefix\url{https://openreview.net/forum?id=RkzWx9Sfcz}.

\bibitem[{Wu et~al.(2023)Wu, Antonova, Kan, Lepert, Zeng, Song, Bohg, Rusinkiewicz and Funkhouser}]{grid68}
Wu J, Antonova R, Kan A, Lepert M, Zeng A, Song S, Bohg J, Rusinkiewicz S and Funkhouser T (2023) {TidyBot: personalized robot assistance with large language models}.
\newblock \emph{Autonomous Robots} 47(8): 1087--1102.
\newblock \doi{10.1007/s10514-023-10139-z}.
\newblock \urlprefix\url{https://doi.org/10.1007/s10514-023-10139-z}.

\bibitem[{Wu et~al.(2026{\natexlab{cd}})Wu, Birchfield, Rabeti, Chang and Wen}]{wu2026gator}
Wu Q, Birchfield S, Rabeti H, Chang AX and Wen B (2026{\natexlab{cd}}) {GATOR: Generative and Agentic 3D Object Reconstruction From Casual Images}.
\newblock \emph{arXiv preprint arXiv:2610.11215} \urlprefix\url{https://arxiv.org/abs/2610.11215}.

\bibitem[{Wu et~al.(2024)Wu, Liu, Yin, Cheng, Zhao, Fang, Yi and Huang}]{wu2024rapiddriving}
Wu S, Liu J, Yin X, Cheng G, Zhao X, Fang M, Yi X and Huang X (2024) {Robust RL with LLM-Driven Data Synthesis and Policy Adaptation for Autonomous Driving}.
\newblock \doi{10.48550/arXiv.2410.12568}.
\newblock \urlprefix\url{https://arxiv.org/abs/2410.12568}.

\bibitem[{Wu et~al.(2026{\natexlab{ce}})Wu, An, Zhao, Guo, Xing, Ma, Zhang, Wu and Tomizuka}]{wu2026galatea}
Wu T, An B, Zhao S, Guo H, Xing W, Ma Y, Zhang K, Wu R and Tomizuka M (2026{\natexlab{ce}}) {Grounding Generated Video Plans in Simulation Towards Versatile Dexterous Controllers}.
\newblock \emph{arXiv preprint arXiv:2609.10050} \doi{10.48550/arXiv.2609.10050}.
\newblock \urlprefix\url{https://arxiv.org/abs/2609.10050}.

\bibitem[{Wu et~al.(2026{\natexlab{cf}})Wu, Liao, Ma, Zhang, Zhou, Chen, Li, Luo, Li, Wu, Xie, Zheng, Cai, Hou and Loy}]{wu2026emboharness}
Wu XM, Liao K, Ma Y, Zhang T, Zhou M, Chen N, Li Z, Luo Y, Li R, Wu Z, Xie T, Zheng X, Cai Z, Hou Z and Loy CC (2026{\natexlab{cf}}) {Embo-Harness: Towards Understanding Frontier-Model-as-Policy Harness}.
\newblock \urlprefix\url{https://arxiv.org/abs/2611.31201}.

\bibitem[{Wu et~al.(2025)Wu, Xiong, Hu, Iyengar, Jiang, Bera, Tan and Jagannathan}]{selp2024}
Wu Y, Xiong Z, Hu Y, Iyengar SS, Jiang N, Bera A, Tan L and Jagannathan S (2025) {SELP: Generating Safe and Efficient Task Plans for Robot Agents with Large Language Models}.
\newblock In: \emph{2025 IEEE International Conference on Robotics and Automation (ICRA)}. pp. 2599--2605.
\newblock \doi{10.1109/icra55743.2025.11128420}.
\newblock \urlprefix\url{https://doi.org/10.1109/icra55743.2025.11128420}.

\bibitem[{Wu et~al.(2026{\natexlab{cg}})Wu, Zhang, Chen, Hu, Yuille and Chen}]{wu2026contactcode}
Wu Y, Zhang Z, Chen Q, Hu H, Yuille A and Chen J (2026{\natexlab{cg}}) {Contact Code: A Typed Program Interface for Measurable and Repairable Contact-Rich Manipulation}.
\newblock CoRL 2026 Workshop on Agentic Robotics.
\newblock \urlprefix\url{https://openreview.net/forum?id=78L8T5nsLM}.

\bibitem[{Wu et~al.(2025{\natexlab{cb}})Wu, Lu, Chen, Liu, Zhuang and Hu}]{xref250204692}
Wu Z, Lu J, Chen Y, Liu Y, Zhuang Y and Hu L (2025{\natexlab{cb}}) {STRIDE: Automating Reward Design, Deep Reinforcement Learning Training and Feedback Optimization in Humanoid Robotics Locomotion}.
\newblock \emph{arXiv preprint arXiv:2502.04692} \urlprefix\url{https://arxiv.org/abs/2502.04692}.

\bibitem[{{X Square Robot}(2026)}]{xsquare2026xrharness}
{X Square Robot} (2026) {XR Harness}.
\newblock \urlprefix\url{https://github.com/X-Square-Robot/xr-harness}.

\bibitem[{{XDOF Team}(2026)}]{xdof2026ember}
{XDOF Team} (2026) {Ember-0: A benchmark for LLMs in the real world}.
\newblock Technical report.
\newblock \urlprefix\url{https://www.xdof.ai/blog/evaluating-llms}.

\bibitem[{Xia et~al.(2026)Xia, Li, Li, Ma, Xu, Liu, Cui, Lin, Ma, Wang, Song and Wei}]{sage2026}
Xia H, Li X, Li Z, Ma Q, Xu J, Liu MY, Cui Y, Lin TY, Ma WC, Wang S, Song S and Wei F (2026) {SAGE: Scalable Agentic 3D Scene Generation for Embodied AI}.
\newblock arXiv:2602.10116.
\newblock \urlprefix\url{https://arxiv.org/abs/2602.10116}.

\bibitem[{Xia et~al.(2025)Xia, Feng, Wang and Hu}]{xia2025phoenix}
Xia W, Feng R, Wang D and Hu D (2025) {Phoenix: A Motion-based Self-Reflection Framework for Fine-grained Robotic Action Correction}.
\newblock \emph{arXiv preprint arXiv:2504.14588} \urlprefix\url{https://arxiv.org/abs/2504.14588}.
\newblock Accepted to CVPR 2025.

\bibitem[{Xiao et~al.(2025)Xiao, Janaka, Hu, Gupta, Li, Yu and Hsu}]{grid59}
Xiao A, Janaka N, Hu T, Gupta A, Li K, Yu C and Hsu D (2025) {Robi Butler: Multimodal Remote Interaction with a Household Robot Assistant}.
\newblock In: \emph{2025 IEEE International Conference on Robotics and Automation (ICRA)}. pp. 4337--4344.
\newblock \doi{10.1109/icra55743.2025.11128329}.
\newblock \urlprefix\url{https://doi.org/10.1109/icra55743.2025.11128329}.

\bibitem[{Xiao et~al.(2026)Xiao, Zhang, Hu and Hsu}]{audit260706501}
Xiao A, Zhang H, Hu T and Hsu D (2026) {Hypothesis-driven Model Expansion under Uncertainty for Open-World Robot Planning}.
\newblock arXiv:2607.06501.
\newblock \urlprefix\url{https://arxiv.org/abs/2607.06501v1}.
\newblock Accepted to RSS 2026.

\bibitem[{Xiao et~al.(2026{\natexlab{cb}})Xiao, Xie, Zhang, Lin, Fu, Xue, Lu, Yang, Dai, Wang, Wu, Wang, Sastry, Goldberg, Fan, Zhu and Shi}]{self260619980}
Xiao W, Xie J, Zhang T, Lin H, Fu LM, Xue H, Lu J, Yang Y, Dai C, Wang Z, Wu J, Wang G, Sastry SS, Goldberg K, Fan LJ, Zhu Y and Shi G (2026{\natexlab{cb}}) {ENPIRE: Agentic Robot Policy Self-Improvement in the Real World}.
\newblock arXiv:2606.19980.
\newblock \urlprefix\url{https://arxiv.org/abs/2606.19980}.

\bibitem[{Xie et~al.(2026)Xie, Ma, Li, Liu, Chen, Zhou and Yang}]{xie2026fastplans}
Xie C, Ma B, Li G, Liu Y, Chen H, Zhou X and Yang J (2026) {Fast Plans, Faithful Actions: Closing the Planning-Execution Gap in Hierarchical Vision-Language-Action Models}.
\newblock \urlprefix\url{https://arxiv.org/abs/2609.30833}.

\bibitem[{Xie(2026)}]{xie2026opengeminiharness}
Xie S (2026) {Open Gemini 4 Argon Robot Harness --- A Zero-Shot Bimanual Handover Case Study}.
\newblock Technical report.
\newblock \urlprefix\url{https://ehehee.github.io/blog/open-gemini-harness/}.

\bibitem[{Xie et~al.(2026{\natexlab{cb}})Xie, Chen, Chen, Adebola, Huang, Ma, Qiu, Yuan, Shah, Sanketi and Goldberg}]{xie2026robovista}
Xie S, Chen K, Chen Z, Adebola S, Huang Y, Ma Z, Qiu T, Yuan W, Shah D, Sanketi PR and Goldberg K (2026{\natexlab{cb}}) {RoboVista}: Evaluating vision language models for diverse robot applications.
\newblock \emph{arXiv preprint arXiv:2607.04610} \urlprefix\url{https://arxiv.org/abs/2607.04610}.
\newblock Accepted to RSS 2026.

\bibitem[{Xie et~al.(2026{\natexlab{cc}})Xie, Chen, Yoo, Zhu, Shen, Drake, Solowjow, Gulich, Millane, Cathomen, Stuyck, Subramanian, Pan, McGowan, Benedict, Fan, Zhu, Huang and Goldberg}]{xie2026vabar}
Xie S, Chen K, Yoo U, Zhu X, Shen M, Drake J, Solowjow E, Gulich L, Millane A, Cathomen R, Stuyck T, Subramanian S, Pan Y, McGowan B, Benedict M, Fan L, Zhu Y, Huang S and Goldberg K (2026{\natexlab{cc}}) {VABAR: A Variational Automation Benchmark for Agentic Robotics}.
\newblock \emph{arXiv preprint arXiv:2611.06881} \urlprefix\url{https://arxiv.org/abs/2611.06881}.

\bibitem[{Xie et~al.(2026{\natexlab{cd}})Xie, Chen, Cao, Lu, Wu and Jiang}]{roboskill2026}
Xie S, Chen Y, Cao H, Lu S, Wu Z and Jiang YG (2026{\natexlab{cd}}) {Explore, Execute, Evolve: A Skill Acquisition and Reuse Loop for Embodied Agents}.
\newblock \urlprefix\url{https://arxiv.org/abs/2609.37810}.

\bibitem[{Xie et~al.(2025)Xie, Wang, Xiao, Wang and Chen}]{xref250104268}
Xie S, Wang H, Xiao Z, Wang R and Chen X (2025) {Robotic Programmer: Video Instructed Policy Code Generation for Robotic Manipulation}.
\newblock \emph{arXiv preprint arXiv:2501.04268} \urlprefix\url{https://arxiv.org/abs/2501.04268}.

\bibitem[{Xie et~al.(2026{\natexlab{ce}})Xie, Wang, Lin, Wang and Li}]{xie2026codimrecon}
Xie S, Wang L, Lin G, Wang Z and Li M (2026{\natexlab{ce}}) {CoDimRecon: Agentic Reconstruction of Sim-Ready 3D Scenes with Deformable Curves, Surfaces, and Volumes}.
\newblock \emph{arXiv preprint arXiv:2609.36024} \urlprefix\url{https://arxiv.org/abs/2609.36024}.

\bibitem[{Xie et~al.(2026{\natexlab{cf}})Xie, Zhang, Wang and Chen}]{si260302623}
Xie S, Zhang Y, Wang R and Chen X (2026{\natexlab{cf}}) {Uni-Skill: Building Self-Evolving Skill Repository for Generalizable Robotic Manipulation}.
\newblock \urlprefix\url{https://arxiv.org/abs/2603.02623}.
\newblock Accepted to ICRA 2026.

\bibitem[{Xie et~al.(2023)Xie, Zhao, Wu, Liu, Luo, Zhong, Yang and Yu}]{text2reward2023}
Xie T, Zhao S, Wu CH, Liu Y, Luo Q, Zhong V, Yang Y and Yu T (2023) {Text2Reward: Reward Shaping with Language Models for Reinforcement Learning}.
\newblock arXiv:2309.11489.
\newblock \urlprefix\url{https://arxiv.org/abs/2309.11489}.
\newblock Accepted to ICLR 2024.

\bibitem[{Xie et~al.(2026{\natexlab{cg}})Xie, Hong, Wang, Ji, Zhou, Wu, Sirejiding, Zhou, Chen, Ling, Ding and Lu}]{audit260920388}
Xie W, Hong M, Wang Y, Ji Y, Zhou Y, Wu S, Sirejiding S, Zhou H, Chen YC, Ling M, Ding Y and Lu H (2026{\natexlab{cg}}) {Navi-Agent: Unlocalized Monocular Navigation Agent}.
\newblock arXiv:2609.20388.
\newblock \urlprefix\url{https://arxiv.org/abs/2609.20388v1}.

\bibitem[{Xie et~al.(2026{\natexlab{ch}})Xie, Xu, Jiang and Zhao}]{xie2026idm}
Xie Z, Xu C, Jiang Y and Zhao L (2026{\natexlab{ch}}) {Predict Before You Modify: Learning Improvement Dynamics for Self-Evolving Agents}.
\newblock \emph{arXiv preprint arXiv:2610.13991} \doi{10.48550/arXiv.2610.13991}.
\newblock \urlprefix\url{https://arxiv.org/abs/2610.13991}.

\bibitem[{Xin et~al.(2026)Xin, Xua, Deghat, Wan, Liu, Hu and Hu}]{audit260828300}
Xin G, Xua J, Deghat M, Wan C, Liu J, Hu Y and Hu Z (2026) {MaCoPlanner: LLM-Assisted Manual-Compiled Task Planning with Proactive Safety Verification for Robotic Industrial Panel Operation}.
\newblock arXiv:2608.28300.
\newblock \urlprefix\url{https://arxiv.org/abs/2608.28300v1}.

\bibitem[{Xin et~al.(2026{\natexlab{cb}})Xin, Li, Jiang, Huang, Wang, Li, Zhu, Qin and Huang}]{grid31}
Xin Z, Li W, Jiang Y, Huang Z, Wang B, Li P, Zhu J, Qin J and Huang S (2026{\natexlab{cb}}) {AgentVLN: Towards Agentic Vision-and-Language Navigation}.
\newblock In: \emph{Computer Vision – ECCV 2026}, Lecture Notes in Computer Science. pp. 114--132.
\newblock \doi{10.1007/978-3-032-37167-6_7}.
\newblock \urlprefix\url{https://doi.org/10.1007/978-3-032-37167-6_7}.

\bibitem[{Xiong et~al.(2025)Xiong, Shen, Li, Zhou, Liu, Wang and Dong}]{xiong2025aicmllm}
Xiong C, Shen C, Li X, Zhou K, Liu J, Wang R and Dong H (2025) Autonomous interactive correction mllm for robust robotic manipulation.
\newblock In: \emph{Proceedings of The 8th Conference on Robot Learning}, \emph{Proceedings of Machine Learning Research}, volume 270. PMLR, pp. 3139--3156.
\newblock \urlprefix\url{https://proceedings.mlr.press/v270/xiong25a.html}.

\bibitem[{{XLANG LAB} et~al.(2026){XLANG LAB}, Cai, Chen, Chen, Cui, Gong, Kang, Liu, Song, Wu, Wu, Zheng, Zhou, Li, Lu, Lu, Pu, Wan and Yu}]{xlanglab2026robowild}
{XLANG LAB}, Cai Y, Chen D, Chen J, Cui K, Gong Z, Kang Z, Liu Y, Song J, Wu H, Wu W, Zheng Y, Zhou Z, Li Y, Lu H, Lu T, Pu Z, Wan Y and Yu T (2026) {RoboWild}: Can frontier models generalize to robot use in the wild?
\newblock \urlprefix\url{https://huggingface.co/datasets/xlangai/RoboWild/blob/main/RoboWild.pdf}.

\bibitem[{Xu et~al.(2026)Xu, Bao, Liu, Zhou, Shang and Ferrara}]{xu2026chunkact}
Xu B, Bao W, Liu X, Zhou J, Shang Y and Ferrara E (2026) {ChunkAct: Action Chunking for Agentic Robotics}.
\newblock \urlprefix\url{https://arxiv.org/abs/2610.18874}.

\bibitem[{Xu et~al.(2026{\natexlab{cb}})Xu, Shang, Dong and Ferrara}]{xu2026safeharness}
Xu B, Shang Y, Dong Z and Ferrara E (2026{\natexlab{cb}}) {Coding Agents with Harness for Safe Robot Control}.
\newblock \emph{arXiv preprint arXiv:2609.20822} \urlprefix\url{https://arxiv.org/abs/2609.20822}.

\bibitem[{Xu et~al.(2026{\natexlab{cc}})Xu, Shang and Ferrara}]{si260816889}
Xu B, Shang Y and Ferrara E (2026{\natexlab{cc}}) {Don't Drop the BATON: Long-Horizon Robot Manipulation via Agentic Subtask Exploration and Transition-aware Memory}.
\newblock \urlprefix\url{https://arxiv.org/abs/2608.16889}.

\bibitem[{Xu et~al.(2024)Xu, Zhang, Li, Liu, Lan and Kong}]{xu2024sinvig}
Xu J, Zhang H, Li X, Liu H, Lan X and Kong T (2024) {SInViG: A Self-Evolving Interactive Visual Agent for Human-Robot Interaction}.
\newblock \doi{10.48550/arXiv.2402.11792}.
\newblock \urlprefix\url{https://arxiv.org/abs/2402.11792}.

\bibitem[{Xu et~al.(2026{\natexlab{cd}})Xu, Zhang, Yan, Deng, Xia, Dong, Chen, Lyu, Gao, Zhang and Wang}]{xu2026gif}
Xu L, Zhang Z, Yan M, Deng S, Xia C, Dong M, Chen J, Lyu J, Gao F, Zhang Z and Wang H (2026{\natexlab{cd}}) {GIF: Agentic Generation of Interactive and Functional Object Compositions for Robot Learning}.
\newblock \emph{arXiv preprint arXiv:2609.05927} \urlprefix\url{https://arxiv.org/abs/2609.05927}.

\bibitem[{Xu et~al.(2023)Xu, Huang, Yu, Liu, Zhang, Niu, Zhang, Xia, Tan and Zhao}]{audit231013065}
Xu M, Huang P, Yu W, Liu S, Zhang X, Niu Y, Zhang T, Xia F, Tan J and Zhao D (2023) {Creative Robot Tool Use with Large Language Models}.
\newblock arXiv:2310.13065.
\newblock \urlprefix\url{https://arxiv.org/abs/2310.13065v1}.

\bibitem[{Xu et~al.(2026{\natexlab{ce}})Xu, Zheng and Mu}]{sref260407774}
Xu P, Zheng J and Mu Y (2026{\natexlab{ce}}) {RoboAgent: Chaining Basic Capabilities for Embodied Task Planning}.
\newblock \urlprefix\url{https://arxiv.org/abs/2604.07774}.
\newblock Accepted to CVPR 2026.

\bibitem[{Xu et~al.(2026{\natexlab{cf}})Xu, Jin, Zhou, Yue, Qiao, Tai, Deng, Jia and Liu}]{xu2026agentchord}
Xu S, Jin R, Zhou H, Yue B, Qiao G, Tai Y, Deng Y, Jia K and Liu G (2026{\natexlab{cf}}) {From Reaction to Anticipation: Proactive Failure Recovery through Agentic Task Graph for Robotic Manipulation}.
\newblock \emph{arXiv preprint arXiv:2605.11951} \urlprefix\url{https://arxiv.org/abs/2605.11951}.
\newblock Accepted to RSS 2026.

\bibitem[{Xu et~al.(2026{\natexlab{cg}})Xu, Wang, Shen, Xu, Jin, Niu, Li, Zhang, Dai, Luo, Chai, Chen, Liu and Yu}]{xu2026theharness}
Xu S, Wang Z, Shen H, Xu Z, Jin W, Niu Y, Li Y, Zhang J, Dai Y, Luo Y, Chai W, Chen X, Liu Z and Yu SX (2026{\natexlab{cg}}) {The Harness: One Agent Harness for Anything that Acts}.
\newblock In: \emph{Conference on Robot Learning, Late Breaking Results}.
\newblock \urlprefix\url{https://web.eecs.umich.edu/~stellayu/publication/doc/2026harnessCoRLL.pdf}.

\bibitem[{Xu et~al.(2026{\natexlab{ch}})Xu, Ren, Yang, Zheng and Qin}]{xu2026rtsafe}
Xu TR, Ren J, Yang Y, Zheng Z and Qin L (2026{\natexlab{ch}}) {RT-Safe: Benchmarking Agent Safety in Real-Time Embodied Environment}.
\newblock \urlprefix\url{https://arxiv.org/abs/2610.09294}.

\bibitem[{Xu et~al.(2024{\natexlab{cb}})Xu, Pei, Yang, Shi, Zhang and Zhao}]{xu2024criticalscenariotesting}
Xu W, Pei H, Yang J, Shi Y, Zhang Y and Zhao Q (2024{\natexlab{cb}}) {Exploring Critical Testing Scenarios for Decision-Making Policies: An LLM Approach}.
\newblock \doi{10.48550/arXiv.2412.06684}.
\newblock \urlprefix\url{https://arxiv.org/abs/2412.06684}.

\bibitem[{Xu et~al.(2025)Xu, Mao, Li, Du, Loz{\'a}no-P{\'e}rez, Kaelbling and Hsu}]{xu2024set}
Xu Y, Mao J, Li L, Du Y, Loz{\'a}no-P{\'e}rez T, Kaelbling LP and Hsu D (2025) “set it up”: Functional object arrangement with compositional generative models.
\newblock \emph{The International Journal of Robotics Research} : 02783649251378198\doi{10.1177/02783649251378198}.
\newblock \urlprefix\url{https://doi.org/10.1177/02783649251378198}.

\bibitem[{Xu et~al.(2026{\natexlab{ci}})Xu, Xu, Chen, Dong, Zhang, Xue, Liu and Zhang}]{xu2026tarsphere}
Xu Y, Xu Z, Chen Z, Dong Y, Zhang T, Xue Z, Liu Y and Zhang J (2026{\natexlab{ci}}) {TARSphere: A Robot-Oriented Full-Stack Framework for Long-Horizon Embodied Agents in Evolving Physical and Social Environments}.
\newblock \emph{arXiv preprint arXiv:2610.30043} \doi{10.48550/arXiv.2610.30043}.
\newblock \urlprefix\url{https://arxiv.org/abs/2610.30043}.

\bibitem[{Yang et~al.(2024)Yang, Su, Gkanatsios, Ke, Jain, Schneider and Fragkiadaki}]{yang2024diffusiones}
Yang B, Su H, Gkanatsios N, Ke TW, Jain A, Schneider J and Fragkiadaki K (2024) {Diffusion-ES: Gradient-free Planning with Diffusion for Autonomous Driving and Zero-Shot Instruction Following}.
\newblock \doi{10.48550/arXiv.2402.06559}.
\newblock \urlprefix\url{https://arxiv.org/abs/2402.06559}.

\bibitem[{Yang et~al.(2026)Yang, Shi, Zhang, Xu, Fan and Wang}]{yang2026tactalent}
Yang C, Shi L, Zhang H, Xu M, Fan J and Wang C (2026) {GPT-6 Astra Is a Native Visuo-Tactile Policy for Contact-Rich Manipulation}.
\newblock \emph{arXiv preprint arXiv:2610.26846} \doi{10.48550/arXiv.2610.26846}.
\newblock \urlprefix\url{https://arxiv.org/abs/2610.26846}.

\bibitem[{Yang et~al.(2026{\natexlab{cb}})Yang, Jiao, Wang, Niu, Liu and Liu}]{xref260204419}
Yang H, Jiao Z, Wang S, Niu Y, Liu S and Liu H (2026{\natexlab{cb}}) {Integrated Exploration and Sequential Manipulation on Scene Graph with LLM-based Situated Replanning}.
\newblock \emph{arXiv preprint arXiv:2602.04419} \urlprefix\url{https://arxiv.org/abs/2602.04419}.
\newblock Accepted to ICRA 2026.

\bibitem[{Yang et~al.(2026{\natexlab{cc}})Yang, Dagli, Zook, Hadfield, Goyal, Birchfield, Ramos and Tremblay}]{yang2026robolabhighfidelitysimulationbenchmark}
Yang JX, Dagli R, Zook A, Hadfield H, Goyal A, Birchfield S, Ramos F and Tremblay J (2026{\natexlab{cc}}) Robolab: A high-fidelity simulation benchmark for analysis of task generalist policies.
\newblock In: \emph{Robotics: Science and Systems XXII}.
\newblock \doi{10.15607/rss.2026.xxii.096}.
\newblock \urlprefix\url{https://doi.org/10.15607/rss.2026.xxii.096}.

\bibitem[{Yang et~al.(2026{\natexlab{cd}})Yang, Khetan and Zhang}]{yang2026astranomy}
Yang JX, Khetan A and Zhang S (2026{\natexlab{cd}}) {Astranomy: An Empirical Study of GPT-6 Astra as a Native Robot Policy}.
\newblock NVIDIA technical report.
\newblock \urlprefix\url{https://research.nvidia.com/labs/srl/projects/astranomy/}.
\newblock Report dated October 14, 2026; accessed October 16, 2026.

\bibitem[{Yang et~al.(2025)Yang, Chen, Zhang, Zhao, Qian, Wang, Wang, Koripella, Movahedi, Li, Ji, Zhang and Zhang}]{embodiedbench}
Yang R, Chen H, Zhang J, Zhao M, Qian C, Wang K, Wang Q, Koripella TV, Movahedi M, Li M, Ji H, Zhang H and Zhang T (2025) {E}mbodied{B}ench: Comprehensive benchmarking multi-modal large language models for vision-driven embodied agents.
\newblock In: \emph{Proceedings of the 42nd International Conference on Machine Learning}, \emph{Proceedings of Machine Learning Research}, volume 267. PMLR, pp. 70576--70631.
\newblock \urlprefix\url{https://proceedings.mlr.press/v267/yang25f.html}.

\bibitem[{Yang et~al.(2024{\natexlab{cb}})Yang, Zhang, Fernandez-Laaksonen, Ding and Gong}]{yang2024drivingstylealignment}
Yang R, Zhang X, Fernandez-Laaksonen A, Ding X and Gong J (2024{\natexlab{cb}}) {Driving Style Alignment for LLM-powered Driver Agent}.
\newblock \doi{10.48550/arXiv.2403.11368}.
\newblock \urlprefix\url{https://arxiv.org/abs/2403.11368}.

\bibitem[{Yang et~al.(2024{\natexlab{cc}})Yang, Du, Ghasemipour, Tompson, Kaelbling, Schuurmans and Abbeel}]{yang2024unisim}
Yang S, Du Y, Ghasemipour K, Tompson J, Kaelbling L, Schuurmans D and Abbeel P (2024{\natexlab{cc}}) Learning interactive real-world simulators.
\newblock In: \emph{International Conference on Learning Representations}.
\newblock \urlprefix\url{https://arxiv.org/abs/2310.06114}.

\bibitem[{Yang et~al.(2024{\natexlab{cd}})Yang, Luo, Pani and Yang}]{audit240208212}
Yang S, Luo Q, Pani A and Yang Y (2024{\natexlab{cd}}) {BBSEA: An Exploration of Brain-Body Synchronization for Embodied Agents}.
\newblock \emph{arXiv preprint arXiv:2402.08212} \doi{10.48550/arXiv.2402.08212}.
\newblock \urlprefix\url{https://arxiv.org/abs/2402.08212}.

\bibitem[{Yang et~al.(2026{\natexlab{ce}})Yang, Bhatt, Wang, Tetteh, Wang and Topcu}]{si260605395}
Yang Y, Bhatt NP, Wang K, Tetteh S, Wang Z and Topcu U (2026{\natexlab{ce}}) {VASO: Formally Verifiable Self-Evolving Skills for Physical AI Agents}.
\newblock \urlprefix\url{https://arxiv.org/abs/2606.05395}.

\bibitem[{Yang et~al.(2026{\natexlab{cf}})Yang, Chen, Gao and Zhang}]{ora2026}
Yang Y, Chen S, Gao J and Zhang H (2026{\natexlab{cf}}) {ORA}: Open robot-use agents.
\newblock Hao AI Lab blog.
\newblock \urlprefix\url{https://haoailab.com/blogs/ora/}.

\bibitem[{Yang et~al.(2025{\natexlab{cb}})Yang, Gu, Wen, Li, Zhao, Wang and Liu}]{audit251011660}
Yang Y, Gu K, Wen Y, Li H, Zhao Y, Wang T and Liu X (2025{\natexlab{cb}}) {ManiAgent: An Agentic Framework for General Robotic Manipulation}.
\newblock \emph{arXiv preprint arXiv:2510.11660} \doi{10.48550/arXiv.2510.11660}.
\newblock \urlprefix\url{https://arxiv.org/abs/2510.11660}.

\bibitem[{Yang et~al.(2025{\natexlab{cc}})Yang, Jia, Zhang and Huang}]{sceneweaver2025}
Yang Y, Jia B, Zhang S and Huang S (2025{\natexlab{cc}}) {SceneWeaver: All-in-One 3D Scene Synthesis with an Extensible and Self-Reflective Agent}.
\newblock arXiv:2509.20414.
\newblock \urlprefix\url{https://arxiv.org/abs/2509.20414}.
\newblock Accepted to NeurIPS 2025.

\bibitem[{Yang et~al.(2023)Yang, Sun, Weihs, VanderBilt, Herrasti, Han, Wu, Haber, Krishna, Liu, Callison-Burch, Yatskar, Kembhavi and Clark}]{holodeck2023}
Yang Y, Sun FY, Weihs L, VanderBilt E, Herrasti A, Han W, Wu J, Haber N, Krishna R, Liu L, Callison-Burch C, Yatskar M, Kembhavi A and Clark C (2023) {Holodeck: Language Guided Generation of 3D Embodied AI Environments}.
\newblock arXiv:2312.09067.
\newblock \urlprefix\url{https://arxiv.org/abs/2312.09067}.
\newblock Accepted to CVPR 2024.

\bibitem[{Yang et~al.(2026{\natexlab{cg}})Yang, Tang, Ma, Fan, Wang, Li and Yang}]{jevevaluation2026}
Yang Y, Tang H, Ma Y, Fan X, Wang W, Li J and Yang T (2026{\natexlab{cg}}) {Code Owns the Simulation, Jev Owns the Evaluation}.
\newblock \emph{arXiv preprint arXiv:2610.01834} \urlprefix\url{https://arxiv.org/abs/2610.01834}.

\bibitem[{Yang et~al.(2026{\natexlab{ch}})Yang, Tong, Zhang, Ding, Yang, Zhang, Li, Yang and Xin}]{sref260623280}
Yang Y, Tong Y, Zhang Z, Ding X, Yang N, Zhang Y, Li H, Yang K and Xin M (2026{\natexlab{ch}}) {Causal Reward World Models: Zero-shot Reward Design for Automated Skill Generation}.
\newblock \urlprefix\url{https://arxiv.org/abs/2606.23280}.

\bibitem[{Yang et~al.(2024{\natexlab{ce}})Yang, Yang, Zhou, Chen, Zhang, Du and Gan}]{yang20243dmem}
Yang Y, Yang H, Zhou J, Chen P, Zhang H, Du Y and Gan C (2024{\natexlab{ce}}) {3D-Mem: 3D Scene Memory for Embodied Exploration and Reasoning}.
\newblock \urlprefix\url{https://arxiv.org/abs/2411.17735}.

\bibitem[{Yang et~al.(2025{\natexlab{cd}})Yang, Chen, Zhou, Yan, Song, Liu, Li, Zhang, Zhou, Chen and Sun}]{agenticrobot}
Yang Z, Chen Y, Zhou X, Yan J, Song D, Liu Y, Li Y, Zhang Y, Zhou P, Chen H and Sun L (2025{\natexlab{cd}}) {Agentic Robot: A Brain-Inspired Framework for Vision-Language-Action Models in Embodied Agents}.
\newblock arXiv:2505.23450.
\newblock \urlprefix\url{https://arxiv.org/abs/2505.23450}.

\bibitem[{Yang et~al.(2026{\natexlab{ci}})Yang, Li, Hu, Liu, Huang, Sun, Li, Wu, Huang, Huang, Sun, Liu, Tong, Huang, Hu, Huang, Bai, Wang, Lin, Tang, Zhang, Chen, Song, Dai, Chen, Ji, Zhan, Hu, Xue, Zhang, Hu, Ho and Guo}]{yang2026robotworld}
Yang Z, Li C, Hu X, Liu Y, Huang W, Sun J, Li H, Wu Z, Huang Y, Huang F, Sun H, Liu J, Tong J, Huang M, Hu S, Huang S, Bai T, Wang X, Lin Y, Tang Z, Zhang Z, Chen Z, Song X, Dai J, Chen B, Ji J, Zhan F, Hu M, Xue W, Zhang Y, Hu H, Ho TY and Guo Y (2026{\natexlab{ci}}) {RobotWorld: Benchmarking Multimodal Agents for Robot Use Across Diverse Tasks and Embodiments}.
\newblock \emph{arXiv preprint arXiv:2610.10409} \urlprefix\url{https://arxiv.org/abs/2610.10409}.

\bibitem[{Yang et~al.(2026{\natexlab{cj}})Yang, Liang, Zhou, Jiang, Cao, Chen, Wang, Sun, Xiang, Li, Mu, Chi, Tong, Li, Ma, Yang, Ren, Ma, Zhu, Ji, Wang, Wang, Liu, Zixiang, Zhang, Liu, Feng, Tang and Qin}]{yang2026proagentphysical}
Yang Z, Liang J, Zhou X, Jiang Y, Cao X, Chen Z, Wang L, Sun H, Xiang Y, Li C, Mu R, Chi B, Tong Z, Li M, Ma H, Yang Y, Ren B, Ma Z, Zhu Z, Ji J, Wang Z, Wang Q, Liu X, Zixiang W, Zhang S, Liu M, Feng X, Tang D and Qin B (2026{\natexlab{cj}}) {Proactive Physical AI through Continuous Observation and Timely Intervention}.
\newblock \urlprefix\url{https://arxiv.org/abs/2610.16967}.

\bibitem[{Yang et~al.(2023{\natexlab{cb}})Yang, Raman, Shah and Tellex}]{audit230909919}
Yang Z, Raman SS, Shah A and Tellex S (2023{\natexlab{cb}}) {Plug in the Safety Chip: Enforcing Constraints for LLM-driven Robot Agents}.
\newblock arXiv:2309.09919.
\newblock \urlprefix\url{https://arxiv.org/abs/2309.09919v1}.

\bibitem[{Yao et~al.(2026)Yao, Liu, Li and Savvides}]{regot2025}
Yao C, Liu X, Li C and Savvides M (2026) {Reward Evolution with Graph-of-Thoughts: A Bi-Level Language Model Framework for Reinforcement Learning}.
\newblock In: \emph{IEEE International Conference on Robotics and Automation}. pp. 4571--4578.
\newblock \doi{10.1109/icra57385.2026.11696350}.
\newblock \urlprefix\url{https://doi.org/10.1109/icra57385.2026.11696350}.

\bibitem[{Yao et~al.(2026{\natexlab{cb}})Yao, Lin, Du and Mu}]{yao2026sceneevolve}
Yao H, Lin C, Du R and Mu Y (2026{\natexlab{cb}}) {Learning to Code the World: Harness Evolution for Compositional 3D Reconstruction}.
\newblock \doi{10.48550/arXiv.2610.37046}.
\newblock \urlprefix\url{https://arxiv.org/abs/2610.37046}.

\bibitem[{Yao et~al.(2025)Yao, Wen, Gao, Guo, Liu and Huang}]{yao2025homesafebench}
Yao J, Wen H, Gao S, Guo Y, Liu Z and Huang H (2025) {HomeSafeBench: Benchmarking Embodied Vision-Language Models in Free-Exploration Home Safety Inspection}.
\newblock \urlprefix\url{https://arxiv.org/abs/2509.23690}.

\bibitem[{Yao et~al.(2026{\natexlab{cc}})Yao, Zou, Jiao, Gao, Gao, Wang, Chen, Zhu, Li, Yan, Wang, Ma, Guo, Zhang and Shen}]{yao2026frontier}
Yao Y, Zou Y, Jiao G, Gao X, Gao N, Wang Y, Chen J, Zhu Z, Li X, Yan N, Wang S, Ma J, Guo H, Zhang W and Shen C (2026{\natexlab{cc}}) {Profiling a Frontier Agent with EBench: From Direct Control to Harness with Policy}.
\newblock \urlprefix\url{https://arxiv.org/abs/2610.35832}.

\bibitem[{Ye(2026)}]{ye2026dexterousastra}
Ye J (2026) {Dexterous Astra}.
\newblock Software / demo.
\newblock \urlprefix\url{https://github.com/jianglongye/dexterous-astra}.

\bibitem[{Ye et~al.(2025)Ye, Mao, Cui, Yu, Zhai, Chen, Zhou, Xiong and Wang}]{ye2025etpr1}
Ye S, Mao S, Cui Y, Yu X, Zhai S, Chen W, Zhou S, Xiong R and Wang Y (2025) {ETP-R1: Evolving Topological Planning with Reinforcement Fine-tuning for Vision-Language Navigation in Continuous Environments}.
\newblock \doi{10.48550/arXiv.2512.20940}.
\newblock \urlprefix\url{https://arxiv.org/abs/2512.20940}.

\bibitem[{Ye et~al.(2025{\natexlab{cb}})Ye, Liu, Ding, Gao, Rybkin and Abbeel}]{ye2025video2policy}
Ye W, Liu F, Ding Z, Gao Y, Rybkin O and Abbeel P (2025{\natexlab{cb}}) {Video2Policy: Scaling up Manipulation Tasks in Simulation through Internet Videos}.
\newblock \emph{arXiv preprint arXiv:2502.09886} \urlprefix\url{https://arxiv.org/abs/2502.09886}.

\bibitem[{Ye et~al.(2025{\natexlab{cc}})Ye, Lu, Ye, Lin, Yang, Yan and Zhao}]{ye2025robofac}
Ye Z, Lu W, Ye M, Lin T, Yang S, Yan J and Zhao B (2025{\natexlab{cc}}) {RoboFAC: A Comprehensive Framework for Robotic Failure Analysis and Correction}.
\newblock \emph{arXiv preprint arXiv:2505.12224} \urlprefix\url{https://arxiv.org/abs/2505.12224}.

\bibitem[{Yeo et~al.(2026)Yeo, Weerakoon, Weerakoon and Misra}]{si260206366}
Yeo T, Weerakoon D, Weerakoon D and Misra A (2026) {Towards Adaptive Environment Generation for Training Embodied Agents}.
\newblock \urlprefix\url{https://arxiv.org/abs/2602.06366}.
\newblock Accepted to AAAI 2026 Bridge Program: Making Embodied AI Reliable with Testing and Formal Verification.

\bibitem[{Yi et~al.(2026)Yi, Zhang, Zhou, Wang, Sun, Dong, Xu, Shen, Zhang, Liu and Zhang}]{yi2026easg}
Yi J, Zhang S, Zhou Z, Wang H, Sun H, Dong Y, Xu X, Shen X, Zhang T, Liu Y and Zhang J (2026) {VLMs are Native Action Models: Unleashing the Manipulation Capabilities of VLMs with Explicit Action-Space Grounding}.
\newblock \urlprefix\url{https://arxiv.org/abs/2610.12832}.

\bibitem[{Yi et~al.(2026{\natexlab{cb}})Yi, Ma, Xu, Hao, Gan, Li and Zhong}]{sref260305185}
Yi P, Ma Y, Xu W, Hao Y, Gan S, Li W and Zhong S (2026{\natexlab{cb}}) {Critic in the Loop: A Tri-System VLA Framework for Robust Long-Horizon Manipulation}.
\newblock \urlprefix\url{https://arxiv.org/abs/2603.05185}.

\bibitem[{Yidou-Weng et~al.(2026)Yidou-Weng, Sun, Ma, Dogan, Wang, Peng, Van~den Broeck and Cui}]{weng2026metactrl}
Yidou-Weng G, Sun E, Ma T, Dogan MA, Wang B, Peng A, Van~den Broeck G and Cui Y (2026) {Meta-Ctrl: Guaranteed Plan Generation by Decoupling Syntactic and Semantic Constraints}.
\newblock arXiv:2608.22149.
\newblock \urlprefix\url{https://arxiv.org/abs/2608.22149}.

\bibitem[{Yin et~al.(2026)Yin, Huang, Yang, Wang, Zhao, Xu, Sun, Hou, Li, Wu, Liu, Xiao, Zhang, Bao, Feng, Pang, Li, Wang and Yao}]{geniesim2026}
Yin C, Huang D, Yang D, Wang J, Zhao N, Xu C, Sun W, Hou L, Li Z, Wu J, Liu Z, Xiao Z, Zhang S, Bao L, Feng R, Pang Z, Li J, Wang Q and Yao M (2026) {Genie Sim 3.0 : A High-Fidelity Comprehensive Simulation Platform for Humanoid Robot}.
\newblock \emph{arXiv preprint arXiv:2601.02078} \urlprefix\url{https://arxiv.org/abs/2601.02078}.
\newblock Agent-driven scene-generation implementation: \url{https://github.com/AgibotTech/genie_sim}.

\bibitem[{Yin et~al.(2026{\natexlab{cb}})Yin, Liang, Zhang, Liu, Li, Zhang and Wang}]{yin2026alldaynav}
Yin H, Liang Y, Zhang J, Liu J, Li M, Zhang Z and Wang H (2026{\natexlab{cb}}) {AllDayNav: Lifelong Navigation via Real-World Reinforcement Learning}.
\newblock \emph{arXiv preprint arXiv:2606.10927} \doi{10.48550/arXiv.2606.10927}.
\newblock \urlprefix\url{https://arxiv.org/abs/2606.10927}.

\bibitem[{Yin et~al.(2024)Yin, Pang, Ding, Chen, Bi, Xiong, Huang, Xiang, Shao and Chen}]{yin2024safeagentbench}
Yin S, Pang X, Ding Y, Chen M, Bi Y, Xiong Y, Huang W, Xiang Z, Shao J and Chen S (2024) {SafeAgentBench: A Benchmark for Safe Task Planning of Embodied LLM Agents}.
\newblock \urlprefix\url{https://arxiv.org/abs/2412.13178}.

\bibitem[{Yin et~al.(2026{\natexlab{cc}})Yin, Mei, Zheng, Yamane, Wang, Sceats, Bateman, Zha, Badithela, Shorinwa and Majumdar}]{yin2026playworld}
Yin T, Mei Z, Zheng Z, Yamane M, Wang D, Sceats J, Bateman SM, Zha L, Badithela A, Shorinwa O and Majumdar A (2026{\natexlab{cc}}) {PlayWorld: Learning Robot World Models from Autonomous Play}.
\newblock \emph{arXiv preprint arXiv:2603.09030} \doi{10.48550/arXiv.2603.09030}.
\newblock \urlprefix\url{https://arxiv.org/abs/2603.09030}.

\bibitem[{Yoneda et~al.(2023)Yoneda, Fang, Li, Zhang, Jiang, Lin, Picker, Yunis, Mei and Walter}]{yoneda2023statler}
Yoneda T, Fang J, Li P, Zhang H, Jiang T, Lin S, Picker B, Yunis D, Mei H and Walter MR (2023) {Statler: State-Maintaining Language Models for Embodied Reasoning}.
\newblock \emph{arXiv preprint arXiv:2306.17840} \urlprefix\url{https://arxiv.org/abs/2306.17840}.
\newblock Accepted to ICRA 2024.

\bibitem[{Yoo et~al.(2026)Yoo, Choi, Kim and Zhang}]{yoo2026onestepplanner}
Yoo Y, Choi JW, Kim D and Zhang BT (2026) {One-Step Planner: Unified Observation and Decision-Making with Vision-Language Models}.
\newblock ICRA 2026 Workshop on Rigorous Perception.
\newblock \urlprefix\url{https://icra2026-rigorous-perception.github.io/pdf/yoo2026.pdf}.

\bibitem[{Yu et~al.(2026)Yu, Chen, Rao, Hu, Zhu, Zhao and Ju}]{audit260908402}
Yu B, Chen Z, Rao J, Hu Y, Zhu Z, Zhao Y and Ju R (2026) {Towards Embodied Air-Ground Cooperative Object Search: Benchmark, Dataset and Agentic Method}.
\newblock arXiv:2609.08402.
\newblock \urlprefix\url{https://arxiv.org/abs/2609.08402v1}.

\bibitem[{Yu et~al.(2026{\natexlab{cb}})Yu, Yuan, Li, Kasaei and Cao}]{grid35}
Yu B, Yuan Q, Li K, Kasaei H and Cao M (2026{\natexlab{cb}}) {Co-NavGPT: Multirobot Cooperative Visual Semantic Navigation Using Vision Language Models}.
\newblock \emph{IEEE Robotics and Automation Letters} 11(2): 2122--2129.
\newblock \doi{10.1109/lra.2025.3645650}.
\newblock \urlprefix\url{https://doi.org/10.1109/lra.2025.3645650}.

\bibitem[{Yu(2026)}]{si260807555}
Yu H (2026) {You Don't Need To Stay in The Loop: An Agentic Robotics Loop for Robot-Policy Improvement}.
\newblock \urlprefix\url{https://arxiv.org/abs/2608.07555}.

\bibitem[{Yu et~al.(2026{\natexlab{cc}})Yu, Li, Li, Yan, Peng, Zhong, Xin, Yao, Tian and Li}]{yu2026robodoctor}
Yu H, Li H, Li S, Yan J, Peng S, Zhong W, Xin T, Yao Y, Tian X and Li Y (2026{\natexlab{cc}}) {RoboDoctor: Training-Free Recursive Self-Improvement for Robot Policies}.
\newblock \emph{arXiv preprint arXiv:2610.35604} \doi{10.48550/arXiv.2610.35604}.
\newblock \urlprefix\url{https://arxiv.org/abs/2610.35604}.

\bibitem[{Yu et~al.(2026{\natexlab{cd}})Yu, Meng, Zhang, Jing, Yang, Hu, Ge, Hou, Xin, Ye, Qin, Li, Zhou, Qiu, Sun, Zhang, Nie, Luo and Zhao}]{yu2026physladder}
Yu H, Meng J, Zhang Z, Jing B, Yang L, Hu Y, Ge C, Hou Y, Xin C, Ye D, Qin H, Li J, Zhou W, Qiu J, Sun J, Zhang D, Nie Z, Luo P and Zhao D (2026{\natexlab{cd}}) {LLM4Robot-Bench: How Far Can GPT-6 Astra Go in Robotic Manipulation? A Hierarchical, Paired Sim-Real Benchmark}.
\newblock \doi{10.48550/arXiv.2610.34527}.
\newblock \urlprefix\url{https://arxiv.org/abs/2610.34527}.

\bibitem[{Yu et~al.(2026{\natexlab{ce}})Yu, Dabhi, Li, Li, Jeni and Liu}]{yu2026maniskillformer}
Yu P, Dabhi M, Li S, Li B, Jeni L and Liu C (2026{\natexlab{ce}}) {ManiSkillFormer: Demonstration-Free Compositional Manipulation via Geometric Contracts and Agentic Skill Graph}.
\newblock \emph{arXiv preprint arXiv:2609.16331} \urlprefix\url{https://arxiv.org/abs/2609.16331}.

\bibitem[{Yu et~al.(2023)Yu, Gileadi, Fu, Kirmani, Lee, Gonzalez~Arenas, Chiang, Erez, Hasenclever, Humplik, Ichter, Xiao, Xu, Zeng, Zhang, Heess, Sadigh, Tan, Tassa and Xia}]{language2rewards2023}
Yu W, Gileadi N, Fu C, Kirmani S, Lee KH, Gonzalez~Arenas M, Chiang HTL, Erez T, Hasenclever L, Humplik J, Ichter B, Xiao T, Xu P, Zeng A, Zhang T, Heess N, Sadigh D, Tan J, Tassa Y and Xia F (2023) Language to rewards for robotic skill synthesis.
\newblock In: \emph{Proceedings of The 7th Conference on Robot Learning}, \emph{Proceedings of Machine Learning Research}, volume 229. PMLR, pp. 374--404.
\newblock \urlprefix\url{https://proceedings.mlr.press/v229/yu23a.html}.

\bibitem[{Yu et~al.(2026{\natexlab{cf}})Yu, Zhu, Li, Wang, Zhou and Chen}]{xref261015814}
Yu Y, Zhu B, Li J, Wang J, Zhou S and Chen J (2026{\natexlab{cf}}) {ReTrace: Cross-Stage Failure Attribution for Self-Improving Robotic Manipulation}.
\newblock \emph{arXiv preprint arXiv:2610.15814} \urlprefix\url{https://arxiv.org/abs/2610.15814}.

\bibitem[{Yu et~al.(2026{\natexlab{cg}})Yu, Wang, Liu, Ren, Yu, Bai, Zhang, Jia, Jia, Zhang, Zhang, Yujing, Chen, Zhou, Wang and Yang}]{f4r2026}
Yu Z, Wang J, Liu T, Ren Y, Yu P, Bai C, Zhang Z, Jia Y, Jia J, Zhang Y, Zhang X, Yujing S, Chen Y, Zhou C, Wang T and Yang J (2026{\natexlab{cg}}) {F4R: Failure-Driven Recognition, Reconstruction, Refinement, and Redeployment for Continual Robot Self-Improvement}.
\newblock \emph{arXiv preprint arXiv:2609.35575} \urlprefix\url{https://arxiv.org/abs/2609.35575}.

\bibitem[{Yu et~al.(2026{\natexlab{ch}})Yu, Yuan and Zhang}]{yu2026memora}
Yu Z, Yuan X and Zhang C (2026{\natexlab{ch}}) {MEMORA: Embodied Action Memory from Egocentric Videos for Reasoning and Planning}.
\newblock \emph{arXiv preprint arXiv:2607.14252} \urlprefix\url{https://arxiv.org/abs/2607.14252}.
\newblock Accepted to EMNLP 2026.

\bibitem[{Yuan et~al.(2025)Yuan, Bai, Fu, Zhou, Feng, Xu, Zhan, Karlsson and Lu}]{audit250312533}
Yuan H, Bai Y, Fu Y, Zhou B, Feng Y, Xu X, Zhan Y, Karlsson BF and Lu Z (2025) {Being-0: A Humanoid Robotic Agent with Vision-Language Models and Modular Skills}.
\newblock arXiv:2503.12533.
\newblock \urlprefix\url{https://arxiv.org/abs/2503.12533v1}.

\bibitem[{Yuan et~al.(2026)Yuan, Zuo, Huang, Zhang, Park, Yang, Zhang, Chen, Ouyang, He, Lv, Wu, Liu, Zhao, Luo, Li, Qian, Ge, Gao, Yang, Deng, Xu, Huang, Zhou, Xu, Liu, Abbeel and Ding}]{yuan2026unirobotagent}
Yuan J, Zuo Y, Huang X, Zhang W, Park K, Yang J, Zhang T, Chen R, Ouyang Y, He B, Lv X, Wu H, Liu J, Zhao J, Luo T, Li H, Qian C, Ge J, Gao Ha, Yang K, Deng J, Xu Q, Huang S, Zhou B, Xu M, Liu Z, Abbeel P and Ding N (2026) {Democratizing Robot-Use Agents: A Unified Harness with Open Models}.
\newblock arXiv:2611.01560.
\newblock \urlprefix\url{https://arxiv.org/abs/2611.01560}.
\newblock \href{https://www.ishannagpal.com/robot}{LIBERO distillation experiment}.

\bibitem[{Yuan et~al.(2025{\natexlab{cb}})Yuan, Ma, Yao, Yao, Tomizuka and Ding}]{xref250322122}
Yuan P, Ma A, Yao Y, Yao H, Tomizuka M and Ding M (2025{\natexlab{cb}}) {REMAC: Self-Reflective and Self-Evolving Multi-Agent Collaboration for Long-Horizon Robot Manipulation}.
\newblock \emph{arXiv preprint arXiv:2503.22122} \urlprefix\url{https://arxiv.org/abs/2503.22122}.

\bibitem[{Yuan et~al.(2025{\natexlab{cc}})Yuan, Duan, Blukis, Pumacay, Krishna, Murali, Mousavian and Fox}]{yuan2025robopoint}
Yuan W, Duan J, Blukis V, Pumacay W, Krishna R, Murali A, Mousavian A and Fox D (2025{\natexlab{cc}}) {RoboPoint}: A vision-language model for spatial affordance prediction in robotics.
\newblock In: \emph{Proceedings of The 8th Conference on Robot Learning}, \emph{Proceedings of Machine Learning Research}, volume 270. pp. 4005--4020.
\newblock \urlprefix\url{https://proceedings.mlr.press/v270/yuan25c.html}.

\bibitem[{Yuan et~al.(2025{\natexlab{cd}})Yuan, Cui, Chen, Dong, Ni, Kou, Liu, Li, Zheng and Hao}]{audit250508548}
Yuan Y, Cui H, Chen Y, Dong Z, Ni F, Kou L, Liu J, Li P, Zheng Y and Hao J (2025{\natexlab{cd}}) {From Seeing to Doing: Bridging Reasoning and Decision for Robotic Manipulation}.
\newblock arXiv preprint arXiv:2505.08548.
\newblock \urlprefix\url{https://arxiv.org/abs/2505.08548}.
\newblock Accepted to ICLR 2026.

\bibitem[{Yuan et~al.(2026{\natexlab{cb}})Yuan, Huang, Yao, Li, Zhang, Han, Li, Sun, Jia, Zhang, Liu, Liao, Hu, Wu, Li, Dong, Ni, Zheng, Gu, Ma, Tang, Hu and Hao}]{yuan2026embodiedr15}
Yuan Y, Huang Y, Yao X, Li Y, Zhang S, Han L, Li P, Sun J, Jia W, Zhang Z, Liu Y, Liao R, Hu Y, Wu Q, Li Y, Dong Z, Ni F, Zheng Y, Gu S, Ma Y, Tang H, Hu H and Hao J (2026{\natexlab{cb}}) {Embodied-R1.5: Evolving Physical Intelligence via Embodied Foundation Models}.
\newblock \urlprefix\url{https://arxiv.org/abs/2606.11324}.

\bibitem[{{YueBit}(2026)}]{robodiagharness2026}
{YueBit} (2026) {RoboDiag Harness}.
\newblock GitHub repository.
\newblock \urlprefix\url{https://github.com/YueBit/robodiag-harness}.

\bibitem[{Zandonati et~al.(2026)Zandonati, Kang, Lozano-Pérez and Kaelbling}]{zandonati2026sprout}
Zandonati B, Kang Y, Lozano-Pérez T and Kaelbling LP (2026) {Generalized Planning Interfaces for Agentic Robotics}.
\newblock CoRL 2026 Workshop on Agentic Robotics.
\newblock \urlprefix\url{https://openreview.net/forum?id=sU18hA2zbD}.

\bibitem[{Zeng et~al.(2022)Zeng, Attarian, Ichter, Choromanski, Wong, Welker, Tombari, Purohit, Ryoo, Sindhwani, Lee, Vanhoucke and Florence}]{audit220400598}
Zeng A, Attarian M, Ichter B, Choromanski K, Wong A, Welker S, Tombari F, Purohit A, Ryoo M, Sindhwani V, Lee J, Vanhoucke V and Florence P (2022) {Socratic Models: Composing Zero-Shot Multimodal Reasoning with Language}.
\newblock \emph{arXiv preprint arXiv:2204.00598} \doi{10.48550/arXiv.2204.00598}.
\newblock \urlprefix\url{https://arxiv.org/abs/2204.00598}.

\bibitem[{Zeng et~al.(2024)Zeng, Zhou, Liang, Liu, Li, Huang, Li, Hu and Sun}]{video2reward2024}
Zeng R, Zhou D, Liang Q, Liu J, Li H, Huang C, Li J, Hu X and Sun F (2024) {Video2Reward: Generating Reward Function from Videos for Legged Robot Behavior Learning}.
\newblock In: \emph{Proceedings of the 27th European Conference on Artificial Intelligence}, \emph{Frontiers in Artificial Intelligence and Applications}, volume 392. pp. 4369--4376.
\newblock \doi{10.3233/faia241014}.
\newblock \urlprefix\url{https://doi.org/10.3233/faia241014}.

\bibitem[{Zeng et~al.(2025)Zeng, Zhou, Li, Shi, Li, Chen, Ren and Li}]{zeng2025vifailback}
Zeng X, Zhou X, Li Y, Shi J, Li T, Chen L, Ren L and Li YL (2025) {Diagnose, Correct, and Learn from Manipulation Failures via Visual Symbols}.
\newblock \urlprefix\url{https://arxiv.org/abs/2512.02787}.

\bibitem[{Zeng et~al.(2026)Zeng, Li, Huang, WuDunn, Kovacs, Harseno, Lin, Hunley, Shentu, Wu, Li and Yan}]{zeng2026lexp}
Zeng Y, Li Y, Huang K, WuDunn M, Kovacs A, Harseno M, Lin K, Hunley P, Shentu F, Wu P, Li Z and Yan Z (2026) {Learning from Physical Experience: Context-Space Adaptation for Robotic Manipulation}.
\newblock \urlprefix\url{https://arxiv.org/abs/2610.41320}.

\bibitem[{Zeng et~al.(2024{\natexlab{cb}})Zeng, Mu and Shao}]{zeng2024selfalignment}
Zeng Y, Mu Y and Shao L (2024{\natexlab{cb}}) {Learning Reward for Robot Skills Using Large Language Models via Self-Alignment}.
\newblock In: \emph{Proceedings of the 41st International Conference on Machine Learning}, \emph{Proceedings of Machine Learning Research}, volume 235. pp. 58366--58386.
\newblock \urlprefix\url{https://proceedings.mlr.press/v235/zeng24d.html}.

\bibitem[{Zeng et~al.(2026{\natexlab{cb}})Zeng, Ding, Yang, Li and Liao}]{zeng2026dexsim2real}
Zeng Z, Ding F, Yang H, Li X and Liao Y (2026{\natexlab{cb}}) {DexSim2Real: Foundation Model-Guided Sim-to-Real Transfer for Generalizable Dexterous Manipulation}.
\newblock \emph{arXiv preprint arXiv:2605.05241} \urlprefix\url{https://arxiv.org/abs/2605.05241}.

\bibitem[{Zha et~al.(2023)Zha, Cui, Lin, Kwon, Gonzalez~Arenas, Zeng, Xia and Sadigh}]{zha2023droc}
Zha L, Cui Y, Lin LH, Kwon M, Gonzalez~Arenas M, Zeng A, Xia F and Sadigh D (2023) {Distilling and Retrieving Generalizable Knowledge for Robot Manipulation via Language Corrections}.
\newblock \emph{arXiv preprint arXiv:2311.10678} \urlprefix\url{https://arxiv.org/abs/2311.10678}.

\bibitem[{Zhan et~al.(2025)Zhan, Wang, Liu, Peng, Wang, Wang, Ruan, Shi, Cao, Yang, Ni, Wang, Zhang, Shao, Li and Zhu}]{zhan2025sentinel}
Zhan SS, Wang P, Liu Y, Peng Y, Wang Z, Wang Q, Ruan Z, Shi X, Cao X, Yang F, Ni Z, Wang K, Zhang R, Shao H, Li M and Zhu Q (2025) {SENTINEL: A Multi-Level Formal Framework for Safety Evaluation of Foundation Model-based Embodied Agents}.
\newblock \urlprefix\url{https://arxiv.org/abs/2510.12985}.

\bibitem[{Zhan et~al.(2024)Zhan, Yu, Yu and Tan}]{xref240510620}
Zhan Z, Yu L, Yu S and Tan G (2024) {MC-GPT: Empowering Vision-and-Language Navigation with Memory Map and Reasoning Chains}.
\newblock \emph{arXiv preprint arXiv:2405.10620} \urlprefix\url{https://arxiv.org/abs/2405.10620}.

\bibitem[{Zhang et~al.(2026)Zhang, Yang, Wang, Shi, Zhou, Zhang and Wang}]{zhang2026usdcraft}
Zhang C, Yang Z, Wang D, Shi L, Zhou D, Zhang R and Wang Z (2026) {USDCraft: Geometrically Grounded Programmatic Modeling of Articulated 3D Assets for Simulation}.
\newblock \emph{arXiv preprint arXiv:2610.11322} \doi{10.48550/arXiv.2610.11322}.
\newblock \urlprefix\url{https://arxiv.org/abs/2610.11322}.

\bibitem[{Zhang et~al.(2026{\natexlab{cb}})Zhang, Zhang, He, Yang, Wang, Liu, Sycara and Xie}]{zhang2026morse}
Zhang C, Zhang J, He J, Yang X, Wang P, Liu H, Sycara K and Xie Y (2026{\natexlab{cb}}) {MORSE: Optimizing Streaming Memory Design via Open-World Auto-Research}.
\newblock \emph{arXiv preprint arXiv:2611.30011} \urlprefix\url{https://arxiv.org/abs/2611.30011}.

\bibitem[{Zhang et~al.(2023)Zhang, Du, Shan, Zhou, Du, Tenenbaum, Shu and Gan}]{coela}
Zhang H, Du W, Shan J, Zhou Q, Du Y, Tenenbaum JB, Shu T and Gan C (2023) {Building Cooperative Embodied Agents Modularly with Large Language Models}.
\newblock arXiv:2307.02485.
\newblock \urlprefix\url{https://arxiv.org/abs/2307.02485}.
\newblock Accepted to ICLR 2024.

\bibitem[{Zhang et~al.(2025)Zhang, Huang, Solak and Ajoudani}]{zhang2025omnivic}
Zhang H, Huang WH, Solak G and Ajoudani A (2025) {OmniVIC: A Self-Improving Variable Impedance Controller with Vision-Language In-Context Learning for Safe Robotic Manipulation}.
\newblock \doi{10.48550/arXiv.2510.17150}.
\newblock \urlprefix\url{https://arxiv.org/abs/2510.17150}.

\bibitem[{Zhang et~al.(2026{\natexlab{cc}})Zhang, Ge, Yoo, Fu, Yang, Liu, Saravanan, Yin, Yu, Niu, Wang, Herzig, Goldberg, Bai, Chan, Stoica, Kanazawa, Lei, Feng and Darrell}]{rats2026}
Zhang J, Ge J, Yoo H, Fu L, Yang Z, Liu Y, Saravanan R, Yin S, Yu J, Niu D, Wang Z, Herzig R, Goldberg K, Bai Y, Chan DM, Stoica I, Kanazawa A, Lei J, Feng H and Darrell T (2026{\natexlab{cc}}) {Playful Agentic Robot Learning}.
\newblock \emph{arXiv preprint arXiv:2606.19419} \urlprefix\url{https://arxiv.org/abs/2606.19419}.

\bibitem[{Zhang et~al.(2024)Zhang, Lampe, Abdolmaleki, Springenberg and Riedmiller}]{gameon2024}
Zhang J, Lampe T, Abdolmaleki A, Springenberg JT and Riedmiller M (2024) {Game On: Towards Language Models as RL Experimenters}.
\newblock \urlprefix\url{https://arxiv.org/abs/2409.03402}.

\bibitem[{Zhang et~al.(2026{\natexlab{cd}})Zhang, Wang and Lin}]{zhang2026debugweights}
Zhang J, Wang K and Lin L (2026{\natexlab{cd}}) {Can LLMs Debug Robots by Editing Their Weights?}
\newblock \emph{arXiv preprint arXiv:2611.09376} \urlprefix\url{https://arxiv.org/abs/2611.09376}.

\bibitem[{Zhang et~al.(2026{\natexlab{ce}})Zhang, Wu, Shi, Diller, Sukhatme and Kumar}]{zhang2026eflux}
Zhang J, Wu Y, Shi G, Diller J, Sukhatme GS and Kumar V (2026{\natexlab{ce}}) {EFLUX: Elastic Multi-Robot Formation Navigation and Adaptation with Agentic LLMs}.
\newblock arXiv:2607.12050.
\newblock \urlprefix\url{https://arxiv.org/abs/2607.12050}.

\bibitem[{Zhang et~al.(2026{\natexlab{cf}})Zhang, Xu, Yu, Guo, Hu, Bao, Zhou and Peng}]{zhang2026supernav}
Zhang J, Xu J, Yu Y, Guo J, Hu R, Bao H, Zhou X and Peng S (2026{\natexlab{cf}}) {SuperNav: An Agentic Navigation System for Any Task in Any Scene}.
\newblock \emph{arXiv preprint arXiv:2610.12126} \urlprefix\url{https://arxiv.org/abs/2610.12126}.

\bibitem[{Zhang et~al.(2025{\natexlab{cb}})Zhang, Zhang, Xiao and Hsu}]{zhang2025robot}
Zhang J, Zhang H, Xiao A and Hsu D (2025{\natexlab{cb}}) Robot operation of home appliances by reading user manuals.
\newblock \emph{arXiv preprint arXiv:2505.20424} .

\bibitem[{Zhang et~al.(2023{\natexlab{cb}})Zhang, Zhang, Pertsch, Liu, Ren, Chang, Sun and Lim}]{audit231010021}
Zhang J, Zhang J, Pertsch K, Liu Z, Ren X, Chang M, Sun SH and Lim JJ (2023{\natexlab{cb}}) Bootstrap your own skills: Learning to solve new tasks with large language model guidance.
\newblock In: \emph{Proceedings of The 7th Conference on Robot Learning}, \emph{Proceedings of Machine Learning Research}, volume 229. PMLR, pp. 302--325.
\newblock \doi{10.48550/arXiv.2310.10021}.
\newblock \urlprefix\url{https://proceedings.mlr.press/v229/zhang23a.html}.

\bibitem[{Zhang et~al.(2026{\natexlab{cg}})Zhang, Zhang, Song, Zhou, Song and Wang}]{zhang2026re0}
Zhang J, Zhang X, Song R, Zhou H, Song C and Wang H (2026{\natexlab{cg}}) {RE-0: Verified Recursive Improvement of Embodied Code-as-Policy Agents through Local On-Policy Distillation}.
\newblock \emph{arXiv preprint arXiv:2609.32416} \urlprefix\url{https://arxiv.org/abs/2609.32416}.

\bibitem[{Zhang et~al.(2026{\natexlab{ch}})Zhang, Zhao, Yang, Foerster, Clune, Jiang, Devlin and Shavrina}]{si260319461}
Zhang J, Zhao B, Yang W, Foerster J, Clune J, Jiang M, Devlin S and Shavrina T (2026{\natexlab{ch}}) {Hyperagents}.
\newblock \urlprefix\url{https://arxiv.org/abs/2603.19461}.

\bibitem[{Zhang et~al.(2026{\natexlab{ci}})Zhang, Zhou, Yin, Huang, Lei, Peng, Yuan, Zhang, Guo, Chen, Yang, Huang, Yang, Lin, Liu, Zhou, Yu, Fan, Liang, Lin, Wang, Li, Chen, Yan, Xu, Li, Hu, Zhang, Li, Xiao, Bai, Ren, Lv, Wu and Chen}]{audit260618112}
Zhang J, Zhou G, Yin H, Huang Y, Lei Z, Peng Q, Yuan H, Zhang J, Guo X, Chen X, Yang A, Huang F, Yang Z, Lin J, Liu D, Zhou J, Yu Z, Fan J, Liang Z, Lin P, Wang Y, Li H, Chen A, Yan K, Xu X, Li J, Hu L, Zhang M, Li S, Xiao W, Bai S, Ren X, Lv C, Wu C and Chen XH (2026{\natexlab{ci}}) {Qwen-RobotNav Technical Report: A Scalable Navigation Model Designed for an Agentic Navigation System}.
\newblock arXiv:2606.18112.
\newblock \urlprefix\url{https://arxiv.org/abs/2606.18112}.

\bibitem[{Zhang et~al.(2026{\natexlab{cj}})Zhang, Ma, Lv, Wang, Ruan, Si, Zhang, Chen, Chen, Liu, Zhang, Deng, Xie and Zhou}]{zhang2026datafarm}
Zhang K, Ma J, Lv G, Wang D, Ruan F, Si R, Zhang C, Chen Y, Chen H, Liu Y, Zhang Y, Deng Y, Xie E and Zhou D (2026{\natexlab{cj}}) {PyRUA-DataFarm: LLM Agent as a Cost-Efficient Teleoperator for Real-Ready Robot Data}.
\newblock \emph{arXiv preprint arXiv:2611.05320} \urlprefix\url{https://arxiv.org/abs/2611.05320}.

\bibitem[{Zhang et~al.(2025{\natexlab{cc}})Zhang, Xu, Ren, Lin, Wu, Lin and Liang}]{grid44}
Zhang K, Xu R, Ren P, Lin J, Wu H, Lin L and Liang X (2025{\natexlab{cc}}) {RoBridge: A Hierarchical Architecture Bridging Cognition and Execution for General Robotic Manipulation}.
\newblock In: \emph{2025 IEEE/CVF International Conference on Computer Vision (ICCV)}. pp. 14590--14601.
\newblock \doi{10.1109/iccv51701.2025.01354}.
\newblock \urlprefix\url{https://doi.org/10.1109/iccv51701.2025.01354}.

\bibitem[{Zhang et~al.(2025{\natexlab{cd}})Zhang, Dong, Bai, Ni, Fu, Marton, Chen and Zhang}]{zhang2025responsiblerobotbench}
Zhang L, Dong J, Bai K, Ni M, Fu Y, Marton ZC, Chen Z and Zhang J (2025{\natexlab{cd}}) {ResponsibleRobotBench: Benchmarking Risk-Aware Task-Level Decision Making for Robotic Manipulation with Large Multimodal Models}.
\newblock \doi{10.48550/arXiv.2512.04308}.
\newblock \urlprefix\url{https://arxiv.org/abs/2512.04308}.

\bibitem[{Zhang et~al.(2026{\natexlab{ck}})Zhang, Tang, Zhan, Chen, Chen, Han, Zhu, Xu, Xu, Wu, Lin and Liang}]{zhang2026atomicvla}
Zhang L, Tang T, Zhan Z, Chen X, Chen Z, Han J, Zhu J, Xu P, Xu H, Wu H, Lin L and Liang X (2026{\natexlab{ck}}) {AtomicVLA: Unlocking the Potential of Atomic Skill Learning in Robots}.
\newblock \emph{arXiv preprint arXiv:2603.07648} \doi{10.48550/arXiv.2603.07648}.
\newblock \urlprefix\url{https://arxiv.org/abs/2603.07648}.
\newblock Accepted to CVPR 2026.

\bibitem[{Zhang and Shah(2026)}]{zhang2026veritas}
Zhang M and Shah D (2026) {Visual Verification Enables Inference-time Steering and Autonomous Policy Improvement}.
\newblock \emph{arXiv preprint arXiv:2606.18247} \urlprefix\url{https://arxiv.org/abs/2606.18247}.

\bibitem[{Zhang et~al.(2025{\natexlab{ce}})Zhang, Xie, Zhang, Chen, Tan, Wan and Li}]{zhang2025adadrive}
Zhang R, Xie J, Zhang W, Chen W, Tan X, Wan X and Li G (2025{\natexlab{ce}}) {AdaDrive: Self-Adaptive Slow-Fast System for Language-Grounded Autonomous Driving}.
\newblock \doi{10.48550/arXiv.2511.06253}.
\newblock \urlprefix\url{https://arxiv.org/abs/2511.06253}.
\newblock Accepted to ICCV 2025.

\bibitem[{Zhang et~al.(2025{\natexlab{cf}})Zhang, Tian, Zhu, Huang, Yang and Zhang}]{zhang2025drivegen}
Zhang S, Tian J, Zhu Z, Huang S, Yang J and Zhang W (2025{\natexlab{cf}}) {DriveGen: Towards Infinite Diverse Traffic Scenarios with Large Models}.
\newblock \doi{10.48550/arXiv.2503.05808}.
\newblock \urlprefix\url{https://arxiv.org/abs/2503.05808}.

\bibitem[{Zhang et~al.(2023{\natexlab{cc}})Zhang, Wicke, {\c S}enel, Figueredo, Naceri, Haddadin, Plank and Sch{\"u}tze}]{audit231012020}
Zhang S, Wicke P, {\c S}enel LK, Figueredo L, Naceri A, Haddadin S, Plank B and Sch{\"u}tze H (2023{\natexlab{cc}}) {LoHoRavens: A Long-Horizon Language-Conditioned Benchmark for Robotic Tabletop Manipulation}.
\newblock \emph{arXiv preprint arXiv:2310.12020} \doi{10.48550/arXiv.2310.12020}.
\newblock \urlprefix\url{https://arxiv.org/abs/2310.12020}.

\bibitem[{Zhang et~al.(2026{\natexlab{cl}})Zhang, Li and Ding}]{zhang2026benchmend}
Zhang T, Li Y and Ding M (2026{\natexlab{cl}}) {BenchMend: Agent-Guided Instruction Repair for Robotics Benchmarks}.
\newblock Project essay and evaluation release.
\newblock \urlprefix\url{https://tianqi-zh.github.io/BenchMend/}.
\newblock Accessed October 16, 2026.

\bibitem[{Zhang et~al.(2026{\natexlab{cm}})Zhang, Qu, Li, Wu, Hutter, Li and Shi}]{zhang2026despite}
Zhang T, Qu K, Li Z, Wu J, Hutter M, Li M and Shi F (2026{\natexlab{cm}}) {Using large language models for embodied planning introduces systematic safety risks}.
\newblock \urlprefix\url{https://arxiv.org/abs/2604.18463}.

\bibitem[{Zhang et~al.(2026{\natexlab{cn}})Zhang, Wang, Ouyang, Huang, Li, Su, Jin, Chai, Liang, Dou, Chen and Chen}]{roboprobe2026}
Zhang W, Wang K, Ouyang Y, Huang X, Li L, Su K, Jin W, Chai W, Liang H, Dou Z, Chen Y and Chen T (2026{\natexlab{cn}}) {An Unexpected Robot Policy: Early Evaluations of GPT-6 Astra on RoboDojo and Beyond}.
\newblock \emph{arXiv preprint arXiv:2609.24170} \urlprefix\url{https://arxiv.org/abs/2609.24170}.

\bibitem[{Zhang et~al.(2026{\natexlab{co}})Zhang, Huang and Du}]{audit260727599}
Zhang X, Huang R and Du Y (2026{\natexlab{co}}) {World Action Planner: Generalizable Robot Decision-Making with Action-Conditioned World Models}.
\newblock arXiv:2607.27599.
\newblock \urlprefix\url{https://arxiv.org/abs/2607.27599}.

\bibitem[{Zhang et~al.(2024{\natexlab{cb}})Zhang, Qin, Wang, Dong and Li}]{zhang2024lammap}
Zhang X, Qin H, Wang F, Dong Y and Li J (2024{\natexlab{cb}}) {LaMMA-P}: Generalizable multi-agent long-horizon task allocation and planning with {LM}-driven {PDDL} planner.
\newblock arXiv preprint arXiv:2409.20560.
\newblock \urlprefix\url{https://arxiv.org/abs/2409.20560}.
\newblock Accepted to ICRA 2025.

\bibitem[{Zhang et~al.(2026{\natexlab{cp}})Zhang, Wang, Bauer, Yu, Huang, Wu and Pavone}]{zhang2026robocop}
Zhang X, Wang M, Bauer E, Yu HX, Huang H, Wu J and Pavone M (2026{\natexlab{cp}}) {Co-Evolving Robot Orchestrators and Policies through Deployment}.
\newblock \urlprefix\url{https://arxiv.org/abs/2610.09228}.

\bibitem[{Zhang et~al.(2026{\natexlab{cq}})Zhang, Wang, Li, Yao and Li}]{zhang2026commcp}
Zhang X, Wang Z, Li Z, Yao J and Li J (2026{\natexlab{cq}}) {CommCP}: Efficient multi-agent coordination via {LLM}-based communication with conformal prediction.
\newblock arXiv preprint arXiv:2602.06038.
\newblock \urlprefix\url{https://arxiv.org/abs/2602.06038}.
\newblock Accepted to ICRA 2026.

\bibitem[{Zhang et~al.(2026{\natexlab{cr}})Zhang, Huang, Chang, Su, Zhou, Xu, Chen, Zhou, Wang, Zhang, Wei, Li, Deng, Li, Chen and Li}]{waa}
Zhang Y, Huang H, Chang Y, Su J, Zhou B, Xu Y, Chen W, Zhou T, Wang C, Zhang T, Wei Y, Li W, Deng S, Li Y, Chen YC and Li Z (2026{\natexlab{cr}}) {World Action Agent: Harnessing VLMs for Robot Manipulation via World Action Rehearsal}.
\newblock arXiv:2609.29964.
\newblock \urlprefix\url{https://arxiv.org/abs/2609.29964}.

\bibitem[{Zhang et~al.(2026{\natexlab{cs}})Zhang, Liu and Zhang}]{zhang2026localrecovery}
Zhang Y, Liu X and Zhang Y (2026{\natexlab{cs}}) {Recursive Self-Improvement of Visuomotor Policies through Local Recovery Supervision}.
\newblock \emph{arXiv preprint arXiv:2610.05151} \urlprefix\url{https://arxiv.org/abs/2610.05151}.

\bibitem[{Zhang et~al.(2026{\natexlab{ct}})Zhang, Wu, Gao, Zhao, Ren, Xu, Liao, Wang, Fan, Wu, Feng, Li, Che, Liu and Tang}]{robogene2026}
Zhang Y, Wu K, Gao Z, Zhao Z, Ren P, Xu Z, Liao F, Wang X, Fan S, Wu D, Feng Q, Li M, Che Z, Liu C and Tang J (2026{\natexlab{ct}}) {RoboGene: Diversity-Driven Agentic Task Generation Framework for VLA Pre-training}.
\newblock arXiv:2602.16444.
\newblock \urlprefix\url{https://arxiv.org/abs/2602.16444}.
\newblock Accepted to CoRL 2026.

\bibitem[{Zhang et~al.(2025{\natexlab{cg}})Zhang, Yang, Bai, Wu, Li, Wang and Li}]{zhang2025read}
Zhang Y, Yang S, Bai C, Wu F, Li X, Wang Z and Li X (2025{\natexlab{cg}}) Towards efficient {LLM} grounding for embodied multi-agent collaboration.
\newblock In: \emph{Findings of the Association for Computational Linguistics: ACL 2025}. pp. 1663--1699.
\newblock \doi{10.18653/v1/2025.findings-acl.84}.
\newblock \urlprefix\url{https://aclanthology.org/2025.findings-acl.84/}.

\bibitem[{Zhang et~al.(2026{\natexlab{cu}})Zhang, Zhang, Gao, Li, Liu, Nie, Zhu, Qiu, Yan, Liu, Tang, Rao, Fang, Wei, Wang, Ding and Yu}]{harnessvla}
Zhang Y, Zhang H, Gao F, Li X, Liu Z, Nie Y, Zhu C, Qiu J, Yan Y, Liu J, Tang W, Rao J, Fang Z, Wei C, Wang Y, Ding W and Yu C (2026{\natexlab{cu}}) {Harness VLA: Steering Frozen VLAs into Reliable Manipulation Primitives via Memory-Guided Agents}.
\newblock arXiv:2607.08448.
\newblock \urlprefix\url{https://arxiv.org/abs/2607.08448}.

\bibitem[{Zhang et~al.(2026{\natexlab{cv}})Zhang, Zhang, Li, Wang, Jin, Liu and Zhao}]{zhang2026simplearm}
Zhang Y, Zhang Y, Li M, Wang J, Jin Z, Liu S and Zhao D (2026{\natexlab{cv}}) {Simple Agentic Memory for Generalist Robot Policies}.
\newblock CoRL 2026 Workshop on Memory for Robotics.
\newblock \urlprefix\url{https://arxiv.org/abs/2609.36595}.

\bibitem[{Zhang et~al.(2026{\natexlab{cw}})Zhang, Jin, Wang, Zhang, Diao, Wang and Lu}]{audit260919554}
Zhang Z, Jin J, Wang Y, Zhang Z, Diao H, Wang L and Lu H (2026{\natexlab{cw}}) {VABench: Measuring Embodied Spatial Intelligence through Visual Demonstrations, Active Perception, and Metric Control}.
\newblock arXiv:2609.19554.
\newblock \urlprefix\url{https://arxiv.org/abs/2609.19554v1}.

\bibitem[{Zhang et~al.(2025{\natexlab{ch}})Zhang, Yue, Xu, Liao, Qi, Gao, Wang and Zhao}]{zhang2025robochemist}
Zhang Z, Yue C, Xu H, Liao M, Qi X, Gao Ha, Wang Z and Zhao H (2025{\natexlab{ch}}) {RoboChemist: Long-Horizon and Safety-Compliant Robotic Chemical Experimentation}.
\newblock \doi{10.48550/arXiv.2509.08820}.
\newblock \urlprefix\url{https://arxiv.org/abs/2509.08820}.
\newblock Accepted to CoRL 2025.

\bibitem[{Zhang et~al.(2026{\natexlab{cx}})Zhang, Zhu, Chen, Tuya and James}]{zhang2026bigym2}
Zhang Z, Zhu Z, Chen Z, Tuya Z and James S (2026{\natexlab{cx}}) {BiGym 2.0: Benchmarking Learned and Agent-Developed Policies for Humanoid Household Manipulation}.
\newblock \urlprefix\url{https://arxiv.org/abs/2610.07594}.

\bibitem[{Zhao et~al.(2026)Zhao, Yue, He and Wang}]{zhao2026roboace}
Zhao D, Yue Z, He J and Wang Y (2026) {RoboACE: Adversarial Co-Evolution for Robot Self-Improvement}.
\newblock \urlprefix\url{https://arxiv.org/abs/2611.13034}.

\bibitem[{Zhao et~al.(2026{\natexlab{cb}})Zhao, Tang, Chen, Kuang, Tian and Li}]{audit260826645}
Zhao G, Tang Z, Chen X, Kuang Z, Tian Y and Li G (2026{\natexlab{cb}}) {FLARE: A Failure-Aware Framework for Autonomous Correction and Recovery in Visual-Language Robotic Manipulation}.
\newblock arXiv:2608.26645.
\newblock \urlprefix\url{https://arxiv.org/abs/2608.26645v1}.
\newblock Accepted to CVPR 2026.

\bibitem[{Zhao et~al.(2024)Zhao, Wang, Zhu, Chen, Huang, Bao and Wang}]{zhao2024drivedreamer2}
Zhao G, Wang X, Zhu Z, Chen X, Huang G, Bao X and Wang X (2024) {DriveDreamer-2: LLM-Enhanced World Models for Diverse Driving Video Generation}.
\newblock \doi{10.48550/arXiv.2403.06845}.
\newblock \urlprefix\url{https://arxiv.org/abs/2403.06845}.

\bibitem[{Zhao et~al.(2025)Zhao, Zeng, Zhuang, Zhao, Xue, Wang, Zhao, Li, Li, Huang, Chen, Li, Zhao and Zou}]{robosimgs2025}
Zhao H, Zeng C, Zhuang L, Zhao Y, Xue S, Wang H, Zhao X, Li Z, Li K, Huang S, Chen M, Li X, Zhao D and Zou H (2025) {High-Fidelity Simulated Data Generation for Real-World Zero-Shot Robotic Manipulation Learning with Gaussian Splatting}.
\newblock arXiv:2510.10637.
\newblock \urlprefix\url{https://arxiv.org/abs/2510.10637}.

\bibitem[{Zhao et~al.(2025{\natexlab{cb}})Zhao, Zhang, Song, Ding and Wang}]{zhao2025vla2}
Zhao H, Zhang J, Song W, Ding P and Wang D (2025{\natexlab{cb}}) {VLA$^2$: Empowering Vision-Language-Action Models with an Agentic Framework for Unseen Concept Manipulation}.
\newblock \doi{10.48550/arXiv.2510.14902}.
\newblock \urlprefix\url{https://arxiv.org/abs/2510.14902}.

\bibitem[{Zhao et~al.(2026{\natexlab{cc}})Zhao, Huang, Mao, Liu, Kochenderfer and Wong}]{zhao2026retriever}
Zhao L, Huang H, Mao J, Liu W, Kochenderfer M and Wong LLS (2026{\natexlab{cc}}) {Retriever: Composing the Perception-Reasoning-Action Loop for Long-Horizon Manipulation}.
\newblock \emph{arXiv preprint arXiv:2607.17213} \urlprefix\url{https://arxiv.org/abs/2607.17213}.

\bibitem[{Zhao et~al.(2026{\natexlab{cd}})Zhao, Xie, Zhao, Zhang, Wang, Wang, Liu and Zhang}]{audit260911561}
Zhao S, Xie H, Zhao W, Zhang C, Wang H, Wang C, Liu Q and Zhang S (2026{\natexlab{cd}}) {Memory as Plans: World-Action Modeling with Memory-Grounded Planning}.
\newblock \emph{arXiv preprint arXiv:2609.11561} \doi{10.48550/arXiv.2609.11561}.
\newblock \urlprefix\url{https://arxiv.org/abs/2609.11561}.

\bibitem[{Zhao et~al.(2024{\natexlab{cb}})Zhao, Weber and Wermter}]{zhao2024agenticskilldiscovery}
Zhao X, Weber C and Wermter S (2024{\natexlab{cb}}) {Agentic Skill Discovery}.
\newblock \emph{arXiv preprint arXiv:2405.15019} \urlprefix\url{https://arxiv.org/abs/2405.15019}.

\bibitem[{Zhao et~al.(2026{\natexlab{ce}})Zhao, Zhou, Lin, Zhou, Li, Yan, Liu, van~den Hengel and Wu}]{zhao2026navharness}
Zhao X, Zhou J, Lin S, Zhou G, Li Z, Yan X, Liu J, van~den Hengel A and Wu Q (2026{\natexlab{ce}}) {NavHarness: Towards Lifelong Embodied Navigation}.
\newblock arXiv:2609.34276.
\newblock \urlprefix\url{https://arxiv.org/abs/2609.34276}.

\bibitem[{Zhao et~al.(2026{\natexlab{cf}})Zhao, Ni, Chen, Zhang and Zhu}]{zhao2026hymes}
Zhao Y, Ni Z, Chen H, Zhang R and Zhu Q (2026{\natexlab{cf}}) {Skills in Weights, Memory in Code: Hybrid Learning for Memory-Dependent Robot Manipulation}.
\newblock \emph{arXiv preprint arXiv:2608.09410} \urlprefix\url{https://arxiv.org/abs/2608.09410}.

\bibitem[{Zhao et~al.(2026{\natexlab{cg}})Zhao, Hua, Berto, Ma, Lee, Li and Park}]{evonav2026}
Zhao Z, Hua C, Berto F, Ma Z, Lee K, Li J and Park J (2026{\natexlab{cg}}) {EvoNav: Evolutionary Reward Function Design for Robot Navigation with Large Language Models}.
\newblock arXiv:2605.11859.
\newblock \urlprefix\url{https://arxiv.org/abs/2605.11859}.

\bibitem[{Zheng et~al.(2026)Zheng, Ban, Zhang, Zheng and Zhou}]{audit260310682}
Zheng G, Ban Y, Zhang M, Zheng J and Zhou B (2026) {OnFly: Onboard Zero-Shot Aerial Vision-Language Navigation toward Safety and Efficiency}.
\newblock arXiv:2603.10682.
\newblock \urlprefix\url{https://arxiv.org/abs/2603.10682v1}.

\bibitem[{Zhi et~al.(2024)Zhi, Zhang, Zhao, Han, Zhang, Li, Jiao, Jia and Huang}]{zhi2024comerobot}
Zhi P, Zhang Z, Zhao Y, Han M, Zhang Z, Li Z, Jiao Z, Jia B and Huang S (2024) {Closed-Loop Open-Vocabulary Mobile Manipulation with GPT-4V}.
\newblock \emph{arXiv preprint arXiv:2404.10220} \urlprefix\url{https://arxiv.org/abs/2404.10220}.
\newblock Accepted to ICRA 2025.

\bibitem[{Zhong et~al.(2026)Zhong, Wu, Yu, Jiang, Li, Xiao, Dai, Wang, Zou, Cui, Zheng, Xiong, Zhang, Gao, Morley, Nakayama, Xu, Shen, Gui, Celi, Yin, Wu and Yang}]{zhong2026discoveryrsi}
Zhong J, Wu C, Yu Z, Jiang X, Li X, Xiao H, Dai X, Wang J, Zou J, Cui Z, Zheng B, Xiong J, Zhang Z, Gao Q, Morley M, Nakayama L, Xu Q, Shen L, Gui T, Celi LA, Yin Z, Wu Y and Yang L (2026) {DiscoveryRSI: Recursively Learning to Discover Valuable Hypotheses}.
\newblock \emph{arXiv preprint arXiv:2610.23839} \urlprefix\url{https://arxiv.org/abs/2610.23839}.

\bibitem[{Zhong et~al.(2025)Zhong, Huang, Li, Zhang, Chen, Guan, Zeng, Lui, Ye, Liang, Yang and Chen}]{zhong2025dexgraspvla}
Zhong Y, Huang X, Li R, Zhang C, Chen Z, Guan T, Zeng F, Lui KN, Ye Y, Liang Y, Yang Y and Chen Y (2025) {DexGraspVLA: A Vision-Language-Action Framework Towards General Dexterous Grasping}.
\newblock \doi{10.48550/arXiv.2502.20900}.
\newblock \urlprefix\url{https://arxiv.org/abs/2502.20900}.

\bibitem[{Zhou et~al.(2026)Zhou, Wu, Zhou, Huo, Gao, Zhu, Zhang and Chen}]{zhou2026simuguard}
Zhou A, Wu J, Zhou X, Huo Y, Gao H, Zhu D, Zhang Z and Chen Y (2026) {When the Simulator Fails, Not the Policy: Auditing Contact Artifacts in Simulated Robot Benchmarks}.
\newblock \urlprefix\url{https://arxiv.org/abs/2610.41857}.

\bibitem[{Zhou et~al.(2026{\natexlab{cb}})Zhou, Chen, Huang, Ruan, Huang, Zhang, Li, Li, Yu, Song, Wu, Li, Song, Xu, Qiu, Wei, Zhang, Zhou, Li and Cheng}]{zhou2026roboaware}
Zhou B, Chen X, Huang E, Ruan W, Huang H, Zhang Y, Li Z, Li W, Yu Q, Song Z, Wu L, Li J, Song M, Xu X, Qiu Z, Wei Y, Zhang T, Zhou K, Li Y and Cheng J (2026{\natexlab{cb}}) {RoboAware: Learning to Coordinate Embodied Skills from Counterfactual Outcomes}.
\newblock \doi{10.48550/arXiv.2610.11480}.
\newblock \urlprefix\url{https://arxiv.org/abs/2610.11480}.

\bibitem[{Zhou et~al.(2025)Zhou, Su, Chi, Zhang, Wang, Huang, Sheng and Wang}]{zhou2024codeasmonitor}
Zhou E, Su Q, Chi C, Zhang Z, Wang Z, Huang T, Sheng L and Wang H (2025) {Code-as-Monitor: Constraint-aware Visual Programming for Reactive and Proactive Robotic Failure Detection}.
\newblock In: \emph{Proceedings of the IEEE/CVF Conference on Computer Vision and Pattern Recognition}. pp. 6919--6929.
\newblock \urlprefix\url{https://openaccess.thecvf.com/content/CVPR2025/html/Zhou_Code-as-Monitor_Constraint-aware_Visual_Programming_for_Reactive_and_Proactive_Robotic_Failure_CVPR_2025_paper.html}.

\bibitem[{Zhou et~al.(2023)Zhou, Hong and Wu}]{xref230516986}
Zhou G, Hong Y and Wu Q (2023) {NavGPT: Explicit Reasoning in Vision-and-Language Navigation with Large Language Models}.
\newblock \emph{arXiv preprint arXiv:2305.16986} \urlprefix\url{https://arxiv.org/abs/2305.16986}.

\bibitem[{Zhou et~al.(2026{\natexlab{cc}})Zhou, Hong, Zhang, Zhao, Zhou, Lei, Wang, Zhao, Chen, Gould, Hengel and Wu}]{zhou2026navgpt3}
Zhou G, Hong Y, Zhang J, Zhao X, Zhou J, Lei Z, Wang Z, Zhao C, Chen X, Gould S, Hengel Avd and Wu Q (2026{\natexlab{cc}}) {NavGPT-3: Harnessing Context in a Hierarchical Navigation Runtime}.
\newblock \emph{arXiv preprint arXiv:2610.10787} \urlprefix\url{https://arxiv.org/abs/2610.10787}.

\bibitem[{Zhou et~al.(2026{\natexlab{cd}})Zhou, Gao, Wang, Lynch, Zhao, Ma, Goyal and Zhang}]{audit261002161}
Zhou H, Gao D, Wang H, Lynch B, Zhao B, Ma Q, Goyal R and Zhang J (2026{\natexlab{cd}}) {DuoMind: Enabling Distributed Multi-Robot Coordination with Semantic Communication}.
\newblock arXiv:2610.02161.
\newblock \urlprefix\url{https://arxiv.org/abs/2610.02161v1}.

\bibitem[{Zhou et~al.(2026{\natexlab{ce}})Zhou, Lin, Li, Fu, Zhou and Wu}]{zhou2026agentcanvas}
Zhou J, Lin S, Li J, Fu S, Zhou G and Wu Q (2026{\natexlab{ce}}) {Automating the Design of Embodied Agent Architectures}.
\newblock \emph{arXiv preprint arXiv:2606.30111} \urlprefix\url{https://arxiv.org/abs/2606.30111}.
\newblock Accepted to CoRL 2026.

\bibitem[{Zhou et~al.(2026{\natexlab{cf}})Zhou, Zhao, Zhou, Li, Lin, Liu and Wu}]{grid30}
Zhou J, Zhao X, Zhou G, Li Z, Lin S, Liu J and Wu Q (2026{\natexlab{cf}}) {Embodied Agents Take Control: Minimal-Interface Zero-Shot Agents Rival Industrial-Scale Policies in Vision-and-Language Navigation}.
\newblock arXiv:2607.26148.
\newblock \urlprefix\url{https://arxiv.org/abs/2607.26148}.

\bibitem[{Zhou et~al.(2026{\natexlab{cg}})Zhou, Li, Lyu, Song, Huang, Zheng, Rupprecht, Vedaldi and Wu}]{articraft2026}
Zhou M, Li R, Lyu X, Song Z, Huang Z, Zheng C, Rupprecht C, Vedaldi A and Wu S (2026{\natexlab{cg}}) {Articraft: An Agentic System for Scalable Articulated 3D Asset Generation}.
\newblock \urlprefix\url{https://arxiv.org/abs/2605.15187}.

\bibitem[{Zhou et~al.(2026{\natexlab{ch}})Zhou, Ma, Zhang, Chen, Wan, Wang, Zhao and You}]{zhou2026rea}
Zhou P, Ma Y, Zhang Z, Chen H, Wan Z, Wang H, Zhao W and You Y (2026{\natexlab{ch}}) {REA: A Routed Embodied Agent Runtime for Asynchronous Brain-Cerebellum Control}.
\newblock \urlprefix\url{https://arxiv.org/abs/2610.42053}.

\bibitem[{Zhou et~al.(2026{\natexlab{ci}})Zhou, Wu, Li, Zheng, Zhang, Yu, Yang, Liu, Sun, Yang, Jiang, Su, Huang and Tian}]{zhou2026evowam}
Zhou S, Wu X, Li W, Zheng S, Zhang J, Yu S, Yang Y, Liu J, Sun H, Yang S, Jiang L, Su J, Huang H and Tian Z (2026{\natexlab{ci}}) {EVO-WAM: Evolving World Action Models through Video-Action Verification}.
\newblock \urlprefix\url{https://arxiv.org/abs/2609.38057}.

\bibitem[{Zhou et~al.(2026{\natexlab{cj}})Zhou, Cai, Chen, Yuan, Mu, Ma, Dong, Lu, Bai and Yang}]{zhou2026navigent}
Zhou X, Cai W, Chen Y, Yuan S, Mu Z, Ma B, Dong J, Lu Y, Bai J and Yang J (2026{\natexlab{cj}}) {Navigent: A Geometry-Grounded Agentic Framework for Vision Language Navigation}.
\newblock \urlprefix\url{https://arxiv.org/abs/2611.34249}.

\bibitem[{Zhou et~al.(2026{\natexlab{ck}})Zhou, Cai, Wang, Ju, Ren, Liu, Zhang, Luo, Chen, Zhou, Ma and Yang}]{zhou2026imitatorgame}
Zhou X, Cai Y, Wang F, Ju R, Ren H, Liu R, Zhang Y, Luo Q, Chen F, Zhou P, Ma Y and Yang Y (2026{\natexlab{ck}}) {The Imitator Game: Benchmarking Robot Imitative Ability Beyond Action Prediction}.
\newblock \emph{arXiv preprint arXiv:2608.22301} \urlprefix\url{https://arxiv.org/abs/2608.22301}.

\bibitem[{Zhou et~al.(2026{\natexlab{cl}})Zhou, Liu, Xiao, Feng, Fu, Meng, Wang, Han, Yu, Du, Sui and Su}]{holoagent}
Zhou X, Liu L, Xiao T, Feng W, Fu F, Meng X, Wang X, Han J, Yu B, Du Y, Sui W and Su Z (2026{\natexlab{cl}}) {HoloAgent-0: A Unified Embodied Agent Framework with 3D Spatial Memory}.
\newblock arXiv:2606.23565.
\newblock \urlprefix\url{https://arxiv.org/abs/2606.23565}.

\bibitem[{Zhou et~al.(2026{\natexlab{cm}})Zhou, Sun, Gong, Tie, Zhou, Sun and Chen}]{embodiedclaw2026}
Zhou X, Sun Y, Gong X, Tie G, Zhou P, Sun L and Chen Y (2026{\natexlab{cm}}) {EmbodiedClaw: Conversational Workflow Execution for Embodied AI Development}.
\newblock arXiv:2604.13800.
\newblock \urlprefix\url{https://arxiv.org/abs/2604.13800}.

\bibitem[{Zhou et~al.(2026{\natexlab{cn}})Zhou, Wang, Li, Hu, Tie, Wan, Liu, Zhou, Sun and Chen}]{zhou2026embodiedtool}
Zhou X, Wang Z, Li Q, Hu Y, Tie G, Wan L, Liu Y, Zhou P, Sun L and Chen Y (2026{\natexlab{cn}}) {Enabling Extensible Embodied Capabilities with Tools}.
\newblock \emph{arXiv preprint arXiv:2605.26637} \urlprefix\url{https://arxiv.org/abs/2605.26637}.

\bibitem[{Zhou et~al.(2026{\natexlab{co}})Zhou, Hou, Fu, Lyu, Ye, Shan, Guo, Liu, He, Shan, Zhang, Yang and Lu}]{zhou2026gptgear}
Zhou Y, Hou B, Fu Y, Lyu Y, Ye Y, Shan S, Guo C, Liu J, He R, Shan C, Zhang S, Yang J and Lu Z (2026{\natexlab{co}}) {Large Multimodal Models for Direct Robot Control: GPT-6 as a Beginning}.
\newblock \emph{arXiv preprint arXiv:2610.28259} \doi{10.48550/arXiv.2610.28259}.
\newblock \urlprefix\url{https://arxiv.org/abs/2610.28259}.

\bibitem[{Zhou et~al.(2025{\natexlab{cb}})Zhou, Atreya, Lee, Walke, Mees and Levine}]{soar2024}
Zhou Z, Atreya P, Lee A, Walke HR, Mees O and Levine S (2025{\natexlab{cb}}) Autonomous improvement of instruction following skills via foundation models.
\newblock In: \emph{Proceedings of The 8th Conference on Robot Learning}, \emph{Proceedings of Machine Learning Research}, volume 270. PMLR, pp. 4805--4825.
\newblock \urlprefix\url{https://proceedings.mlr.press/v270/zhou25b.html}.

\bibitem[{Zhou et~al.(2026{\natexlab{cp}})Zhou, Luo, Cao, Xiao, Peng, Li, Tian, Lian, Wang, Ma, Ivanovic, Pavone and Ding}]{zhou2026verifine}
Zhou Z, Luo R, Cao Y, Xiao C, Peng C, Li B, Tian T, Lian Z, Wang Y, Ma J, Ivanovic B, Pavone M and Ding W (2026{\natexlab{cp}}) {VeriFine: Scaling Verification for Self-Improvement in Embodied Reasoning}.
\newblock \emph{arXiv preprint arXiv:2610.08761} \doi{10.48550/arXiv.2610.08761}.
\newblock \urlprefix\url{https://arxiv.org/abs/2610.08761}.

\bibitem[{Zhu et~al.(2025)Zhu, Tanaka, Goldberg and Hong}]{auracurriculum2025}
Zhu A, Tanaka Y, Goldberg A and Hong D (2025) {AURA: Autonomous Upskilling with Retrieval-Augmented Agents}.
\newblock arXiv:2506.02507.
\newblock \urlprefix\url{https://arxiv.org/abs/2506.02507}.

\bibitem[{Zhu et~al.(2025{\natexlab{cb}})Zhu, Zhou, Ji and Zhao}]{audit250601538}
Zhu G, Zhou R, Ji W and Zhao S (2025{\natexlab{cb}}) {LAMARL: LLM-Aided Multi-Agent Reinforcement Learning for Cooperative Policy Generation}.
\newblock \emph{arXiv preprint arXiv:2506.01538} \doi{10.48550/arXiv.2506.01538}.
\newblock \urlprefix\url{https://arxiv.org/abs/2506.01538}.
\newblock Accepted to IEEE Robotics and Automation Letters.

\bibitem[{Zhu et~al.(2026)Zhu, Li, Liu, Guo, Lin, Cai, Chen, Lv, Wang, Oh and Zhang}]{grid33}
Zhu H, Li Z, Liu Z, Guo K, Lin Z, Cai Y, Chen G, Lv C, Wang W, Oh J and Zhang J (2026) {SysNav: Multi-Level Systematic Cooperation Enables Real-World, Cross-Embodiment Object Navigation}.
\newblock \emph{IEEE Robotics and Automation Letters} 11(9): 10577--10584.
\newblock \doi{10.1109/lra.2026.3713730}.
\newblock \urlprefix\url{https://doi.org/10.1109/lra.2026.3713730}.

\bibitem[{Zhu et~al.(2026{\natexlab{cb}})Zhu, Zhao, Cheng, Li and Fragkiadaki}]{zhu2026skillweaver}
Zhu H, Zhao L, Cheng KM, Li S and Fragkiadaki K (2026{\natexlab{cb}}) {SkillWeaver: Agentic Exploration over Neural Interaction Skills for Scalable Robot Data Generation}.
\newblock \doi{10.48550/arXiv.2609.36171}.
\newblock \urlprefix\url{https://arxiv.org/abs/2609.36171}.

\bibitem[{Zhu et~al.(2024)Zhu, Cano, Bermudez and Drozdzal}]{audit240205188}
Zhu JY, Cano CG, Bermudez DV and Drozdzal M (2024) {InCoRo: In-Context Learning for Robotics Control with Feedback Loops}.
\newblock arXiv:2402.05188.
\newblock \urlprefix\url{https://arxiv.org/abs/2402.05188v1}.

\bibitem[{Zhu et~al.(2024{\natexlab{cb}})Zhu, Zhu, Li, Wen, Xu, Che, Shen, Peng, Liu, Feng and Tang}]{zhu2024rfst}
Zhu M, Zhu Y, Li J, Wen J, Xu Z, Che Z, Shen C, Peng Y, Liu D, Feng F and Tang J (2024{\natexlab{cb}}) {Language-Conditioned Robotic Manipulation with Fast and Slow Thinking}.
\newblock \doi{10.48550/arXiv.2401.04181}.
\newblock \urlprefix\url{https://arxiv.org/abs/2401.04181}.
\newblock Accepted to ICRA 2024.

\bibitem[{Zhu et~al.(2026{\natexlab{cc}})Zhu, Zhao, Sun, Guo, Jiang, Zhang, Fan and Huang}]{zhu2026leash}
Zhu Q, Zhao W, Sun K, Guo X, Jiang Y, Zhang R, Fan L and Huang K (2026{\natexlab{cc}}) {LEASH}: Stay safe in agentic robotics.
\newblock \emph{arXiv preprint arXiv:2611.14115} \urlprefix\url{https://arxiv.org/abs/2611.14115}.

\bibitem[{Zhu et~al.(2026{\natexlab{cd}})Zhu, Guo, Mei, Pang, Zhang, He, Ji and Liu}]{si260514504}
Zhu Z, Guo L, Mei Y, Pang B, Zhang Z, He X, Ji R and Liu J (2026{\natexlab{cd}}) {When Robots Do the Chores: A Benchmark and Agent for Long-Horizon Household Task Execution}.
\newblock \urlprefix\url{https://arxiv.org/abs/2605.14504}.

\bibitem[{Zhu et~al.(2024{\natexlab{cc}})Zhu, Wu, Zhang, Han, Liu and Wu}]{zhu2024earbench}
Zhu Z, Wu B, Zhang Z, Han L, Liu Q and Wu B (2024{\natexlab{cc}}) {EARBench}: Towards evaluating physical risk awareness for task planning of foundation model-based embodied ai agents.
\newblock \urlprefix\url{https://arxiv.org/abs/2408.04449}.

\bibitem[{Zi et~al.(2026)Zi, Cheng, Song, Chen, Ji, Dai, Peng, Li, Qiu, Xu, Bai, Zhang, Li and Li}]{zi2026selfimprovingrobotic}
Zi C, Cheng S, Song W, Chen J, Ji Y, Dai L, Peng S, Li C, Qiu Z, Xu H, Bai Y, Zhang Z, Li H and Li J (2026) {Toward Self-Improving Robotic Agents: From Interaction Experience to Reusable Manipulation Capabilities}.
\newblock Preprints.org.
\newblock \doi{10.20944/preprints202610.3088.v1}.
\newblock \urlprefix\url{https://www.preprints.org/manuscript/202610.3088}.

\bibitem[{Zook et~al.(2024)Zook, Sun, Spjut, Blukis, Birchfield and Tremblay}]{grs2024}
Zook A, Sun FY, Spjut J, Blukis V, Birchfield S and Tremblay J (2024) {GRS: Generating Robotic Simulation Tasks from Real-World Images}.
\newblock arXiv:2410.15536.
\newblock \urlprefix\url{https://arxiv.org/abs/2410.15536}.

\end{thebibliography}
